%% file: main.tex
\documentclass{article}
\usepackage[preprint]{neurips_2020}

\usepackage{natbib}
 \bibpunct[, ]{(}{)}{,}{a}{}{,}%
 \def\bibfont{\small}%
\usepackage{soul}
\usepackage{graphicx}				% Use pdf, png, jpg, or eps§ with pdflatex; use eps in DVI mode
\usepackage{float}
\usepackage[font=small]{subcaption}
\usepackage{xcolor}
\usepackage{transparent}		
\graphicspath{{./diagrams/}{./figures/}}
\usepackage{amssymb, amsmath, bm}
\usepackage{mathabx}
\usepackage{bbold}
\usepackage{tabu, multirow}
\usepackage{boldline}
\usepackage{rotating}
\usepackage{hyperref}
\usepackage{arydshln}
\usepackage{algorithm}
\usepackage[noend]{algpseudocode}
\usepackage{booktabs}
\usepackage{verbatim}
\usepackage{enumitem}

\usepackage{thmtools}
\usepackage{amsthm}

\newtheorem{lemma}{Lemma}	
\newtheorem{assumption}{Assumption}
\newtheorem{definition}{Definition}

\DeclareMathOperator*{\argmax}{arg\,max}
\DeclareMathOperator*{\argmin}{arg\,min}

\newcommand{\appropto}{\mathrel{\vcenter{
			\offinterlineskip\halign{\hfil$##$\cr
				\propto\cr\noalign{\kern2pt}\sim\cr\noalign{\kern-2pt}}}}}
				
\algnewcommand{\IIf}[1]{\State\algorithmicif\ #1\ \algorithmicthen}
\algnewcommand{\EndIIf}{\unskip\ \algorithmicend\ \algorithmicif}

\title{Exact Pareto Optimal Search for Multi-Task Learning and Multi-Criteria Decision-Making}

\author{%
  Debabrata Mahapatra
    \\
  Department of Computer Science\\
  School of Computing \\
  National University of Singapore\\
  \texttt{debabrata@u.nus.edu} \\
   \And
   Vaibhav Rajan \\
   Department of Information Systems and Analytics \\
   School of Computing \\
   National University of Singapore\\
   \texttt{vaibhav.rajan@nus.edu.sg} \\
}

\begin{document}

\maketitle
% \ABSTRACT{
\begin{abstract}
Given multiple non-convex objective functions and objective-specific weights, 
Chebyshev scalarization (CS) is a well-known approach to obtain an Exact Pareto Optimal (EPO), i.e., a solution on the Pareto front (PF) that intersects the ray defined by the inverse of the weights.
First-order optimizers that use the CS formulation to find EPO solutions encounter practical problems of oscillations and stagnation that affect convergence. 
Moreover, when initialized with a PO solution, they do not guarantee a 
% descent 
controlled trajectory that lies completely on the PF.
These shortcomings lead to modeling limitations and computational inefficiency in multi-task learning (MTL) and multi-criteria decision-making (MCDM) methods that utilize CS for their underlying non-convex multi-objective optimization (MOO).
To address these shortcomings, we design a new MOO method, EPO Search.
We prove that EPO Search converges to an EPO solution and empirically illustrate its computational efficiency and robustness to initialization.
When initialized on the PF, EPO Search can trace the PF and converge to the required EPO solution at a linear rate of convergence.
Using EPO Search we develop new algorithms -- PESA-EPO, that approximates the PF for a posteriori 
% MOO,
MCDM, and GP-EPO for preference elicitation in interactive MCDM; experiments on benchmark datasets confirm their advantages over competing alternatives.
EPO Search scales linearly with the number of decision variables which enables its use for training deep networks. %\red{further, our extension to constrained MOO allows regularization to prevent overfitting}.
Empirical results on real data from personalized medicine, e-commerce and hydrometeorology demonstrate the efficacy of EPO Search for deep MTL.

\end{abstract}

% }%

% Sample
%\KEYWORDS{deterministic inventory theory; infinite linear programming duality;
%  existence of optimal policies; semi-Markov decision process; cyclic schedule}

% Fill in data. If unknown, outcomment the field
% \KEYWORDS{Chebyshev scalarization,
%  gradient descent, multi-objective optimization, Pareto front,
%  multi-criteria decision-making,  preference elicitation,
%  deep learning, multi-task learning} 
%\HISTORY{This paper wasfirst submitted on April 12, 1922 and has been with the authors for 83 years for 65 revisions.}

% \maketitle
% \begin{bibunit}
\section{Introduction}\label{sec:intro}
Multi-objective optimization (MOO) has numerous real-world applications ranging from engineering design to public sector planning \citep{stewart2008real}.
%A Pareto optimal (PO) solution is one where no  objective value can be improved further without degrading some other objectives.
A MOO problem can have multiple, possibly infinite, Pareto optimal (PO) solutions, represented by the Pareto front (PF).
%, each solution with a different trade-off between the conflicting objectives.
A MOO problem is often solved by scalarization that transforms it to a single objective optimization (SOO) problem.
A widely used technique, 
in optimization, 
decision analysis
and more recently, in artificial intelligence (see, e.g., \cite{Miettinen1998,REEVES19991311,ozbeykarwan2014,daulton2022multi}),
is the weighted Chebyshev (or Tchebycheff) scalarization (CS).
Given $m$ objective functions $f_j(\mathbf{x})$, for $j\in[m]$, on a decision or (feasible) solution space $\mathbb{X}$ and an input {\it weight} vector $\mathbf{r}\in \mathbb{R}_+^m$, CS minimizes the objective with maximum relative weighted value:
\begin{align}\label{eq:chebychev}
	    \mathbf{x}^*_\mathbf{r} = \argmin_{\mathbf{x}\in \mathbb{X}} \  \|\mathbf{r} \odot \mathbf{f}(\mathbf{x})\|_\infty = \argmin_{\mathbf{x}\in \mathbb{X}} \ \max_{j\in [m]}\, r_j f_j(\mathbf{x}),
	\end{align}
where $\odot$ is the element-wise product operator.
% The CS criterion is preferred to the Linear scalarization (LS) approach (which finds \red{eqn}) because LS cannot obtain Pareto optimal solutions for all preferences when the objectives are non-convex  and a uniform choice of weights does not necessarily find a uniform set of Pareto optimal solutions \citep{Das1997}.
A key advantage of CS, over alternative scalarizations, is that it
satisfies the necessary and sufficient conditions for modeling {\it all}  PO solutions of a non-convex MOO problem --
the complete PF can be obtained by varying the weight values $\mathbf{r}$ \citep{doi:10.1287/mnsc.39.10.1255,KALISZEWSKI1995439}. %including when the objectives are non-convex.
The solution to \eqref{eq:chebychev}, in general (weak PO solutions are an exception, see \S \ref{sec:moo}), lies at the intersection of the PF  and
the ray
$\mathbf{r}^{-1}=(1/r_1, \cdots, 1/r_m)$ as shown in Figure \ref{fig:descent_only}. 
%(except for weak PO solutions, discussed in \S \ref{sec:moo}). 
We call this an \textbf{E}xact \textbf{P}areto \textbf{O}ptimal (EPO) solution.

%\red{connect to existing literature - prev works theoretical results - they don't indicate/resolve the following practical problems?}
In this paper, our focus is on
first order methods, which can scale to high-dimensional solution spaces, to find EPO solutions for differentiable $f_j$'s.
To the best of our knowledge, extant literature
does not provide 
a robust first-order iteration strategy with convergence guarantees to find EPO solutions. 
%Na\"ively applying 
A first order method like gradient descent to solve \eqref{eq:chebychev}
has the following practical and theoretical limitations.
%leads to several 
%problems.
%in practice.
%oscillations around the $\mathbf{r}^{-1}$ ray, which adversely affects convergence. 
%\deb{We cannot cite Tanabe, they don't use CS}.
% In this setting, gradient-based iterative optimization using the min-max formulation
% of \eqref{eq:chebychev} has the following limitations.
First, it uses gradients of only one of the objectives in each iteration, which changes frequently around the $\mathbf{r}^{-1}$ ray.  
As a result, there are oscillations in the trajectory,
% around the $\mathbf{r}^{-1}$ ray 
which slows convergence during descent.
Second, 
% during descent, 
if the gradient magnitude vanishes for 
% one of the objectives, 
the objective with highest weighted value, 
descent stagnates. In such cases, movement in each iteration
is negligibly small.
Third, when initialized with a PO solution, it does not guarantee that the 
% descent 
trajectory to the required EPO solution remains on the PF. 
Further, the non-differentiable $\max$\
function in the $\ell_\infty$ norm of \eqref{eq:chebychev} makes convergence rate analysis for gradient-based methods non-trivial.
These shortcomings lead to
computational inefficiency and modeling limitations in methods, such as those outlined below, that utilize CS in solving their underlying non-convex MOO problems.

Consider neural multi-task learning (MTL) where each objective is a loss function (usually non-convex) for a task and a MOO solution corresponds to trained neural network parameters.
Linear scalarization (that solves $\argmin_{\mathbf{x} \in \mathbb{X}}\, \mathbf{r}^T\,\mathbf{f}(\mathbf{x})$) is commonly used to train MTL models, where the weights specify relative priorities among tasks.
CS is theoretically advantageous for non-convex functions and is also more interpretable (see \S \ref{sec:scalarization}, \ref{sec:mtl_back}).
%allows us to specify relative priorities among tasks.
However, first order solvers that optimize the min-max formulation in \eqref{eq:chebychev} face challenges during training of deep networks due to the aforementioned problems of oscillations and stagnation.
%They use 
Since the gradient of only one of the objectives is used in each iteration,
%As a result, within each iteration, 
optimization effectively leads to just single-task learning which ultimately deteriorates the model's predictive accuracy.
The same problem occurs when a relatively high priority is given to a task -- the other tasks are completely ignored during training.
%always chooses that task ignoring others.
%and so, advantages of a multi-task model may never be realized.
This may be alleviated through second order methods but they are not scalable to high-dimensional parameter spaces in deep networks.
%\red{The second problem occurs when the gradient magnitude vanishes for one of the objectives during descent. As a result, that objective function is no longer effective during descent which prematurely stagnates. Thus, extant solvers for CS-based MOO do not provide a robust first order iteration strategy to converge to the EPO solution.} 

As another example, consider multi-criteria decision-making (MCDM), where the decision maker (DM) has to choose the most suitable PO solution of the underlying (non-convex) MOO problem.
%This is in the form of preferences specified by the DM, which can be provided in many different ways.
We study two approaches -- (i) a posteriori methods, where multiple PO solutions are computed that collectively provide an approximate view of the PF and enables the 
%are computed from which the 
DM to select one desired solution
and (ii) interactive methods,
where the DM progressively articulates preferences among solutions 
% and, along with 
% obtained from the MOO solver, 
and 
% iteratively 
proceeds towards a satisfactory solution 
while interacting with the MOO solver.
The DM's preferences are assumed to follow an (unknown) utility function that can score and order PO solutions, and is monotonic, i.e., a solution that Pareto dominates another has higher utility.
In both these cases, there are multiple calls to the MOO solver, each time after a PO solution is obtained.
If the min-max formulation \eqref{eq:chebychev} is used,
% to find each solution, and 
% the 
% descent 
and the trajectory between consecutive solutions is not on the PF, 
the MCDM approach has high computational burden.
Moreover, such solvers also need to be re-initialized and re-started at PF discontinuities where they may halt prematurely.
%where re-initialization is required. 
%and the solver has to be re-started.

%To address the limitations of MOO methods that use \eqref{eq:chebychev} to obtain EPO solutions, 
We design a new approach, called EPO Search, to efficiently find EPO solutions for non-convex MOO, which addresses these limitations. 
Using EPO Search we develop techniques that advance the state-of-the-art in first order methods for (a) PF approximation for a posteriori MCDM, (b) preference elicitation in interactive MCDM and (c) training deep multi-task neural networks.
The four main contributions of this paper are as follows.
%\red{TODO: make each point precise, should clarify contribution wrt SOTA}

1. 
We design and analyze search direction strategies that {\it balance} the dual goals of moving towards the PF as well as towards the $\mathbf{r}^{-1}$ ray, which equip us to combine gradient descent
with carefully controlled ascent in objectives with less relative weights to avoid their local minima.
%and a stopping strategy that does not halt at any PO solution, but only at an EPO solution.
%We design and analyze the properties of \textit{balancing} search directions to simultaneously move towards the PF and the $r^{-1}$ ray in iterative first-order multi-objective gradient-based solvers. %(\S \ref{sec:proportionality}).
By using a linear combination of all objective gradients, while moving in a balancing search direction, EPO Search avoids the problems of oscillations and stagnation.
%faced with gradient descent using CS. % (\S \ref{sec:QP}).
%Assuming only differentiability of the objective functions (they need not be convex), and 
%When the initialization is random,
%the iterative procedure of 
We prove, under mild assumptions and without assuming convexity, that,  from a random initialization,
EPO Search converges 
to the EPO solution, or, if an exact solution does not exist, to a PO solution closest to the $r^{-1}$ ray.
% (\S \ref{sec:epo_rand}).
When initialized at an arbitrary PO solution,
we prove that EPO Search converges to the desired EPO solution  
% while guaranteeing a descent trajectory 
with a trajectory close to the PF and % (\S \ref{sec:epo_trace}). 
under mild regularity conditions, 
%without assuming convexity, 
we prove that the convergence rate, even for non-convex objectives, 
%of EPO search, when initialized from a PO solution, 
is linear. % (\S \ref{sec:conv_rate}). 
EPO Search scales linearly with the gradient dimension per iteration and thus,
can efficiently find (local) EPO solutions in high-dimensional solution spaces. % (\S \ref{sec:time_complexity}). %
We extend EPO Search 
% to find EPO solutions 
for solving constrained MOO problems
% with box and equality constraints on the parameters 
without compromising on its computational efficiency. % (\S \ref{sec:constrained_moo}). 
Our empirical results on benchmark MOO problems support the theoretical claims of
scalability and accuracy
of EPO Search. % (\S \ref{sec:exp_toy_moo}).

2. 
%A posteriori methods using \eqref{eq:chebychev} have to independently solve optimization problems for each input $\mathbf{r}^{-1}$ ray to obtain an approximation of the Pareto front.
%They also face challenges at Pareto front discontinuities where re-initialization is required and the solver has to be re-started.
%As a result, they are computationally inefficient especially with increasing number of objectives.
%Given multiple input weight vectors, the tracing ability of EPO Search enables it to efficiently find multiple PO solutions and yield an approximation of the PF.
Using PESA \citep{Stanojevic2020} to generate a diverse set of weight vectors, 
%in the $m-1$ dimensional simplex $\mathcal{S}^m$, 
we develop the algorithm PESA-EPO to approximate the PF. Leveraging
the PF tracing ability of EPO Search,
PESA-EPO efficiently finds multiple PO solutions
without 
%which does not require 
multiple optimizer calls, and without premature halts at PF discontinuities.
 % (\S \ref{sec:pesa-epo}).
%Thus, given an input set of preferences, EPO Search can efficiently move from one EPO Solution to another, and present  solutions on the (approximate) Pareto front delimited by the input preferences.
%We evaluate the tracing capability of EPO-Search on the PF, using 
On several benchmark convex and non-convex MOO problems, both with and without constraints, % (\S \ref{sec:pf_trace_exp}, \ref{sec:appox_pf}).
%Our results demonstrate that 
%the use of EPO Search 
PESA-EPO
leads to better or comparable PF approximation in lower execution time, compared to competing gradient-based as well as evolutionary MOO algorithms.

3. In probabilistic preference elicitation, a Gaussian Process (GP), which can model any (including non-linear and non-convex) function,
is used to learn the DM's unknown utility.
Pairwise comparisons from the DM are used to 
% iteratively
interactively
learn the GP parameters in a Bayesian active learning framework, where, in 
%to our knowledge, 
each interaction, the DM specifies her preference between the two presented PO solutions. 
%all 
%previous approaches %obtain PO samples by solving an optimization problem over a discrete subset of the solution space  
To reduce their computational burden,
%of sampling from the entire solution space, 
previous approaches, e.g., \cite{8618894,10.5555/3237383.3237920}, sample these PO solutions from a discrete subset of the solution space (see \S \ref{sec:interactive_back}), which affects their accuracy of preference learning.  
Further, since sampling from the GP does not guarantee a PO solution, 
%to ensure that only PO solutions are presented to the DM for comparison,
previous methods impose monotonicity constraints during or employ postprocessing heuristics after sampling from the GP; these steps either deteriorate performance or increase computational time.
%\red{Both these steps are computationally demanding and increase the time between queries presented to the DM.}
We address these limitations by developing GP-EPO where we explore the PF at high resolution, by
sampling $\mathbf{r}^{-1}$ rays (in a lower $m$-dimensional space, instead of the entire solution space) and then efficiently find PO solutions through EPO Search to present to the DM. Our approach 
obviates the need to explicitly model monotonicity constraints.
Further, over the interactions, GP-EPO moves from one EPO solution to another with linear convergence rate.
%developing a mapping from the higher dimensional solution space to lower dimensional simplex $\mathcal{S}^m$ and using EPO Search to efficiently find EPO solutions. % (\S \ref{sec:gp-epo}). 
%GP-EPO obtains a more computationally efficient solution compared to other GP-based PE strategies for MODM.
Evaluation on benchmark problems show that GP-EPO 
learns the utility with 
%achieves 
substantially 
better accuracy,
%lower regret, 
compared to extant GP-based methods, in just a few interactions.
%, even when compared to strong baseline approaches that assume knowledge of the Pareto front. % (\S \ref{sec:pref_eli_exp}).

4. In MOO-based neural MTL, which
models tradeoffs among objectives,
%during network training, 
%where the MOO solution corresponds to trained network parameters.
previous methods either
do not use task-specific priorities or yield multiple PO solutions for an input set of diverse relative priorities.
An EPO solution models task priorities specified by the $\mathbf{r}^{-1}$ ray, thus prioritizing tasks that are challenging to learn; without losing the benefits of MTL that allows shared learning from other datasets and tasks.
Compared to gradient descent to solve \eqref{eq:chebychev} for network training,
EPO Search offers a more robust iterative procedure that overcomes the problems of oscillation and stagnation; % (\S \ref{sec:mtl}). 
further, its ability to use the gradients of all objectives and escape  minima of lower priority objectives leads to improved MTL. 
%The computational efficiency of EPO Search enables its use for MTL using highly parameterized deep neural networks.
The per-iteration complexity of EPO Search remains linear in the gradient dimensions (similar to the best previous methods that  neither use input priorities nor allow regularization constraints) enabling its use for deep MTL networks.
%Thus, the use of EPO Search for network training enables prioritizing training for tasks that are challenging (by specifying task-specific weights), without losing the benefits of the MTL framework that allows shared learning other datasets and tasks.
We evaluate the efficacy of EPO Search for MTL on three real datasets from different domains: personalized medicine, e-commerce and hydrometeorology. % (\S \ref{sec:expt_real}, \ref{sec:more_results}).
In all cases, the use of EPO Search leads to
%we demonstrate the advantages of our MTL formulation leads to we obtain
higher predictive accuracy compared to single-task learning, the direct use of linear and Chebyshev scalarization during training and competing MTL models.

The rest of the paper is organized as follows.
Background and related work are presented in \S \ref{sec:related}.
We then describe our theory and algorithms for EPO Search in \S\ref{sec:epo_search}.
Algorithms PESA-EPO, GP-EPO and EPO Search for MTL are described in \S \ref{sec:epo_mcdm_mtl}.
Experimental results are in \S \ref{sec:expt}, followed by our concluding discussion in \S \ref{sec:concl}.

\section{Background and Related Work}
\label{sec:related}
We describe relevant concepts and related work from three streams of literature --  multi-objective optimization (MOO), multi-criteria decision making (MCDM) and multi-task learning (MTL).
%-- and highlight our contributions relative to extant literature.
% \deb{Should we add a DAG/table of reading sequence to aid the reader, backgraound $\rightarrow$ method $\rightarrow$ experiments, for all the contributions: MOO, a posterior, interactive and MTL? }

% \begin{table}[h]
% \centering
% \caption{Reading sequence}
% \label{tab:my-table}
% \begin{tabular}{ccccc}
%                          &             & Background       $\rightarrow$               & Contribution      $\rightarrow$            & Experiment \\ \toprule
% \multicolumn{2}{c}{MOO}                & \S \ref{sec:moo}               & \S \ref{sec:epo_search}       &  \S \ref{sec:exp_toy_moo}, \S \ref{sec:pf_trace_exp}  \\ \midrule
% \multirow{2}{*}{MCDM}   & A posterior &  \S \ref{sec:posterior_back}     & \S \ref{sec:epo_trace}, \S \ref{sec:pesa-epo}  &   \S \ref{sec:pf_trace_exp},\S \ref{sec:appox_pf}\\ \cmidrule{2-5}
%                         & Interactive  &  \S \ref{sec:interactive_back}  &  \S \ref{sec:epo_trace}, \S \ref{sec:gp-epo}            &  \S \ref{sec:pref_eli_exp} \\ \midrule
% \multicolumn{2}{c}{MTL}                &   \S \ref{sec:mtl_back}         &  \S \ref{sec:epo_rand}, \S \ref{sec:mtl}  &  \S \ref{sec:expt_real}          \\ \bottomrule
% \end{tabular}
% \end{table}

\subsection{Multi-Objective Optimization}
\label{sec:moo}

We consider a multi-objective optimization (MOO) problem with $m$ non-negative differentiable objective functions, $f_j:\mathbb{X}\rightarrow \mathbb{R}_+$ for $j\in[m]$, where $\mathbb{X}\subset \mathbb{R}^n$. This formulation is fairly general, since problems with different specifications can be converted to this form. For instance, if an objective $f_j$ is negative at its minimizer, i.e., $\min_{\mathbf{x}\in \mathbb{X}} f_j(\mathbf{x}) = f^*_j \leq 0$, then, to make it non-negative, one can reformulate as $f_j(\mathbf{x}) := f_j(\mathbf{x}) - f^{**}_j$, where $f^{**}_j$ is a lower bound on the minimum: $f^{**}_j\leq f^*_j$. The vector $\mathbf{f}^{**}$ consisting of the lower bounds of all objectives is called as a ``utopia'' point, which makes $\mathbf{f} - \mathbf{f}^{**}$ a non-negative vector valued function. We develop MOO algorithms for unconstrained problems in the main paper, and extend them to solve constrained MOO problems in $\S \ref{sec:constrained_moo}$.

We use $\mathbf{f}$ to denote both a vector valued %\st{loss}
function and a point in the \textit{Objective Space} $\mathbb{R}^m$, which should be unambiguous from the context. The range of $\mathbf{f}$, denoted by $\mathcal{O}$, is a subset of the positive cone $\mathbb{R}^m_+ := \left\lbrace \mathbf{f}\in \mathbb{R}^m \;\middle|\; f_j \ge 0 \ \forall j \in [m]\right\rbrace$.
% 	\begin{align}
% 	\mathbb{R}^m_+ := \left\lbrace \mathbf{f}\in \mathbb{R}^m \;\middle|\; f_j \ge 0 \ \forall j \in [m]\right\rbrace.
% 	\end{align} 
	The partial ordering for any two points $\mathbf{f}^1, \mathbf{f}^{2} \in \mathbb{R}^m$, denoted by $\mathbf{f}^1 \succcurlyeq \mathbf{f}^2$ is defined by $\mathbf{f}^1-\mathbf{f}^2 \in \mathbb{R}^m_+$, which implies $f^1_j \ge f^2_j$ for every $j\in [m]$ and strict inequality $\mathbf{f}^1 \succ \mathbf{f}^2$ occurs when there is at least one $j$ for which $f^1_j > f^2_j$. Geometrically, $\mathbf{f}^1\succ \mathbf{f}^2$ means that $\mathbf{f}^1$ lies in the positive cone pivoted at $\mathbf{f}^2$, i.e., $\mathbf{f}^1 \in \{\mathbf{f}^2\} + \mathbb{R}^m_+ := \left\{ \mathbf{f}^2 + \mathbf{f} \;\middle|\; \mathbf{f}\in \mathbb{R}^m_+ \right\}$, and $\mathbf{f}^1 \neq \mathbf{f}^2$. 
	% A multi-objective value $l\in \mathcal{O}$ is said to be minimal if there exists no other point $l' \in \mathcal{O}$ such that $l' \succcurlyeq l$.
	
	For a minimization problem, a solution  ${\mathbf{x}^1 \in \mathbb{X}}$ is (weakly) \textit{dominated} by another solution $\mathbf{x}^2\in \mathbb{X}$ if ($\mathbf{f}(\mathbf{x}^1) \succcurlyeq \mathbf{f}(\mathbf{x}^2)$) $\mathbf{f}(\mathbf{x}^1) \succ \mathbf{f}(\mathbf{x}^2)$.
    Note that $\mathbf{f}(\mathbf{x}^1) \nsucc \mathbf{f}(\mathbf{x}^2)$ if $\mathbf{x}^1$ is not dominated by $\mathbf{x}^2$, i.e. $\mathbf{f}(\mathbf{x}^1) \notin \{\mathbf{f}(\mathbf{x}^2)\} + \mathbb{R}^m_+$.
	%, we denote their relation as . 
	A solution $x^*$ is  \textbf{Pareto optimal (PO)} if it is not dominated by any other solution. Weak PO solutions are weakly dominated by other PO solutions.
    The set of all global PO solutions is $\mathcal{P}_{glo} := \left\lbrace \mathbf{x}^* \in \mathbb{X} \;\middle|\; \forall \mathbf{x} \in \mathbb{X} \backslash \{\mathbf{x}^*\},\ \mathbf{f}(\mathbf{x}^*) \nsucc \mathbf{f}(\mathbf{x})\right\rbrace$,
% 	\begin{align}
% 	\label{eq:pareto_set}
% 	\mathcal{P}_{glo} := \left\lbrace \mathbf{x}^* \in \mathbb{X} \;\middle|\; \forall \mathbf{x} \in \mathbb{X} \backslash \{\mathbf{x}^*\},\ \mathbf{f}(\mathbf{x}^*) \nsucc \mathbf{f}(\mathbf{x})\right\rbrace.
% 	\end{align}
	%The Pareto set in \eqref{eq:pareto_set} characterizes all the global Pareto optimal points. Similarly, we can also define a set of solutions, which are 
	%We are interested in 
    and the set of local PO solutions is:
	\begin{align}
	\mathcal{P} := \left\lbrace \mathbf{x}^* \in \mathbb{X} \; \middle|\;
	\ \exists\  \mathcal{N}_\epsilon(\mathbf{x}^*) \subset \mathbb{X} \ \text{for an } \epsilon>0, \text{ s.t. } \forall \mathbf{x} \in \mathcal{N}_\epsilon(\mathbf{x}^*) \backslash \{\mathbf{x}^*\}, \  \mathbf{f}(\mathbf{x}^*) \nsucc \mathbf{f}(\mathbf{x})
%	\begin{tabular}{@{}l@{}}
%	$\exists\ \mathcal{N}(x^*) \subset \mathbb{X}$ \textbar \\
%	$\forall x \in \mathcal{N}(x^*) \backslash \{x^*\}$, \\ 
%	$l(x^*) \nsucc l(x)$
%	\end{tabular}
	\right\rbrace,
	\end{align}
	where $\mathcal{N}_\epsilon(\mathbf{x}^*)=\{\mathbf{x}\in \mathbb{X} \,|\, \|\mathbf{x}-\mathbf{x}^*\| < \epsilon\}$ is an open neighbourhood of $\mathbf{x}^*$ in $\mathbb{X}$. Note that $\mathcal{P}_{glo}\subset \mathcal{P}$. %In this paper, we are interested in local Pareto Optimal solutions. So 
	%We use $\mathcal{P}$ to denote both local and global solutions.
	The set of multi-objective values of the PO solutions, $\mathbf{f}(\mathcal{P}) \subset \mathcal{O}$, is called the \textbf{Pareto front (PF)}. 
%Simultaneously optimizing multiple, possibly conflicting criteria in multi-objective optimization problems has been studied extensively.
%in Multiobjective Optimization (MOO). 
Excellent surveys on MOO can be found in, e.g., 
\cite{gandibleux2006multiple,Wiecek2016}.

	\subsubsection{Descent Methods for Converging to the Pareto Front.}\label{sec:grad_moo}
% 	\subsubsection*{Gradient-based MOO.}
	%

	Gradient-based MOO solvers, such as in %Bento2012
 \cite{fliege2000steepest,Vieira2012}, find a PO solution by starting from an arbitrary initialization $\mathbf{x}^0 \in \mathbb{R}^n$ and iteratively obtaining the next solution $\mathbf{x}^{t+1}$ that dominates the previous one $\mathbf{x}^t$ (i.e., $\mathbf{f}^{t+1} \preccurlyeq \mathbf{f}^t$, where $\mathbf{f}^t := \mathbf{f}(\mathbf{x}^t)$), by moving \textit{against} a direction $\mathbf{d}\in \mathbb{R}^n$ with a step size $\eta>0$ as $\mathbf{x}^{t+1} = \mathbf{x}^t - \eta \mathbf{d}$, such that there is descent in every objective, $f^{t+1}_j \le f^{t}_j \,\,\forall j \in [m]$. This can happen only if $\mathbf{d}$ has positive angles with the gradients of every objective function at $\mathbf{x}^t$:
	\begin{align}\label{eq:des_dir}
	 \mathbf{d}^{T} \,\nabla_{\!\!\mathbf{x}}f_{j}^{t} \geq 0, \quad \text{for all } j\in[m]\, ;
	 \end{align}
    we call such a $\mathbf{d}$ a \textit{descent direction}. Note, by convention, the move is against $\mathbf{d}$, i.e., along $-\mathbf{d}$.

	\citet{Desideri2012} showed that descent directions can be found in the \textit{Convex~Hull} of the gradients, 
% 	defined by $\mathcal{CH}_{\mathbf{x}} :=\left\lbrace \sum_{j=1}^{m} \nabla_{\!\!\mathbf{x}} f_{j} \, \beta_{j} \;\middle|\; \bm{\beta} \in \mathcal{S}^m \right\rbrace$
	\begin{align} \label{eq:ch_grads}
	    \mathcal{CH}_{\mathbf{x}} :=\left\lbrace \sum_{j=1}^{m} \nabla_{\!\!\mathbf{x}} f_{j} \, \beta_{j} \;\middle|\; \bm{\beta} \in \mathcal{S}^m \right\rbrace
	\end{align}
	where $\mathcal{S}^m := \left\lbrace \bm{\beta} \in \mathbb{R}_+^m \,\middle|\, \sum_{j=1}^m \beta_j = 1, \ \text{and } \ \beta_{j} \geq 0 \ \ \forall j \in [m]\right\rbrace$
% 	\begin{align}\label{eq:simplex}
% 	    \mathcal{S}^m := \left\lbrace \bm{\beta} \in \mathbb{R}_+^m \,\middle|\, \sum_{j=1}^m \beta_j = 1, \ \text{and } \ \beta_{j} \geq 0 \ \ \forall j \in [m]\right\rbrace
% 	\end{align}
% 	\begin{align}
% 	\mathcal{CH}_{\mathbf{x}} &:= \left\lbrace \sum_{j=1}^{m} \nabla_{\!\!\mathbf{x}} f_{j} \, \beta_{j} \;\middle|\; \sum_{j=1}^m \beta_j = 1, \ \text{and } \ \beta_{j} \geq 0 \ \ \forall j \in [m] \right\rbrace. \label{eq:cv}
% 	\end{align}
    is the $m-1$ dimensional simplex.
	Their multiple gradient descent algorithm (MGDA) converges to a local PO by iteratively using the descent direction: $\mathbf{d}^* = \arg\min_{\mathbf{d} \in \mathcal{CH}_{\mathbf{x}}} \|\mathbf{d}\|_2^2\,.$ Although descent-based methods provide convergence guarantees \citep{Tanabe2022}, they cannot find EPO solutions.

 \subsubsection{Scalarization.}	\label{sec:scalarization}
	
\begin{figure}[t]
	    \centering
	    \begin{subfigure}[t]{0.48\textwidth}
	        \centering
	        \includegraphics[page=3,trim=13cm 3cm 13cm 5cm, clip,width=0.7\linewidth]{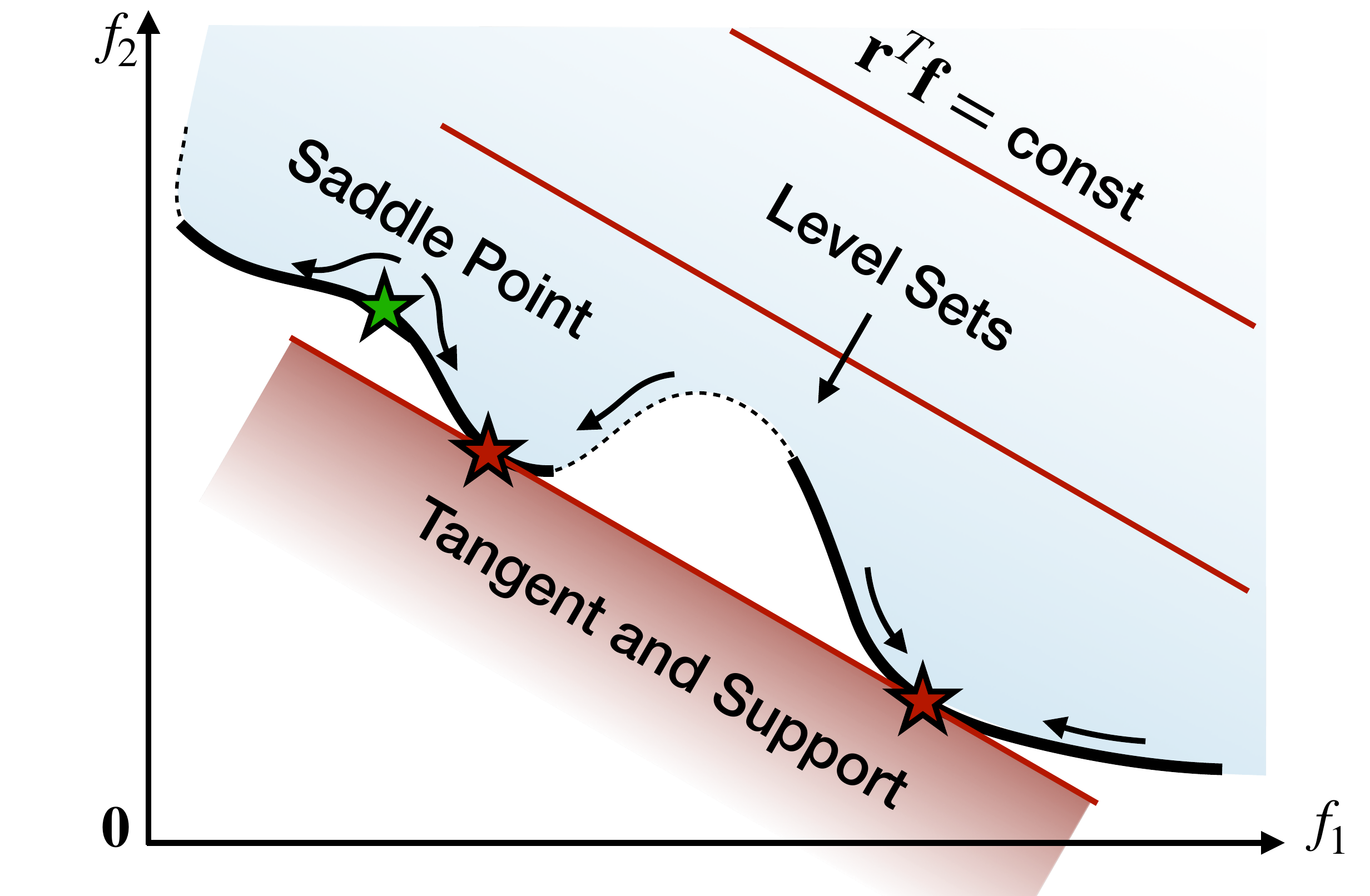}
	        \caption{Linear Scalarization}
	        \label{fig:ls_limit}
	    \end{subfigure}~
	    \begin{subfigure}[t]{0.48\textwidth}
	        \centering
	        \includegraphics[page=4,trim=13cm 3cm 13cm 5cm, clip,width=0.7\linewidth]{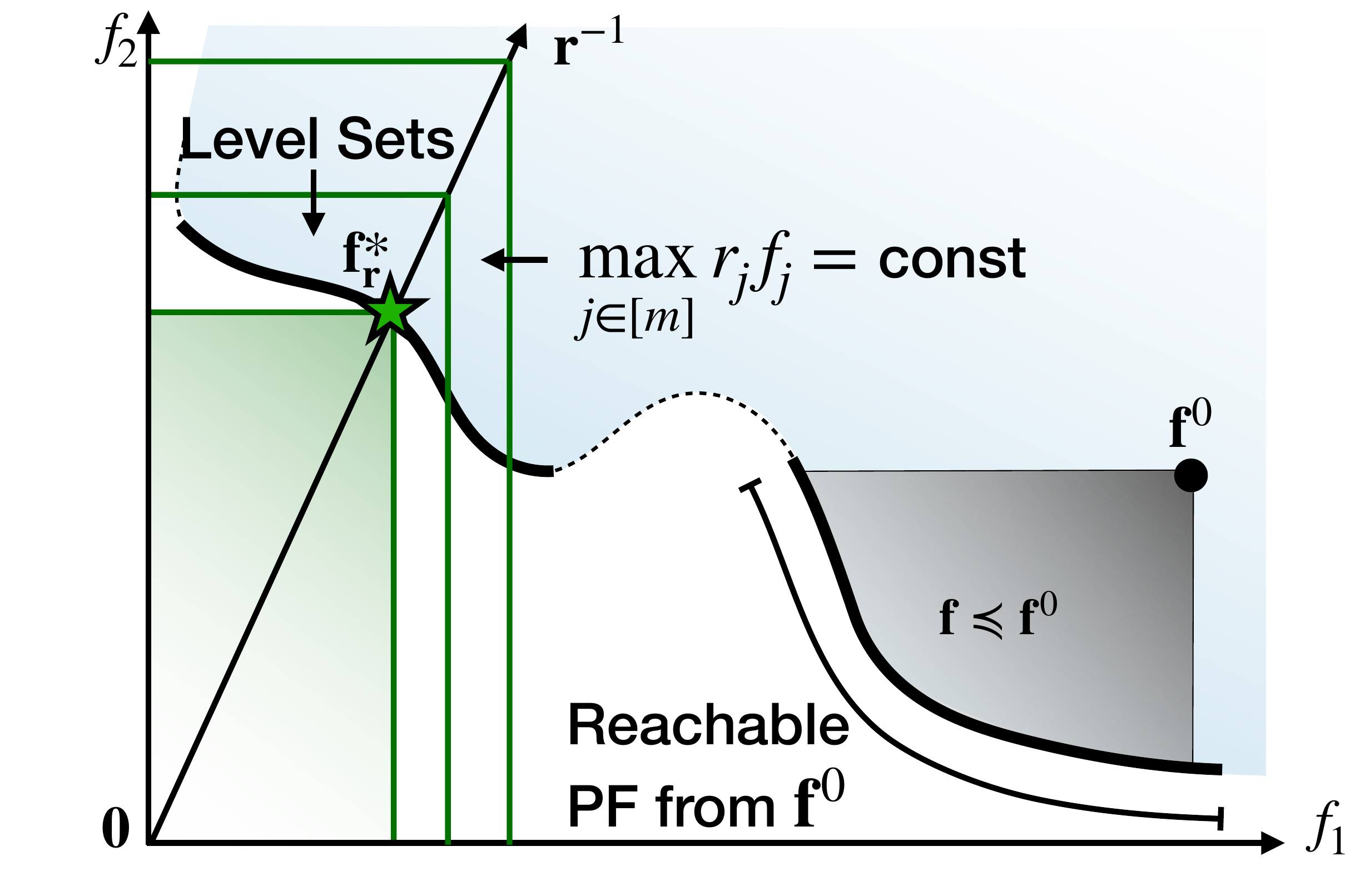}
	        \caption{Chebyshev Scalarization \& Descent method}
	        \label{fig:descent_only}
	    \end{subfigure}
	    \vspace{1mm}
	    \caption{(Color Online) 
	    Figure \ref{fig:ls_limit} shows how linear scalarization can have non-unique Pareto optimal points (red) for the same weight $\mathbf{r}$; among the 3 optima, the green one, a saddle point for LS \eqref{eq:soo}, cannot be attained. Figure \ref{fig:descent_only} shows Chebyshev scalarization \eqref{eq:chebychev} can attain the green optimum; it also illustrates how a \textit{descent-only} search cannot find the preferred optimal $\mathbf{f}_\mathbf{r}^*$ starting from a random initialization $\mathbf{x}^0$, where $\mathbf{f}^0=\mathbf{f}(\mathbf{x}^0)$. Any descent direction $\mathbf{d}$ will keep the objective vector $\mathbf{f}^{t+1}$ of next iterate $\mathbf{x}^{t+1} = \mathbf{x}^t - \eta \mathbf{d}$ in the black shaded region.
		\label{fig:moo_illus}}
	\end{figure}

	The popular approach of linear scalarization (LS) of a MOO problem with an input weight vector $\mathbf{r}\in \mathbb{R}_+^m$ finds
	\begin{align}
		\label{eq:soo}
		\mathbf{x}^*(\mathbf{r}) = \argmin_{\mathbf{x} \in \mathbb{X}} \ \langle \mathbf{r},\, \mathbf{f}(\mathbf{x})\rangle \, = \argmin_{\mathbf{x} \in \mathbb{X}}\, \mathbf{r}^T\,\mathbf{f}(\mathbf{x}).
	\end{align}
	%and obtain many Pareto optimal solutions using different $\mathbf{r}$. 
%	Without loss of generality, we assume that $\sum_{j=1}^m r_j = 1$. 
	If the range of %\st{vector valued objective} 
 $\mathbf{f}$, i.e. $\mathcal{O}$, is convex then for every $\bar{\mathbf{x}}^* \in \mathcal{P}$, there exists an $\mathbf{r}$ such that $\bar{\mathbf{x}}^* = \mathbf{x}^*(\mathbf{r})$. 
% 	However, if $\mathcal{O}$ is non-convex then it may not be possible to reach every optimal point in $\mathcal{P}$ using linear scalarization (see \cite{boyd_vandenberghe_2004}[Ch 4.7]).
	The weight vector $\mathbf{r}$ is an element in the dual of the objective space \citep{Luenberger1997}, and represents a hyperplane in the objective space. The hyperplane of $\mathbf{r}$ at $\mathbf{f}_\mathbf{r}^* = \mathbf{f}(\mathbf{x}^*(\mathbf{r}))$ has to be both a tangent and a support to the PF, i.e., $\mathbf{r}^T\mathbf{f} \geq \mathbf{r}^T\mathbf{f}_\mathbf{r}^*$ for all $\mathbf{f} \in \mathcal{O}$. 
 %Specifying trade-offs by elements in the dual space has limitations. 
 If any of the objectives %costs 
	is non-convex, i.e., the range $\mathcal{O}$ is a non-convex set, LS cannot guarantee to reach every optimal point in the PF by varying the weights (\cite{boyd_vandenberghe_2004}[Ch 4.7]), 
 %The reason is that 
 because the tangent hyperplane at a point on the PF is not necessarily a support if $\mathcal{O}$ is non-convex. 
 %Therefore, the trade-off specification provided by the preferences is not guaranteed to be satisfied in the final solution.
 Moreover, even a single vector $\mathbf{r}$ can have non-unique PO solutions; see
 %Moreover, a single weight vector specification can have non-unique Pareto optimal solutions, as illustrated in 
 Figure \ref{fig:ls_limit}.
 
%\blue{In general, a solution to the CS criterion \eqref{eq:chebychev}, is an EPO solution} defined as follows.
\begin{definition}\label{def:epo}
    	Given a {\it weight vector} $\mathbf{r}\in \mathbb{R}_+^m$,
	%$\mathbf{x}_\mathbf{r}^* \in \mathcal{P}$ is Pareto optimal {\it and} satisfies the property:
	%for any given $\mathbf{r} \in \mathbb{R}^m_+$ 
% 	{\color{blue}
% 	if $r_j \ge r_{j'}$, then 
% 	%the Pareto optimal solution $\mathbf{x}_\mathbf{r}^* \in \mathcal{P}$ satisfies
% 	$f_j(\mathbf{x}_\mathbf{r}^* )\le f_{j'}(\mathbf{x}_\mathbf{r}^*) \quad \forall j,j' \in [m]$}.
	an \textbf{Exact Pareto Optimal} (EPO) solution %with respect to a preference vector %$\mathbf{r}\in \mathbb{R}_+^m$ 
	belongs to the set: 
	%want a solution from the set
	\begin{align}
		\label{eq:epo}
		\mathcal{P}_\mathbf{r} = \left\{\mathbf{x}^* \in \mathcal{P} \;\middle|\; r_1f_1^* = \cdots = r_j f_j^* = \cdots = r_mf_m^*\right\}, 
        \quad \text{where} \ f_j^* = f_j(\mathbf{x}^*).
	\end{align}
	% where $f_j^* = f_j(\mathbf{x}^*)$. 
\end{definition}
	For any EPO solution $\mathbf{x}_\mathbf{r}^*$, 
	$\mathbf{f}_\mathbf{r}^*=\mathbf{f}(\mathbf{x}_\mathbf{r}^*)$ is a point on the PF intersecting the ray towards $\mathbf{r}^{-1} := (1/r_1, \cdots, 1/r_m) $ (see Figure \ref{fig:descent_only}), i.e.,
 $\mathbf{f}_\mathbf{r}^*$ is perfectly {\it proportional} to the $\mathbf{r}^{-1}$ ray.
%{\color{blue}\st{, that we shall use,}} the objective with maximum relative value is minimized:
% 	\begin{align}\label{eq:chebychev}
% 	    \mathbf{x}^*(r) = \argmin_{\mathbf{x}\in \mathbb{X}} \  \|\mathbf{r} \odot \mathbf{f}(\mathbf{x})\|_\infty = \argmin_{\mathbf{x}\in \mathbb{X}} \ \max_{j\in [m]}\, r_j f_j(\mathbf{x}),
% 	\end{align}
% 	where $\odot$ is the element-wise product operator. 
 CS, i.e., a solution to \eqref{eq:chebychev}, is an EPO solution, as illustrated in Figure \ref{fig:descent_only}, because, its $c-$level set $\mathcal{L}_c=\{\mathbf{f} \in \mathbb{R}^m \, | \, \max_{j\in[m]} r_jf_j \leq c\}$ can also be written as $\{c \mathbf{r}^{-1}\} - \mathbb{R}_{+}^m$, the negative orthant pivoted  at $c \mathbf{r}^{-1}$. Therefore, if $\mathcal{L}_c\cap \mathcal{O}$ is a singleton set then the corresponding solution is an EPO point w.r.t. 
 %the objective weights 
 $\mathbf{r}$.
	%Although its optimal solution models the EPO point, the CS method doesn't provide a robust strategy to iterate using the objective gradients. Practical aspects of an optimization procedure, such as step-wise iteration and random initialization, lead to oscillation and premature stagnation (see section \ref{sec:exp_toy_moo}) in CS.
% 	based approach, since it uses only one gradient, i.e. from maximum relative objective, in each iteration.
	
Although for convex MOO problems LS and CS may be considered equivalent \citep{doi:10.1287/mnsc.1060.0595}, they differ for non-convex MOO. Unlike LS, CS provides both necessary and sufficient condition for weak Pareto optimality of non-convex MOO problems.
CS has extensions such as augmented CS \citep{10.1007/978-3-642-74919-3_8,Miettinen1998} that eliminates the corner case of weak PO solutions.
%wherein some of the objective values can be further improved without degrading others. 
However, such extensions lose the exactness property of CS for its strong PO solutions,
%, wherein no further improvement in any objective is possible without degrading others. 
and we leverage the exactness property for our algorithms.
    These scalarization methods, their variants and generalizations \citep{Gembicki1975ApproachTP,Pascoletti1984,Wierzbicki1986,marler2004survey} specify a desired PO solution through their parameters. However, to our knowledge, there is no
    %they do not provide a 
    robust first-order iterative
	strategy that guarantees convergence to a desired EPO %\st{Pareto optimal} 
 solution starting from a random initialization.
 
 \subsection{Multi-Criteria Decision-Making (MCDM)} \label{sec:mcdm_back}
MCDM aims to solve decision-making problems involving multiple conflicting objectives \citep{10.2307/20122479,koksalan2012multiple}.
%koksalan2011multiple,
We focus on continuous MCDM problems, 
%also known as Multi-Objective Decision Making (MODM), 
with differentiable objectives, where the solution alternatives are (or can relaxed to be) in a continuous space. 
% When formulated as MOO problems, which 
% %A Pareto optimal (PO) solution is one where no  objective value can be improved further without degrading some other objectives.
% can have multiple (possibly infinite) PO solutions, each solution with a different trade-off between the conflicting objectives.
% To find the most suitable among the PO solutions, additional preference information from the Decision Maker (DM) is required.
Based on how a DM participates in the solution process 
MCDM approaches can be categorized into \citep{hwang1979methods,Miettinen1998}: % three types:
(i) a priori methods where preferences are specified before the MOO problem is solved
(ii) a posteriori methods where multiple PO solutions are computed from which the DM selects one and 
(iii) interactive methods
where the DM progressively articulates preferences  and, along with the MOO solver, iteratively proceeds towards a satisfactory solution.

In each of these categories, there are various ways to articulate preferences  %and PO solutions may be obtained 
\citep{Miettinen1998}.
%Preference information may be provided by comparison of objectives, e.g., as $m$ scalar weights that reflect relative importance of objectives \cite{} or by classifying the objectives into multiple classes based on various criteria \cite{}.
Preferences may be articulated by comparison of PO solutions, e.g., through pairwise comparisons (e.g., \cite{korhonen1984solving,koksalan1995interactive}) or selection from a set of solutions (e.g., \cite{greco2010interactive}).
% Preference models, such as utility (or value) functions \cite{keeney1993decisions}, may be used to evaluate PO solutions, and can be of various types.
A utility function $u:\mathbb{R}^m\rightarrow \mathbb{R}$ that assigns a scalar value to every objective vector \citep{keeney1993decisions}, may be used to model the DM's preferences and create
    %which creates 
    a total ordering among the PO solutions.
    The utility is commonly assumed \citep{10.2307/2628674} to satisfy the monotonicity property: for any two alternatives $\mathbf{x}^1$ and $\mathbf{x}^2$, if $\mathbf{x}^1$ dominates $\mathbf{x}^2$ (i.e. $\mathbf{f}^1
    %\mathbf{f}(\mathbf{x}^1) 
    \preccurlyeq \mathbf{f}^2$),
    %=\mathbf{f}(\mathbf{x}^2)$, 
    then $u(\mathbf{f}^1) \geq u(\mathbf{f}^2)$, where $\mathbf{f}^i = \mathbf{f}(\mathbf{x}^i)$.
    The oracle solution maximizes the DM's (unknown) utility:
    \begin{align} \label{eq:pe_ideal}
        \mathbf{x}_{orc} = \argmax_{\mathbf{x}\in \mathbb{X}}\ u\left(\mathbf{f}(\mathbf{x})\right).
    \end{align}
    %However the DM's utility function is unknown. }\deb{these sentence can be used in the initial description of MCDM.}  
    %Preference elicitation techniques are used to estimate this utility by comparing (pairwise in our case) different solutions. 
%E.g., scalar function of all objectives such as deterministic value functions or stochastic utility functions.
The DM's utility may be learnt in both a posteriori and interactive MCDM, e.g., in 
\cite{JING2019123}, a utility function is constructed using the PF approximation from an a posteriori method; 
%The parameters of the function may be specified by the DM or may have to be learnt through 
preference elicitation (PE) methods may be used in interactive MCDM (see \S \ref{sec:interactive_back}). 

% \deb{We can cite the MS 2007 paper \cite{doi:10.1287/mnsc.1060.0595} on  “Equivalent Information for Multiobjective Interactive Procedures”.Our earlier notes:
% \begin{itemize}
% \item Discusses
% relations among the different kinds of information (local
% weights, local trade-offs, reference points, etc.) to allow the DM to choose what kind of questions he
% wants to answer, and to provide him with enough information to give such answers. 
% \item Defines
% equivalent information—that is, different kinds of information—that produces the same solution when used in
% their corresponding interactive schemes.
% \end{itemize}
% }

	\subsubsection{A Posteriori Methods and Pareto Front Approximation.}\label{sec:posterior_back}
% 	\subsubsection*{Generating Diverse Solutions and Pareto Front Approximation.}
% 	\red{Uses, recent methods, limitations}
	In a posteriori methods, 
 %\red{ \citep{Messac2002}, first a diverse set of PO solutions are first generated that approximate the Pareto Front, then the DM selects the one best suited for the application \blue{, e.g., \cite{JING2019123} where a utility function is constructed from the  approximation set.} \red{some other OR/MS journal?}
  %{\color{gray}
	%These methods are useful when a priori preference articulation is difficult. 
 %In these scenarios,} 
 the ability of an algorithm to quickly and evenly approximate the PF is crucial \cite{Das1997,Das1998}.
	Many Multi-Objective Evolutionary Algorithms (MOEA) %\citep{Emmerich2018} 
 have been developed to approximate the PF, 
 %leveraging their inherent design of iterating with a population of solutions,
 e.g., \citep{zhang2007moea,4633340,10.1162/EVCO_a_00109,7347861,996017,6600851,10.1007/s00500-017-2609-4,8413136}. 
 %  \red{\cite{zhang2007moea} developed MOEA/D that evenly decomposes the objective space into several regions and solves a SOO problem parametrically formulated for each region. Many variants of MOEA/D have been developed since then %,
	% \citep{4633340,10.1162/EVCO_a_00109,7347861}.
	% %  e.g., MOEA/DDE\citep{4633340}, MOEA/D-AWA \citep{10.1162/EVCO_a_00109}, and MOEA/D-UD \citep{7347861}. 
	% The popular NSGA-II \citep{996017} method was improved by incorporating the decomposition idea in NSGA-III \citep{6600851} and recent extensions include MOEA/D-ABD \citep{10.1007/s00500-017-2609-4} that handles discontinuous Pareto fronts, and CTAEA \citep{8413136} that allows for constraints. These methods are gradient-free and can also handle objectives without functional forms (black box functions). 
	% %But they are beyond the scope of this study; in this work we limit our focus to the MOOPs with differentiable objectives. 
	% Although MOEA methods do not provide theoretical convergence guarantees,} 
 They are effective in practice and scale well to many objectives \citep{Emmerich2018}. However, they do not scale well to high-dimensional solution spaces.
	%due to computationally infeasible requirements in population sizes.
	%requirement makes them computationally infeasible.
	
	In gradient-based MOO, a scalarization method is run several times with a diverse set of parameters to generate well spread out PF approximations. The simple LS method fails to generate evenly spread solutions from a set of evenly spread weights  \citep{Das1997}.
	%for a detailed analysis. 
	To address this problem, \cite{Das1998} developed a direction-based scalarization method called Normalized Boundary Intersection (NBI) that finds PO solutions along many directions normal to the Convex Hull of the Individual Minima (CHIM) of the objective functions. 
	Several improvements of NBI have been proposed \citep{Ismail2002,Shukla2007,10.1007/s00158-012-0797-1} but they remain computationally inefficient as they require re-initializations at PF discontinuities and do not scale well to many objectives.
	%It was improved in \cite{Ismail2002,Shukla2007} to remove non-Pareto solutions. NBI methods thus far required solving several SOO problems to approximate the PF, one for each direction vector, making the whole process time consuming. \cite{10.1007/s00158-012-0797-1} addressed it with modified NBI where the SOO problems are initialized with an existing Pareto optimal and the iteration history of each optimization is included in the final approximation set. However, the iterations in the SOO prematurely stops if there are discontinuities in the PF, needing re-initialization of additional SOO problems. Moreover, this method also struggles to generate an even spread of PO solutions for more than two objectives \cite{10.1007/s00158-012-0797-1} as it does not use evenly spread directions to begin with. 
	%Recently 
 \cite{Stanojevic2020} proposed a recursive sampling strategy Pattern Efficient Set Algorithm (PESA) that samples weights/directions from the simplex spread uniformly, and scales well with number of objectives.
	%generalizes to higher dimensions (i.e. many objectives). 
	They used an NBI type scalarization called Targeted Direction Model (TDM) to find direction specific solutions. However, similar to NBI, their method also has to solve many SOO problems, one for each direction ray. 
% We address these computational limitations by developing PESA-EPO that combines the recursive sampling of PESA with the efficient PF tracing ability of EPO Search.
	
% 	\red{The tracing ability of EPO Search enables it to overcome these computational difficulties.
% 	Given a set of input preference vectors (e.g., from PESA), EPO Search can trace the PF from one EPO Solution to another, collecting Pareto optimal solutions during the process.
% 	This enables it to 
% % 	In this research, we extend the proposed EPO search algorithm to approximate the PF by inheriting the virtues of PESA and modified NBI. We use a set of evenly spread directions and include the iteration history of each optimization to 
% generate an approximation set of the PF in a computationally efficient manner. Moreover, the controlled ascent aspect of our algorithm guarantees, under mild conditions, that the iterations do not stop prematurely for disconnected PF thereby avoiding unnecessary re-initialization. 
% %We also highlight limitations of our approach by formalizing the corner cases where the iterations do get stuck. 
% }
	
% 	We require the gradient information. The Pareto navigation methods that uses gradient can be compared with ours. But ones that assumes black box objectives \cite{Hartikainen2019} are not comparable to ours.

% 	\cite{COCCHI2021100008} consider problems with convex constraints.
	
\subsubsection{Preference Elicitation for Interactive MCDM.} \label{sec:interactive_back}
% In interactive MCDM \citep{Xin2018}, a DM is progressively queried to compare a few alternatives to 
% find his/her most preferred solution among a larger set of alternatives. 
%In this research, 
%There are several interactive methods that are primarily developed for solving discrete MCDM problems, also known as Multi-Attribute Decision Making (MADM) (\cite{MI2019205}), such as 
%It is assumed that a DM's preference can be represented as a scalar valued utility function (\cite{board2009preferences}), 
    %We assume
    %In interactive MCDM, 
    %the otherwise incomparable Pareto optimal solutions.
    Preference Elicitation (PE) aims to 
%\red{iteratively finds optimal solution to} 
infer the DM's unknown utility function using the DM's responses to queries in interactive MCDM.
We assume the queries to the DM are in the form of pairwise comparison of alternatives, which is cognitively less demanding compared to other query types (such as value assignment) \citep{NIPS1988_a8baa565,Forgas1995}.
%For example, 
Methods such as 
Conjoint Analysis \citep{rao2010conjoint,ANGUR1996195} and Best-Worst Method \citep{oztas2021} 
%have been developed 
are well-known
for discrete MCDM problems.
%the focus of our research.
%Previous methods for PE,
In continuous MCDM, the utility is a fixed parametric function of the objective values, e.g., Chebyshev utility that uses CS \citep{10.1007/978-3-642-74919-3_8,doi:10.1287/mnsc.39.10.1255,dellkarwan1990,ozbeykarwan2014,REEVES19991311}; and in Preference Robust Optimization (PRO), where \cite{https://doi.org/10.48550/arxiv.2003.01899} model the utility as a linear function and \cite{haskell2018preference} model it as a quasi-concave function. These methods cannot model non-convex PE problems, where either the MOO problem or the utility function could be non-convex. 
See Appendix \ref{sec:pe_det} for an extended discussion.

In contrast to these methods, the Bayesian active learning framework for PE 
\citep{eric2007active,10.5555/3237383.3237920,10.1007/978-3-030-67664-3_28}
%does not assume a fixed function and 
uses Gaussian Process (GP) to model the utility that can model {\it any} class of functions.
%\citet{10.1145/1102351.1102369} proposed the use of Gaussian Process (GP) for PE. 
 %Therefore, the utility need not be restricted to {\it any} class of functions. %(quasi-) concave functions. 
%The other advantages of GP are that 
Further, this can model uncertainty in the DM's stated preferences and 
works well with less data -- requiring just a few interactions with the DM to learn preferences
\citep{deisenroth2013gaussian}.
%In Gaussian Process (GP) based Preference Elicitation (PE) 
%(\cite{10.1145/1102351.1102369,eric2007active,8618894,10.5555/3237383.3237920}), 
Appendix \ref{sec:gp} has an overview of Bayesian optimization with GP.
% If the observational data $\mathcal{D}$ consists of direct (noisy) measurements of utility function at different alternatives, then the posterior distribution of utility function, $P(u|\mathcal{D}) \propto P(u) P(\mathcal{D}|u)$, would also be a GP with updated mean and covariance functions. However, in case of PE,

The prior for the unknown utility function, $P(u)$, is modelled as $\mathrm{GP}(\mu, \kappa)$, where $\mu:\mathbb{R}^m \rightarrow \mathbb{R}$ and $\kappa:\mathbb{R}^m \times \mathbb{R}^m \rightarrow \mathbb{R}_+$ are the mean and covariance functions respectively.
Learning proceeds by alternating between querying the DM based on the current GP and updating the GP posterior after obtaining the resulting comparison from the query.
Thus, the observational data from which the GP is learnt is $\mathcal{D}_t=\{c_1, \cdots, c_{t'}\}$, where $t$ is the number of pairwise queries to the DM, and an ordered pair $c = (\mathbf{x}^i, \mathbf{x}^j)$ represents the DM's \textit{comparison} of two alternatives as ``$\mathbf{x}^i$ is preferred to $\mathbf{x}^j$'', suggesting $u(\mathbf{f}^i) \geq u(\mathbf{f}^j)$.
If the DM values both alternatives equally, then the dataset can include both $(\mathbf{x}^i, \mathbf{x}^j)$ and $(\mathbf{x}^j, \mathbf{x}^i)$, making $|\mathcal{D}_t| = t' \geq t$.
% \red{The posterior distribution, $P(u|\mathcal{D}_{t+1}) \propto P(u) P(\mathcal{D}_{t}|u)$, 
% would also be a 
% %with updated mean and covariance functions.
% $\mathrm{GP}(\mu_{t+1}, \kappa_{t+1})$,
% %$ = P(u | \mathcal{D}_t)$, 
% where $\mu_{t+1}$ and $\kappa_{t+1}$ are the updated mean and covariance functions after collecting $t+1$ comparisons from the DM.}\deb{Posterior will not be a GP.}
Inconsistencies in DM's response are modelled as additive noise: $u(\mathbf{f}^i) + \epsilon_i \geq u(\mathbf{f}^j) + \epsilon_j$, where $\epsilon_i,\epsilon_j\sim \mathcal{N}(0, \sigma)$. 
For such noisy pairwise comparison data, \citet{10.1145/1102351.1102369} developed a probit likelihood model for $P(\mathcal{D}_t | u)$, wherein the posterior, $P(u|\mathcal{D}_{t+1}) \propto P(u) P(\mathcal{D}_{t}|u)$, is analytically non-tractable, but can be approximated to a GP using Laplace approximation. %\cite{brochu2010tutorial}. 
%The approximated posterior can be written as $\mathrm{GP}(\mu_{t}, \kappa_{t}) = P(u | \mathcal{D}_t)$, where $\mu_t$ and $\kappa_t$ are the updated mean and covariance functions after collecting $t$ comparisons from the DM.
In the beginning, when there is no comparison data, $\mathcal{D}_0 = \phi$, the GP prior is usually initialized with a zero mean function.
% The query to be presented to the DM is selected in a manner that balances {\it exploration} (i.e., of items with high uncertainty in utility) with {\it exploitation} (i.e., where items with high expected utility are presented).
% Such a selection is enabled through acquisition functions.

Let $\ddot{\mathbb{X}}_{\mathcal{D}_t}$ be the discrete set of alternatives presented to the DM, and $\mathbf{x}_{inc}^t \in \ddot{\mathbb{X}}_{\mathcal{D}_t}$ be the incumbent solution, where $\mathbf{x}_{inc}^t$ is preferred to all other solutions in $\ddot{\mathbb{X}}_{\mathcal{D}_t}$. 
In each iteration, selection of a new alternative to create the next query is done using an \textit{acquisition function} $\alpha^t:\mathbb{R}^m\rightarrow \mathbb{R}$.
%that enables balancing {\it exploration} (of alternatives with high variance) with {\it exploitation} (of alternatives with high expected utility).
%To find a better alternative than the incumbent, first, the unknown $u$ is approximated to an 
%instead of the unknown utility 
% that is formulated using the updated $\mu_t$ and $\kappa_t$, e.g., Expected Improvement {\color{blue} (cite)} and Upper Confidence Bound {\color{blue} (cite)}. Then, instead of \eqref{eq:pe_ideal}, 
A commonly used acquisition function in previous works on interactive PE, e.g., \cite{eric2007active}, is the \textit{Expected Improvement} (\cite{10.1007/3-540-07165-2_55}) function which
is optimized to \textit{suggest} a new alternative:
\begin{align} \label{eq:pe_prac}
    \mathbf{x}_{sug}^t = \argmax_{\mathbf{x} \in \ddot{\mathbb{X}}} \ \alpha^t\left(\mathbf{f}(\mathbf{x})\right),
\end{align}
where $\ddot{\mathbb{X}}$ is a discrete subset of $\mathbb{X}$. 
%and Upper Confidence Bound {\color{blue} (cite)}.
% In interactive PE (\cite{10.5555/3237383.3237920,10.1007/978-3-030-67664-3_28}), 
The DM is then asked to compare between $\mathbf{x}^{inc}_t$ and $\mathbf{x}^{sug}_t$.
The new comparison datapoint $(\mathbf{x}_{inc}^t, \mathbf{x}_{sug}^t)$
is included in the dataset, and the posterior is updated to $GP(\mu^{t+1}, \kappa^{t+1})$ to approximate $P(u|\mathcal{D}_{t+1})$. This iterative procedure 
%of iteratively querying the DM and updating GP 
is continued until, either $\|\mathbf{f}_{sug}^t - \mathbf{f}_{inc}^t\| \leq \epsilon$ for a small $\epsilon > 0$ or the DM is satisfied with $\mathbf{x}_{sug}^t$.

%While they offer the flexibility of modeling {\it any} class of utility functions, 
Previous GP-based approaches have two important limitations.
First, %to our knowledge, all the previous works 
they optimize over a discrete subset, $\ddot{\mathbb{X}}$, of the entire search space, %Discretization is practiced 
because $\mu^t$ and $\kappa^t$ can be highly non-linear, which renders the global optimization of $a^t$ \eqref{eq:pe_prac} computationally expensive, especially if $\mathbb{X}$ is high dimensional.
Decreasing the size of the discrete subset improves computational efficiency but reduces accuracy of preference learning. 
Second, additional monotonicity constraints need to be enforced because sampling from a GP does not guarantee a PO solution. 
%unless all the solutions in $\ddot{\mathbb{X}}$ are PO, a monotonicity prior has to be enforced explicitly in the GP to avoid %suggesting dominated  solutions. 
%In the sampling step (i) 
%GPs cannot enforce such monotonicity constraints since solutions are sampled from the entire solution domain.
Several heuristics to address this problem have been proposed
\citep{8618894,10.5555/3237383.3237920,10.1007/978-3-030-67664-3_28} that either deteriorate performance or
% include \red{linear priors \cite{8618894}}, data augmentation \cite{10.5555/3237383.3237920}
% and heuristics to prune out non-dominating solutions \cite{10.1007/978-3-030-67664-3_28}.
% The use of linear priors deteriorates performance after a few %initial
% queries \cite{10.5555/3237383.3237920} while other strategies 
increase the computational burden. 
%per-step computational burden of learning.
%We use the tracing ability of EPO Search, to develop GP-EPO,
%our proposed algorithm, GP-EPO, 
%that addresses these limitations.
%by reformulating the problem, which enables optimization without discretization and obviates the need for additional monotonicity constraints on the GP.
%by ensuring that only PO solutions are chosen during sampling.

% We improve the modeling and efficiency by re-formulating the optimization problem on the $m-1$ dimensional simplex, and mapping the samples to the solution space for presentation to the DM.
% This is facilitated by our unique approach of solving the underlying MOO problem with EPO Search.
% Moreover,
% In contrast, our proposed approach obviates the need to model additional monotonicity constraints on the GP by ensuring that only Pareto optimal solutions are chosen during sampling.

%In practice, however, instead of enforcing the monotonicity, it is easier to remove the dominated solutions from $\ddot{\mathbb{X}}$ (e.g. \cite{10.1007/978-3-030-67664-3_28}) making it a discrete approximation to the Pareto frontier. 

\subsection{Multi-Task Learning} \label{sec:mtl_back}
Multi-Task Learning (MTL) has been studied extensively 
in machine learning
(see, e.g., 
\cite{zhang2021survey}). %provide a general survey and \cite{ruder2017overview} gives an overview of neural MTL models. 
Learning multiple tasks together leads to inductive bias towards hypotheses that can explain 
%(or internal representations that are predictive of) 
more than one task and has been found to 
improve model generalization in machine learning \citep{caruana1997multitask}.

\begin{figure}[!h]
    \centering
    \includegraphics[page=1,trim=0 11.5cm 0 12cm, clip,width=\linewidth]{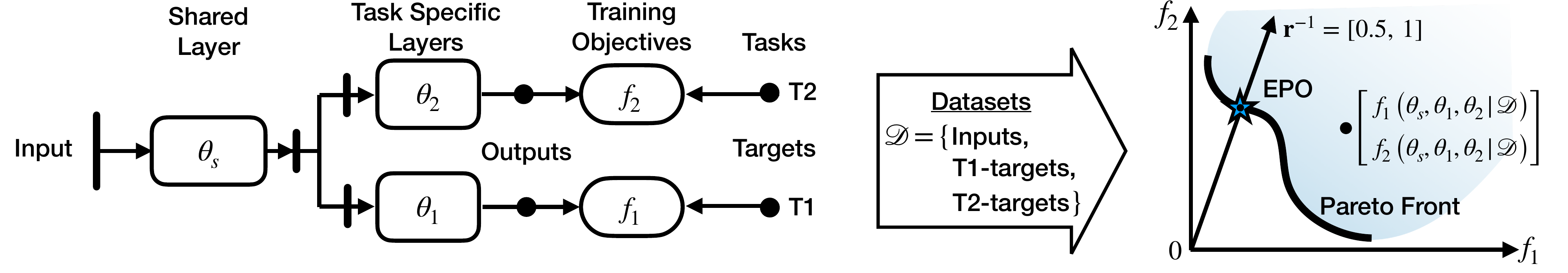}
    \caption{
    Left: Schematic of a MTL DNN for 2 tasks;
    Right: MOO-based network training yields multiple PO solutions for $f_1,f_2$ forming a PF, 
    %Illustration of the task specific cost functions of a typical MTL-DNN model forming a Pareto Front, and the 
    EPO solution (trained network parameters) for priorities $r_1 =2, r_2=1$.}%  $\mathbf{r}=2:1$.}
    \label{fig:mtl_moo}
\end{figure}

A multi-task neural network is trained for multiple tasks simultaneously and inductive transfer is enabled
through shared parameters, most commonly through fixed layers common to all tasks \citep{ruder2017overview} (Appendix \ref{sec:nn} has an overview of %optimization in training 
neural networks).
%Corresponding to each task, the network has a loss function that has to be minimized during training.
%is trained on a collection of input data points and target variables per task.
% The most common approach for MTL in DNNs is through 
% hard parameter sharing in fixed layers that are common to all tasks.
% %Such models have been used for state-of-the-art systems in many applications.
%These models are trained using gradient-based optimization
% and most commonly involve LS or its variants, e.g., with adaptive weights \citep{guo2018dynamic,chen2018gradnorm,kendall2018multi}.
%,heydari2019softadapt}.
Figure \ref{fig:mtl_moo} (left) shows a schematic of a MTL network for two tasks T1, T2 with objective (loss) functions $f_1, f_2$ respectively.
The network has one or more shared layers with parameters $\theta_s$ and task-specific layers with parameters $\theta_1,\theta_2$ that are learnt during training.
% For such networks, it has been shown that overfitting the shared parameters 
%or soft parameter sharing (where each task has its own layers but are constrained to be similar during training). \red{EXPAND}. 
% Model parameters ($\theta_1,\theta_2,\theta_s$ in Figure \ref{fig:mtl_moo}) are learned by solving an optimization problem
% %, usually by gradient descent,
% based on the objective (or loss) functions for each task, where the loss functions are determined by tasks and their target variables.
For conflicting tasks, we cannot assume that  parameters learned through SOO are effective across all tasks
%because this %LS may not model 
as trade-offs among the tasks are not explicitly modeled.
In such cases,
PO solutions from
MOO 
%multi-objective optimization (MOO), 
yield better models, as shown by \cite{NeurIPS2018_Sener_Koltun}.
%MGDA \citep{Desideri2012} reduced the dimension of the 
	%{\it inner} optimization problem of finding the steepest descent direction, in each iteration of gradient descent, 	from that of solution space (network parameters) to objective space (loss functions).
 %  , which is useful for 
	% %steepest descent in 
	% training models where the dimension of objective space (number of loss functions) is typically less than that of the solution space (number of network parameters). 
%Pareto optimal solutions with different trade-offs were first modeled in MTL through gradient-based MOO by \cite{NeurIPS2018_Sener_Koltun}.
% Gradient-based MOO algorithms, such as MGDA (described in \S \ref{sec:moo}), 
% Thus, they 
% are a natural choice for MOO-based MTL: task-specific gradients are used to update the neural network weights during training in both the method of \cite{NeurIPS2018_Sener_Koltun} and Pareto MTL (PMTL) \citep{NIPS2019pmtl}.
%\cite{NeurIPS2018_Sener_Koltun} 
They extended MGDA to handle high-dimensional gradients, thereby making it usable for deep MTL models.
However, their method finds a single arbitrary PO solution 
	%(figure \ref{fig:comparison}(b)) %
	%from the Pareto set 
	and cannot be used by MTL designers to explore solutions with different trade-offs. 
	%This limitation was recognized by 	
%	In MGDA and MGDA-based algorithms (e.g., \citep{NeurIPS2018_Sener_Koltun}), 
	%using a descent-only direction, we can only find a solution that dominates the previous one, without any control over moving towards the preference. Thus, depending on the initialization $\mathbf{x}^0$, the algorithm may reach \textit{any} local Pareto optimal. This has also been empirically verified by \citet{NIPS2019pmtl}.
%	starting from a random initialization,	iterating along descent directions can lead to a Pareto optimal solution, but it may not be a preference-specific optimal. 
 This is illustrated in Figure \ref{fig:descent_only}.

Current MTL approaches which use MOO for training, do not explicitly model task-specific priorities.
While LS has been used in SOO settings, to our knowledge, CS, has not been used in MTL.
Task-specific priorities through CS can be effected through EPO Search for
%, e.g., through a CS-based criterion 
model training. 
	The related, Pareto MTL (PMTL) algorithm by \citet{NIPS2019pmtl}, finds multiple solutions on the PF through a decomposition strategy and may be modified to find EPO solutions. They use several reference vectors $\mathbf{u}^k, k=1,\ldots,K$, each of unit magnitude, to
	partition the solution space into $K$ sub-regions $\Omega_k := \left\{\mathbf{x} \in \mathbb{R}^n \;\middle|\; \langle \mathbf{u}^k,\, \mathbf{f(x)} \rangle \, \ge\, \langle \mathbf{u}^{k'},\, \mathbf{f(x)} \rangle, \ \forall k'\neq k \right\}$.
	% \begin{align}\label{eq:decomp}
	%    \Omega_k := \left\{\mathbf{x} \in \mathbb{R}^n \;\middle|\; \langle \mathbf{u}^k,\, \mathbf{f(x)} \rangle \, \ge\, \langle \mathbf{u}^{k'},\, \mathbf{f(x)} \rangle, \ \forall k'\neq k \right\}
	% \end{align}
% 	$\Omega_k := \left\{\mathbf{x} \in \mathbb{R}^n \;\middle|\; \langle \mathbf{u}^k,\, \mathbf{f(x)} \rangle \, \ge\, \langle \mathbf{u}^{k'},\, \mathbf{f(x)} \rangle, \ \forall k'\neq k \right\}$. 
	With this decomposition, if $\mathbf{u}^k = \mathbf{r}^{-1}$, then the EPO solution $\mathbf{x}_\mathbf{r}^* \in \Omega_k$. There are two phases in their algorithm. In phase one, starting from a random initialization, they find a point $\mathbf{x}^0_\mathbf{r} \in \Omega_k$, such that the corresponding $\mathbf{u}^k = \mathbf{r}^{-1}$.
	In phase two, they iterate using  descent-only directions to reach a PO $x^* \in \mathcal{P}$.
	However, their method does not guarantee that the outcome of second phase $x^*$ also lies in $\Omega_k$. 
  Moreover,
 to reach a desired EPO solution, they have to increase	
	the number of reference vectors $\mathbf{u}^k$ exponentially with increase in number of objectives $m$.
 %, making it practically infeasible.
	%$Thus, although their reference vectors can be based on user preferences, 
 Their method, by design, does not reach an EPO solution but only in the sub-regions of the PF between the references (see \S \ref{sec:exp_toy_moo} and Appendix \ref{sec:pmtl_info}).
\section{Exact Pareto Optimal Search Algorithms}\label{sec:epo_search}
    Our key idea is to gain control over the trajectory of objective vectors in the objective space $\mathbb{R}^m$ so that an iterative algorithm can efficiently reach the desired solution. To \textit{anchor} the iterations, we design suitable vector fields on $\mathbb{R}^m$ by defining their direction:
    %, called as \textit{anchoring direction}. 
    \begin{definition}[Anchoring Direction]\label{def:anchoring_dir}
        We call an element of the tangent space $\mathcal{T}$ of the objective space $\mathbb{R}^m$  an \textit{Anchoring direction}, denoted by $\mathbf{a}(\mathbf{f}) \in \mathcal{T}_{\mathbb{R}^m}(\mathbf{f})$, where $\mathbf{f}\in \mathbb{R}^m$ is the footpoint.
        %and $\mathcal{T}$ denotes the tangent space.
        %$\mathbf{a}(\mathbf{f})$ is the direction.
    \end{definition}
    % We design many anchoring directions in this research (to obtain search directions $\mathbf{d} \in \mathbb{R}^n$ in the solution space) that \textit{anchor} the search procedure towards the EPO solution.   
    
    To find the EPO solution by an iterative procedure, it is not sufficient to advance only along the descent directions \eqref{eq:des_dir} because it  leads to an arbitrary solution in the PF. Moreover, even if $\mathbf{f}^{t+1} \prec \mathbf{f}^{t}$ for every $t$, the solution may not lie on the $\mathbf{r}^{-1}$ ray (figure \ref{fig:descent_only}), and hence will not satisfy the condition in \eqref{eq:epo}. 
    Therefore, apart from a descent direction that moves the objective vector $\mathbf{f}^{t}$ closer to the PF, we also need to consider a search direction $\mathbf{d} \in \mathbb{R}^{n}$ (tangent space of $\mathbb{X}$) that moves the objective vector $\mathbf{f}^{t}$ ``closer'' to the $\mathbf{r}^{-1}$ ray.     To find such a direction, in \S \ref{sec:proportionality}, first we define a general \textit{Proportionality Gauge} (PG) to measure the closeness between $\mathbf{f}^{t}$ and $\mathbf{r}^{-1}$ and analyze its properties. Then we present three specific examples of PGs 
    % in \S \ref{sec:mu_kld}, \S \ref{sec:mu_sci} and \S \ref{sec:mu_li} 
    whose (scaled) gradient fields, called as \textit{balancing anchor directions}, can advance the iterates $\{\mathbf{f}^t\}_{t=1,2, \cdots}$ closer towards the $\mathbf{r}^{-1}$ ray, while maintaining $\mathbf{f}^{t+1} \nsucc \mathbf{f}^{t}$. 
    In \S \ref{sec:descending_dir}, we present \textit{descending anchor direction} to advance closer to the PF with $\mathbf{f}^{t+1} \prec \mathbf{f}^{t}$. We then develop two MOO algorithms in \S \ref{sec:epo_rand} and \S \ref{sec:epo_trace} to reach an EPO solution, starting from a random initialization in $\mathbb{X}$ and $\mathcal{P}$ respectively, by using both balancing and descending anchor directions.

    % We introduce the notion of \textit{Anchoring directions} $\mathbf{a} \in \mathbb{R}^{m}$ (tangent plane of $\mathcal{O}$) that anchor the desired changes in consecutive objective vectors: whether to a) make $\mathbf{f}^{t+1}$ more proportional to the $\mathbf{r}^{-1}$ ray than $\mathbf{f}^{t}$ while $\mathbf{f}^{t+1} \nsucc \mathbf{f}^{t}$ (developed in \S \ref{sec:proportionality}), or b) advance closer to the Pareto Front with $\mathbf{f}^{t+1} \prec \mathbf{f}^{t}$ (discussed in \S \ref{sec:descending_dir}). All proofs are in Appendix \ref{sec:proofs}.

    % To find such a direction, we first construct an anchoring direction $\mathbf{a} \in \mathbb{R}^{m}$ (tangent plane of $\mathcal{O}$) according to the desired change in the consecutive objective vectors: $\mathbf{f}^{t+1}$ should be more proportional to the $\mathbf{r}^{-1}$ ray than $\mathbf{f}^{t}$ in addition to  $\mathbf{f}^{t+1} \nsucc \mathbf{f}^{t}$.
    % We define and analyze three specific ways of quantifying the proportionality in \S \ref{sec:mu_kld}, \S \ref{sec:mu_sci} and \S \ref{sec:mu_li}.
    % Before that, in \S \ref{sec:proportionality} %\ref{sec:baldir}
    % we describe the properties required from such a measure to gauge proportionality that allow us to characterize the search direction in terms of the anchor direction.
    % All proofs are in Appendix \ref{sec:proofs}.

    \subsection{Proportionality of Vectors and Balancing Direction} \label{sec:proportionality}
    % \subsubsection{A Balancing Search Direction}
    % \label{sec:baldir}
    % \deb{Convert it to a formal definition}
    %We propose the following definition to measure the proportionality between the two vectors.
    \begin{definition}[Proportionality Gauge]\label{def:prop_gauge}
    A function $\omega: \mathrm{R}^{m}_{+} \times \mathbb{R}^{m}_{++} \rightarrow \mathbb{R}_{+}$ is called a \textit{Proportionality Gauge}, if, for any given inputs $\mathbf{f} \in \mathbb{R}^{m}_{+}$ and $\mathbf{r}\in \mathbb{R}^{m}_{++}$, we have:
    \begin{enumerate}
    \item $\omega(\mathbf{f}, \mathbf{r}^{-1}) = 0$ only when $\mathbf{f}$ is a positive scalar multiple of $\mathbf{r}^{-1}$, and\label{num:prop+}
    \item \label{num:mtct}$\omega_{\mathbf{r}}(\cdot) = \omega(\cdot,\ \mathbf{r}^{-1})$ is differentiable w.r.t $\mathbf{f}$ and increases monotonically along $(1-\lambda)\mathbf{r}^{-1} + \lambda \mathbf{f}$, for $\lambda \geq 0$, with increment in $\lambda$.\label{num:prop_smth_mntnc}
    \end{enumerate}
    \end{definition}
    
    % Let $\omega: \mathrm{R}^{m}_{+} \times \mathrm{R}^{m}_{++} \rightarrow \mathrm{R}_{+}$ be a function that measures the degree of proportionality between two vectors, such that, for any given $\mathbf{f} \in \mathbb{R}^{m}_{+}$ and $\mathbf{r}\in \mathrm{R}^{m}_{++}$
    % \begin{enumerate}
    % \item $\omega(\mathbf{f}, \mathbf{r}^{-1}) = 0$ only when $\mathbf{f}$ is a positive scalar multiple of $\mathbf{r}^{-1}$, and\label{num:prop+}
    % \item \label{num:mtct}$\omega_{\mathbf{r}}(\cdot) = \omega(\cdot,\ \mathbf{r}^{-1})$ is differentiable w.r.t $\mathbf{f}$ and increases monotonically along $(1-\lambda)\mathbf{r}^{-1} + \lambda \mathbf{f}$, for $\lambda \geq 0$.\label{num:prop_smth_mntnc}
    % \end{enumerate}

    For a given weight vector $\mathbf{r}$ and a point $\mathbf{f}$, we define an anchor direction as $\mathbf{a}(\mathbf{f}) := s\nabla_{\!\!\mathbf{f}}\, \omega_{\mathbf{r}}$ for some $s>0$. We use it to characterize a search direction $\mathbf{d}\in \mathbb{R}^{n}$ to move the objective vector $\mathbf{f}(\mathbf{x})$ closer to $\mathbf{r}^{-1}$ ray. 
    
    \begin{restatable}{lemma}{ancr}
    \label{th:anchor}
    If all the objective functions are differentiable, then for any direction $\mathbf{d}\in \mathbb{R}^n$ satisfying $\mathbf{a}^{T}\mathrm{F}\,\mathbf{d} \geq 0$, where $\mathrm{F}$ is the Jacobian of $\mathbf{f}$ at $\mathbf{x}$, and $\max_{j} \{\mathbf{d}^{T}\,\nabla_{\!\mathbf{x}}\!f_{j}\} > 0$, there exists a step size  $\eta_0 > 0$ such that for all $\eta \in [0,\eta_0]$
        \begin{subequations}
            \begin{align}	\omega\!\left(\mathbf{f}(\mathbf{x} - \eta \mathbf{d}), \, \mathbf{r}^{-1} \right) &\leq \omega\!\left(\mathbf{f}(\mathbf{x}),\, \mathbf{r}^{-1}\right), \; \text{ and } \;
			\label{eq:mu+<mu} \\
			\mathbf{f}(\mathbf{x} - \eta \mathbf{d}) &\nsucc \mathbf{f}(\mathbf{x}). \label{eq:f+n>f}
		\end{align}
        \end{subequations}
    \end{restatable}

    A move against the search direction $\mathbf{d}$ of Lemma \ref{th:anchor} reduces the variations in relative objective values $f_{j}r_{j}$ to make them equal: brings {\it balance} among the values of $\mathbf{f} \odot \mathbf{r} = [f_{1}r_{1}, \cdots, f_{m}r_{m}]$. Therefore, we call this $\mathbf{d}$ a \textit{Balancing Search Direction}, and $\mathbf{a}$ a \textit{Balancing Anchor Direction}. 
    We call a balancing anchor direction $\mathbf{a}$  \textit{Scale Invariant} to $\mathbf{r}$ if ${\nabla_{\!\!\mathbf{f}}\, \omega_{s\mathbf{r}} = \nabla_{\!\!\mathbf{f}}\, \omega_{\mathbf{r}}}$ for all $s>0$. Using the scale invariant property, we further narrow down the characteristics of a balancing search direction.
    
    % qualify the $\omega$ function by enforcing scale invariance in the anchor direction.

    % Using Lemma \ref{th:anchor} and \ref{th:fa>0>rinva} we narrow down the characteristics of a balancing search direction.
    \begin{restatable}{theorem}{fdeqa}\label{th:Fd=a}
    If a balancing anchor direction $\mathbf{a}$ is scale invariant to $\mathbf{r}$ and all the objective functions are differentiable at $\mathbf{x}^{t}$, then moving against a direction $\mathbf{d}\in\mathbb{R}^{n}$ with $\mathrm{F}\mathbf{d} = s \mathbf{a}$, for some $s > 0$, yields a non-dominated solution $\mathbf{x}^{t+1}$ such that $\mathbf{f}(\mathbf{x}^{t+1})$ is closer to the $\mathbf{r}^{-1}$ ray than $\mathbf{f}(\mathbf{x}^{t})$. 
    \end{restatable}

    Note that for a small step size $\eta$, the difference between the consecutive objective vectors can be approximated as $\Delta \mathbf{f} = \mathbf{f}^{t+1} - \mathbf{f}^{t} \approx -\eta\mathrm{F}\mathbf{d}$ from the first order Taylor series expansion. So, the search direction in Theorem \ref{th:Fd=a} moves the objective vector \textit{against} the balancing anchor direction $\mathbf{a}$ (i.e., along $-\mathbf{a}$) in the objective space.
    In the following, we propose three different functions -- using Cauchy–Schwarz inequality, Lagrange's identity and KL divergence -- for gauging the proportionality between two vectors and analyze them based on their respective balancing anchor directions.

    %\citet{Mahapatra2020} defined a function $\omega: \mathrm{R}^{m}_{+} \times \mathrm{R}^{m}_{+} \rightarrow \mathrm{R}_{+}$, based on KL divergence \eqref{eq:non-uniformity}, to measure the proportionality between $l(x)$ and $r^{-1}$ ray. They also proposed a direction of movement along which one is guaranteed to improve the proportionality measured by $\omega$.
    %In this paper, we propose two more functions, inspired from Cauchy–Schwarz inequality and Lagrange's identity, to gauge the proportionality between $l$ and $r^{-1}$ and provide directions of movement to improve it. We analyze these three functions and compare the properties, pros and cons  of their respective directions.

    \begin{figure}[t]
	\centering
	\begin{subfigure}{0.25\textwidth}
		\includegraphics[width=\linewidth]{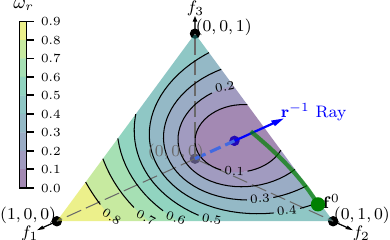}
		\caption{\label{fig:cs_simplex}% $\omega$ from 
  \hspace{-0.5mm}CS inequality \eqref{eq:cs_prop}}
	\end{subfigure}~
	\begin{subfigure}{0.25\textwidth}
		\includegraphics[width=\linewidth]{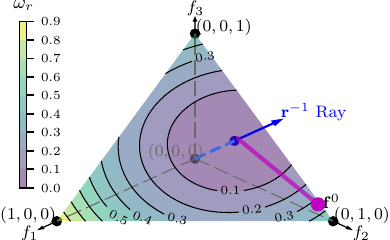}
		\caption{\label{fig:lagrange_simplex}%$\omega$ from 
  \hspace{-0.5mm}Lagrange identity \eqref{eq:lgrn_prop}}
	\end{subfigure}~
    \begin{subfigure}{0.25\textwidth}
		\includegraphics[width=\linewidth]{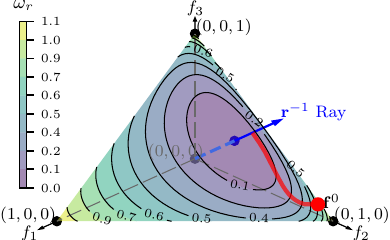}
		\caption{\label{fig:kld_simplex}%$\omega$ from 
  \hspace{-0.5mm}KL divergence \eqref{eq:non-uniformity}}
	\end{subfigure}~
    \begin{subfigure}{0.08\textwidth}
		% \centering
    \includegraphics[width=\linewidth]{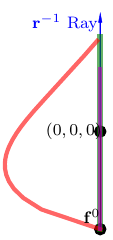}
		\caption{\label{fig:topview}Top}
	\end{subfigure}~
	\begin{subfigure}{0.135\textwidth}
		% \centering
		\includegraphics[width=\linewidth]{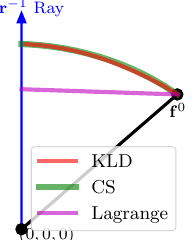}
		\caption{\label{fig:sideview}\hspace{-0.5mm}Side view}
	\end{subfigure}
%	\vspace{-.5cm}
	\caption{ (Color Online) Contour plots in \ref{fig:kld_simplex}, \ref{fig:cs_simplex} and \ref{fig:lagrange_simplex} show the variation of $\omega_\mathbf{r}$ in proportionality measuring functions \eqref{eq:non-uniformity}, \eqref{eq:cs_prop} and \eqref{eq:lgrn_prop} respectively on the $2-$d simplex $\mathcal{S}^3$. These sub-figures use the same weight vector, ${\mathbf{r}= [0.6, 0.2, 0.2]}$, and the same view, perspective projection on $\mathcal{S}^3$ from $(1,1,1)$ direction towards origin. The trajectories from $\mathbf{f}^0$ to $\mathbf{r}^{-1}$ ray are obtained by solving $\frac{d\mathbf{f}(t)}{dt} = -\mathbf{a}$ with same initial condition $\mathbf{f}(0) = \mathbf{f}^0 = [0.01, 0.9, 0.09]$ and time interval $t \in [0, 1.5]$ (integrated numerically in $150$ steps), where $\mathbf{a}$ is the corresponding anchor direction in \eqref{eq:kl_anchor}, \eqref{eq:cs_anchor} and \eqref{eq:lgrn_anchor} respectively. Figures \ref{fig:sideview} and \ref{fig:topview} show the orthogonal projections of these trajectories on the planes spanned by $\left\{\mathbf{r}^{-1},\ \mathbf{f}^0\right\}$ and $\left\{\mathbf{r}^{-1}-\mathbf{f}^0, \ (\mathbf{r}^{-1}-\mathbf{f}^0) \times (\mathbf{r}^{-1}+\mathbf{f}^0)\right\}$ respectively, where $\times$ is the vector cross product operator.
 	}
	\label{fig:prop}
    \end{figure}

    % \printinunitsof{in}\prntlen{\textwidth}
    \subsubsection{Proportionality Gauge from Cauchy–Schwarz (CSZ) Inequality:\quad}\label{sec:cs_prop}
	% \label{sec:mu_sci}
	The CSZ inequality of our non-zero vectors $\mathbf{f}, \mathbf{r}^{-1} \in \mathrm{R}_{+}^{m}$, $\langle \mathbf{f}, \,\mathbf{r}^{-1} \rangle^{2} \leq \|\mathbf{f}\|^{2} \, \|\mathbf{r}^{-1}\|^{2}$,
	% \begin{align} \label{eq:csineq}
	% 	\langle \mathbf{f}, \,\mathbf{r}^{-1} \rangle^{2} = \left(\sum_{j=1}^{m}  f_{j} r^{-1}_{j}\right)^{2} \leq \left( \sum_{j=1}^{m} f_{j}^{2} \right) \left( \sum_{j=1}^{m} r^{-2}_{j} \right) = \|\mathbf{f}\|^{2} \, \|\mathbf{r}^{-1}\|^{2},
	% \end{align} 
	% is tight
    is tight (equal)
    % , i.e. equality is attained, 
    when both the vectors are proportional to each other. Rearranging the terms to one side, we get the following
	\begin{align}
	\omega(\mathbf{f},\, \mathbf{r}^{-1}) = \frac{1}{2} \left(1 - \frac{\langle \mathbf{f}, \,\mathbf{r}^{-1} \rangle^{2}}{\|\mathbf{f}\|^{2} \, \|\mathbf{r}^{-1}\|^{2}}\right), \label{eq:cs_prop}
	\end{align}
	which satisfies properties \ref{num:prop+} and \ref{num:prop_smth_mntnc} of a proportionality gauge. Figure \ref{fig:cs_simplex} shows the corresponding $\omega_\mathbf{r}$ in case of $3$ objectives and a particular weight vector. 
    % A scaled (simplified) form of its anchoring direction, up to a constant factor of $\nabla_{\!\mathbf{f}}\omega_\mathbf{r}$, is given by
    We formulate its anchoring direction as
	%  \begin{align}
	% \mathbf{a} = \frac{\langle \mathbf{f}, \,\mathbf{r}^{-1} \rangle}{\|\mathbf{f}\| \, \|\mathbf{r}^{-1}\|} \mathbf{f} - \frac{\|\mathbf{f}\|}{\|\mathbf{r}^{-1}\|} \mathbf{r}^{-1}. \label{eq:cs_anchor}
	% \end{align}
    \begin{align}
	\mathbf{a}(\mathbf{f}) = \langle \overrightarrow{\mathbf{f}}, \,\overrightarrow{\mathbf{r}^{-1}} \rangle^2 \overrightarrow{\mathbf{f}} - \langle \overrightarrow{\mathbf{f}}, \,\overrightarrow{\mathbf{r}^{-1}} \rangle \overrightarrow{\mathbf{r}^{-1}}, \label{eq:cs_anchor}
	\end{align}
    where $\overrightarrow{\mathbf{v}}$ is the $\ell_2$ normalization of a vector $\mathbf{f}$. This $\mathbf{a}$ is scale invariant to $\mathbf{r}^{-1}$ and has the same direction as the gradient $\mathbf{a} = \|\mathbf{f}\|\nabla_\mathbf{f}\omega_\mathbf{r}$. Its main benefit is drawn from the following.  
	% Much like the previous anchoring direction we can state the relation between $\mathbf{a}$ and $\mathbf{f}$:
	\begin{restatable}{claim}{csaf}\label{th:fa_cs=0}
	The anchor direction $\mathbf{a}$ in \eqref{eq:cs_anchor} is always orthogonal to the objective vector $\mathbf{f}$: $\mathbf{a}^{T} \, \mathbf{f}= 0$.
	\end{restatable}
    The change in consecutive objective vectors $\Delta \mathbf{f} = \mathbf{f}^{t+1} - \mathbf{f}^{t}$  is approximately aligned to $-\mathbf{a}^{t}$. Therefore, Claim \ref{th:fa_cs=0} suggests that $\Delta\mathbf{f}^{T}\, \mathbf{f}^{t} \approx 0$. In other words, since $\mathbf{f}$ is an all positive vector, changes in some objectives $\Delta f_{j}$ are positive and others are negative. The advantage of an orthogonal anchoring direction lies in its ability to simultaneously ascend and descend which helps in escaping a PO solution that is not an EPO solution (formalized in Theorem \ref{th:d_nd_unconst}).
    
	% Therefore this anchoring direction can also escape local non-EPO %local Pareto optimal
	% solutions. 
    Since $\mathbf{a}$ is a linear combination of $\mathbf{f}$ and $\mathbf{r}^{-1}$, the trajectory of the objective vectors lies in the span of $\left\{\mathbf{r}^{-1}, \mathbf{f}^{0} \right\}$. However, it does not result in the shortest trajectory to reach $\mathbf{r}^{-1}$ ray (see Figure \ref{fig:sideview}).
	The shortest path between a point $\mathbf{f}^{0}$ and the $\mathbf{r}^{-1}$ ray is the line segment from $\mathbf{f}^{0}$ orthogonal to the $\mathbf{r}^{-1}$ ray, wherein every $\Delta\mathbf{f}$, and hence
    % change $\delta \mathbf{f}$, and hence every 
    $\mathbf{a}(\mathbf{f})$, 
    % ($\mathbf{a}\approx \Delta\mathbf{f}$), 
    should be orthogonal to the $\mathbf{r}^{-1}$ ray at all $\mathbf{f}\in \mathbb{R}^m$. 
    
 %    \subsubsection{Proportionality Gauge from Cauchy-Schwarz Inequality:} In \S \ref{sec:cs_prop}, we present this proportionality gauge,
 %    % based on the \textit{Cauchy-Schwarz Inequality}, 
 %    which is based on the cosine distance between the two vectors.
 %    % measures the square of sine of angle between the two vectors; 
 %    Figure \ref{fig:cs_simplex} shows the corresponding $\omega_\mathbf{r}$ in case of 3 objectives. Although its anchor direction \eqref{eq:cs_anchor} lies in the span of $\left\{\mathbf{r}^{-1}, \mathbf{f}^{0}\right\}$, it still follows a curved path, as shown in Figure \ref{fig:sideview}.
 % %    \deb{Remove the next line after regenerating the Figures.} Therefore it does not reach as close to the $\mathbf{r}^{-1}$ ray as the other alternatives (discussed in the following sections) 
	% % %when obtained 
 % %    using in the same number of iterations (steps in numerical integration).
    
 	\subsubsection{Proportionality Gauge from Lagrange's Identity:}
 	\label{sec:mu_li}
	% Instead of using the ratio between the right hand side (RHS) and left hand side (LHS) in \eqref{eq:csineq}, we can also take the difference between them to quantify the proportionality between $\mathbf{f}$ and $\mathbf{r}^{-1}$.
    % The residue (RHS - LHS) is given by the Lagrange's identity (scaled by $1/\|\mathbf{r}^{-1}\|^2$):
    The difference between 
    both sides of CSZ inequality,
    % squared magnitudes and squared inner product, 
    known as Lagrange's Identity, can be a proportionality gauge:
	\begin{align}
		\omega(\mathbf{f},\, \mathbf{r}^{-1}) = \frac{1}{2\|\mathbf{r}^{-1}\|^2}\left(\|\mathbf{f}\|^{2} \, \|\mathbf{r}^{-1}\|^{2} \,- \,\langle \mathbf{f}, \,\mathbf{r}^{-1}  \rangle^{2} \right)\ =\ \frac{1}{\|\mathbf{r}^{-1}\|^2}\frac{1}{4}\sum_{j=1}^{m}\sum_{k=1}^{m} \left( f_{j}r^{-1}_{k} - f_{k}r^{-1}_{j} \right)^{2}. \label{eq:lgrn_prop}
	\end{align}
    % Without the scaling factor $1/\|\mathbf{r}\|^2$, \eqref{eq:lgrn_prop} is called Lagrange's identity. 
    It satisfies both conditions \ref{num:prop+} and \ref{num:prop_smth_mntnc} of a proportionality measuring function. 
    % The 
    % % simplified form of its 
    % anchoring direction
    % % , up to a constant factor of $\nabla_{\!\mathbf{f}}\omega_\mathbf{r}$, 
    % is given by
    The factor $1/\|\mathbf{r}^{-1}\|^2$ in \eqref{eq:lgrn_prop} makes its anchoring direction scale invariant to $\mathbf{r}$ and we equate to the gradient $\nabla_\mathbf{f}\omega_\mathbf{r}$:
	\begin{align}
	\mathbf{a}(\mathbf{f}) = \mathbf{f} - \frac{\langle \mathbf{f}, \mathbf{r}^{-1}\rangle}{\|\mathbf{r}^{-1}\|^{2}} \, \mathbf{r}^{-1}. \label{eq:lgrn_anchor}
	\end{align}
    % The factor $1/\|\mathbf{r}^{-1}\|^2$ in \eqref{eq:lgrn_prop} makes the anchor direction scale invariant to $\mathbf{r}^{-1}$.
	This anchor direction can yield the trajectory of shortest path, as shown in figure \ref{fig:sideview}.
    % due to the following relation:
    \begin{restatable}{claim}{lgnaf}\label{th:lgrnar=0}
	The anchor direction $\mathbf{a}$ in \eqref{eq:lgrn_anchor} is always orthogonal to the $\mathbf{r}^{-1}$ ray: $\mathbf{a}^{T} \mathbf{r}^{-1} = 0$.
    \end{restatable}
	Note, when $\mathbf{f}$ and $\mathbf{r}^{-1}$ are not proportional, $\mathbf{a}$ and $\mathbf{f}$ are not orthogonal as $\mathbf{a}^{T}\mathbf{f} > 0$. 
    As a result, 
    % ascent in some of the objectives can not be guaranteed. Therefore, 
    this anchor direction may not escape a PO solution that is not EPO, because orthogonality of $\mathbf{a}$ and $\mathbf{f}$ is a necessary condition for non-convergence at a non-EPO solution in Theorem \ref{th:d_nd_unconst}. In its proof, we discuss a corner case where the $\mathbf{a}$ in \eqref{eq:lgrn_anchor}  cannot escape a non-EPO solution.
    % \deb{Explain why.}
	%local Pareto optimal.

    % {\color{blue} Table ?? shows the comparison between the three types of balancing anchor directions proposed above.}

    \subsubsection{Proportionality Gauge from KL Divergence:} In \S \ref{sec:mu_kld}, we present this proportionality gauge,
    % based on the \textit{Cauchy-Schwarz Inequality}, 
    which measures the KL divergence between $\frac{\mathbf{f}\odot \mathbf{r}}{\|\mathbf{f}\odot \mathbf{r}\|_1}$ and the uniform vector $\mathbf{1}/m$.
    % measures the square of sine of angle between the two vectors; 
    Figure \ref{fig:kld_simplex} shows the corresponding $\omega_\mathbf{r}$ in case of 3 objectives. 
    However, we do not use it since its anchoring direction neither produces the shortest path nor belongs to $\mathrm{span}(\{\mathbf{r}^{-1}, \mathbf{f}^0\})$, as shown in Figures \ref{fig:kld_simplex} and \ref{fig:topview}. 
    % Although its anchor direction \eqref{eq:cs_anchor} lies in the span of $\left\{\mathbf{r}^{-1}, \mathbf{f}^{0}\right\}$, it still follows a curved path, as shown in Figure \ref{fig:sideview}.
 %    \deb{Remove the next line after regenerating the Figures.} Therefore it does not reach as close to the $\mathbf{r}^{-1}$ ray as the other alternatives (discussed in the following sections) 
	% %when obtained 
 %    using in the same number of iterations (steps in numerical integration).

    We compare the proportionality gauges and their anchor directions in \S \ref{sec:compare_prop}.

\subsection{Descending Direction} \label{sec:descending_dir}
	%for Pure Descent}
% \red{TODO: why}
% Among the three anchor directions discussed above, 
% %we observe that the 
% $\mathbf{a}$ from Cauchy-Schwarz inequality, given by equation \eqref{eq:cs_anchor}, should be used when the initialisation $\mathbf{x}^{0}$ is itself a local Pareto optimal solution and $\mathbf{f}^{0}$ lies on the Pareto front. 
	When the iterate $\mathbf{f}^{t}$ is on or close to the $\mathbf{r}^{-1}$ ray, i.e. $\omega(\mathbf{f}^{t}, \mathbf{r}^{-1}) < \epsilon$ for small $\epsilon>0$, to reach the EPO solution, we require descent for every objective. 
    % The anchor directions proposed earlier do not guarantee descent in every objective.
    % A simple and effective strategy is to consider
    To guarantee descent for all, we choose
	\begin{align}
	\mathbf{a}(\mathbf{f}) =  \mathbf{f}. \label{eq:des_anchor} % \frac{\|\mathbf{f}\|}{\|\mathbf{r}^{-1}\|} \mathbf{r}^{-1}.
	\end{align}
	Because if the search direction satisfies $\mathrm{F}\mathbf{d}=\mathbf{f}$, then $\mathbf{d}^{T}\,\nabla_{\!\mathbf{x}}f_{j} = f_{j} \geq 0$ for all $j\in[m]$ as $\mathbf{f} \in \mathbb{R}_{+}^{m}$, hence $\mathbf{d}$ would be a descent direction. When $\mathbf{a}=\mathbf{f}$, we call it a \textit{Descending Anchor} direction. It can be considered as the gradient field of $\frac{1}{2} \|\mathbf{f}\|^2$.
%	\begin{claim}\label{th:Fd=r^-1}
%	If all the objective functions are differentiable at $\mathbf{x}^{t}$, then moving against a direction $\mathbf{d}\in\mathbb{R}^{n}$ with $\mathrm{F}\mathbf{d} = \lambda \mathbf{a}$, for some $\lambda > 0$, yields a dominating solution $\mathbf{x}^{t+1}\,: \ \mathbf{f}^{t+1} \preccurlyeq  \mathbf{f}^{t} $.
%	\end{claim}
%	So we move the objective vectors in two modes of operation:

    \subsection{Quadratic Program for Modelling the Search Direction}\label{sec:QP}
    % \deb{this section needs to be merged within \ref{sec:epo_rand} and \ref{sec:epo_trace}, and the $\beta$ should be constrained as $\|\beta\|_1 \leq 1$, instead of the current way of $\|\beta\|_\infty \leq 1$.}
    We now develop iterative methods to find the EPO solution w.r.t a weight vector $\mathbf{r}$. In each iteration, we solve a \textit{Quadratic Programming} (QP) problem to obtain a search direction $\mathbf{d} \in \mathcal{T}_\mathbb{X}(\mathbf{x}^t)$, the tangent plane (or cone, if $\mathbb{X}$ is constrained, see \S \ref{sec:constrained_moo}) at $\mathbf{x}^t\in \mathbb{X}$, such that it corresponds to an anchor direction $\mathbf{a}\in \mathcal{T}_\mathcal{O}(\mathbf{f}^t)$, the tangent plane at $\mathbf{f(x}^t)\in \mathcal{O} \subset \mathbb{R}^m$. 
    % We introduce the basic QP problem in this section and then extend it to develop two algorithms. 
    % \red{Algorithm \ref{alg:epo_search_x0_random}, detailed in \S \ref{sec:epo_rand}, finds the EPO solution for any random initialization $\mathbf{x}^0\in \mathbb{X}$ that is not a Pareto optimal solution.
    % Algorithm \ref{alg:epo_search_x0_po}, described in \S \ref{sec:epo_trace}, finds the EPO solution when the initialization is a Pareto optimal solution $\mathbf{x}^0\in \mathcal{P}$ but not exact w.r.t $\mathbf{r}$. Algorithm \ref{alg:epo_search_x0_po} traces the Pareto front from $\mathbf{x}^0$ to $\mathbf{x}^*_\mathbf{r}$.}

    We model the search direction as a linear combination of the objective gradients, i.e., ${\mathbf{d} = \sum_{j=1}^{m} \beta_{j} \nabla_{\!\mathbf{x}}f_{j} = \mathrm{F}^{T}\bm{\beta}}$
    % , where  $\mathrm{F}$ is the Jacobian of objective vector, 
    and compute the optimal coefficients $\bm{\beta}^* \in \mathbb{R}^m$ by solving 
    \begin{align} \label{eq:epo_qp}
        \bm{\beta}^{*} = \argmin_{\|\bm{\beta}\|_1 \leq 1} \ \ & \|\mathrm{F}\mathrm{F}^{T}\bm{\beta} - \mathbf{a}\|^{2},
    \end{align}
    so that $\mathrm{F}\mathbf{d}$ is aligned to the anchor direction as much as possible.
    Note that, unlike \cite{Desideri2012}, we do not restrict $\mathbf{d}$ to the convex hull of the positive gradients $\mathcal{CH}_\mathbf{x}$ \eqref{eq:ch_grads} to model 
    % , which 
    % results in 
    % was used to find 
    only descent directions.
    With coefficients $\bm{\beta}$ in the $\ell_1$ ball,  we facilitate gradient ascent for some of the objectives whenever necessary by allowing $\mathbf{d}$ to be in the convex hull of both positive and negative gradients: 
    \begin{align}\label{eq:ch_pmgrads}
        \mathcal{CH}_\mathbf{x}^{\pm} :=\left\lbrace \sum_{j=1}^{m} \nabla_{\!\!\mathbf{x}} f_{j} \, \beta_{j}^+ + \sum_{j=1}^m -\nabla_{\!\!\mathbf{x}} f_{j} \, \beta_{j}^- \;\middle|\; [\bm{\beta}^+, \bm{\beta}^-]^T \in \mathcal{S}^{2m} \right\rbrace.
    \end{align}
    % As a result, we facilitate gradient ascent for some of the objectives whenever necessary.
    
% 	\begin{align}
% 		\mathbf{d} = \sum_{j=1}^{m} \beta_{j} \nabla_{\!\mathbf{x}}f_{j} = \mathrm{F}^{T}\bm{\beta}, \quad \bm{\beta} \in [-1, 1]^{m} \label{eq:d=Fb}
% 	\end{align}
% 	where 
	% and $\mathrm{F}$ is the Jacobian of objective vector. Note that, unlike \cite{Desideri2012}, we do not restrict $\mathbf{d}$ to the convex hull of the gradients, which was used to find a descent--only direction. As a result, we facilitate gradient ascent for some of the objectives whenever necessary. We obtain the search direction by solving a QP problem %\eqref{eq:qp_cost} 
	% to align $\mathrm{F}\mathbf{d}$ to the anchor direction as much as possible; we minimize the following quadratic cost 	for the coefficients $\bm{\beta} \in [-1, 1]^{m}$.
	% \begin{align}
	% 	\|\mathrm{F}\mathbf{d} - \mathbf{a}\|_2^{2} = \|\mathrm{F}\mathrm{F}^{T}\bm{\beta} - \mathbf{a}\|_2^{2} \label{eq:qp_cost}
	% \end{align}

	% \subsubsection{\red{Modes of Operation.}}\label{sec:modes}
	Depending on the choice of anchor direction in \eqref{eq:epo_qp}, there could be two \textbf{modes of operation}:
	\begin{enumerate}
	\item \textbf{Balance mode}, where a balancing anchor directions, based on either CSZ inequality \eqref{eq:cs_anchor} or Lagrange's identity \eqref{eq:lgrn_anchor}, is used to improve the proportionality between $\mathbf{f}$ and $\mathbf{r}^{-1}$.
	\item \textbf{Descent mode}, where a descending anchor direction \eqref{eq:des_anchor} is used to decrease $f_j\ \forall j\in[m]$. 
        % $f_j$ for all $j\in[m]$.
	\end{enumerate}

    % We describe the main ingredients of the QP problem in this section, %\S \ref{sec:QP}, 
    % and then use it to develop two algorithms. Algorithm \ref{alg:epo_search_x0_random}, detailed in \S \ref{sec:epo_rand}, finds the EPO solution for any random initialization $\mathbf{x}^0\in \mathbb{X}$ that is not a Pareto optimal solution.
    % Algorithm \ref{alg:epo_search_x0_po}, described in \S \ref{sec:epo_trace}, finds the EPO solution when the initialization is a Pareto optimal solution $\mathbf{x}^0\in \mathcal{P}$ but not exact w.r.t $\mathbf{r}$. Algorithm \ref{alg:epo_search_x0_po} traces the Pareto front from $\mathbf{x}^0$ to $\mathbf{x}^*_\mathbf{r}$.
    %, a capability that we use to approximate the Pareto front, described in \S \ref{sec:epo_trace}.
 
	\subsection{EPO Search from Random Initialization}
	\label{sec:epo_rand}
	When the goal is to find the EPO solution for a given $\mathbf{r}$ starting from a random initialization $\mathbf{x}^{0} \in \mathbb{X}$, we use the balance mode with the anchor direction from Lagrange's identity \eqref{eq:lgrn_anchor} for every iteration until $\mathbf{f}^{t}$ is (nearly) proportional to $\mathbf{r}^{-1}$, $\omega(\mathbf{f}^{t}, \mathbf{r}^{-1}) < \epsilon_1$ for a small $\epsilon_1 > 0$. After that, we use the descent mode until convergence, $\|d\| \leq \epsilon_2$ for a small $\epsilon_2 > 0$.
    % i.e., when the magnitude of search direction vanishes.
    We extend the QP in \eqref{eq:epo_qp} (see \S \ref{sec:cepo_rand} for its extension to constrained MOO problem) to solve
\begin{subequations}\label{eq:qp_x0_random_unconst}
    \begin{align}
        \bm{\beta}^{*} = \argmin_{\|\bm{\beta}\|_1 \leq 1} \ \ &\|\mathrm{F}\mathrm{F}^{T}\bm{\beta} - \mathbf{a}\|^{2} \label{eq:qp_x0_random_unconst_cost}\\
        \text{s.t.} \quad & \bm{\beta}^T\mathrm{F} \, \nabla f_{j} \geq 0 \quad \forall\ j \in \mathrm{J} = \begin{cases}
            \ \mathrm{J}^*\ \quad \text{in balance mode} \\
            [m] \quad \text{in descent mode}
        \end{cases}, \label{eq:J_index} \\
        \text{where}\quad & \mathrm{J}^* = \left\{j \in [m]\ \middle|\ j = \arg\max_{j'\in[m]} \, f_{j'} r_{j'}\right\}  \label{eq:qp_J*_unconst}
    \end{align}
    \end{subequations}
    % \begin{align}
    %     \text{where}\quad & \mathrm{J}^* = \left\{j \in [m]\ \middle|\ j = \arg\max_{j'\in[m]} \, f_{j'} r_{j'}\right\}  \label{eq:qp_J*_unconst}
    % \end{align}
    is the index set of maximum relative objective values. 
    % We extend the above QP \eqref{eq:qp_x0_random_unconst} in \S \ref{sec:cepo_rand} to find the EPO solution of an MOO problem with constrained domain $\mathbb{X}$. 
    % The constraint \eqref{eq:qp_J*_unconst} ensures that maximum relative value decreases. 
    % Note, if $\mathbf{a} \in \mathcal{CH}_\mathbf{x}^{\pm}$, then 
    % In each iteration and for any 
    We call the resulting $\mathbf{d}_{nd} = \mathrm{F}^T\bm{\beta}^*$ a \textit{Non-Dominating Search Direction}, because it can yield a solution $\mathbf{x}^{t+1}$ that is not dominated by $\mathbf{x}^t$, i.e. $\mathbf{f}^{t+1} \nsucc \mathbf{f}^t$. We state this formally for the balance mode in Lemma \ref{th:bal}  and descent mode in Lemma \ref{th:des} with a regularity assumption. In the terminology of differentiable maps, $\mathbf{x}^t$ is a \textit{Regular Point} of the vector valued function $\mathbf{f}$, if its Jacobian $\mathrm{F}(\mathbf{x}^t)$ is full rank. 
    \begin{restatable}{lemma}{dndbal}\label{th:bal}
        If $\mathbf{x}^t$ is a regular point of the differentiable vector function $\mathbf{f}$ in a balance mode, i.e. $\omega(\mathbf{f}^t, \mathbf{r}^{-1}) > \epsilon_1$, then the non-dominating direction obtained from QP \eqref{eq:qp_x0_random_unconst} makes
        \begin{enumerate}
            \item \hspace{-1mm}non-negative \hspace{-0.3mm}angles \hspace{-0.3mm}with \hspace{-0.3mm}the \hspace{-0.3mm}gradients \hspace{-0.3mm}of \hspace{-0.3mm}maximum \hspace{-0.3mm}relative \hspace{-0.3mm}objectives: \hspace{-0.3mm}${\mathbf{d}_{nd}^T\nabla_{\!\mathbf{x}^t}\!f_j\!\geq\!0\ \forall\!j\!\in\!\mathrm{J}^*}$ \hspace{-1mm} \eqref{eq:qp_J*_unconst},
            \item \hspace{-1mm}a positive angle with the balancing anchor direction \eqref{eq:lgrn_anchor} in the objective space: $\mathbf{a}^T\mathrm{F}\mathbf{d}_{nd} > 0$.
        \end{enumerate}
	\end{restatable}
    \begin{restatable}{lemma}{dnddes}\label{th:des}
        If $\mathbf{x}^t$ is a regular point of the differentiable vector function $\mathbf{f}$ in a descent mode, i.e. $\omega(\mathbf{f}^t, \mathbf{r}^{-1}) \leq \epsilon_1$, 
	    then the non-dominating direction obtained from QP \eqref{eq:qp_x0_random_unconst} makes a non-negative angle with every gradient,  $\mathbf{d}_{nd}^T\, \nabla_{\!\mathbf{x}}f_j^t \geq 0\ \forall j\in [m]$, and a positive angle with at least one gradient.
	\end{restatable}

    A positive angle with the gradient means moving against $\mathbf{d}_{nd}$ will reduce the corresponding objective value. Lemmas \ref{th:bal} and \ref{th:des} are true even without the constraint \eqref{eq:J_index} (\eqref{eq:qp_J*} for constrained MOO) if $\mathbf{a} \in \left\{\mathrm{F}\mathbf{d} \,\middle|\, \mathbf{d} \in \mathcal{CH}_\mathbf{x}^{\pm} \right\}$, where $\mathcal{CH}_\mathbf{x}^{\pm}$ is defined in \eqref{eq:ch_pmgrads}. Also, the Lemmas are true at certain \textit{irregular} points $\mathbf{x}\in\mathbb{X}$ whose Jacobian matrices are not full rank (discussed in the proof). 
    
    % beyond the regularity assumption, these Lemmas are true at certain $\mathbf{x}\in\mathbb{X}$ with rank deficient Jacobian as well (discussed in proof). 
    
    We summarize our EPO Search procedure for random initialization in Algorithm~\ref{alg:epo_search_x0_random} .
	A practically useful variation to improve the descent mode is  discussed in Appendix \ref{sec:restricted_descent}.
	\begin{algorithm}[t]
        \caption{EPO Search for Random initialization}\label{alg:epo_search_x0_random}
		\begin{algorithmic}[1]
		\State \textbf{Input:} $\mathbf{x}^0\in\mathbb{X}$, $\mathbf{r}\in\mathbb{R}^m$, $\eta$, $\epsilon_1, \epsilon_2$ \Comment{$\mathbf{x}^0\notin \mathcal{P}$}
		\While{maximum iterations not reached}
% 		\State Compute the Jacobians $\mathrm{F, H, G}$ at $\mathbf{x}^t$ \Comment{$\mathrm{H, G}$ are empty if no active constraints}
		\If{$\omega_\mathbf{r}(\mathbf{f(x}^t)) \leq \epsilon_1$} %\Comment{Lagrange $\omega$ \eqref{eq:lgrn_prop}}
		\quad $\mathbf{a}$ = Lagrange anchor from \eqref{eq:lgrn_anchor} \Comment{Balance mode}
		\Else
		\quad $\mathbf{a}$ = Descending anchor from \eqref{eq:des_anchor} \Comment{Descent mode}
		\EndIf
		\State $\mathbf{d}_{nd}=\mathrm{F}^{T}\bm{\beta}^{*}$, \ $\bm{\beta}^{*}$ obtained by solving the QP \eqref{eq:qp_x0_random_unconst} \Comment{\eqref{eq:qp_x0_random} for constrained MOO}
		% \State $\mathbf{x}^{t+1} = \pi(\mathbf{x}^{t} - \eta \mathbf{d}_{nd})$ \Comment{Project to preserve boundary constraints \eqref{eq:bd_project}}
		\If{$\|\mathbf{d}_{nd}\| \leq \epsilon_2$} {\bf break} \Comment{Check for convergence} \EndIf
        \State $\mathbf{x}^{t+1}=\mathbf{x}^t - \eta \mathbf{d}_{nd}$
		\EndWhile
		\State {\bf Output:} $(\mathbf{x}^{t},\ \mathbf{f}^{t})$
		\end{algorithmic}
	\end{algorithm}
    \subsubsection{Convergence.}\label{sec:epo_rand_conv}
% 	Assuming the problem satisfies regularity condition, we prove the convergence of EPO search for both the goals. In the first goal, where $\mathbf{x}^0$ is randomly initialized, we show that every iteration is associated with a set that contains the EPO solution, and that this sequence of sets converges to $\mathcal{P}_\mathbf{r}$, the set of exact Pareto optimal solutions \eqref{eq:epo}.
    
	We prove the convergence of
	Algorithm \ref{alg:epo_search_x0_random}
	in two steps. First we define an admissible set $\mathcal{A}^\mathbf{r}_{\mathbf{f}^t} \subset \mathbb{R}^m$ that contains potential objective vectors $\mathbf{f}^{t+1} = \mathbf{f}(\mathbf{x}^{t+1})$ to which the EPO Search in Algorithm \ref{alg:epo_search_x0_random} can reach. Then we prove that the sequence of sets $\left\{\mathcal{A}^\mathbf{r}_{\mathbf{f}^t}\right\}_t$ converges to $\mathcal{P}_r$, the set containing the EPO solutions. The %theoretical 
    results hold true for both constrained and unconstrained MOO problems, as detailed in the proofs.
	
	To characterize the properties of $\mathbf{x}^{t+1}$  obtained by moving against $\mathbf{d}_{nd}$, we define some sets in $\mathbb{R}^m$ that are illustrated in Figure~\ref{fig:admissible_set}. The set of all attainable objective vectors that dominate the $\mathbf{f}^t$ is denoted as $\mathcal{V}_{\preccurlyeq \mathbf{f}^t}$ \eqref{eq:V<ft}.
	% 	\begin{align}
% 		\mathcal{V}_{\preccurlyeq \mathbf{f}^t} = \left\{ \mathbf{f} \in\mathcal{O} \;\middle|\; \mathbf{f} \preccurlyeq \mathbf{f}^t \right\}.
% 	\end{align}
	The set of all attainable objective vectors that have better proportionality than $\mathbf{f}^t$ is denoted as $\mathcal{M}^\mathbf{r}_{\mathbf{f}^t}$ \eqref{eq:Mrft}.
% 	\begin{align}
% 		\mathcal{M}^\mathbf{r}_{\mathbf{f}^t} = \{\left. \mathbf{f} \in\mathcal{O} \ \right|\ \omega_r(\mathbf{f}) \le \omega_r(\mathbf{f}^t)\}. 
% 	\end{align} 
	During a descent mode $\mathbf{f}^{t+1} \in \mathcal{V}_{\preccurlyeq \mathbf{f}^t}$, and in a balance mode $\mathbf{f}^{t+1} \in \mathcal{M}^\mathbf{r}_{\mathbf{f}^t}$. For the $t^\text{th}$ iteration, we define a point $\widecheck{\mathbf{f}}^{t}\in\mathbb{R}^m_+$ as in \eqref{eq:fcheckt}, where $\lambda^t$ is the maximum relative objective value. 
% 	\begin{align}
% 		\widecheck{\mathbf{f}}^t = \lambda^t (1/r_1, \cdots, 1/r_m),
% 		\text{ where } \lambda^t = \max \; \{f_j^tr_j\;|\; j \in [m]\}.
% 	\end{align}
	% $\lambda^t$, and hence $\widecheck{\mathbf{f}}^t$, are bounded as each $r_j$ is positive. 
    Finally, using 
    $\widecheck{\mathbf{f}}^t$, 
    % $\mathcal{V}_{\preccurlyeq \widecheck{\mathbf{f}}^t}$, which is similarly defined as $\mathcal{V}_{\preccurlyeq \mathbf{f}^t}$,
    we define the admissible set\footnote{The admissible set of an iteration in the CS \eqref{eq:chebychev} is $\left\{\mathbf{f} \in\mathcal{O} \;\middle|\; f_{j^*} \leq f^t_{j^*}\right\}$, where $j^* =\argmax_{j\in [m]} f^t_jr_j$.} of an iteration in EPO Search Algorithm \ref{alg:epo_search_x0_random} as $\mathcal{A}^\mathbf{r}_{\mathbf{f}^t}$ \eqref{eq:Arft} 
    for any mode, balance or descent.
% 	\begin{align}
% 		\mathcal{A}^\mathbf{r}_{\mathbf{f}^t} = \left\{\mathbf{f} \in\mathcal{O} \;\middle|\; \mathbf{f} \preccurlyeq \widecheck{\mathbf{f}}^t \right\},
% 	\end{align}
	% which is also bounded. 
 
	\begin{minipage}{0.51\textwidth}
	\begin{figure}[H]
		\centering
		\def\svgwidth{0.8\linewidth}	% if the document class is double column, use \columnwidth
		\begingroup\makeatletter\def\f@size{9}\check@mathfonts	% f@size decides the fontsize of "math" text in Figure
		\def\maketag@@@#1{\hbox{\m@th\large\normalfont#1}}%
		% \fontsize{9pt}{11pt}\selectfont % decides the overall fontsize
		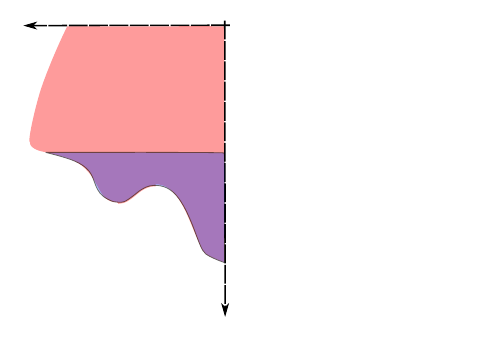
		\endgroup\vspace{-.2cm}
		\caption{(Color Online). Illustration of the sets associated with the admissible set $\mathcal{A}^\mathbf{r}_{\mathbf{f}^t}$ in the objective space $\mathbb{R}^2$, at iteration $t$. 
			This admissible set contains the objective vector $\mathbf{f}^{t+1}$ of next iteration. 
			\label{fig:admissible_set}
			\vspace{0.5cm}
		}
	\end{figure}
	\end{minipage}~
	\begin{minipage}{0.45\textwidth}
	\begin{align}
		\mathcal{V}_{\preccurlyeq \mathbf{f}^t} &= \left\{ \mathbf{f} \in\mathcal{O} \;\middle|\; \mathbf{f} \preccurlyeq \mathbf{f}^t \right\} \label{eq:V<ft}\\[0.7cm]
		\mathcal{M}^\mathbf{r}_{\mathbf{f}^t} &= \{\left. \mathbf{f} \in\mathcal{O} \ \right|\ \omega_r(\mathbf{f}) \le \omega_r(\mathbf{f}^t)\} \label{eq:Mrft}\\[0.5cm]
		\widecheck{\mathbf{f}}^t\ &= \lambda^t (1/r_1, \cdots, 1/r_m),\label{eq:fcheckt}\\
		\text{where}\ \lambda^t &= \max \; \{f_j^tr_j\;|\; j \in [m]\} \nonumber \\[0.4cm]
		\mathcal{A}^\mathbf{r}_{\mathbf{f}^t} &= 
        \left\{\mathbf{f} \in\mathcal{O} \;\middle|\; \mathbf{f} \preccurlyeq \widecheck{\mathbf{f}}^t \right\} \\
        &\supset \ \underbrace{\left(
        % \mathcal{V}_{\preccurlyeq \widecheck{\mathbf{f}}^t}
        \mathcal{A}^\mathbf{r}_{\mathbf{f}^t}
        \cap \mathcal{M}^\mathbf{r}_{\mathbf{f}^t} \right)}_\text{balance mode} \ \ \cup \underbrace{\mathcal{V}_{\preccurlyeq \mathbf{f}^t}}_\text{descent mode} \nonumber
        \label{eq:Arft}
	\end{align}
	\vspace{0.5cm}
	\end{minipage}
% 	\begin{figure}[h]
% 		\centering
% 		\def\svgwidth{0.5\linewidth}	% if the document class is double column, use \columnwidth
% 		\begingroup\makeatletter\def\f@size{9}\check@mathfonts	% f@size decides the fontsize of "math" text in Figure
% 		\def\maketag@@@#1{\hbox{\m@th\large\normalfont#1}}%
% 		% \fontsize{9pt}{11pt}\selectfont % decides the overall fontsize
% 		\input{diagrams/admissible_sets.pdf_tex}
% 		\endgroup\vspace{-.2cm}
% 		\caption{(Color Online) Illustration of 
% 			the sets associated with the admissible set $\mathcal{A}^\mathbf{r}_{\mathbf{f}^t}$ in the objective space $\mathbb{R}^2$, at iteration $t$. 
% 			This admissible set contains the objective vector $\mathbf{f}^{t+1}$ of next iteration. 
% 			\label{fig:admissible_set}
% 			% 		\vspace{-.2cm}
% 		}
% 	\end{figure}
 %    Note the relation between $\mathcal{A}^\mathbf{r}_{\mathbf{f}^t}$ and $\mathcal{V}_{\preccurlyeq l^t}$:
	% \begin{restatable}{lemma}{vlta}\label{th:V<A}
	% 	When $\mathbf{f}^t$ and the $\mathbf{r}^{-1}$ ray are not proportional, i.e. $\omega(\mathbf{f}^t, \mathbf{r}^{-1}) > 0$, the set of dominating objective vectors is a subset of the admissible set $\mathcal{V}_{\preccurlyeq \mathbf{f}^t} \subsetneq \mathcal{A}^\mathbf{r}_{\mathbf{f}^t}$. But when proportional, i.e. ${\omega(\mathbf{f}^t, \mathbf{r}^{-1}) = 0}$, both sets are equal $\mathcal{V}_{\preccurlyeq \mathbf{f}^t} = \mathcal{A}^\mathbf{r}_{\mathbf{f}^t}$.
	% \end{restatable}
	% Using Lemmas \ref{th:df_j*>0} and \ref{th:V<A}, we can show:
    Clearly, the admissible set contains all the points in $\mathcal{O}$ that dominate the $\mathbf{f}^{t}$, $\mathcal{V}_{\preccurlyeq \mathbf{f}^t}\subset \mathcal{A}^\mathbf{r}_{\mathbf{f}^t}$. Moreover, when $\omega_\mathbf{r}(\mathbf{f}^t)>0$, it also has points with better proportionality than $\mathbf{f}^t$, ${\mathcal{A}^\mathbf{r}_{\mathbf{f}^t}\cap M^\mathbf{r}_{\mathbf{f}^t} \neq \phi}$.
	% Therefore, the admissible set contains the required solution for the next iteration, satisfying both proportionality to $\mathbf{r}^{-1}$ and dominating properties.
    Therefore, using Lemmas \ref{th:bal} and \ref{th:des}, we can state the following. %show:
	\begin{restatable}{lemma}{adms}
		\label{th:admissible_set}
		If $\mathbf{x}^t$ is a regular point of $\mathbf{f}$, there exists a step size $\eta_0>0$, such that for every $\eta \in [0, \eta_0]$, the objective vector of 
        % the new solution point 
        $\mathbf{x}^{t+1} = \mathbf{x}^t-\eta \mathbf{d}_{nd}$ lies in the $t^\text{th}$ admissible set: $\mathbf{f}\left(\mathbf{x}^{t+1}\right) \in \mathcal{A}^\mathbf{r}_{\mathbf{f}^t}$.
		% \begin{align}	
  % \mathbf{f}\left(\mathbf{x}^{t+1}\right) &\in \mathcal{A}^\mathbf{r}_{\mathbf{f}^t}.
		% \end{align}
	\end{restatable}
	% A natural consequence of 
 %    % Lemma~\ref{th:df_j*>0} and 
 %    Lemma~\ref{th:admissible_set} is the monotonicity of $\mathcal{A}^r_{\mathbf{f}^{t+1}}$, i.e. $\mathcal{A}^r_{\mathbf{f}^{t+1}} \subset \mathcal{A}^r_{\mathbf{f}^t}$.
 % % \red{make this a Theorem}
 %   is regular everywhere in its domain
   We extend the regularity assumption of $\mathbf{f}$ to all points in $\mathbb{X}\backslash \mathcal{P}$ to state the convergence result.
   % assuming the objective to be regular at all points in its domain, we can state the convergence result.
	\begin{restatable}{theorem}{mntnct}\label{th:monotonicity}
		If $\mathbf{f}$ is a differentiable regular vector valued objective, then the sequence of 
        admissible sets $\{\mathcal{A}^r_{\mathbf{f}^t}\}$,
        % maximum relative objective values $\{\lambda^t\}$
        which correspond to the solutions $\{\mathbf{x}^t\}$ produced according to Lemma~\ref{th:admissible_set} starting from a non-Pareto Optimal point $\mathbf{x}^0 \notin \mathcal{P}$, 
        % monotonically decreases, i.e., ${\lambda^{t+1} \le \lambda^t}$, which means 
        converges by decreasing monotonically
        % decreases 
        $\mathcal{A}^r_{\mathbf{f}^{t+1}} \subset \mathcal{A}^r_{\mathbf{f}^t}$.
        % and converges to $\mathcal{P}_\mathbf{r}$.
        % and there sequence of bounded sets $\{\mathcal{A}^r_{\mathbf{f}^t}\}$ converges.
	\end{restatable}
    When $\mathcal{P}_\mathbf{r}$ is a singleton set then $\{\mathcal{A}^r_{\mathbf{f}^t}\}$ converges to $\mathcal{P}_\mathbf{r}$.  Note, if an EPO solution does not exist for the given preference $\mathcal{P}_\mathbf{r}=\phi$, i.e. $\mathbf{r}^{-1}$ ray does not intersect the PF $\mathcal{P}$, then EPO search finds the intersection point between $\mathbf{r}^{-1}$ ray and $\partial \mathcal{O}$, the boundary of the attainable objective vectors. If the $\mathbf{r}^{-1}$ ray does not intersect $\partial \mathcal{O}$, then EPO search finds the point in $\partial \mathcal{O}$ that is maximally proportional to the $\mathbf{r}^{-1}$ ray.

    \subsection{EPO Search for Tracing the Pareto Front from $\mathbf{x}^0 \in \mathcal{P}$}
    \label{sec:epo_trace}
    If the initialization $\mathbf{x}^0$ itself is a PO solution, then we can modify the EPO search of Algorithm \ref{alg:epo_search_x0_random} to trace the PF from $\mathbf{x}^0$ to $\mathbf{x}^*_\mathbf{r}$, an EPO solution w.r.t. weight vector $\mathbf{r}$, and obtain new PO solutions along the trajectory.
    In the modified algorithm we use CSZ inequality-based anchor direction \eqref{eq:cs_anchor} in the balance mode because it is orthogonal to the objective vector (see Claim \ref{th:fa_cs=0}), and guarantees to escape the local PO at $\mathbf{x}^t$ to a new point (justified in the proof of Theorem \ref{th:d_nd_unconst}).
    However, if we use only the balance mode until $\omega(\mathbf{f}^t, \mathbf{r}^{-1}) \geq \epsilon_1$ for a small $\epsilon_1>0$, the trajectory may drift away from the PF, especially for convex objective functions (see \S \ref{sec:trace_wo_des}). Therefore, to ensure that the objective vectors $\mathbf{f}^{t}$ in EPO search trajectory stay close to the PF, we alternate between the balance mode and descent mode in every iteration. To  ensure the proportionality gauge \eqref{eq:cs_prop} does not increase in the descent mode, we extend the QP \eqref{eq:epo_qp} (see \S \ref{sec:cepo_trace} for its extension to constrained MOO)
    \begin{subequations}\label{eq:qp_x0_po_unconst} as follows
    \begin{align}
        \bm{\beta}^{*} &= \argmin_{\|\bm{\beta}\|_1 \leq 1} \ \ \|\mathrm{F}\mathrm{F}^{T}\bm{\beta} - \mathbf{a}\|^{2} \\
        \text{s.t.} \quad & \mathbb{1}_{des}\bm{\beta}^T\mathrm{F}\mathrm{F}^T \mathbf{a}^{bal} \geq 0, \label{eq:des_pos} \\
        \text{and} \quad & \mathbb{1}_{des}\bm{\beta}^T\mathrm{F} \, \nabla f_{j} \geq 0 \quad \forall\ j \in [m],
    \end{align}
    \end{subequations}
    where $\mathbf{a}^{bal}$ is the balancing anchor direction \eqref{eq:cs_anchor}, and $\mathbb{1}_{des}$ is an indicator variable for descent mode, which applies the constraint \eqref{eq:des_pos} only in the descent mode.
    But the stopping criteria is checked only in the balance mode: the goal is to keep {\it balancing} the relative objective vector $\mathbf{r}\odot\mathbf{f}$ until it is proportional to uniformity $\mathbf{1}$. 
    Note, this balance mode ignores the constraints in \eqref{eq:J_index} for objectives of $\mathrm{J}^*$ \eqref{eq:qp_J*_unconst} because, if $\mathbf{f}^t$ is on (close to) the PF, then an $f_{j^*}^t$ for $j^*\in \mathrm{J}^*$ 
    % objective with maximum relative value 
    may need to increment to move towards the EPO solution. 
    %This EPO Search is summarized in 
    Algorithm~\ref{alg:epo_search_x0_po} summarizes the entire method. 

    \begin{algorithm}[t]
        \caption{EPO Search for Pareto Optimal Initialization}\label{alg:epo_search_x0_po}
		\begin{algorithmic}[1]
		\State {\bf Input:} $\mathbf{x}^0\in\mathbb{X}$, $\mathbf{r}\in\mathbb{R}^m$, $\eta$, $\epsilon_1$,$\epsilon_2$ \Comment{$\mathbf{x}^0\in \mathcal{P}$}
		\State $mode = 0$ \Comment{$0$ for balance mode, $1$ for descent mode}
		\While{$\omega_\mathbf{r}(\mathbf{f}(\mathbf{x}^t))> \epsilon_1$ \ \textbf{or} \ $t<$ maximum iterations} \Comment{CSZ inequality-based $\omega$ \eqref{eq:cs_prop}}
% 		\State Compute the Jacobians $\mathrm{F, H, G}$ at $\mathbf{x}^t$ \Comment{$\mathrm{H, G}$ are empty if no active constraints}
		\If{$mode=0$}
		\quad $\mathbf{a}$ = CSZ inequality-based anchor  \eqref{eq:cs_anchor} \Comment{Balance mode}
		\Else
		\quad \quad $\mathbf{a}$ = descent anchor from \eqref{eq:des_anchor} \Comment{Descent mode}
		\EndIf
		\State $\mathbf{d}_{nd}=\mathrm{F}^{T}\bm{\beta}^{*}$, \ $\bm{\beta}^{*}$ obtained by solving the QP \eqref{eq:qp_x0_po_unconst} \Comment{\eqref{eq:qp_x0_po} for constrained MOO}
		% \State $\mathbf{x}^{t+1} = \pi(\mathbf{x}^{t} - \eta \mathbf{d}_{nd})$ \Comment{Project to confirm boundary constraints \eqref{eq:bd_project}}
		\If{$mode = 0$ {\bf and} $\|\mathbf{d}_{nd}\| \leq \epsilon_2$} {\bf break} \Comment{Check for convergence only in balance mode} \EndIf
		\State $\mathbf{x}^{t+1}=\mathbf{x}^t - \eta \mathbf{d}_{nd}$, $mode = 1 - mode$ \Comment{Alternates mode while tracing the PF}
        % \State $\mathbf{x}^{t+1}=\mathbf{x}^t - \eta \mathbf{d}_{nd}$
		\EndWhile
		\State {\bf Output:} $\{ \mathbf{x}^{i}\}_{i=0}^{t}$ \Comment{Trace of EPO Search on $\mathcal{P}$}
		\end{algorithmic}
	\end{algorithm}
	
	%    {\color{red} 
	%    How does the performance of previous methods \cite{NeurIPS2018_Sener_Koltun,NIPS2019pmtl} scale with m and n?} {\color{blue} Pareto MTL solves an $m$d optimization with $m+K-1$ constraints in every iteration, where $K$ is the number of reference vectors; $K$ grows exponential with $m$.}
	%    
	% For every iteration, there can be at most $2m+1$ constraints in the LP \eqref{eq:lp}: $m+1$ for the simplex constraint $\beta \in \mathcal{S}^m$, and others for \eqref{eq:J_bar} and \eqref{eq:J_start}. In the balancing mode, maximum number of constraint is applicable when $J$ is empty, which means none of the $c_j$s make a positive angle with the adjustment vector $a$. In the pure descent mode, there are exactly $2m+1$ constraints; if we add the restriction in \eqref{eq:res_des} then there are $2m+2$ restrictions. So, apart from computing the objective $l(x^t)$ values and gradients $G$, the extra time that our method requires in every iteration is {\color{red} $O($ time complexity of an LP of m variables and 2m+2 constraints$)$}. Computation of the adjustements $a$ takes $O(m)$ time.
	
	% {\color{red} 
	% How does EPO's performance vary with increasing parameter space (high-dimensional gradients and number of tasks?}
	
	% {\color{blue}The dimension of parameter/solution space only affects the computation of $C=G^TG$ matrix. As $G$ is an $n\times m$ matrix, computation of $C$ is $O(m^2n)$, which is linear w.r.t the number of parameter $n$, and qudratic w.r.t. number of tasks $m$. Computation of all the constraints in \eqref{eq:lp} and \eqref{eq:res_des} are of $O(m)$.}
    
	\subsubsection{Convergence.}\label{sec:convergence_x0_po}
	We prove the contra-positive: convergence is not achieved until the iterate $\mathbf{x}^t$ is close to $\mathbf{x}_\mathbf{r}^*\in \mathcal{P}_\mathbf{r}$, and $\omega_\mathbf{r}(\mathbf{f}(\mathbf{x}^t))$ keeps decreasing.
    At an $\mathbf{x}^t$, the set of descent directions is given by $\mathcal{D}_\mathbb{X}^\mathbf{f}(\mathbf{x}^t) = \left\{\mathbf{d}\in\mathcal{T}_\mathbb{X}(\mathbf{x}^t) \,\middle|\, \mathbf{d^T}\nabla_{\!\mathbf{x}^t}f_j \geq 0, \, \forall \, j \in [m]\right\}.$
    A necessary condition to check if $\mathbf{x}^t$ is a PO solution, i.e.,  $\mathcal{D}_\mathbb{X}^\mathbf{f}(\mathbf{x}^t) = \{\mathbf{0}\}$, is given by Pareto Criticality: 
    % $\text{there exists a } \bm{\beta} \in \mathcal{S}^m \ \text{s.t. }\ \mathrm{F}^T\bm{\beta} = \mathbf{0}$. 
	\begin{align}\label{eq:criticality_unconst}
	    \text{there exists a } \bm{\beta} \in \mathcal{S}^m, \ \text{s.t. }\ \mathrm{F}^T\bm{\beta} = \mathbf{0}.
	\end{align}
    The Jacobian at an $\mathbf{x}^*\in \mathcal{P}$ is not full rank. However, $\mathbf{x}^*$ is called a \textit{Regular Pareto Optimal} solution if its Jacobian $\mathrm{F}(\mathbf{x}^*)$ has rank $m-1$ \citep{4358761}.   
	Previous gradient-based methods \citep{fliege2000steepest,Desideri2012} use the Pareto criticality condition as a stopping criterion, since they use descent directions in every iteration. Therefore, 
 %these MOO methods 
 they stop at any local PO solution. In contrast, our method is not designed to find $\bm{\beta}$ for a descent direction, hence does not stop prematurely at any local PO solution; it traces the Pareto front until an EPO solution is found.

    % We make a regularity assumption. 
    % An $\mathbf{x}^*$ is called as a \textit{Regular Pareto Optimal} solution if its Jacobian $\mathrm{F}$ is of rank $m-1$ (\citep{4358761}). 

 %    For constrained MOO problems, we introduce a problem dependent assumption to guarantee non-convergence at a non-EPO point when there are active constraints .
 %    We assume %we make an assumption: 
	% there exists an $\eta_0 > 0$ such that ${\mathbf{f}^* + \eta \overrightarrow{\mathbf{f}^*} \in \mathrm{Int}(\mathcal{O})}$ for all $\eta \in [0, \eta_0]$, where $\mathbf{f}^* = \mathbf{f}(\mathbf{x}^*)$. In other words, an infinitesimal step along the direction of objective vector $\overrightarrow{\mathbf{f}^*}$ starting from $\mathbf{f}^*\in\partial\mathcal{O}$ will take it to the interior of $\mathcal{O}$. We call this as $\overrightarrow{\mathbf{f}^*}$ \textit{penetrates} $\mathcal{O}$. We consider it to be mild assumption and explain why in \S \ref{sec:assum_pen}.
 
 % the penetration assumption mild because
	% when the range set $\mathcal{O}$ is $m$ dimensional and its boundary $\partial\mathcal{O}$ is $m-1$ dimensional, almost all points in $\partial\mathcal{O}$ that violate the penetration assumption are not Pareto optimal, not even locally. 
	% This is discussed further in \S \ref{sec:assum_pen}.

    \begin{restatable}{theorem}{unconstdndnc}		\label{th:d_nd_unconst}
		If $\mathbf{x}^* \in \mathcal{P}$ is a regular Pareto Optimal solution, and its non\nobreakdash-dominating direction $\mathbf{d}_{nd} = \mathrm{F}^T\bm{\beta}^*$ is obtained by the QP \eqref{eq:qp_x0_po_unconst} with Cauchy-Schwarz inequality-based balancing anchor \eqref{eq:cs_anchor}, then $\mathbf{d}_{nd} = \mathbf{0}$ if and only if $\mathbf{x}^* \in \mathcal{P}_r$. 
	\end{restatable}
	
% 	Let ${\mathrm{F}\mathcal{F}(\mathbf{x^*}):=\{\mathrm{F}\!(\mathbf{x^*})\,\mathbf{d}\ |\  \mathbf{d} \in \mathcal{F}(\mathbf{x^*})\}}$ be the set of feasible directions on the objective space $\mathbb{R}^m$ against which one can move to remain within $\mathcal{O}$. We assume that this set contains the objective vector at $\mathbf{x}^*$ in its relative interior:
%     \begin{align}\label{eq:finFF} 
% 	-\mathbf{f}^* = -\mathbf{f}(\mathbf{x}^*)  \in \mathrm{RelInt}(\mathrm{F}\mathcal{F}(\mathbf{x^*})).
% 	\end{align}
% 	In other words, an infinitesimal step along the direction of objective vector $\overrightarrow{\mathbf{f}^*}$ starting from $\mathbf{f}^*\in\partial\mathcal{O}$ will take it to the interior of $\mathcal{O}$. Formally, we assume there exists an $\eta_0 > 0$ such that ${\mathbf{f}^* + \eta \mathbf{f}^* \in \mathrm{Int}(\mathcal{O})}$ for all $\eta \in [0, \eta_0]$. With this assumption we formalize the non-convergence Theorem.
	
	% \red{Theorem~\ref{th:d_nd_unconst} shows that when an EPO solution exists, the algorithm does not stop prematurely at a local Pareto Optimal solution; it traces the Pareto front until an EPO solution is found.} 
    Theorem~\ref{th:d_nd_unconst} holds for certain cases when the regularity assumptions are not satisfied (discussed in the proof). In our extension to solving constrained MOO problems, we make an additional assumption (see \S \ref{sec:assum_pen}) and restate the above result in Theorem \ref{th:d_nd}.

\subsection{Convergence Rate and Iteration Complexity}\label{sec:conv_rate}
% We analyze the convergence rates of the proposed gradient based EPO Search Algorithms for an unconstrained (non-linear) non-convex MOO problem. 
The convergence rate of a gradient-based SOO method to reach a stationary point is well known to be sub-linear for non-convex problems and linear for convex problems \citep{Nesterov2004}. \citet{Tanabe2022} extended this result for descent-based MOO methods to reach a Pareto stationary point: sub-linear for non-convex and linear for convex MOO problems. 

EPO Search has two modes of operation. In Algorithm~\ref{alg:epo_search_x0_random}, when the random initialization is not a PO solution, first the balance mode iterations decrease the value of $\omega$ down towards a small value $\epsilon$ to align the objective vector to the $\mathbf{r}^{-1}$ ray. Then, the descent mode iterations decrease each objective to reach an EPO solution that also satisfies the Pareto critical condition \eqref{eq:criticality_unconst}. 
%\red{The descent mode is similar to existing gradient based MOO methods, hence its sub-linear convergence rate can be proven similar to that of \cite{Tanabe2022}.} \blue{
The descent mode is similar to solving a non-convex SOO problem of the objective $\frac{1}{2}\|\mathbf{f}(\mathbf{x})\|^2$, with sub-linear convergence rate. %}
Here, we show that the balance mode has linear convergence rate. This result is significant because it shows that the EPO Search Algorithm \ref{alg:epo_search_x0_po}, where the initialization is a PO solution, can converge at linear rate to the desired PO solution, even for a non-convex MOO problem. Note, Algorithm \ref{alg:epo_search_x0_po} traces the PF by alternating between balance and descent modes, where the descent mode does not increase $\omega$ but balance mode decreases $\omega$. 
% down towards $\epsilon$. 

% To prove linear convergence of the balance mode for optimizing non-convex objectives $\mathbf{f}$, we harness the convex nature of $\omega_\mathbf{r}$. In particular, 
We analyze the composite objective function $\omega_\mathbf{r} \circ \mathbf{f}$ to prove linear convergence of the balance mode when optimizing non-convex objectives $\mathbf{f}$. The following Polyak-{\L}ojasiewicz (P\L) type  inequality \citep{POLYAK1963864,10.2307/1997218,10.1007/978-3-319-46128-1_50} is instrumental in our analysis.
\begin{restatable}{lemma}{plineq}
    \label{th:pl_ineq}
    There exists a $\tau > 0$ such that the proportionality gauges $\omega_\mathbf{r}$ \eqref{eq:cs_prop},\eqref{eq:lgrn_prop} and their respective anchor directions $\mathbf{a}$ in \eqref{eq:cs_anchor}, \eqref{eq:lgrn_anchor} satisfy 
    \begin{align} \label{eq:pl_ineq}
          \frac{1}{2}\|\mathbf{a}(\mathbf{f}) \|^2  \geq \tau \omega_\mathbf{r}(\mathbf{f}) \quad \forall \tau \leq \begin{cases}
            \langle\overrightarrow{\mathbf{f}_0}, \overrightarrow{\mathbf{r}^{-1}}\rangle^2 \\
            1
        \end{cases}\!\!
        \text{and} \quad
        \forall \mathbf{f} \in
        \begin{cases}
            \mathcal{M}^\mathbf{r}_{\mathbf{f}^0} \quad \text{if Cauchy-Schwarz inequality
            % \eqref{eq:non-uniformity}  \eqref{eq:kl_anchor}
            } \\
            \mathcal{O} \qquad \text{if Lagrange's identity 
            % \eqref{eq:lgrn_prop} \eqref{eq:lgrn_anchor}
            },
        \end{cases}
    \end{align}
    where $\mathbf{f}^0=\mathbf{f}(\mathbf{x}^0)$ is the initialization, $\overrightarrow{\mathbf{v}}$ is the $\ell_2$ normalization of vector $\mathbf{v}$, and $\mathcal{M}^\mathbf{r}_{\mathbf{f}^0}$ is as in \eqref{eq:Mrft}.
    % for both KL divergence ($\omega_\mathbf{r}$=\eqref{eq:non-uniformity}, $\mathbf{a}$=\eqref{eq:kl_anchor}) and Lagrange's identity ($\omega_\mathbf{r}$ = \eqref{eq:lgrn_prop}, $\mathbf{a}$=\eqref{eq:lgrn_anchor}). 
\end{restatable}
The inequality \eqref{eq:pl_ineq} suggests that the magnitude of anchor direction grows quadratically w.r.t. the value of the proportionality gauge. Our additional assumptions are as follows.
\begin{assumption}[Compactness]\label{asm:comp}
The image $\mathcal{O}$ of the objectives is a compact set in $\mathbb{R}^m$, upper bounded by a nadir point $\mathbf{f}^{ndr}$ and lower bounded by a minimum magnitude $M>0$: 
% $\mathbf{f} \preccurlyeq \mathbf{f}^{ndr}$ and lower bounded by a positive minimum magnitude 
$M \leq \|\mathbf{f}\| \leq \|\mathbf{f}^{ndr}\|\ \forall  \mathbf{f} \in \mathcal{O}$. 
The anchor directions \eqref{eq:cs_anchor} and \eqref{eq:lgrn_anchor} are bounded by $W$: $\|\mathbf{a}(\mathbf{f})\| \leq W \ \forall \mathbf{f} \in \mathcal{O}$.
\end{assumption}
\begin{assumption}[Smoothness]\label{asm:smth}
The gradients of each objective function is Lipschitz smooth: $\|\nabla f_j(\mathbf{x}^1) - \nabla f_j (\mathbf{x}^2)\| \leq L_j \|\mathbf{x}^1 - \mathbf{x}^2\|$ where $L_j>0$ for all $j\in [m]$; and $L_\mathbf{f} = \max_{j\in [m]} L_j$. The gradient of $\omega$ is also assumed to be Lipschitz smooth: $\|\nabla \omega_\mathbf{r}(\mathbf{f}^1) - \nabla \omega _\mathbf{r}(\mathbf{f}^2)\| \leq L_\omega \|\mathbf{f}^1 - \mathbf{f}^2\|$.
\end{assumption}

Since the Jacobian is $\mathrm{F}:\mathbb{X} \rightarrow \mathbb{R}^{m\times n}$ is a smooth function, the singular values of $\mathrm{F}$, 
$\sigma_j:\mathbb{X} \rightarrow \mathbb{R}_+$ for $j\in [m]$, where $ \sigma_1 \leq \sigma_2 \leq \cdots \leq \sigma_m$, are also smooth functions. Note, $\sigma_1(\mathbf{x}^*)=0$ for all $\mathbf{x}^*\in\mathcal{P}$ due to the Pareto criticality condition \eqref{eq:criticality_unconst}. So far, we assumed Jacobian $\mathrm{F}(\mathbf{x})$ is full rank if $\mathbf{x} \notin \mathcal{P}$ and rank $m-1$ if $\mathbf{x}\in \mathcal{P}$, as the regularity condition. We make this assumption more specific by considering a $\delta$-neighbourhood around the PF as 
% $\mathcal{P}^\delta := \left\{\mathbf{x} \in \mathbb{X} \,\middle|\, \exists \mathbf{x}^* \in \mathcal{P}\ \text{s.t.}\  {\lambda_1(\alpha\mathbf{x}^* + (1-\alpha) \mathbf{x}) < \delta}\ \forall \alpha \in [0,1]\right\}.$
\begin{align} \label{eq:P_delta}
    \mathcal{P}^\delta := \left\{\mathbf{x} \in \mathbb{X} \,\middle|\, \exists \mathbf{x}^* \in \mathcal{P}\ \text{s.t.}\  {\sigma_1(\nu\mathbf{x}^* + (1-\nu) \mathbf{x}) < \delta}\ \forall \nu \in [0,1]\right\}.
\end{align}
$\mathbb{X}\backslash \mathcal{P}^\delta$ can be considered as the operating region for the balance mode of EPO Search Algorithm~\ref{alg:epo_search_x0_random}, and $\mathcal{P}^\delta$ as the operating region of EPO Search Algorithm \ref{alg:epo_search_x0_po} that traces the PF.
\begin{assumption}[Regularity]\label{asm:reg} There exists a $\delta > 0$ such that the smallest singular value $\sigma_1(\mathbf{x}) \geq \delta$ for all $\mathbf{x} \in \mathbb{X} \backslash \mathcal{P}^\delta$ and the second smallest singular value $\sigma_2(\mathbf{x}) \geq \delta$ for all $\mathbf{x} \in \mathcal{P}^\delta$.
% {\color{blue}The non-zero singular values of $F(\mathbf{x})$ for all $\mathbf{x}\in \mathbb{X}$ is lower bounded by $\sigma$.}
\end{assumption}
With the above three assumptions,
% , Lemma \ref{th:pl_ineq} and an additional Lemma \ref{th:afd_angle},
we state the linear convergence rate result.
\begin{restatable}[Convergence Rate, Iteration Complexity]{theorem}{cvnrt}\label{th:conv_rate}
If the stepsize is any $\eta\in(0, \eta_0)$, where  $\eta_0=\frac{c_0}{c_0 + \max \{2c_1,3c_2,4c_3\}}$, ${c_0=\frac{\delta^2}{s_0\sqrt{m}W^2}}$, $c_1 = \frac{1}{2} (L_\omega+\frac{L_\mathbf{f}m}{s_1})$, $c_2= \frac{1}{2} L_\omega L_\mathbf{f}m^2$, $c_3 =\frac{1}{8}L_\omega L_\mathbf{f}^2m^2$, ${s_0=s_1=1}$ when $\omega_\mathbf{r}=$\eqref{eq:cs_prop}, and $s_0=\|\mathbf{f}^{ndr}\|$ and $s_1=M$ when $\omega_\mathbf{r}=$\eqref{eq:lgrn_prop}, 
then the balance mode iterations, using either \eqref{eq:cs_anchor} or \eqref{eq:lgrn_anchor} as anchor direction, decrease $\omega_\mathbf{r}$ linearly:
\begin{align}
    \omega_\mathbf{r}(\mathbf{f}^{t+1}) \leq \left(1- 2\tau\, \eta\, p(\eta)\right) \omega_\mathbf{r}(\mathbf{f}^{t})\leq \cdots \leq \left(1- 2\tau\, \eta\, p(\eta)\right)^{t+1} \omega_\mathbf{r}(\mathbf{f}^{0}),
\end{align}
where the polynomial $p(\eta) = c_0 - c_1\eta - c_2 \eta^2 - c_3 \eta^3$ is positive for all $\eta \in (0, \eta_0)$.
Consequently, the maximum number of iterations (iteration complexity) required to decrease $\omega_\mathbf{r}$ down to $\epsilon$ is $O(\log(\frac{1}{\epsilon}))$.
\end{restatable}

% \red{This Theorem is useful and an exciting discovery for non-convex optimization due to its following implications.  }
    \subsubsection{Time Complexity per Iteration:}
    \label{sec:time_complexity}
% 	, since the computation of $C=G^TG$ has runtime $O(m^2n)$.
    Both the $m$ dimensional QP problems \eqref{eq:qp_x0_random_unconst} and \eqref{eq:qp_x0_po_unconst} are convex, having linear constraints and are independent of the dimension $n$ of the solution space. 
    %Let $s$ denote the number of inequality constraints.
    The former has at most $s=3m$ inequality constraints, and the latter has at most $s=3m+1$ inequality constraints; both have $2m$ inequality constraints for $\ell_1$ ball. 
    % For unconstrained MOO, $s = 2m+1$ in the former and $s = m$ in the latter and $q=0$. 
    The QPs can be solved efficiently, e.g., using interior point methods \citep{cai2013complexity,zhang2021wide} in $O(m^2s)=O(m^3)$ time.
    %Note that these QP problems are  
    %The most time-consuming step is the 
    The complexity of the Jacobian matrix multiplication $\mathrm{F}\mathrm{F}^T$ is $O(nm^2)$.
    Thus, the per-iteration time complexity of both Algorithms \ref{alg:epo_search_x0_random} and \ref{alg:epo_search_x0_po} is $O(nm^2 + m^3)$.
    % $O(nm^2 + m^2(s+q))$.

%\vfill
%\pagebreak
\section{EPO Search for MCDM and MTL}
\label{sec:epo_mcdm_mtl}
% \red{We present three different \red{applications} of EPO Search in MCDM and MTL.
% We illustrate the ability of tracing the Pareto front for a posteriori MCDM in \S \ref{sec:pesa-epo}.
% EPO Search can be used to make preference elicitation for interactive MCDM more efficient as discussed in \S \ref{sec:gp-epo}.
% Finally, we present the use of EPO Search in training deep neural networks for multi-task learning (MTL) in \S \ref{sec:mtl}.
% }

\subsection{Pareto Front Approximation for A posteriori MCDM}\label{sec:pesa-epo}
%\subsection{EPO for Diverse Preference vectors to Approximate the Pareto Front}
% \section{Approximating the Pareto Front with EPO Search}
% \red{In a posterior MCDM, since the preferences over different solutions are unknown a priori, the goal is to first get a diverse set Pareto optimal solutions $\widehat{\mathcal{P}}$ that approximates the Pareto Front $\mathcal{P}$, then pick one overall best among them based on several heuristics \cite{JING2019123}. Here we address the first goal of approximating the Pareto Front.
% }
    With the tracing capability of our algorithm we can move from one PO solution to another, and discover new ones in the path. We use this feature to generate a diverse set of optimal solutions by tracing towards the EPO solutions for different weight vectors. We adopt the \textit{Pattern Efficient Set} Algorithm (PESA) by \cite{Stanojevic2020} for generating a diverse set of weight vectors in the $m-1$ dimensional Simplex $\mathcal{S}^m$. 
    %\subsubsection{PESA-EPO:} 
    % In PESA, the weight vectors are deterministically sampled from $\mathcal{S}^m$ in a recursive manner to progressively approximate the simplex.
    % fill the gaps among existing optimal solutions in the Pareto front. 
    PESA is a recursive sampling procedure to approximate the simplex $\mathcal{S}^m$.
    Instead of sampling $\mathbf{r}$, we directly sample the  rays $\mathbf{r}^{-1}$ %, named as preference rays henceforth, 
    since it is directly (not inversely) associated with the anchor directions. The sampling process in PESA is as follows: given a set of $m$ rays $\mathrm{R}^0 = \left\{\mathbf{r}_k^{-1}\in \mathcal{S}^m \middle|\ k \in [m]\right\}$, the next new ray $\mathbf{r}_{new}^{-1}$ is sampled as a convex combination of rays in $\mathrm{R}^0$, i.e. $\mathbf{r}_{new}^{-1} = \frac{1}{m}\sum_{k=1}^m \mathbf{r}_k^{-1}$.
    % \begin{align}
    %     \mathbf{r}_{new}^{-1} = \frac{1}{m}\sum_{k=1}^m \mathbf{r}_k^{-1}.
    % \end{align}
    This new ray creates $m$ more sets, $\mathrm{R}^{0j} = \mathrm{R}^0 \backslash \{\mathbf{r}_{j}^{-1}\} \cup \{\mathbf{r}_{k+1}^{-1}\} $ for all $j \in [m]$, and one can recursively sample rays from these $m$ sets. Thus, the convex hull of the original set $\mathrm{R}^0$ is ``filled". If $\mathrm{R}^0$ consists of the axes of positive orthant in $\mathbb{R}^m$, then this recursive sampling process approximates the entire simplex.

    We integrate this sampling rule with EPO Search algorithm and develop the PESA-EPO Search Algorithm \ref{alg:pesa-epo} for approximating the PF. Instead of a set of $\mathbf{r}^{-1}$ rays, we maintain a set of PO solutions $\mathrm{R}=\{\mathbf{x}^*_k\}_{k=1}^m$, and sample the next ray as 
    % Given a set of $m$ Pareto optimal objective vectors $\mathrm{R} = \{\mathbf{f}^*_k\}_{k=1}^m$, first we sample the next preference ray as
    \begin{align}\label{eq:pesa-epo-rnew}
        \mathbf{r}^{-1}_{new} = \frac{1}{|\mathrm{R}|} \sum_{\mathbf{x} \in \mathrm{R}} \frac{\mathbf{f}(\mathbf{x})}{\|\mathbf{f}(\mathbf{x})\|_1}.
    \end{align}
    We then run EPO search (Algorithm \ref{alg:epo_search_x0_po}) to trace the PF
    %, i.e., Algorithm \ref{alg:epo_search_x0_po}, 
    from $\mathbf{f}(\mathbf{x}_k^*)$ to $\mathbf{r}^{-1}_{new}$ ray for each $\mathbf{x}_k^* \in \mathrm{R}$. 
    This generates $m$ trajectories on the PF.
    % , wherein each end point is (approximately) 
    %a Pareto optimal 
    % an EPO solution. 
    % Let ${}^k\mathbf{f}_k^*$ be the end point of a trajectory starting at $\mathbf{f}_k^*$. If the EPO solution exists, then all of ${}^k\mathbf{f}_k^*$ will be close to each other near to $\mathbf{r}^{-1}_{k+1}$ ray, i.e. $\omega(\mathbf{r}^{-1}_{k+1},\ {}^k\mathbf{f}_k^*) \sim 0$. If not, then instead of $\mathbf{r}^{-1}_{k+1}$, we use the end points ${}^k\mathbf{f}_k^*$  to build $m$ new sets: $\mathrm{R} \,\bigcup\, \{{}^k\mathbf{f}_k^*\} \backslash \{\mathbf{f}_k^*\}$ for all $k\in[m]$.
    Let ${}^k\mathbf{x}_{\mathbf{r}_{new}}^*$ be the end point of the trajectory starting at $\mathbf{x}_{k}^*$. This creates creates $m$ new sets, $R \backslash \{\mathbf{x}_{k}^*\} \cup \{{}^k\mathbf{x}_{\mathbf{r}_{new}}^*\}$ for all $k\in [m]$, and the same procedure is repeated recursively. 
    % The exact same procedure is recursively performed on these new $m$ sets. 
    The procedure starts with an initial set $\mathrm{R}$ consisting of the optimal solutions of individual objectives, and the recursion stops at a given input depth. In the collection of points obtained from the trajectories, there could be solutions that are dominated by others. In a post-processing step, the dominated solutions are removed. 
    % to finally present the set of non-dominated solutions. 
    %We call this method PESA-EPO.
    
    % We conjecture that if the initial set of objective vectors are (closest to the) EPO solutions for the extreme preference rays, i.e. positive axes of $\mathbb{R}^m$, and the boundary of attainable objective vectors $\partial\mathcal{O}$ is connected, then one can arbitrarily approximate the Pareto Front  by increasing the depth of the above recursive procedure. 
    Note that, the PF $\mathbf{f}(\mathcal{P})\subset \partial \mathcal{O}$ may be disconnected. But the trajectories of EPO Search can move between different portions of $\mathcal{P}$ if $\mathcal{O}$ is one connected component, which for unconstrained MOO is trivially true, and for constrained MOO is true when the domain $\mathbb{X}$ is a connected component. 
    % due to the connected boundary assumption the trajectories of EPO Search can move between different portions of $\mathcal{P}$.
    In case of a bi-objective optimization with a connected PF, a recursion depth of just $1$ can approximate the PF. Because for $m=2$, the PF will be at most a $1$-dimensional manifold, and the trajectories of traced EPO Search are also $1$-dimensional. This is further clarified in our empirical results (\S \ref{sec:appox_pf}).

    \begin{algorithm}[!h]
    \caption{PESA-EPO for Pareto Front approximation}\label{alg:pesa-epo}
    \begin{algorithmic}[1]
    \State {\bf Input:} $MaxDepth$ \Comment{Depth of maximum recursion for PESA-EPO}
    \Procedure{PESA\_EPO}{$\mathrm{R}$, $depth$, $MaxDepth$}
    \State $PF=\phi$
    \If{$depth\leq MaxDepth$}
    \State Compute $\mathbf{r}^{-1}_{new}$ from \eqref{eq:pesa-epo-rnew}
    %$\mathbf{r}^{-1} = \frac{1}{|R|} \sum_{p \in R} \frac{p.\mathbf{f}}{\|p.\mathbf{f}\|_1}$
    \For{$\mathbf{x}$ in $\mathrm{R}$} %\Comment{for all $p=(\mathbf{x}, \mathbf{f}(\mathbf{x}))$ in $R$}
    \State $Traces=$ Output of EPO Search Algorithm \ref{alg:epo_search_x0_po} run for $\mathbf{r}$ starting from $p.\mathbf{x}$
    \State $\mathbf{x}_{new} = $ last solution in $Traces$ %\Comment{$p_{new}\simeq(\mathbf{x}_\mathbf{r}, \mathbf{f}(\mathbf{x}_\mathbf{r}))$ Approximate EPO}
    \State $\mathrm{R}_{new} = \{\mathbf{x}_{new}\} \cup \mathrm{R}  \backslash \{\mathbf{x}\} $
    \State $PF = PF \cup Traces \ \cup $ PESA\_EPO$(\mathrm{R}_{new},\ depth + 1,\ MaxDepth)$
    \EndFor
    \EndIf
    \State \Return PF
    \EndProcedure
    \State $\mathrm{R} = \{\mathbf{x}_j^*\}_{j=1}^m$ \Comment{where $\mathbf{x}_j^* = \argmin_{\mathbf{x} \in \mathbb{X}} f_j(\mathbf{x})$ for all $j\in [m]$}
    \State {\bf Output:} $PF =$ PESA\_EPO($R, 1, MaxDepth$)
    \end{algorithmic}
\end{algorithm}

\subsection{Preference Elicitation in Interactive MCDM}
\label{sec:gp-epo}

    % In interactive MCDM also, the DM's preferences are not known a priori. It is assumed 

%\subsubsection{GP-EPO:}
%We consider the active learning framework where the DM's unknown utility function is modelled as a Gaussian Process $\mathrm{GP}(\mu, \kappa)$. In each iteration, there is an incumbent solution $\mathbf{x}_{inc}^t$ which is preferred to all other solutions in $\ddot{\mathbb{X}}_{\mathcal{D}_t}$, the discrete set of alternatives presented so far to the DM. The DM is asked to compare between $\mathbf{x}_{inc}^t$ and $\mathbf{x}_{sug}^t$, the latter obtained by solving \eqref{eq:pe_prac}, an optimization problem over $\ddot{\mathbb{X}}$, a discrete subset of the feasible solution set $\mathbb{X}$.

The key idea of our approach is to operate in the domain of the $m-1$ dimensional simplex $\mathcal{S}^m$ instead of the high-dimensional solution space $\mathbb{X}$ (or its discrete subset that is used in \eqref{eq:pe_prac} by previous methods). In our PE, the {\it data} on which the GP is learnt consists of $\mathbf{r}^{-1} \in \mathcal{S}^m$ instead of $\mathbf{x} \in \mathbb{X}$. Using a mapping $\psi_\mathbf{f}$ defined below and EPO Search, we find PO solutions and
return to $\mathbb{X}$.
%to present alternatives to the DM. 
% \blue{In this section, we abuse the notation $\mathbf{r}^{-1}$ with $\mathbf{r}$ for simplicity.}

Let $\psi_\mathbf{f}: \mathcal{S}^m \rightarrow \widebar{\mathbb{R}}^m$, where $\widebar{\mathbb{R}} = \mathbb{R} \cup \{\infty\}$ is the extended real line,  be a mapping from the simplex $\mathcal{S}^m$ to the objective space $\mathcal{O}$
% , whose range covers the PF in the objective space, 
defined as
\begin{align}\label{eq:psi}
    \psi_\mathbf{f}(\mathbf{r}^{-1}) = s^*\, \mathbf{r}^{-1}\quad \forall \mathbf{r}^{-1}\in \mathcal{S}^m, \quad
    \text{where}\quad s^* &= \min_{s \geq 0} s  \quad \text{s.t. } s\mathbf{r}^{-1} \in \mathcal{O}.
\end{align}
% The function 
$\psi_\mathbf{f}$ maps (scales with factor $s^*$) a point $\mathbf{r}^{-1}\in \mathcal{S}^m$ in the simplex to a point on  $\partial \mathcal{O}$, the boundary of the image $\mathcal{O}$, if the $\mathbf{r}^{-1}$ ray intersects $\mathcal{O}$. If it does not intersect $\mathcal{O}$, then it is mapped to a vector at $\infty$, since the scaling factor $s^*$ will be $\infty$. 
Clearly, the image of $\mathcal{S}^m$ under $\psi_\mathbf{f}$ covers all the PO solutions, $\psi_\mathbf{f}(\mathcal{S}^m) \supseteq \mathbf{f}(\mathcal{P})$. Using EPO search Algorithm~\ref{alg:epo_search_x0_po}, for a given $\mathbf{r}^{-1}\in \mathcal{S}^m$, we can reach $\psi_\mathbf{f}(\mathbf{r}^{-1})$ and the corresponding EPO solution $\mathbf{x}^*_\mathbf{r}$ if it exists.
Non-existence can be detected through the value of proportionality gauge, i.e. $\omega_\mathbf{r}(\mathbf{f}(\widehat{\mathbf{x}}_\mathbf{r})) > \epsilon$, where $\widehat{\mathbf{x}}_\mathbf{r}$ is the output of Algorithm~\ref{alg:epo_search_x0_po}. 
%\red{explain what s and s* mean, what the mapping does in words}

\begin{algorithm}[t]
    \caption{GP-EPO for Preference Elicitation}\label{alg:iepo}
    \begin{algorithmic}[1]
    \State {\bf Input:} $\mathbf{x}^{0},\ \mathbf{x}^{1}$ \Comment{$\mathbf{x}^{0}, \mathbf{x}^{1} \in \mathcal{P}$}
    \State {\bf Initialize:} $\mathbf{x}_{inc}=\mathbf{x}^0$, $\mathbf{x}_{sug}=\mathbf{x}^1$, $\widehat{\mathcal{D}} = \phi$, $\mathrm{GP}$ \Comment{Initialize GP.$\hat{\mu}$ as zero function}
    \While{maximum queries not reached}
    \State $\mathbf{r}^{-1}_{inc} = \frac{\mathbf{f}_{inc}}{\|\mathbf{f}_{inc}\|_1}$, $\mathbf{r}^{-1}_{sug} = \frac{\mathbf{f}_{sug}}{\|\mathbf{f}_{sug}\|^1}$
    \State response = QueryDM$(\mathbf{f}_{inc}, \mathbf{f}_{sug})$
    \If{response = ``incumbent preferred to suggestion"} \; 
    $\widehat{\mathcal{D}} = \widehat{\mathcal{D}} \cup \{ (\mathbf{r}^{-1}_{inc}, \mathbf{r}^{-1}_{sug})\}$
    \ElsIf{response = ``suggestion preferred to incumbent"}
    \State $\widehat{\mathcal{D}} = \widehat{\mathcal{D}} \cup \{(\mathbf{r}^{-1}_{sug}, \mathbf{r}^{-1}_{inc})\}$, $\mathbf{x}_{inc}=\mathbf{x}_{sug}$, $\mathbf{r}^{-1}_{inc}=\mathbf{r}^{-1}_{sug}$
    \Else \; $\widehat{\mathcal{D}} = \widehat{\mathcal{D}} \cup \{(\mathbf{r}^{-1}_{inc}, \mathbf{r}^{-1}_{sug}), (\mathbf{r}^{-1}_{sug}, \mathbf{r}^{-1}_{inc})\}$
    \EndIf
    \State Update GP.$\hat{\mu}$ and GP.$\hat{\kappa}$ with $\widehat{\mathcal{D}}$ \Comment{Approximates the posterior $P(\hat{u} | \widehat{\mathcal{D}})$}
    \State Get $\mathbf{r}^{-1}_{sug}$ by maximizing the updated acquisition function GP.$\hat{a}$ \eqref{eq:pe_ours}
    \State Get $\mathbf{x}_{sug},\ \mathbf{f}_{sug}$ from EPO Search Algorithm \ref{alg:epo_search_x0_po} starting from $\mathbf{x}_{inc}$
    \If{$\mu(\mathbf{f}_{sug}, \mathbf{r}_{sug}) > \epsilon$} \; 
    $\widehat{\mathcal{D}} = \widehat{\mathcal{D}} \cup \{ (\mathbf{r}^{-1}_{inc}, \mathbf{r}^{-1}_{sug})\}$ \Comment{$\psi(\mathbf{r}^{-1}_{sug})$ in \eqref{eq:psi} is unreachable}
    \EndIf
    \If{($\|\mathbf{f}_{inc} - \mathbf{f}_{sug}\| < \epsilon$) or (DM satisfied)} {\bf break} \EndIf
    \EndWhile
    \State {\bf Output:} $\mathbf{x}_{inc}$
    \end{algorithmic}
\end{algorithm}

Instead of modeling the GP prior $u$ for the entire $\mathcal{O}$, we model $\hat{u}:\mathcal{S}^m \rightarrow \mathbb{R}$, where $\hat{u} = u \circ \psi_\mathbf{f}$ with $\mathrm{GP}(\hat{\mu}, \hat{\kappa})$, where $\hat{\mu}:\mathcal{S}^m \rightarrow \mathbb{R}$ and $\hat{\kappa}: \mathcal{S}^m\times \mathcal{S}^m \rightarrow \mathbb{R}$.
We interactively estimate the utility $\hat{u}$ for PE by maintaining a parallel dataset $\widehat{\mathcal{D}}_t = \{\hat{c}_1, \cdots, \hat{c}_{t'}\}$, where the ordered pair $\hat{c} = ({\mathbf{r}^{-1}}^{i}, {\mathbf{r}^{-1}}^j)$ corresponds to the DM's comparison of two PO solutions $c=(\mathbf{x}^i, \mathbf{x}^j)$ (see \S \ref{sec:interactive_back})
% \ref{sec:pe_back}
by using the inverse mapping of $\psi_\mathbf{f}$: ${\mathbf{r}^{-1}}^i=\frac{\mathbf{f}(\mathbf{x}^i)}{\|\mathbf{f}(\mathbf{x}^i)\|_1}$, and similarly for ${\mathbf{r}^{-1}}^j$. 
Thus, we convert the PE for $\mathbb{X}\xrightarrow{\mathbf{f}}\mathbb{R}^m \xrightarrow{u} \mathbb{R}$ into a PE for $\mathcal{S}^m \xrightarrow{\psi_\mathbf{f}} \mathbb{R}^m \xrightarrow{u} \mathbb{R}$.
Analogous to $\ddot{\mathbb{X}}_{\mathcal{D}_t}$ in \S \ref{sec:interactive_back},
% \ref{sec:pe_back}, 
we define $\ddot{\mathcal{S}}^m_{\mathcal{D}_t}$ to be the discrete set of $\mathbf{r}^{-1}$s whose corresponding $\mathbf{x}$s have been presented to the DM. Similarly, the incumbent $\mathbf{r}^{-1}_{inc}$ corresponds to $\mathbf{x}_{inc}$. After updating the GP to approximate the posterior $P(\hat{u}|\widehat{\mathcal{D}}_t)$, we obtain  
\begin{align}\label{eq:pe_ours}
    \mathbf{r}^{-1}_{sug} = \argmax_{\mathbf{r}^{-1} \in \mathcal{S}^m} \, \hat{\alpha}^t(\mathbf{r}^{-1})
\end{align}
as the next suggestion, where $\hat{\alpha}_t$ is the acquisition function formulated using the updated $\hat{\mu}_t$  and $\hat{\kappa}_t$. Then we run EPO Search Algorithm \ref{alg:epo_search_x0_po} starting from $\mathbf{x}_{inc}$ to find $\mathbf{x}_{sug}$, the EPO solution for $\mathbf{r}^{-1}_{sug}$.
% with $\mathbf{x}^{inc}$ as the starting point and $\mathbf{r}^{sug}$ as the preference ray to obtain $\mathbf{x}^{sug}$. 
If $\omega_\mathbf{r}(\mathbf{f}(\mathbf{x}_{sug})) > \epsilon$ is detected, i.e. the $\mathbf{r}^{-1}$ ray does not intersect $\mathcal{O}$ and $\psi(\mathbf{r}^{-1})$ maps to a point at infinity, we augment the dataset $\widehat{D}_t$ with $(\mathbf{r}^{-1}_{inc}, \mathbf{r}^{-1}_{sug})$, declaring that $\mathbf{r}^{-1}_{sug}$ is inferior to $\mathbf{r}^{-1}_{inc}$.
% This scenario can be easily handled by checking if the output of EPO Search $\mathbf{f}(\mathbf{x}^{sug})$ does not align with $\mathbf{r}^{sug}$ and augmenting the dataset $\widehat{D}_t$ with $(\mathbf{r}^{inc}, \mathbf{r}^{sug})$, declaring that $\mathbf{r}^{sug}$ is inferior to $\mathbf{r}^{inc}$.
Thus, we avoid optimization on the high dimensional domain of solutions $\mathbb{X}$, and instead 
%As a result, instead of performing interactive PE from the high dimensional solution space $\mathbb{X}$, we can now 
do it from the domain of $m-1$ dimensional $\mathcal{S}^m$. We summarize our method
%interactive EPO search 
in Algorithm \ref{alg:iepo}.

GP-EPO does not estimate the utility of non-PO solutions.
Although $u$ is defined for all possible alternatives, its estimation for the entire range $\mathcal{O}$ of $\mathbf{f}$ is unnecessary due to the monotonicity property of a utility function. 
Since the DM's preferred solution is assumed to be PO, it suffices to estimate the utility only for solutions on the PF. 
Note that for any non-PO solution $\mathbf{x}$, we can employ Algorithm~\ref{alg:epo_search_x0_random} to obtain the EPO solution $\mathbf{x}^*_\mathbf{r}$ for $\mathbf{r}^{-1} = \frac{\mathbf{f}(\mathbf{x})}{\|\mathbf{f}(\mathbf{x})\|_1}$, which is preferred over $\mathbf{x}$. 
%\blue{The only limitation of this approach is that the utility of non-Pareto optimal solution cannot be estimated. However, under the assumption that $u$ is a monotonic function and the DM is interested only on the Pareto optimal solution, this limitation becomes irrelevant. Because, for a given any non-Pareto optimal solution $\mathbf{x}$, we can employ Algorithm~\ref{alg:epo_search_x0_random} to obtain the EPO solution $\mathbf{x}^*_\mathbf{r}$ for $\mathbf{r} = \frac{\mathbf{f}(\mathbf{x})}{\|\mathbf{f}(\mathbf{x})\|_1}$, which is more preferred to the DM than $\mathbf{x}$. }

Our approach obviates the need to model additional monotonicity constraints on the GP, done in previous GP-based PE methods 
\citep{10.5555/3237383.3237920,10.1007/978-3-030-67664-3_28}.
Moreover, unlike \eqref{eq:pe_prac}, we do not discretize the domain in \eqref{eq:pe_ours}, since global optimization of non-linear objectives is computationally feasible for low-dimensional problems by running multiple threads initialized at several seeds in $\mathcal{S}^m$. Therefore, we can explore %access 
the PF at its highest possible resolution. This is facilitated by the linear convergence rate of EPO Search Algorithm \ref{alg:epo_search_x0_po} (see \S \ref{sec:conv_rate}) to efficiently obtain the next alternative solution $\mathbf{x}^{sug}$ %on the fly 
while interacting with the DM. Note that a CS--based GP procedure cannot achieve this, since CS cannot trace the PF requiring a re-initialization for every query. This can have a sub-linear convergence rate at best, akin to a non-convex SOO algorithm.
% \red{advantage over GP-CS}

\subsection{Deep Multi-Task Learning}
\label{sec:mtl}

%In this section we present an application of EPO Search in training deep neural networks (DNN) for supervised multi-task learning (MTL).
%Here \red{preferences} among objectives are provided as (fixed) inputs and hence, illustrates the use of EPO Search in a-priori MOO.

%MTL setup

In many MTL applications, 
%As discussed above, 
model builders may require %explore 
%models with %\textit{specific} 
trade-offs in the form of priorities 
%or \red{preferences} 
among the tasks. 
%The need for such a selection arises in several MTL applications.
%For instance, 
	%\red{CHANGE} in their multi-task recommender system \cite{milojkovic2019multi} optimize semantic relevance, content quality and revenue.
	%In different applications, we may want models %from the Pareto set
	%that prioritize 
	%with varying priorities for relevance, quality and revenue.
% if a MTL model is designed for the tasks of drug efficacy prediction and side-effect prediction, the drug discovery team may prefer a model that prioritizes drug efficacy prediction over side-effect prediction, while the pharmacovigilance team may prefer the opposite.
%This would enable the stakeholder to not only leverage the advantages of %MOO-based MTL in terms of improved generalizability but also obtain a model that is focussed on the individual task.
Consider $m
$ tasks for MTL, indexed by $1 \le i,j \le m$.
We assume priority specification for each task by numeric values, with higher values indicating higher task priority.
Let $r_i$ and $f_i$ denote the priority and loss function for the $i^\text{th}$ task.
%The requirement %for \red{preference}-driven MTL can be stated as follows: 
For any two tasks,
%Among $n$ considered tasks for MTL, for any two tasks -- the $i^\text{th}$ and $j^\text{th}$ tasks where $i,j \in 1,\ldots,n$, if
if the priority for the $i^\text{th}$ task is higher than that of the $j^\text{th}$ task, i.e., if $r_i \ge r_j$, then we want
%preference weights $r_i \ge r_j$,  
% then the corresponding training losses should follow $l_i \le l_j$, i.e., 
the network to be trained better for the $i^\text{th}$ task, i.e.,
%If the preferences are specified using real numbers, %between 0 and 1, then for the $i^\text{th}$ and $j^\text{th}$ tasks, if preferences are $r_i \ge r_j$, 
we want the corresponding training losses to follow 
$f_i \le f_j$. 
To the best of our knowledge, current MOO-based MTL methods do not model such priorities. %\deb{What to do for $\theta$?}

This can be achieved by training the network using scalarized MOO and
we propose the use of CS, which
%, which, to our knowledge, has not been used in MTL. %\citep{10.1007/978-3-642-74919-3_8} for \red{preference}-driven MTL.
overcomes the limitations of LS (see \S \ref{sec:related}) and 
%offers multiple advantages:
% unlike LS, it provides both necessary and sufficient condition for (weak) Pareto optimality. 
% It 
satisfies the required inverse relationship between priorities and objective values exactly at the EPO solution, i.e., $r_1f_1 = \cdots = r_j f_j = \cdots = r_mf_m$, for all $m$ objectives as shown in Figure \ref{fig:mtl_moo} (right).
%Thus, EPO Search can be use to train neural networks for such priority-based MTL.
Both the problems of oscillation and premature stagnation (especially when there are tasks with low priority weights) are effectively addressed by EPO Search. 
%By using a linear combination of all the objectives’ gradients, EPO Search descends along the preference ray without oscillations and without systematically choosing only objective. Moreover, the ability to ascend enables EPO Search to escape the minima of less preferred objectives making it robust to initialization.
%Both the problems of CS and the efficacy of EPO Search are illustrated through an example in \S\ref{sec:exp_toy_moo}.
Further, 
%many deep MTL models require constraints on the weights that provide various forms of 
regularization, in the form of constraints on parameters, may be required in deep MTL models to prevent over-fitting. EPO Search can handle both the unconstrained case and cases of equality, inequality and box constraints (see \S \ref{sec:constrained_moo}). 

These advantages in EPO Search are achieved without compromising on its efficiency. 
% – the per-iteration complexity of EPO Search (see \S \ref{sec:time_complexity}) remains linear in the gradient dimensions (similar to the method of 
% \cite{NeurIPS2018_Sener_Koltun} that neither uses input priorities nor handles constraints).
%Note that the the per-iteration complexity of EPO Search, for $n$ network weights and $m$ tasks, is  %  $O(nm^2 + m^2(s+q))$.
In deep MTL, the number of DNN parameters ($n$) is typically much greater than the number of tasks ($m$).
    The most time-consuming step in EPO Search in such cases is the Jacobian matrix multiplication.
    Thus, the per-iteration complexity of EPO Search is linear in $n$ and quadratic in $m$.
    This is comparable to the method of 
\cite{NeurIPS2018_Sener_Koltun} that neither uses input priorities nor handles constraints.
In comparison, gradient descent with CS and LS scale linearly with the number of objectives.
However, PMTL scales exponentially with $m$, as the number of reference vectors required for decomposing  the objective space  increases exponentially with $m$ (see \S \ref{sec:exp_toy_moo}).
%for an empirical comparison).

Priority weights are assumed to be provided as inputs during model training.
These weights may be determined based on domain knowledge, data-related factors and application-specific requirements.
For instance, tasks that are more difficult due to, e.g., lesser training data, may be given higher priority (as done in our case study \S \ref{sec:expt_real}).
Priorities may be considered as hyperparameters and automated  tuning techniques \citep{yang2020hyperparameter} may be used.
Many are based on Bayesian optimization and use GP to model the unknown generalization performance of the model.
%, and are similar to GP-EPO.

    % {\color{blue}Computationally both CS and LS scale linearly with the number of objectives. EPO Search scales quadratically as it solves a QP problem of same dimension as the number of objectives. All these $3$ methods are linear with respect to the dimension of solution space n. In MTL applications, usually $n>>m$; the number of neural network parameters are a lot more the number of tasks. However, PMTL scales exponentially with $m$, as the number of reference vectors required for decomposing  the objective space (as discussed in \ref{sec:po_prefer}) increases exponentially with $m$. }

\section{Experimental Results}
\label{sec:expt}

    \subsection{Advantages of EPO Search for gradient descent: A synthetic MOO problem}
    \label{sec:exp_toy_moo}
        \begin{minipage}{0.5\linewidth}
    \begin{figure}[H]
    \centering
    \includegraphics[width=\linewidth]{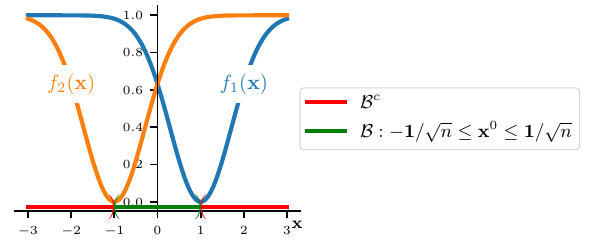}%\vspace{-.3cm}
    \caption{$1$d solution space of \eqref{eq:moo_toy} in hyper-box \eqref{eq:hyperbox}
    \vspace{0.5cm}}
    \label{fig:synth_moo}
    \end{figure}
    \end{minipage}~
    \begin{minipage}{0.49\linewidth}
    % \vspace{-2cm}
    \begin{subequations}\label{eq:moo_toy}
    % \footnotesize
    \begin{align}
        f_1(\mathbf{x}) &= 1 - \exp\left(\!- \left\|\mathbf{x} - \frac{\mathbf{1}}{\sqrt{n}}\right\|_2^2\right)\\
        f_2(\mathbf{x}) &= 1 - \exp\left(\!- \left\|\mathbf{x} + \frac{\mathbf{1}}{\sqrt{n}}\right\|_2^2\right)
    \end{align}
    \end{subequations}
    \begin{align}\label{eq:hyperbox}
        \mathcal{B} = \left\{ \mathbf{x} \in \mathbb{R}^n \;\middle|\; -\mathbf{1}/\sqrt{n} \preccurlyeq_n \mathbf{x} \preccurlyeq_n \mathbf{1}/\sqrt{n} \right\}
    \end{align}
    % where $\mathbf{x} \in \mathbb{R}^n$; Figure~\ref{fig:synth_moo} shows this function for $n=1$. The set of attainable objective values $\mathcal{O}$ is also non\nobreakdash-convex in the objective space $\mathbb{R}^2$.
    %\vspace{0.5cm}
    \end{minipage}
    %\subsubsection{A 2-objective problem}
    
    We use the problem introduced by
    \cite{fonseca1995multiobjective} to 
    show the advantage of EPO Search (Algorithm \ref{alg:epo_search_x0_random}) over competing  approaches: LS, PMTL and CS (where \eqref{eq:chebychev} is solved).
    %Linear Scalarization (LS), Chebychev Scalarization (CS), Pareto MTL (PMTL). %\cite{NIPS2019pmtl}.
    This problem consists of two non-convex objective functions \eqref{eq:moo_toy} that are to be minimized over $\mathbf{x} \in \mathbb{R}^n$; Figure~\ref{fig:synth_moo} shows the functions for $n=1$. 
    In this problem, the set of PO solutions $\mathcal{P}$ is a subset of the hyper-box defined in \eqref{eq:hyperbox},
    % \begin{align}\label{eq:hyperbox}
    %     \mathcal{B} = \left\{ \mathbf{x} \in \mathbb{R}^n \;\middle|\; -\mathbf{1}/\sqrt{n} \preccurlyeq_n \mathbf{x} \preccurlyeq_n \mathbf{1}/\sqrt{n} \right\},
    % \end{align}
    where $\preccurlyeq_n$ denotes the partial ordering induced by the positive cone $\mathbb{R}^n_+$ in the solution space. Note that for $n=1$, we have $\mathcal{B}=\mathcal{P}$ as shown in Figure \ref{fig:synth_moo}. We evaluate each MOO algorithm in two scenarios, when the initialization is: (a) inside this hyper-box, i.e., $\mathbf{x}^0\in \mathcal{B}$, and (b) outside this hyper-box, i.e., $\mathbf{x}^0 \in \mathcal{B}^c$. The latter is more difficult, especially when the EPO is far from the initialization $\mathbf{x}^0$. For instance, in Figure \ref{fig:synth_moo}, if the desired optimal is $x^*=-0.5$ and the initialization is at $x^0=2$, the iterate has to escape the minimum of objective $f_1$, i.e., $x=1$, to reach $x^*$. In other words, without ascending in $f_1$, it is not possible to reach $x^*$ from $x^0$ in a continuous trajectory, i.e., using a gradient-based iterative algorithm.
    Each algorithm is tested with four weight vectors, spread uniformly over the first quadrant. The number of iterations, step size, and random initializations are the same for all algorithms. 
    
%    \begin{figure}[!h]
%        \centering
%        \begin{subfigure}[b]{0.49\linewidth}
%        	\centering
%            \includegraphics[width=0.5\linewidth]{EPO_easy_init.pdf}~\hspace{-0.02\linewidth}~
%            \includegraphics[width=0.5\linewidth]{Chebychev_easy_init.pdf}\\
%            \includegraphics[width=0.5\linewidth]{PMTL_easy_init.pdf}
%            \caption{Easy initialization\label{fig:easy_init}}
%        \end{subfigure}~\hspace{0.03\linewidth}~
%        \begin{subfigure}[b]{0.49\linewidth}
%        	\centering
%            \includegraphics[width=0.5\linewidth]{EPO_hard_init.pdf}~\hspace{-0.02\linewidth}~
%            \includegraphics[width=0.5\linewidth]{Chebychev_hard_init.pdf}\\
%            \includegraphics[width=0.5\linewidth]{PMTL_hard_init.pdf}
%            \caption{Hard intialization \label{fig:hard_init}}
%        \end{subfigure}
%      \caption{ (Color Online) 
%      Trajectories of EPO Search (left) and PMTL (right) in $\mathbb{R}^2$ with $n=20$ dimensional solution space, when initializations are near Pareto optimal, $\mathbf{x}^0 \in \mathcal{B}$ (fig \ref{fig:easy_init}), and are far, $\mathbf{x}^0 \notin \mathcal{B}$ (fig \ref{fig:hard_init}).
%      \label{fig:moo_toy_result}}
%    \end{figure}
    \begin{figure}[!h]
    % 	\centering
    	\begin{subfigure}[b]{\linewidth}
    		\centering
    		\includegraphics[width=0.25\linewidth]{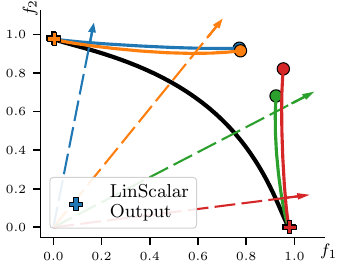}~\hspace{-0.02\linewidth}~
    		\includegraphics[width=0.25\linewidth]{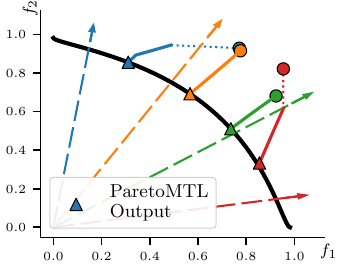}~\hspace{-0.02\linewidth}~
    		\includegraphics[width=0.25\linewidth]{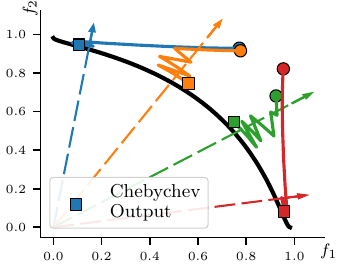}~\hspace{-0.02\linewidth}~
    		\includegraphics[width=0.25\linewidth]{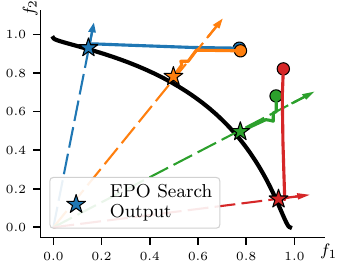}
    		\caption{Easy initialization, $\mathbf{x}^0 \in \mathcal{B}$\label{fig:easy_init}}
    	\end{subfigure}\\
    	\begin{subfigure}[b]{\linewidth}
    		\centering
    		\includegraphics[width=0.25\linewidth]{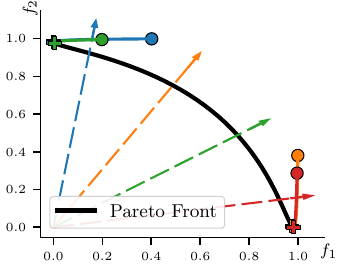}~\hspace{-0.02\linewidth}~
    		\includegraphics[width=0.25\linewidth]{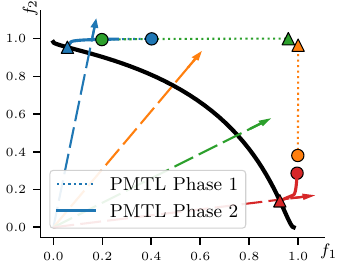}~\hspace{-0.02\linewidth}~
    		\includegraphics[width=0.25\linewidth]{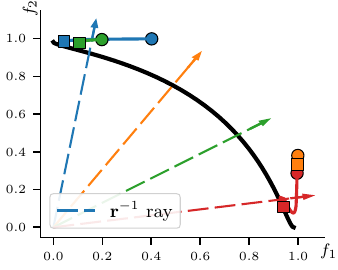}~\hspace{-0.02\linewidth}~
    		\includegraphics[width=0.25\linewidth]{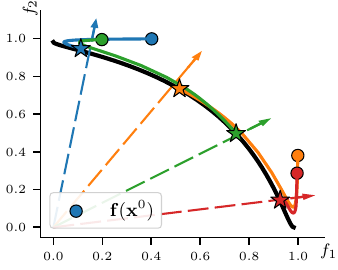}
    		\caption{Hard initialization, $\mathbf{x}^0 \in \mathcal{B}^c$ \label{fig:hard_init}}
    	\end{subfigure}
    	\caption{Trajectories of MOO algorithms in $\mathbb{R}^2$ with $n=20$ dimensional solution space, when initialized (a) inside and (b) outside the hyper-box \eqref{eq:hyperbox}. 
    	%(fig \ref{fig:easy_init}), and outside (fig \ref{fig:hard_init}). 
    	In both rows, the results of LS, PMTL, CS, and EPO search are presented from left to right. 
    	Each weight vector is shown in a different color, and matched to the color of each algorithm's trajectory towards the corresponding EPO solution. Legend for PF, $\mathbf{r}^{-1}$, $f(\mathbf{x}^0)$ are for all subfigures.
    	%Color of a trajectory indicates the preference for which the MOO problem was solved.
    % 		EPO Search (left) and PMTL (right) in $\mathbb{R}^2$ with $n=20$ dimensional solution space, when initializations are near Pareto optimal, $\mathbf{x}^0 \in \mathcal{B}$ (fig \ref{fig:easy_init}), and are far, $\mathbf{x}^0 \notin \mathcal{B}$ (fig \ref{fig:hard_init}).
    \label{fig:moo_toy_result}}
    \end{figure}

    \begin{figure}[b]
    \begin{minipage}{0.6\textwidth}
    \centering
		\begin{subfigure}{0.45\linewidth}
			% \centering
			\includegraphics[width=\linewidth]{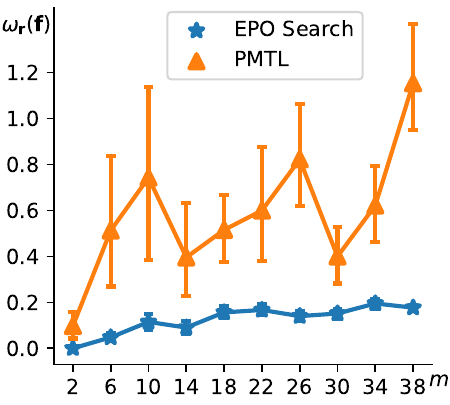}
			\caption{\label{fig:sim}}
		\end{subfigure}~
		\begin{subfigure}{0.45\linewidth}
			% \centering
			\includegraphics[width=\linewidth]{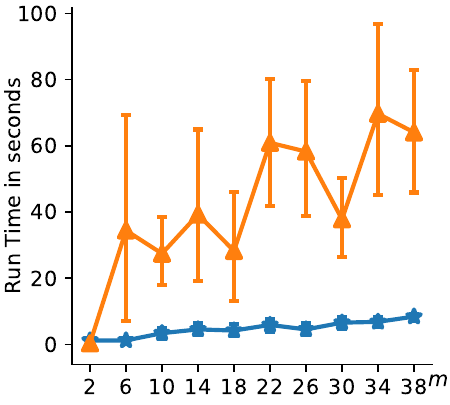}
			\caption{\label{fig:run_time}}
		\end{subfigure}
    \end{minipage}~
	\begin{minipage}{0.38\textwidth}
	    \caption{\label{fig:synthetic}Scalability of PMTL and EPO~Search with increasing number of objective functions ($m$). 
		%Figure \ref{fig:sim}
		(a): quality of EPO solution ($\omega_r$), 
		%figure \ref{fig:run_time}
		(b): run time for $200$ iterations (in seconds). }
	\end{minipage}
	\end{figure}
    Figure \ref{fig:moo_toy_result} shows the results. 
    LS does not reach any of the EPO solutions, owing to its theoretical limitations for non-convex MOO problems. In its phase 1, PMTL enters the vicinity of the EPO, and in phase 2 it descends to the PF, although not exactly to the goal. However, PMTL converges to the PF only when initialized near the EPO, and diverges otherwise, e.g., in the green and yellow trajectories of Figure \ref{fig:hard_init} when $\mathbf{x}^0 \in \mathcal{B}^c$. The CS method theoretically has the EPO as its solution, but in practice, when an iterative procedure is used with a step size, it oscillates around the $\mathbf{r}^{-1}$ ray (Figure \ref{fig:easy_init}), without reaching the goal exactly
    because %This happens because, in CS, 
    only one objective (with the maximum relative value \eqref{eq:chebychev}) is considered in each iteration. Although the oscillations could be reduced with a smaller step size, it would demand more number of iterations to reach the EPO.
    Using only one objective becomes more problematic if it's gradient magnitude vanishes. For instance, the green and yellow trajectory in Figure \ref{fig:hard_init} does not make substantial progress. This is similar to the $1$d scenario in Figure \ref{fig:synth_moo} discussed above, i.e., if $x^*=-0.5$ and $x^0=2$, only $f_2$ will be considered whose gradient is close to zero.
    On the other hand, EPO search reaches very close to all the EPO solutions. Unlike CS, it uses a linear combination of all the objectives' gradients. As a result, it descends along the $\mathbf{r}^{-1}$ ray without oscillations. Moreover, the ability to ascend enables EPO Search to escape the minima of less preferred objectives in Figure \ref{fig:hard_init} making it robust to initialization. 
    
    % When initialization is inside the hyperbox  (figure~\ref{fig:easy_init}), we observe 
    % % in Figure \ref{fig:easy_init} 
    % that PMTL, that does not use any notion of proportionality, descends in every step and reaches a Pareto optimal but not at the preference vectors.
    % On the other hand, EPO search reaches the preference-specific solutions. 
    % When initialization is outside the hyperbox (figure~\ref{fig:hard_init}), we see that EPO Search both descends and ascends in a controlled manner. It traces the Pareto front to find the required solutions, which makes it robust to initialization. 
    % No updates are seen in phase 2 of PMTL, when initialization is outside the hyper-box and far from the preference vectors. It reaches the Pareto front only in 2 out of 4 runs. 

    % \red{compare time complexity of EPO, LS, CS}
    We extend the above example to create $m$ objectives functions and compare PMTL with EPO Search, with respect to their scalability.
    %in Appendix \ref{sec:epo_scale}.
    %\subsection{Scalability to Many Objectives}
    %\label{sec:epo_scale}
    % \subsection{Synthetic Experiments: Many Objectives} % Robustness
	%We test how our algorithm scale with increasing number of objectives and compare that with Pareto MTL. 
	The objective functions are defined as
% 	We create $m$ loss functions as
	$f_j(\mathbf{x}) = 1 - \exp \left(-\left\|\mathbf{x} - \hat{\mathbf{x}}^j\right\|_2^2\right)$, for $j \in [m]$,
% 	\begin{align}\label{eq:multi_obj}
% 	f_j(\mathbf{x}) = 1 - \exp \left(-\left\|\mathbf{x} - \hat{\mathbf{x}}^j\right\|_2^2\right), \quad j \in [m],
% 	\end{align}
	where the entries of $\hat{\mathbf{x}}^j \in \mathbb{R}^n$ are sampled uniformly in $[-1/n, 1/n]$. For every $m$, we run both the algorithms for $20$ different $n$ (dimension of solution space), randomly sampled within $20$ and $100$. We randomly select a weight vector in $\mathbb{R}^m_+$ for every $(m, n)$ pair. In addition %to a preference vector, the 
    PMTL requires $K$ reference vectors;
	%, which, according to the authors, should be increased exponentially with the increasing $m$. However, 
	for a fair comparison, we provide $K=2m$ (maximum number of constraints in EPO search in this problem) reference vectors, which are again randomly selected in $\mathbb{R}^m_+$. 
    
	% \begin{figure}[t]
	% \centering
	% 	\begin{subfigure}{0.3\linewidth}
	% 		% \centering
	% 		\includegraphics[width=\linewidth]{simulation.pdf}
	% 		\caption{\label{fig:sim}}
	% 	\end{subfigure}~
	% 	\begin{subfigure}{0.3\linewidth}
	% 		% \centering
	% 		\includegraphics[width=\linewidth]{run_time.pdf}
	% 		\caption{\label{fig:run_time}}
	% 	\end{subfigure}
	% 	\caption{Scalability of PMTL and EPO Search with increasing number of objective functions ($m$). 
	% 	%Figure \ref{fig:sim}
	% 	(a): quality of EPO solution ($\omega_r$), 
	% 	%figure \ref{fig:run_time}
	% 	(b): run time for $200$ iterations (in seconds).}
	% 	\label{fig:synthetic}
	% \end{figure}	
	We use $\omega_r$ from Lagrange identity \eqref{eq:lgrn_prop} as a measure of the quality of %preference-specific Pareto optimal 
	the solutions found. % by the algorithms.
	For every $(m,n)$ pair, both the algorithms were run for $200$ iterations with equal step size.
	Figure \ref{fig:sim} and \ref{fig:run_time} show the quality and run time, respectively, for different number of objectives ($m$). 
	%The comparison of overall run time (in seconds) is shown in Figure \ref{fig:run_time}. 
 Compared to PMTL,
	EPO search scales better with increasing number of objectives and produces better quality solutions. 

        \begin{figure}[b]
        \centering
        \begin{subfigure}{0.24\textwidth}
            \centering
            \includegraphics[width=\linewidth, height=0.8\linewidth]{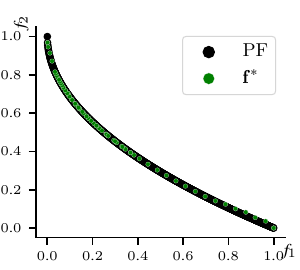}
            \caption{\label{fig:zdt1}ZDT1}
        \end{subfigure}
        \begin{subfigure}{0.24\textwidth}
            \centering
            \includegraphics[width=\linewidth, height=0.8\linewidth]{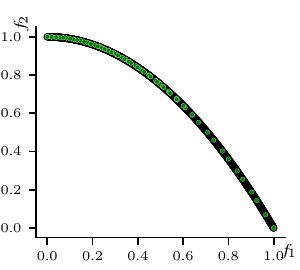}
            \caption{\label{fig:zdt2}ZDT2}
        \end{subfigure}
        \begin{subfigure}{0.24\textwidth}
            \centering
            \includegraphics[width=\linewidth, height=0.8\linewidth]{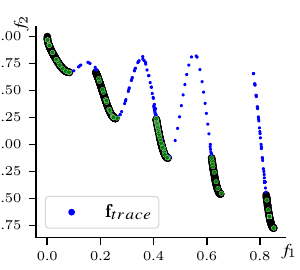}
            \caption{\label{fig:zdt3}ZDT3}
        \end{subfigure}
    %   \caption{
    %   \label{fig:zdt} Result of applying PESA-EPO to approximate/trace the Pareto front. 
    %   }
    % \end{figure}
    % \begin{figure}[h]
    %     \centering
        \begin{subfigure}{0.24\textwidth}
            \centering
            \includegraphics[width=\linewidth]{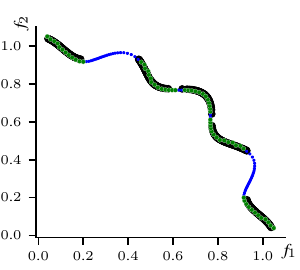}
            \caption{\label{fig:tnk}TNK}
        \end{subfigure}
        \begin{minipage}{0.66\textwidth}
        \begin{subfigure}{0.49\textwidth}
            \centering
            \includegraphics[width=\linewidth]{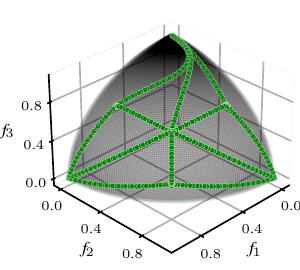}
            \caption{\label{fig:dtlz2}\small DTLZ2}
        \end{subfigure}~
        \begin{subfigure}{0.49\textwidth}
            \centering
            \includegraphics[width=\linewidth, height=0.8\linewidth]{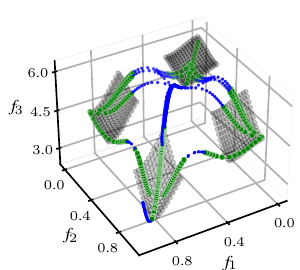}
            \caption{\label{fig:dtlz7}\small DTLZ7}
        \end{subfigure}
        \end{minipage}~
        \begin{minipage}{0.33\textwidth}
        \caption{
        \label{fig:trace-pf} PF traced by EPO Search Algorithm \ref{alg:epo_search_x0_po} on six benchmark problems. $\mathbf{f}_{trace}$ is the unfiltered iteration history of EPO Search trajectory from one EPO to another. $\mathbf{f}^*$ is obtained after removing the non-dominated points from $\mathbf{f}_{trace}$.}
      \end{minipage}
    \end{figure}
    \subsection{Pareto front tracing by EPO Search: Illustrations on benchmark problems} \label{sec:pf_trace_exp}
    % \subsubsection{Evaluation on Benchmark MOO Problems.}
    We show the tracing ability of EPO Search Algorithm \ref{alg:epo_search_x0_po} on 6 benchmark MOO problems: ZDT1, ZDT2, ZDT3 \citep{Zitzler2000}, DTLZ2 and DTLZ7 \citep{Deb2005}, and TNK \citep{537993}. We use PESA \citep{Stanojevic2020} (see \S \ref{sec:pesa-epo}) to generate $\mathbf{r}^{-1}$ rays. %and trace between the corresponding EPO solutions.
    %\subsubsection{Pareto front generated by EPO Search.}
    %\label{sec:pf_illus}
 
    The ZDT series of problems minimize two objectives: $f_1(\mathbf{x})=x_1$ and $f_2(\mathbf{x}) = g(\mathbf{x}) h(f_1(\mathbf{x}), g(\mathbf{x}))$, where 
    $g(\mathbf{x}) = 1 + \frac{9}{n-1} \sum_{i=2}^n x_i$, 
    $h(f_1, g)=1 - \sqrt{f_1/g}$ for ZDT1, $h(f_1, g) = 1 - (f_1/g)^2$ for ZDT2 and $h(f_1, g)= 1 - \sqrt{f_1/g} - (f_1/g) \sin (10\pi f_1)$ for ZDT3.
    % \begin{align*}
    %     h(f_1, g) &= 
    %     \begin{cases}
    %     1 - \sqrt{f_1/g} \quad \text{for ZDT1}, \\
    %     1 - (f_1/g)^2 \quad \text{for ZDT2}, \\
    %     1 - \sqrt{f_1/g} - (f_1/g) \sin (10\pi f_1) \quad \text{for ZDT3},
    %     \end{cases}
    % \end{align*}
    The argument $\mathbf{x}$ is bounded as $0 \leq x_i \leq 1$ for $i = 1, \cdots, n$. Figures \ref{fig:zdt1}, \ref{fig:zdt2} and \ref{fig:zdt3} show the results. 
    %that traces the Pareto Front of these problems starting from the EPO solutions of extreme preference vectors (see section \ref{sec:pesa-epo}). 
    In these results, the depth of PESA procedure is 1 for ZDT1 and ZDT2, and 2 for ZDT3. Note that ZDT3 has a disconnected PF, but its boundary of attainable objective vectors $\partial\mathcal{O}$ is connected. Therefore, while tracing, EPO Search connects the disconnected segments of the PF
    by tracing the boundary $\partial\mathcal{O}$ through $\mathbf{f}_{trace}$.
    %in the pursuit of the EPO solution. 
    This is made possible due to the controlled gradient ascent within EPO Search. 
    %For the disconnected in fig \eqref{fig:zdt3}, the algorithm moves from one segment of PF to another .
 In the TNK problem, $m=n=2$. The two objectives to minimize are $f_1(\mathbf{x}) = x_1$ and $f_2(\mathbf{x}) = x_2$ with 
    bounds $0 \leq x_i \leq \pi$ and 
    the inequality constraints $x_1^2 + x_2^2 - 1 - 0.1\cos(16 \arctan\left(x_1/x_2\right)) \geq 0$, and $(x_1 - 0.5)^2 + (x_2 - 0.5)^2 \leq 0.5$.
    % \begin{align*}
    %     x_1^2 + x_2^2 - 1 - 0.1\cos(16 \arctan\frac{x_1}{x_2}) \geq 0, \quad
    %     (x_1 - 0.5)^2 + (x_2 - 0.5)^2 \leq 05.
    % \end{align*}
    It has a discontinuous PF.
    Figure \ref{fig:tnk} shows the tracing result for this problem.

    The DTLZ series of problems have more than two objectives to minimize. Let the last $n-m+1$ variables of $\mathbf{x}\in [0,1]^n$ be denoted as $\mathbf{x}_m$. The $m$ objective functions in DTLZ2 are defined as
    \begin{align*}
        % \min \quad& f_1(\mathbf{x}) = (1 + g(\mathbf{x}_m))  \cos(x_1\pi/2)\cdots\cos(x_{m-2}\pi/2)\cos(x_{m-1}\pi/2) \\
        % \min \quad& f_2(\mathbf{x}) = (1 + g(\mathbf{x}_m)) \cos(x_1\pi/2)\cdots\cos(x_{m-2}\pi/2)\sin(x_{m-1}\pi/2) \\
        % \min \quad& f_3(\mathbf{x}) = (1 + g(\mathbf{x}_m)) \cos(x_1\pi/2)\cdots\sin(x_{m-2}\pi/2) \\
        % \vdots \quad\quad & \quad \vdots  \\
        % \min \quad& f_m(\mathbf{x}) = (1 + g(\mathbf{x}_m)) \sin(x_1\pi/2) \\
        f_j(\mathbf{x}) &= (1 + g(\mathbf{x}_m)) \prod_{i=1}^{m-j}  \cos(x_{i}\pi / 2) \times
        \begin{cases}
            1, \quad \text{if}\ j = 1 \\
            \sin(x_{m-j+1}\pi/2), \quad \text{if}\ j=2,\cdots, m
        \end{cases}
        % \\
        % \text{with} \quad g(\mathbf{x}_m) &= \sum_{x_i \in \mathbf{x}_m} (x_i - 0.5)^2.
    \end{align*}
    with $g(\mathbf{x}_m) = \sum_{x_i \in \mathbf{x}_m} (x_i - 0.5)^2$.
    In DTLZ7, the first $m-1$ objective vectors are $f_j = x_j$ for $j=1, \cdots, m-1$. The last objective is defined as $f_m(\mathbf{x}) = (1+g(\mathbf{x}_m))\,h(f_1, f_2, \cdots, f_{m-1}, g)$, where $g(\mathbf{x}_m) = 1 + \frac{9}{n-m+1} \sum_{x_i \in x_m} x_i$, and $h(f_1, f_2, \cdots, f_{m-1}, g) = m - \sum_{j=1}^{m-1} \frac{f_j}{1 + g} (1 + \sin(3 \pi f_j))$.
    % \begin{align*}
    %     f_m(\mathbf{x}) &= (1+g(\mathbf{x}_m))\,h(f_1, f_2, \cdots, f_{m-1}, g), \\
    %     \text{where}\quad g(\mathbf{x}_m) &= 1 + \frac{9}{n-m+1} \sum_{x_i \in x_m} x_i, \\
    %     h(f_1, f_2, \cdots, f_{m-1}, g) &= m - \sum_{j=1}^{m-1} \frac{f_j}{1 + g} (1 + \sin(3 \pi f_j)).
    % \end{align*}
    Figures \ref{fig:dtlz2} and \ref{fig:dtlz7} shows the tracing results for DTLZ problems. Note that DTLZ7 has disconnected PF but has a connected boundary of attainable objective vectors $\partial\mathcal{O}$. Therefore, akin to ZDT3, here also EPO Search connects the PFs while tracing. For clarity of presentation, we show the tracing results for $\mathbf{r}^{-1}$ rays generated up to a recursion depth of 3 in PESA.

    %\subsubsection{Evaluation of generated Pareto front.}
    %\label{sec:pf_igd}
    
    \subsection{Evaluation of PESA-EPO for Pareto front approximation} % on benchmark MOO problems} 
    \label{sec:appox_pf}
    %In \S \ref{sec:pf_igd} 
    We numerically compare the efficacy of PESA-EPO Algorithm \ref{alg:pesa-epo} 
    %developed in \S \ref{sec:pesa-epo} 
    in PF approximation %and numerically compare its efficacy 
    with PESA-CS and two state-of-the-art algorithms: %NSGA-II by \cite{996017}, 
    CTAEA \citep{8413136}, %an evolutionary algorithm 
    and PESA-TDM \citep{Stanojevic2020}.
    %, a gradient-based method. 
    % NSGA-II (\textit{Non-dominated Sorting Genetic Algorithm}) and 
    CTAEA ({Constrained Two-Archive Evolutionary Algorithm}) is an evolutionary algorithm, whereas PESA-TDM ({Pattern Efficient Search Algorithm with Targeted Directional Model}) is a gradient-based algorithm. 
    Empirically, 
    CTAEA has been found to outperform 
    %several evolutionary algorithms for constrained MOO: 
    C-MOEA/D, C-NSGA-III \citep{jain2013evolutionary},
    C-MOEA/DD \citep{li2014evolutionary},
    I-DBEA \citep{asafuddoula2014decomposition} 
    and CMOEA \citep{woldesenbet2009constraint}; and
    the performance of PESA-TDM was found to be similar or better than NSGA-II \citep{996017},
    %four variants of MOEA/D ${\text{\citep{zhang2007moea}}}$: 
    MOEA/DDE \citep{4633340}, MOEA/D-AWA \citep{10.1162/EVCO_a_00109}, MOEA/D-UD1 and MOEA/D-UD2 \citep{7347861}.
    %For CTAEA, we used the python package pymoo \citep{pymoo}, and for PESA-TDM we used the authors' source code 
    %provided by the authors \citep{Stanojevic2020}.}
    In PESA-CS, CS (by solving \eqref{eq:chebychev}) is used instead of TDM.
    %\blue{We implemented PESA-CS similar to PESA-TDM, to find the EPO solution for a weight vector, CS is used instead of TDM.}
    
   % We evaluate the tracing ability of EPO Search (Algorithm \ref{alg:epo_search_x0_po}) on the task of Pareto front approximation on benchmark problems, both with and without constraints.
   % To generate preference rays to be used as inputs to EPO Search, we adopt the Pattern Efficient Set Algorithm (PESA) by \cite{Stanojevic2020}.
   % PESA generates a diverse set of 
   % preference vectors sampled from the $m-1$ dimensional Simplex $\mathcal{S}^m$  
   %  % In PESA, the preference vectors are sampled from $\mathcal{S}^m$
   %  recursively to progressively fill the gaps among initial Pareto optimal solutions.
   %  %in the Pareto front. %Instead of sampling $\mathbf{r}$, we directly sample the preference rays $\mathbf{r}^{-1}$ %, named as preference rays henceforth, 
   %  %since it is directly (not inversely) associated with the anchor directions and EPO objective vector.
   %  EPO Search is used to find EPO solutions at the specified preferences (at each level of recursion) that produces trajectories along the Pareto front.
   %  Non-dominated solutions from the trajectories are removed in a post-processing step and 
   %  the set of all Pareto optimal solutions in these trajectories are collectively used as an approximation to the Pareto front.
   %  We call this entire procedure PESA-EPO.
   %  More details are in \S \ref{sec:pesa-epo}.
    %In \S \ref{sec:pf_illus}

To evaluate how closely the obtained solutions approximate the PF,
    we use \textit{Inverted Generational Distance} (IGD) \citep{10.1007/978-3-540-24694-7_71}, defined in \eqref{eq:igd}, 
    %as a performance indicator to measure how closely the obtained solutions approximate the PF. Let 
    where the ground truth PF $\mathcal{P}_g = \{\mathbf{y}^*_1, \cdots, \mathbf{y}^*_{|\mathcal{P}_g|}\}$ is a finely discretized set of solutions from the actual PF $\mathcal{P}$, and $\mathcal{P}_a = \{\mathbf{f}^*_1, \cdots, \mathbf{f}^*_{|\mathcal{P}_a|}\}$ is the set of points found by an algorithm. 
    %Then IGD is defined as in \eqref{eq:igd}.
    % \begin{align} 
    %     \mathrm{IGD}(\mathcal{P}_g, \mathcal{P}_a) = \frac{1}{|\mathcal{P}_g|} \sum_{i = 1}^{|\mathcal{P}_g|} \,d(\mathbf{y}^*_i,\, \mathcal{P}_a)  \quad
    %     \text{where} \quad d(\mathbf{y}^*_i,\, \mathcal{P}_a) = \min_{\mathbf{f}^*_j \in \mathcal{P}_a} \, \|\mathbf{y}^*_i - \mathbf{f}^*_j\|.
    % \end{align}
    Ground truth PFs for the problems (\S \ref{sec:pf_trace_exp}) were obtained from 
    %the JMetal framework 
    \cite{5586354,jmetalcode}.
    %Each algorithm is used to obtain the Pareto front approximation of the 6 benchmark MOO problems mentioned above and the 
    IGD values and time of execution for all the methods  are shown in Table \ref{tab:igd_time_comparison}.
    %the web address \url{http://jmetal. sourceforge.net/problems.html}.
    % We compare the efficacy of our algorithm with two state-of-the-art algorithms: %NSGA-II by \cite{996017}, 
    % CTAEA \citep{8413136} and PESA-TDM \citep{Stanojevic2020}. 
    % % NSGA-II (\textit{Non-dominated Sorting Genetic Algorithm}) and 
    % CTAEA ({Constrained Two-Archive Evolutionary Algorithm}) is an evolutionary algorithm, whereas PESA-TDM ({Pattern Efficient Search Algorithm with Targeted Directional Model}) is a gradient-based algorithm. 
    % Empirically, 
    % CTAEA has been found to outperform several evolutionary algorithms for constrained MOO: 
    % C-MOEA/D, C-NSGA-III \citep{jain2013evolutionary},
    % C-MOEA/DD \citep{li2014evolutionary},
    % I-DBEA \citep{asafuddoula2014decomposition} 
    % and CMOEA \citep{woldesenbet2009constraint}; and
    % the performance of PESA-TDM was found to be similar or better than NSGA-II \citep{996017} and
    % four variants of MOEA/D ${\text{\citep{zhang2007moea}}}$: MOEA/DDE \citep{4633340}, MOEA/D-AWA \citep{10.1162/EVCO_a_00109}, MOEA/D-UD1 and MOEA/D-UD2 \citep{7347861}.
    % For CTAEA, we used the python package pymoo \citep{pymoo}, and for PESA-TDM we used the authors' source code 
    % %provided by the authors 
    % \citep{Stanojevic2020}.
    % Each algorithm is used to obtain the Pareto front approximation of the 6 problems mentioned above and the IGD values and time of execution are shown in Table \ref{tab:igd_time_comparison}.
    \begin{table}[t]
    %Comparison among algorithms based on IGD and time of execution for MOO different problems.}
    \begin{minipage}{0.66\textwidth}
    \centering
% Table generated by Excel2LaTeX from sheet 'pesa_exp'
    \resizebox{\textwidth}{!}{
    \begin{tabular}{ccccccc}
          MOO Problems    &  Metrics   & CTAEA & PESA-TDM & PESA-CS & PESA-EPO \\
    \toprule
    % \multirow{12}[10]{*}{\begin{sideways}Multi-Objective  Problems\end{sideways}} & 
    \multirow{2}{*}{ZDT1 ($m=2, n=30$)} & IGD   & 0.0426 & 0.0088 & 0.0095 & \textbf{0.0016} \\[1mm]
    % \cmidrule{3-7}      
          & Times (s) & 4.01  & 1.71  & 2.1   & \textbf{1.38} \\
    % \cmidrule{2-7}      
    \midrule
    \multirow{2}{*}{ZDT2 ($m=2, n=30$)} & IGD   & 0.0404 & 0.0051 & 0.0059 & \textbf{0.0016} \\[1mm]
    % \cmidrule{3-7}      
          & Times (s) & 4.05  & 9.93  & 11.75 & \textbf{1.14} \\
    % \cmidrule{2-7}
    \midrule
    \multirow{2}{*}{ZDT3 ($m=2, n=30$)} & IGD   & 0.0572 & 0.0217 & 0.062 & \textbf{0.0027} \\[1mm]
    % \cmidrule{3-7}      
          & Times (s) & 4.08  & 20.23 & 38.75 & \textbf{1.85} \\
    % \cmidrule{2-7}      
    \midrule
    \multirow{2}{*}{TNK ($m=n=p=2$)} & IGD   & 0.0922 & 0.0069 & 0.0117 & \textbf{0.0061} \\[1mm]
    % \cmidrule{3-7}      
          & Times (s) & 1.64  & \textbf{0.61} & 6.48  & 0.83 \\
    % \cmidrule{2-7}      
    \midrule
    \multirow{2}{*}{DTLZ2 ($m=3, n=12$)} & IGD   & \textbf{0.0269} & 0.0681 & 0.1214 & 0.0307 \\[2mm]
    % \cmidrule{3-7}      
          & Times (s) & 221.77 & 40.38 & 68.13 & \textbf{2.9} \\
    % \cmidrule{2-7}      
    \midrule
    \multirow{2}{*}{DTLZ7 ($m=3, n=12$)} & IGD   & \textbf{0.0369} & 0.0439 & 0.1532 & 0.0384 \\[1mm]
    % \cmidrule{3-7}      
          & Times (s) & 60.49 & 41.74 & 48.16 & \textbf{2.32} \\
    \bottomrule
    \end{tabular}%
    }
    \end{minipage}~
    \begin{minipage}{0.35\textwidth}
        \caption{PF approximation metric IGD and execution time (lower is better for both) of MOO algorithms CTAEA, PESA-based TDM, CS, and EPO 
    on benchmark MOO problems ($m$ objectives, $n$ variables, $p$ constraints). Row-wise best result is in bold.\label{tab:igd_time_comparison}}
        \begin{align} 
        \mathrm{IGD}
        % (\mathcal{P}_g, \mathcal{P}_a) 
        &= \frac{ \sum_{i = 1}^{|\mathcal{P}_g|} \,d(\mathbf{y}^*_i,\, \mathcal{P}_a) }{|\mathcal{P}_g|} \label{eq:igd}\\ 
        % \text{where} 
        d(\mathbf{y}^*_i,\, \mathcal{P}_a) &= \min_{\mathbf{f}^*_j \in \mathcal{P}_a} \, \|\mathbf{y}^*_i - \mathbf{f}^*_j\|. \nonumber
    \end{align}
    \end{minipage}
    \end{table}
    The results indicate that PESA-EPO is able to efficiently (lower time of execution) achieve close approximation to the PF (lower IGD). CTAEA uses a decomposition technique similar to PMTL (\S \ref{sec:mtl_back} and Appendix \ref{sec:pmtl_info}). Its computational complexity 
    %of decomposition strategy 
    grows exponentially with the number of objectives, as seen in our experiment as well. The time required to reach an IGD value of same scale as that of the competing algorithms is significantly more in DTLZ2 and DTLZ7, where $m=3$, as compared to the other bi-objective problems.  
    PESA-TDM is efficient and suitable when the dimension of solution space is low: in TNK, it achieves as good an approximation as PESA-EPO with lesser execution time. Although PESA-CS is similar to PESA-TDM, it requires more samples of weight vectors from the PESA recursions (see \S \ref{sec:pesa-epo}) since CS stagnates for some weights and does not reach the PF, thereby requiring more time to achieve similar level of IGD values as that of PESA-TDM.
    For high-dimensional solution spaces both PESA-TDM and PESA-CS are inefficient because, for every new weight vector in PESA, they have to solve an optimization problem starting from a random initialization.
    On the other hand, PESA-EPO uses a previously obtained EPO as an initialization to solve the next %optimization
    problem. Moreover, the points in the trajectory of this optimization are PO solutions. As a result, PESA-EPO efficiently achieves very good performance.
    %As a result, PESA-EPO outperforms competing methods. 

    \subsection{Evaluation of GP-EPO for preference elicitation} % on benchmark MOO problems}
    \label{sec:pref_eli_exp}
    To evaluate an interactive PE algorithm, we measure the decrease in regret with the number of queries to the DM. The regret at the $t^\text{th}$ query is defined as the difference between the oracle utility and the incumbent, i.e., the best solution so far: $reg^t = u\left( \mathbf{f}(\mathbf{x}_{orc})\right) - u\left( \mathbf{f}(\mathbf{x}_{inc}^t)\right).$
    % \begin{align}
    %     reg^t = u\left( \mathbf{f}(\mathbf{x}_{orc})\right) - u\left( \mathbf{f}(\mathbf{x}_{inc}^t)\right).
    % \end{align}
    
    \begin{figure}[t]
        \centering
        \begin{subfigure}[b]{0.24\textwidth}
            \centering
            \includegraphics[page=1,width=\linewidth]{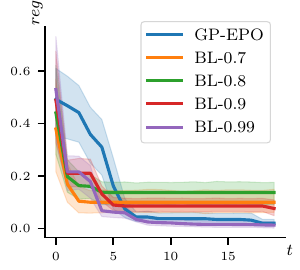}
            \caption{\label{fig:zdt1_pe}ZDT1}
        \end{subfigure}
        \begin{subfigure}[b]{0.24\textwidth}
            \centering
            \includegraphics[page=2,width=\linewidth]{gp-epo.pdf}
            \caption{\label{fig:zdt2_pe}ZDT2}
        \end{subfigure}
        \begin{subfigure}[b]{0.24\textwidth}
            \centering
            \includegraphics[page=3,width=\linewidth]{gp-epo.pdf}
            \caption{\label{fig:zdt3_pe}ZDT3}
        \end{subfigure}
        \begin{subfigure}[b]{0.24\textwidth}
            \centering
            \includegraphics[page=4,width=\linewidth]{gp-epo.pdf}
            \caption{\label{fig:tnk_pe}TNK}
        \end{subfigure}
        \begin{minipage}{0.5\textwidth}
            \begin{subfigure}[b]{0.48\textwidth}
            \centering
            \includegraphics[page=5,width=\linewidth]{gp-epo.pdf}
            \caption{\label{fig:dtlz2_pe}\small DTLZ2}
        \end{subfigure}~
        \begin{subfigure}[b]{0.48\textwidth}
            \centering
            \includegraphics[page=6,width=\linewidth]{gp-epo.pdf}
            \caption{\label{fig:dtlz7_pe}\small DTLZ7}
        \end{subfigure}
        \end{minipage}~
        \begin{minipage}{0.48\textwidth}
            \caption{
            \label{fig:pe} Results of Interactive Preference Elicitation. The 4 baselines (BL) use 70\%, 80\%, 90\% and 99\% of the ground truth PO solutions. 
            The number of ground truth solutions are: problems ZDT1, ZDT2 and ZDT3 -- each 1000, TNK -- 450, DTLZ2 -- 10k, DTLZ7 -- 680.}
        \end{minipage}
    \end{figure}
    We use the same 6 MOO problems described in \S \ref{sec:pf_trace_exp} to evaluate %\ref{alg:iepo} 
    GP-EPO.
    Following \cite{ozbeykarwan2014}, we use the Chebyshev utility function (which is unknown to the PE algorithm) to simulate a virtual DM that compares between two alternatives. 
    Previous GP-based PE methods
    \citep{8618894,10.5555/3237383.3237920,10.1007/978-3-030-67664-3_28}, differ from GP-EPO in the use of \eqref{eq:pe_prac} with a discrete set of points $\ddot{\mathbb{X}}$ (that could, e.g., be generated by a PF approximator like PESA-EPO).
    %One can use an efficient algorithm like PESA-EPO \ref{sec:pesa-epo} that approximates the PF and use them as the discrete set $\ddot{\mathbb{X}}$ in \eqref{eq:pe_prac}. 
    However, we use a stronger baseline by using the ground truth PFs of the 6 MOO problems as the discrete set.
    %In practice, it is unclear how many points in a priori solution points to be considered. 
    We randomly choose (without replacement) $x\%$ of the ground truth PF for the PE, and call it {\it BL-$x$}. 
    For all the PE methods, we use Gaussian kernel and expected improvement as the acquisition function. 
    We run each method for 10 trials, and in each trial, the utility of the virtual DM (parameters of Chebyshev utility) are decided randomly, to test the methods for different oracle solutions. We compare the decrement in their regrets for up to 20 queries.
    % to the (virtual) DM. 
    
    The results are shown in Figure \ref{fig:pe}.
    We observe that for every MOO problem, GP-EPO surpasses the baseline methods that use up to 90 \% of the ground truth PO solutions. 
    Among the baseline results, the regret consistently decreases with increase in size of the discrete set $\ddot{\mathbb{X}}$.
    %increases for the BO in \eqref{eq:pe_prac}. 
    This decrement in regret is more prominent when the PE problem \eqref{eq:pe_ideal} is non-convex, i.e., when the MOO has a non-convex range $\mathcal{O}$ in the objective space. E.g., in the ZDT family, ZDT1 has a convex MOO problem, and the difference between regret curves for 70\%, 80\% and 90\% are not significant. Whereas, ZDT2 and ZDT3 have non-convex $\mathcal{O}$, which, we conjecture, makes the difference between regret curves significant.
    Note that, in the baseline approach, similar to the state-of-the-art for GP-based PE, the discrete set of PO solutions has to be obtained before Bayesian optimization \eqref{eq:pe_prac} can start for PE. However, in practice, it is unclear as to how many PO solutions would suffice to reach a desired level of regret. From the baseline results it is clear that if the discrete set $\ddot{\mathbb{X}}$ represents the PF at a coarser resolution, then the regret may not go lower than a certain level, since there may not be enough samples closer to the oracle solution. 
    Whereas, in GP-EPO, only two PO solutions are required a priory for the first query to start PE. We obtain the first PO solution by solving for $\mathbf{r}^{-1} = [1, \cdots, 1]^T / m$ and the second one by solving for a random $\mathbf{r}^{-1}\in \mathcal{S}^m$.  Since the Bayesian optimization \eqref{eq:pe_ours} is over a continuous domain, GP-EPO can probe the PF virtually at infinite resolution. Therefore, unlike the baseline approach, its regret keeps decreasing without saturating after few initial queries.

    \subsection{Evaluation of EPO Search for Multi-Task Learning on Real Data}
    \label{sec:expt_real}
    
    We demonstrate the efficacy of EPO Search for MTL in three applications from diverse domains.
We discuss application 1 in the following and the other two in Appendix \ref{sec:more_results}.

\subsubsection{Personalized Medicine and Pharmacogenomics.} 
\label{sec:exp_pharma}
    We consider three drug-related tasks from different stages of drug discovery and development, summarized in Table \ref{tab:3tasks}.
    We use data from multiple publicly available databases summarized in Table \ref{tab:dsdetail}.
    More details of these tasks and datasets used are in Appendix \ref{sec:drug_casestudy}.
    These tasks model the effects of drugs in hierarchically increasing levels of complexity.
    Drug-target (DT) prediction models the effect of drugs on specific genes (targets); drug response (DR) prediction models the effect of drugs on cancer cells (or a cancer patient with a given genomic profile); Drug side effect (DS) prediction models the side effects of drugs on patients.
    %for a population-level characterisation. 
    Hence, we expect DT to %is least likely to 
    benefit less from auxiliary signal of the other two tasks.
    Among the three tasks, DR is the most challenging because data for relatively fewer drugs are available and
    %As a result, it is most likely to 
    we expect DR to benefit the most from MTL.
    
    % of drug side effects.
    %We test the efficacy of EPO Search for training an MTL model designed to perform three tasks: Drug Target, Drug Response, and Drug Side-effect prediction. 
    % Drug Response Datasets: PDX \cite{Gao2015-ay}, GDSC \cite{Yang2013-vm}, CCLE \cite{Rees2016-su}. 
    % Drug Side-effect Datasets: SIDER \cite{Kuhn2016-vy}, OFFSIDE \cite{Tatonetti2012-ml}

        {\bf MTL Model}.
    A standard deep neural MTL architecture (DNN-MTL) is used, as 
    %We use the MTL network architecture proposed by \cite{Jiang2020DrugOrchestraJP}, 
    illustrated in Figure \ref{fig:mtl_drugorchestra}, where feature representations of each of the four entities, viz., drug, target gene, cancer cell, and disease,
    %we use the feature representations 
    are used. % as described in \cite{Jiang2020DrugOrchestraJP}.
    These features 
    are first passed through separate feed-forward neural networks (FNN) to get their corresponding embeddings, then the task-specific embedding pairs are concatenated and passed through  task-specific FNN predictors for the final outputs -- classifiers for DT and DS, and regressor for DR. The model is trained with binary cross entropy loss for DT and DS, and mean squared error loss for DR. 
    Each embedding FNN has one hidden layer of $256$ neurons, and each predictor FNN has two hidden layers, $64$ neurons followed by $16$, totaling $730,147$ parameters 
    %(network weights) 
    for the entire model. 
    %We used ReLU activation function for all the layers, except the final layers of the predictors; sigmoid activation is used for DT and DS classifiers, and an identity map for DR regressor.
    % \red{layers, losses}.
    Each sub-network is associated with its parameters denoted by $\theta_i^j$ where $j = e,c,r$ represent embedding, classification and regression respectively; $i = 1,2,3$ represent each of the tasks and $i=s$ represents the shared drug-related features.
    In this MTL-DNN model, the network parameters of drug embedding FNN are shared for all the tasks, making it a suitable testbed for MOO training. 
    %{\bf Features.}
    The dimensions of input feature and embeddings are given in Table \ref{tab:feature}. Additional details are in Appendix \ref{sec:drug_casestudy}.
    
    \begin{table}[h]
        \centering
        % \small
        \caption{Summary of Tasks in our MTL Setup}
        \label{tab:3tasks}
        \resizebox{0.9\textwidth}{!}{
        \begin{tabular}{lllll}
        % {cp{3.35cm}p{2cm}p{4.25cm}p{5cm}}
        \toprule
             & \textbf{Prediction Task} & \textbf{Problem} & \textbf{Inputs} & \textbf{Output(s)}  \\
        \midrule
        1 & Drug Target (DT) & Binary Classification & Drug $d$, Gene $g$ & 1: $g$ is a target of $d$, 0: otherwise \\
        2a & Drug Response (DR) & Regression & Drug $d$, Genomic Profile $p$ & Drug efficacy of $d$ on $p$\\
        2b & Drug Response (DR)& Ranking & Drug list, Genomic Profile $p$ & Top-$k$ most effective drugs for $p$ \\
        3 & Drug Side Effect (DS) & Binary  Classification & Drug $d$, Disease $s$ & 1: $s$ is a side effect of $d$, 0: otherwise \\
        \bottomrule
        \end{tabular}
        }
    %\vspace{-1cm}
    \end{table}
    \begin{minipage}{0.55\textwidth}
    \begin{table}[H]
    \centering
    \caption{Details of the datasets used}
    \label{tab:dsdetail}
    \resizebox{0.8\textwidth}{!}{
    \begin{tabular}{ccccc} \toprule
    \multirow{2}{*}{Task}                                                       & \multirow{2}{*}{Dataset} & \multirow{2}{*}{\begin{tabular}[c]{@{}c@{}}No. of \\ Drugs\end{tabular}} & \multicolumn{2}{c}{No. Sample pairs} \\ \cmidrule{4-5} 
                                                                                &                          &                                                                          & Training         & Test        \\ \midrule
    \multirow{3}{*}{\begin{tabular}[c]{@{}c@{}}Drug\\ Target\end{tabular}}      & STITCH                   & 16K                                                                      & 627596           & 263032            \\ %\cmidrule{2-5} 
                                                                                & DrugBank                 & 6K                                                                       & 158708           & 75820             \\ %\cmidrule{2-5} 
                                                                                & Repur                    & 4K                                                                       & 80080            & 39798             \\ \midrule
    \multirow{2}{*}{\begin{tabular}[c]{@{}c@{}}Drug\\ Response\end{tabular}}    & GDSC                     & 235                                                                      & 128004           & 62849             \\ %\cmidrule{2-5} 
                                                                                & CCLE                     & 483                                                                      & 212516           & 109529            \\ \midrule
    \multirow{2}{*}{\begin{tabular}[c]{@{}c@{}}Drug\\ Side-effect\end{tabular}} & SIDER                    & 1.4K                                                                     & 524395           & 259471            \\ %\cmidrule{2-5} 
                                                                                & OFFSIDE                  & 2.2K                                                                     & 550888           & 278807            \\ \bottomrule
    \end{tabular}
    }
    \\[2mm]
    \caption{Input and embedding dimensions}
    \label{tab:feature}
    \resizebox{0.8\textwidth}{!}{
    \begin{tabular}{ccccc}
    \toprule
               & Drug & Gene-Target & Cell-line & Disease \\ \midrule
    Features   & 300  & 800         & 300       & 300      \\ %\midrule
    Embeddings & 128  & 128         & 128       & 128      \\ \bottomrule
    \end{tabular}
    }
    \end{table}
    \vspace{0.1cm}
\end{minipage}\hspace{-0.5cm}~
    \begin{minipage}{0.4\textwidth}
        \begin{figure}[H]
        % \centering
        \includegraphics[page=2,trim=12.5cm 0 12.5cm 0,clip,width=\linewidth]{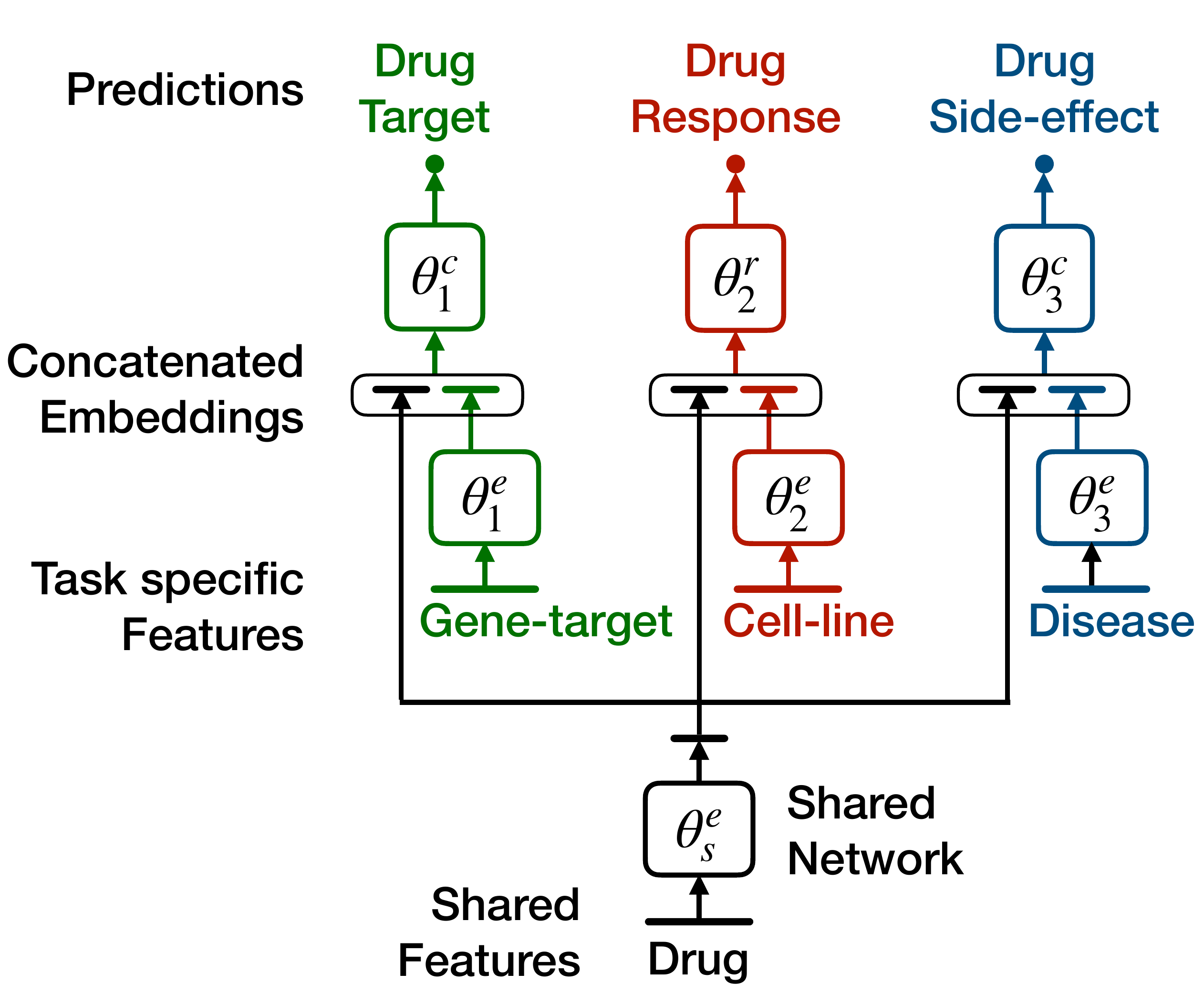}
        \caption{Illustration of MTL-DNN}
        \label{fig:mtl_drugorchestra}
        \end{figure}
        \vspace{0.1cm}
    \end{minipage}

    {\bf Experiment Setting.} %Evaluation Metrics.} 
    In each dataset, 1/3rd of the drugs are randomly chosen to create a held-out test set; the remaining 2/3rd of the drugs are used for training.
    The total number of samples used in train and test sets are given in Table \ref{tab:dsdetail}. 
    %This setting evaluates the model predictions on drugs unseen during training.
    For classification tasks, DT and DS,
    %We used different metrics for performance evaluation on these three tasks. For 
    %Drug Target and Side-effect prediction, we use classification metrics: 
    performance is measured using
    Area Under ROC curve (AUROC) and Area Under PR curve (AUPRC). 
    For DR, we use two metrics:
    %To evaluate the regression model's performance  on the entire test set, we use 
    Mean Squared Error (MSE) and 
    %To evaluate the model from the perspective of clinical use, we use 
    the ranking metric Normalized Discounted Cumulative Gain (NDCG@10) to judge how well the top $10$ most effective drugs are predicted for a cancer patient and thus evaluate the model from the perspective of clinical use. 
    %Conversations with clinicians indicated their  we chose 
    %for $k=5,10$.
    %{\color{red} Why are we not using PDX dataset? Is it because NDCG@5 or 10 won't make sense for PDX? No. of training data 822, no. of valid data 812.}
    %{\bf Baseline Algorithms.}
    
    We compare EPO Search with LS and CS.
    %other MOO-based MTL techniques Linear Scalarization (LS), Chebyshev Scalarization (CS).
    %{\color{red}In each of these methods, we set a relatively higher preference (100) for one task and lower preference (1) for the other two tasks. This is repeated three times, each time one of the tasks gets the higher preference.
     In each of these methods, we determine the trade-off between DR and DS %Drug Response and Drug Side-effect, 
     by having a higher priority (100) for one and lower priority for the other (1). DT plays a supporting role for both these tasks as it is the most specific task and we expect it to benefit the least from MTL. So, we keep its priority fixed at (10) for all scenarios.
     %, irrespective of the preference values for the other two tasks.
    We report the results for %evaluation metrics when 
    the model trained with a maximum priority for the corresponding task.
    In addition, we use
    Single Task Learning (STL) as a baseline, where each task-specific network of the MTL-DNN model is trained with data for each task independently. %We did not compare with PMTL, because, as noted in \cite{NIPS2019pmtl} and observed in our experiment in \S \ref{sec:exp_toy_moo}, it fails to scale for more than two objectives when the solution space is high-dimensional (in this case,
    % %the dimensionality of our model's DNN parameter space is 
    % more than 0.73 million). 
    % The training hyper parameters used are identical for all methods: Adam optimizer with learning rate $0.001$, mini-batch size of $256$ for each training dataset, and $10$K iterations. 
    For every priority setting, training for each method is repeated over 5 runs to randomize over model initialization and mini-batch formation, and the mean and standard deviations of the metrics are reported.
    %The seeds of randomization are different across the $5$ runs, but are the same across the MOO methods, so that, in a run, each method trains the same initial model using the same mini-batch formation. 
    We use paired $t-$test to determine statistical significance at 0.05 significance level.
    If a method is pair-wise better than {\it all} the other $3$ methods, then we mark the result with an asterisk. % in Table \ref{tab:drug3}.
    %This facilitates paired $t-$test for determining statistical significance, and avoids type II error.
    Results are shown in 
    Table \ref{tab:drug3}.

    %\cite{Jiang2020DrugOrchestraJP} empirically showed the three tasks are related and benefit by training simultaneously as opposed to training individually. 
    %However, they trained the MTL-DNN model with equal priority for each tasks. But in practice, depending on the stakeholder, the priority for each task may differ. Therefore, we train the MTL-DNN network for three different scenarios, where each task gets a higher priority as compared to the other two. 

    \begin{table}[h]
    \centering
    \caption{
    Results for Drug Target, Drug Response, and Drug Side-effect Prediction.
    %Results (with preferences of corresponding objectives) obtained for Drug Target (10:1:1), Drug Response (10:100:1), and Drug Side-effect (10:1:100) Prediction.
    Percentage values reported for AUROC, AUPRC and NDCG@10 (higher is better). Lower is better for MSE.
    %For all metrics, except for MSE, percentage value are reported. 
    Row-wise best result is in bold. Statistical significance (from paired $t$-test) is indicated by asterisk.}
    \label{tab:drug3}
    \resizebox{0.8\textwidth}{!}{
    \begin{tabular}{ccccccc}
    \toprule
    \multirow{2}{*}{Task}                                                                         & \multirow{2}{*}{Dataset}                   & \multirow{2}{*}{Metric}      &
    \multicolumn{4}{c}{Algorithms}\\ \cmidrule{4-7}
    &&&STL              & CS                        & LS               & EPO Search                          \\ \midrule
    \multirow{6}{*}{\begin{tabular}[c]{@{}c@{}}Drug \\ Target\\(DT)\end{tabular}}      & \multirow{2}{*}{STITCH}   & AUROC       & 94.90 $\pm$ 0.18 & 95.17 $\pm$ 0.40          & 95.08 $\pm$ 0.20 & \textbf{95.55 $\pm$ 0.27$^*$} \\ %\cmidrule{3-7} 
                                                                                 &                           & AUPRC       & 79.36 $\pm$ 0.46 & 79.14 $\pm$ 0.99          & 79.55 $\pm$ 0.66 & \textbf{79.86 $\pm$ 0.61}     \\ \cmidrule{2-7} 
                                                                                 & \multirow{2}{*}{DrugBank} & AUROC       & 91.95 $\pm$ 0.34 & 92.41 $\pm$ 0.24          & 91.88 $\pm$ 0.25 & \textbf{92.66 $\pm$ 0.21$^*$} \\ %\cmidrule{3-7} 
                                                                                 &                           & AUPRC       & 66.35 $\pm$ 0.60 & 66.42 $\pm$ 0.99          & 65.76 $\pm$ 0.67 & \textbf{67.52 $\pm$ 0.59$^*$} \\ \cmidrule{2-7} 
                                                                                 & \multirow{2}{*}{Repur}    & AUROC       & 90.80 $\pm$ 0.53 & \textbf{91.35 $\pm$ 0.18} & 90.45$\pm$ 0.29  & 91.16 $\pm$ 0.31     \\ %\cmidrule{3-7} 
                                                                                 &                           & AUPRC       & 64.54 $\pm$ 1.59 & 64.97 $\pm$ 0.15          & 64.34 $\pm$ 1.09 & \textbf{65.15 $\pm$ 1.13}     \\ \midrule
    \multirow{4}{*}{\begin{tabular}[c]{@{}c@{}}Drug \\ Response\\(DR)\end{tabular}}    & \multirow{2}{*}{GDSC}     & MSE         & 1.030 $\pm$ 0.02 & 1.012 $\pm$ 0.05          & 1.029 $\pm$ 0.02 & \textbf{0.965 $\pm$ 0.02}     \\ %\cmidrule{3-7} 
                                                                                 %&                           & \multirow{2}{*}{NDCG@\!} & \!5  & 49.45 $\pm$ 3.16 & 53.61 $\pm$ 2.20          & 49.57 $\pm$ 2.95 & \textbf{55.96 $\pm$ 3.04$^*$} \\ \cmidrule{4-8} 
                                                                                 &                           &              NDCG@10    %& \!10 
                                & 52.69 $\pm$ 2.15 & 54.55 $\pm$ 1.55          & 53.05 $\pm$ 2.11 & \textbf{56.62 $\pm$ 1.84$^*$} \\ \cmidrule{2-7} 
                                                                                 & \multirow{2}{*}{CCLE}     & MSE         & 0.854 $\pm$ 0.02 & 0.857 $\pm$ 0.04          & 0.862 $\pm$ 0.02 & \textbf{0.827 $\pm$ 0.04}     \\ %\cmidrule{3-7} 
                                                                                 %&                           & \multirow{2}{*}{NDCG@\!} & \!5  & 43.56 $\pm$ 2.25 & 44.39 $\pm$ 1.73          & 40.98 $\pm$ 1.40 & \textbf{46.52 $\pm$ 4.05}     \\ \cmidrule{4-8} 
                                                                                 &                           &                      NDCG@10     %& \!10 
                                & 48.70 $\pm$ 0.94 & 50.28 $\pm$ 1.31          & 47.80 $\pm$ 1.15 & \textbf{53.67 $\pm$ 0.90$^*$} \\ \midrule
    \multirow{4}{*}{\begin{tabular}[c]{@{}c@{}}Drug \\ Side-effect\\(DS)\end{tabular}} & \multirow{2}{*}{SIDER}    & AUROC       & 77.39 $\pm$ 0.20 & 77.37 $\pm$ 0.15          & 77.46 $\pm$ 0.14 & \textbf{78.60 $\pm$ 0.33$^*$} \\ %\cmidrule{3-7} 
                                                                                 &                           & AUPRC       & 39.02 $\pm$ 1.25 & 40.01 $\pm$ 1.33          & 39.58 $\pm$ 1.07 & \textbf{41.29 $\pm$ 1.25$^*$} \\ \cmidrule{2-7} 
                                                                                 & \multirow{2}{*}{OFFSIDE}  & AUROC       & 80.10 $\pm$ 0.19 & 80.53 $\pm$ 0.20 & 80.14 $\pm$ 0.14 & \textbf{81.23 $\pm$ 0.33$^*$} \\ %\cmidrule{3-7} 
                                                                                 &                           & AUPRC       & 61.43 $\pm$ 0.69 & 62.00 $\pm$ 0.45          & 61.23 $\pm$ 0.37 & \textbf{62.93 $\pm$ 0.58$^*$} \\ \bottomrule
    \end{tabular}
    }
    \end{table}

    {\bf Results.}
    First, we observe that in almost all cases, 
    %preference-specific MOO methods such as 
    CS and EPO Search perform better than STL, which demonstrates the advantages of EPO solutions for MTL, and DR, which is a more challenging problem, is most benefited.
    %by CS and EPO Search as compared to STL or LS. 
    Although LS uses gradient information from all the tasks, it fails to consistently perform better than STL % in our set up, where 
    as the priorities are disproportionate among the tasks. This can be attributed to the non-convexity of loss surface, for which LS gravitates towards an extreme PO solution, as illustrated in \S \ref{sec:exp_toy_moo}. STL can be considered as a MOO method that finds an extreme solution corresponding to one task only. We observe that in some cases, e.g. CCLE, LS performs even worse than STL. In these cases, the simple weighted sum strategy inhibits LS from reaching the PF as close as STL does. 
    Near an extreme solution the gradient directions are opposing and a fixed weight gradient combination reduces the magnitude of the search direction, especially when the gradient magnitude of a less preferred task is high. A reduced magnitude in the search direction decreases the magnitude of resulting network update. With fixed learning rate and number of iterations, % fixed across methods,
    update magnitude finally determines proximity to the PF.
    
    CS benefits from MTL by amortizing its usage of gradient information from different tasks over many iteration, but only after reaching the $\mathbf{r}^{-1}$ ray, as illustrated in \S \ref{sec:exp_toy_moo}. Before that it behaves like STL, since only the maximum relative objective \eqref{eq:chebychev} is minimized. 
    % This results in poorer  its performance, in most cases, is worse than EPO Search. 
    On the other hand, EPO Search adaptively combines the gradient information in every iteration and moves closer to the EPO solution of the training losses as compared to CS. This is reflected, through better performance with respect to the evaluation metrics, on the test data as well. 
    
    %For every experimental setup (a row in Table \ref{tab:drug3}), we compare each pair of MOO methods with a paired $t-test$ to determine
    %if the results are distinct with a 
    %statistical significance.
    %above $95\%$, i.e. $0.05$ p-value.
    %If a method is pair-wise better than all the other $3$ methods, then we report it with asterisks in Table \ref{tab:drug3}. 
    %Note that asterisks indicate that the marked method is significantly better than {\it all} the other 3 methods.
    For DS, EPO Search outperforms other methods in both datasets and both metrics.
    In DR, EPO search outperforms other methods in the clinically important metric, NDCG@10, in both datasets.
    In DT, EPO search outperforms other methods in the DrugBank dataset, on both metrics and in the STITCH dataset on AUROC.
    Overall, 
    %in 10 out 14 cases, EPO search is significantly better, and 
    in 13 out of 14 cases, EPO Search has the best average performance, and in 10 out of the 13 cases, the improvement is statistically significant.

    \subsubsection{Summary of Results on Real Data.}
    In Appendix \ref{sec:more_results} we evaluate EPO Search in two other applications.
    The first, from hydrometeorology, consists of predicting river flow at 8 sites in the Mississippi river network -- a problem with 8 regression tasks.
    The second, from e-commerce, consists of 2 classification problems, predicting the category of multiple fashion product images simultaneously.
    %In both these problems, EPO Search outperforms previous MOO-MTL methods PMTL, Linear scalarization as well as models trained on each task individually.
In all cases, our results demonstrate the advantages of MTL over learning for each task independently, and the superior performance of EPO-Search over competing MTL methods.
 
    % It is clear that our methods finds better Pareto optimal solutions for a given preference than the competing algorithms. Moreover, it spans the Pareto Front in objective (loss) space, which is also reflected in the accuracies of the corresponding tasks.

\section{Conclusion}
\label{sec:concl}

%\blue{One one hand, the CS method can model an EPO solution but cannot provide convergence guarantee. On the other hand, the descent based methods cannot model an EPO solution but can provide convergence guarantees. We develop a EPO Search method that gets the best of both worlds: models the EPO solution with convergence guarantees.}

In this paper we present new first-order iterative algorithms to find EPO solutions, 
% in both the unconstrained case and with box and equality constraints on the parameters.
for both unconstrained and constrained non-convex MOO problems.
EPO Search is designed for problems with high-dimensional solution spaces where it is computationally more efficient than popular evolutionary algorithms. 
From a random initialization, EPO Search converges to an EPO solution or, if an EPO solution does not exist, to the closest PO solution.
We prove its convergence and empirically demonstrate that our approach addresses the shortcomings of oscillations and premature stagnation in previous methods using the min-max formulation of \eqref{eq:chebychev}.
Similar to existing gradient descent methods, the convergence rate of EPO Search is sub-linear when moving from a random point to a Pareto stationary point. 
Interestingly, we show that the convergence rate, from any PO point to the EPO is linear under mild conditions.
The literature on CS has methods to obtain EPO solutions, but without convergence guarantees; while the literature on gradient-based MOO methods presents convergence guarantees but only to reach arbitrary PO solutions.
EPO Search offers both:  a robust iteration strategy to reach the desired EPO and with convergence guarantees.

A direct application of the improved MOO through EPO Search is seen in MTL.
Most previous MTL models use LS to combine task-specific priorities and losses.
While CS is more suitable for non-convex loss functions,
if the the min-max formulation of \eqref{eq:chebychev} is used, learning is dominated by the task with highest priority and the influence of other tasks is either low or absent.
In contrast, EPO Search uses a combination of gradients of all tasks in every iteration and can
escape the minima of low priority objectives through controlled ascent
which improves learning and makes it robust to initialization.
%where the use of gradients of only one objective per iteration (in CS) and imbalanced weight specifications lead to ineffective model training. 
% formulation for Multi-Task Learning (MTL) based on Multi Objective Optimization (MOO) that models conflicting trade-offs as well as task-specific preferences among the task objectives.
% This offers dual advantages of improved generalizability through MTL and task-specific focus based on the decision-maker's preferences.
% We develop new theory and a computationally efficient optimization method, EPO Search, to train large-scale highly parameterized models, such as deep networks, for MTL.
% EPO Search enables MOO-based MTL methods to (i) find PO solutions that satisfy  input  task priorities in highly parameterized deep networks, (ii) model  constraints in the network parameters (iii) provide theoretical guarantees of convergence.
The per-iteration complexity of EPO Search remains linear in the gradient dimensions (similar to the best previous methods that  neither use input priorities nor allow %regularization 
constraints on parameters) enabling its use for deep MTL networks.
%To the best of our knowledge, no previous MOO-based MTL method offers all these advantages.
%We prove that, under mild assumptions, it is guaranteed to converge.
We demonstrate the superior performance of EPO Search over competing approaches in MTL on synthetic and real datasets.
% EPO Search enables a MTL approach
% that can effectively utilize multiple data sources 
% %Predictive analytics across organizations can benefit from data sharing and MTL 
% to build more accurate and generalizable models and extract value from such connected data. 
% %Moreover, through our preference-specific formulation, each stakeholder can tailor their model for the tasks that they prioritize.
% In several tasks, e.g., pharmaceutical applications \citep{kulkov2021role},
% large quantities of labelled data may not be available for all tasks.
EPO Search allows us to prioritize training for tasks that are more challenging while leveraging the MTL framework which enables shared learning from other datasets and tasks.
%MTL offers a distinct modeling advantage for such cases, since data from additional tasks provide auxiliary signal to learn.
We observe this in our own experiments where drug response prediction, which has the least number of drugs for training, benefits from collective learning of drug side effects and drug targets.

When initialized on the PF,
EPO search %is robust to initialization and 
can systematically trace the PF to reach an EPO solution, with a theoretically guaranteed linear rate of convergence. 
%which can be used to generate multiple PO solutions to approximate the Pareto front; using this, we design PESA-EPO to generate a diverse set of Pareto optimal solutions to approximate the Pareto front.
This unique ability makes it a computationally efficient alternative in use cases that require traversing the PF from one EPO solution to another.
We investigate two such use cases in MCDM and develop new algorithms based on EPO Search.
First we develop PESA-EPO for a posteriori MCDM where multiple EPO solutions can be used to approximate a PF.
On benchmark datasets, PESA-EPO is faster than competing alternatives, without compromising on the approximation quality, because it does not require multiple optimizer calls and does not stop prematurely at PF discontinuities.
Second, we develop GP-EPO for PE in interactive MCDM where we
%. Here, we not only utilize the linear convergence rate of EPO Search to provide a more efficient procedure but also 
also address two limitations of previous GP-based PE methods by developing a new formulation. We leverage EPO Search to efficiently find PO solutions given a ray in $S^m$ in GP-EPO that samples in a lower-dimensional space $S^m$, instead of the higher-dimensional solution space which previous methods use; moreover, this obviates the need to explicitly model monotonicity constraints. Numerical experiments on benchmark problems confirm the advantages of GP-EPO in terms of improved regret with very few queries to the DM.

\subsubsection*{Future Directions.} 
The key idea of our convergence rate proof is to compose non-convex objective functions with a well behaved function that satisfies the PL inequality (the proportionality gauges in our case) whose minima lie close to the desired PO. 
This enables reaching the vicinity of the desired solution at a linear rate.
This idea may be utilized in other contexts, e.g., to solve non-convex SOO by (a) additional (convex or non-convex) objective functions and (b) a function satisfying PL inequality that, upon composition, models a solution in a convex neighbourhood of the global optimal of the original non-convex objective.
A limitation of EPO Search, that may be addressed in future work, is that the tracing procedure can approximate the entire PF starting from the extreme solutions only if
the set $\mathcal{O} = \{\mathbf{f}(\mathbf{x})\ | \ \mathbf{x} \in \mathbb{X}\}$ is connected, i.e., there are paths connecting the discontinuous segments of the PF. %$\mathcal{P}$.
Note that this is the case for disconnected PFs, in problems ZDT3, TNK and DTLZ7 in \S \ref{sec:appox_pf}.
%Although the tracing mechanism works for MOPs with disconnected , it has a limitation. All points in the path took by the EPO trace are feasible. Therefore 
%So that there 
If $\mathcal{
O}$ is not connected, e.g. in MOO problems of \cite{Wang2019}, then more initial seed points are required in each connected component of $\mathcal{O}$. 
%This can be addressed in future work by, e.g.,
%As a future work, one can devise 
For instance, techniques to add relevant non-extreme seed points by detecting discontinuities in the PF during tracing in EPO Search could be investigated.
%Data-driven techniques, e.g., based on 
Regularization on the network parameters, through our extension to handle constraints,
may be empirically evaluated for MTL in future work.
Finally, methods to adaptively find the best priorities for an MTL model may be explored in future work. A possibility is to adapt our method GP-EPO such that the DM's role is replaced by validation dataset performance to compare solutions.
\newpage
\bibliographystyle{informs2014} % outcomment this and next line in Case 1
\bibliography{references} % if more than one, comma separated
\newpage
\section*{Appendix}
\appendix
% \section{Notations}
% \label{sec:not}
\begin{table}[h]
\caption{Notations used in sections 
%\ref{sec:proportionality} and 
\S \ref{sec:epo_search} and \S \ref{sec:epo_mcdm_mtl}}
\label{tab:notation-table}
\resizebox{\textwidth}{!}{
\begin{tabular}{cl}
\toprule
\textbf{Notations} &
  \textbf{Description} \\ \midrule
$n$, and $m$ &
  number of variables in the solution space, and number of objectives \\[1mm] 
$\mathbf{f}$ &
  $\mathbb{R}^m-$valued objective function, or a vector in $\mathbb{R}^m$ \\[1mm]
$\mathcal{S}^m$ &
  $m-1$ dimensional simplex \\[1mm] 
$\mathbf{r} \in \mathcal{S}^m, \mathbf{r}^{-1}\in \mathbf{R}^m$ &
  preference vector, and its point-wise inverse \\[1mm] 
$\mathbf{x}^*, \mathbf{x}^*_\mathbf{r} \in \mathbb{R}^n$ &
  a Pareto Optimal solution, an Exact Pareto Optimal solution w.r.t $\mathbf{r}$ \\[1mm]
$\mathbf{x}^t, \mathbf{f}^t$ &
  solution and its objective vector at $t^{th}$ iteration \\[1mm] 
$\mathbf{b}^l, \mathbf{b}^u \in \mathbb{R}^n$ &
  lower and upper bounds (box constraint) on solution variable: $b^l_i \leq x_i \leq b^u_i$ \\[1mm] 
$\pi:\mathbb{R}^n\rightarrow \mathbb{R}^n$ &
  Projection function that brings $\mathbf{x}$ inside the box constraints element-wise \\[1mm] 
$p, q$ &
  number of inequality and equality constraints \\[1mm] 
$\mathbf{g, h}$ &
  $\mathbb{R}^p-$valued inequality and $\mathbb{R}^q$ valued equality constraints \\[1mm] 
$p_a, q$ &
  number of active inequality and equality constraints \\[1mm] 
$\mathrm{F,  G , H}$ &
  Jacobians of objectives, active inequality and and equality constraints \\[1mm] 
$\mathbb{X} \subset \mathbb{R}^n, \mathcal{O} \subset \mathbb{R}^m$ &
  Set of feasible solutions, and range of  $\mathbb{R}^m-$valued objective function $\mathbf{f}$ \\[1mm] 
$\partial \mathbb{X}, \partial \mathcal{O}$ &
  Boundaries of domain and range of $\mathbf{f}$ respectively \\[1mm] 
$\mathrm{Int}(\mathbb{X}),  \mathrm{Int}(\mathcal{O})$ &
  Interior of domain and range of $\mathbf{f}$ respectively \\[1mm] 
$\mathcal{P}, \mathcal{P}_\mathbf{r}$ &
  Set of Pareto optimal solutions, and Exact Pareto Optimal solutions w.r.t $\mathbf{r}$ \\[1mm] 
$\mathcal{T}_\mathbb{X}(\mathbf{x}), \mathcal{F}_\mathbb{X}(\mathbf{x}),  \mathcal{D}_\mathbb{X}^\mathbf{f}(\mathbf{x})$ &
  Tangent plane/cone, set of feasible directions, descent directions at $\mathbf{x}$ \\[1mm] 
$\mathbf{d}, \mathbf{d}_{nd}\in \mathcal{F}_\mathbb{X}(\mathbf{x})$ &
  a general search direction, non-dominating search direction \\[1mm] 
$\bm{\beta}\in \mathbb{R}^m, \bm{\rho}\in\mathbb{R}^{p_a}, \bm{\gamma}\in\mathbb{R}^{q_a}$ &
  coefficients for gradients of objectives, active inequality and equality constraints \\[1mm] 
$\omega_\mathbf{r}(\mathbf{f}) = \omega(\mathbf{f}, \mathbf{r}^{-1})$ &
  A measure of proportionality of $\mathbf{f}$ w.r.t. $\mathbf{r}^{-1}$ \\[1mm] 
$\mathbf{a}(\mathbf{f, r})$ or simply $\mathbf{a}$ &
  Anchor direction at point $\mathbf{f}\in \mathbb{R}^m$ w.r.t. preference $\mathbf{r}$ \\[1mm] 
$\lambda$ &
  maximum relative value of objectives: $\max_j r_jf_j$ \\[1mm] 
$\mathrm{J}^*$ &
  index set of objectives with maximum relative value \\[1mm] 
$\mathcal{A}^\mathbf{r}_\mathbf{f}, \mathcal{V}_{\preccurlyeq \mathbf{f}}$ &
  set of attainable objective vectors dominated by $\lambda^t\mathbf{r}^{-1}$ and $\mathbf{f}$ \\[1mm]
$\mathcal{M}^\mathbf{r}_\mathbf{f}$ &
  set of attainable objective vectors with measure of proportionality $ < \omega_\mathbf{r}(\mathbf{f})$ \\[1mm]
$\mathrm{R}^0, \mathrm{R}^{0ijk}$ &
  discrete sets of $m$ preference vectors at start, and after 3 recursion in PESA \\ 
  $\mu, \kappa, \alpha$ & mean, kernel and acquisition functions of a Gassian Process \\
 $\psi_\mathbf{f}$ & function that maps a point in the Simplex to the PF \\
 $\mathcal{D}_t$ & Pairwise comparison data constructed from the $t$ queries to the DM. \\
 $\ddot{\mathbb{X}}_{\mathcal{D}_t}$ & The discrete set of solutions present in the dataset $\mathcal{D}_t$ \\
  \bottomrule
\end{tabular}
}
\end{table}
% \vfill

\section{Proportionality of Vectors and Balancing Direction}
\subsection{Proportionality Gauge from KL Divergence}\label{sec:mu_kld}
% 	In \cite{Mahapatra2020}, the authors formulated a
	One possible approach to measure the proportionality between $\mathbf{f}$ and $\mathbf{r}^{-1}$ is through KL divergence between the normalized vectors of  $\mathbf{f} \odot \mathbf{r}$ and $\mathbf{1}=[1, \overset{m}{\cdots}, 1]$, i.e. the uniform distribution:
	\begin{align}
		\omega\!\left(\mathbf{f}, \mathbf{r}^{-1}\right) \ =\ \sum_{j=1}^{m} \frac{ f_{j}r_{j}}{\|\mathbf{f} \odot \mathbf{r}\|_{1}} \log\left(\frac{mf_{j}r_{j}}{\|\mathbf{f} \odot \mathbf{r}\|_{1}}\right) \ =\ \mathrm{KL}\!\left( \left. \overline{\mathbf{f} \odot \mathbf{r}} \; \right| \overline{\mathbf{1}}\,\right),\label{eq:non-uniformity}
	\end{align}
	where $\overline{\mathbf{v}}$ is the $\ell_{1}$ normalization of a vector $\mathbf{v}$. This $\omega$ satisfies both conditions \ref{num:prop+} and \ref{num:prop_smth_mntnc} of a proportionality gauge. Figure \ref{fig:kld_simplex} shows the corresponding $\omega_\mathbf{r}$ in case of $3$ objectives and a particular weight vector. 
	% We can obtain the anchoring direction of this $\omega$ by computing its gradient w.r.t $\mathbf{f}$. 
    % However, 
	%for the purpose of analysis, 
	% it is sufficient to analyze a simplified $\mathbf{a}$ that is aligned with, but has a different magnitude compared to, $\nabla_{\!\mathbf{f}}\omega_\mathbf{r}$. 
% 	In fact, the anchor direction proposed in \cite{Mahapatra2020}, given by
    Its anchor direction $\mathbf{a} = \nabla_\mathbf{f}\omega_\mathbf{r}$ is scale invariant to $\mathbf{r}$:
	\begin{align}
	a_{j} = \frac{r_{j}}{ \|\mathbf{f} \odot \mathbf{r}\|_{1}}\left( \log\left(\frac{f_{j}r_{j} / \|\mathbf{f} \odot \mathbf{r}\|_{1}}{1/m}\right) - \omega\!\left(\mathbf{f}, \mathbf{r}^{-1}\right) \right), \quad j \in [m]. \label{eq:kl_anchor}
	\end{align}
	% which is scaled as 
% 	is such that 
	% $\mathbf{a} = \|\mathbf{f} \odot \mathbf{r}\|_{1} \nabla_{\!\mathbf{f}}\omega_\mathbf{r}$.
% 	\begin{claim}
% 	The anchoring direction with elements
% 	\begin{align}
% 	a_{j} = r_{j}\left( \log\left(\frac{f_{j}r_{j} / \|\mathbf{f} \odot \mathbf{r}\|_{1}}{1/m}\right) - \omega\!\left(\mathbf{f}, \mathbf{r}^{-1}\right) \right), \quad j \in [m], \label{eq:kl_anchor}
% 	\end{align}
% 	is such that $\mathbf{a} = \|\mathbf{f} \odot \mathbf{r}\|_{1} \nabla_{\!\mathbf{f}}\omega_\mathbf{r}$.
% 	\end{claim}

	Notice that, unless all $f_{j}r_{j}$ are equal, the anchor elements $a_{j}$ are non-negative for some objectives and negative for the rest. As a result, if we move against the search direction $\mathbf{d}$ in Theorem \ref{th:Fd=a}, we will be descending for the objectives with $\mathbf{d}^{T}\,\nabla_{\!\mathbf{x}}f_{j} = s a_{j} \geq 0$ and ascending for the other objectives, since $s > 0$. This is further clarified by analyzing the relation between $\mathbf{f}$ and $\mathbf{a}$.
	
% 	\begin{claim}[\cite{Mahapatra2020}]\label{th:kl_af}
    \begin{restatable}{claim}{klaf}\label{th:kl_af}
	The anchor direction $\mathbf{a}$ in \eqref{eq:kl_anchor} is always orthogonal to the objective vector $\mathbf{f}$: $\mathbf{a}^{T} \, \mathbf{f}= 0$.
    \end{restatable}

	% However, iterating based on the anchoring direction in \eqref{eq:kl_anchor} has a limitation. 
    However, this anchoring direction \eqref{eq:kl_anchor}  does not move the objective vectors to the $\mathbf{r}^{-1}$ ray in the shortest possible path. The shortest path between $\mathbf{f}^{t}$ and the $\mathbf{r}^{-1}$ ray should lie on the hyperplane containing both these vectors. So, in order for $\mathbf{f}^{t+1} \approx \mathbf{f}^{t}-\eta \mathbf{a}$ to be on the shortest path, a necessary condition is $\mathbf{a}$ should lie in the span of $\mathbf{f}^{t}$ and $\mathbf{r}^{-1}$. This does not happen in anchoring direction of \eqref{eq:kl_anchor}
    % for every $(\mathbf{f}, \mathbf{r}^{-1})$, 
    as shown in the Figure \ref{fig:kld_simplex} and \ref{fig:topview}.
    % with a counter example. 
    The curved trajectory 
    % of objective vectors 
    deviates from the $\mathrm{span}(\left\{\mathbf{r}^{-1}, \mathbf{f}^{0}\right\})$.
    % span of $\left\{\mathbf{r}^{-1}, \mathbf{f}^{0}\right\}$ 
    % and follows a curved path. 

    \subsection{Comparison of Proportionality Gauges} \label{sec:compare_prop}
    Among the three options discussed above, we use CSZ inequality and Lagrange's identity based proportionality gauges as their anchoring directions satisfy the PL inequality (see Lemma \ref{th:pl_ineq}), which we leverage in \S \ref{sec:conv_rate} to prove linear convergence in Theorem \ref{th:conv_rate}.  
    It is non-trivial to design a scaling factor $s$ for the KL divergence based anchor \eqref{eq:kl_anchor} such that $\mathbf{a}=s\nabla_\mathbf{f}\omega_\mathbf{r}$ satisfies the PL inequality in Lemma \eqref{th:pl_ineq}.
    % , which is satisfied by the other two anchor directions, and facilitating linear convergence of the balance mode (see \S \ref{sec:conv_rate}). 
    Therefore, we do not use this in our development.
    We use Lagrange identity based anchoring direction when the initialization is not on the PF, in order to reach the $\mathbf{r}^{-1}$ ray through the shortest path in lesser number of iterations as compared to the CSZ inequality based anchor. However, when the initialization is on the PF, in order to escape the local PO solution, we use the CSZ inequality based anchoring direction.
    % of our algorithm. 
    % \red{When the initialisation is random, the anchor direction from  Lagrange's identity, given by equation \eqref{eq:lgrn_anchor}, should be used since it can yield the trajectory of shortest path. However, if the initialization $\mathbf{x}^{0}$ is itself a local Pareto optimal solution and $\mathbf{f}^{0}$ lies on the Pareto front then $\mathbf{a}$ from KL divergence, given by equation \eqref{eq:kl_anchor}, should be used to escape local non-EPO solutions which the Lagrange anchor may not.} 
    Table \ref{tab:pg_comp} summarizes their properties.

    \begin{table}[h]
    \centering
    \caption{Comparison among the proportionality gauges}
    \label{tab:pg_comp}
    \begin{tabular}{ccccccc}
    \multicolumn{3}{c}{Proportionality Gauge}                               & \multirow{2}{*}{$\ \mathbf{a} \perp?\ $} & \multirow{2}{*}{$\ \mathbf{a} \in \mathrm{span}(\{\mathbf{f}, \mathbf{r}^{-1}\}) ?\ $} & \multirow{2}{*}{\begin{tabular}[c]{@{}c@{}}Trajectory \\ to $\mathbf{r}^{-1}$ ray\end{tabular}} & \multirow{2}{*}{\begin{tabular}[c]{@{}c@{}}PL inequality \\ in Lemma \eqref{th:pl_ineq} \end{tabular}} \\ \cmidrule{1-3}
    based on              & $\omega$                  & $a$                    &                                                                                              &                                          &                                                                                        &                                                                                                 \\ \toprule
    CSZ Inequality     & \eqref{eq:cs_prop}        & \eqref{eq:cs_anchor}    & $\mathbf{f}$                             & Yes                                                                                    & Curved                                                      & Satisfied                                                               \\
    Lagrange Identity & \eqref{eq:lgrn_prop}      & \eqref{eq:lgrn_anchor} & $\mathbf{r}^{-1}$                        & Yes                                                                                    & Straight & Satisfied                                  \\% \midrule
      KL Divergence     & \eqref{eq:non-uniformity} & \eqref{eq:kl_anchor}   & $\mathbf{f}$                             & No                                                                                     & Curved                                         & Not Satisfied                                                                                           \\ %\midrule
    \bottomrule                                   
    \end{tabular}
    \end{table}
    
\section{Constrained Multi-Objective Optimization}\label{sec:constrained_moo}
Our approach for Constrained MOO is similar to that of \cite{Fliege2016}, where the descent based method (discussed in \S \ref{sec:grad_moo}) is used to handle constraints. But their method cannot obtain EPO solutions specific to an $\mathbf{r}$ like ours, which is facilitated by both ascent and descent. 
% \deb{Include this somewhere: Constraints in MOO have also been discussed in, e.g., \cite{Fliege2016}.}
The  constrained domain or \textit{Feasible Solution Set} is defined as
	\begin{align}
		\mathbb{X} :=  \left\lbrace \mathbf{x}\in \mathbb{R}^n \; \middle| \; 
		\begin{tabular}{@{}l@{}}
		$b^l_i \leq x_i \leq b^u_i  \quad\forall i \in [n],$ \\ %[1mm]
		\ \ $g_k(\mathbf{x}) \leq 0 \quad \, \forall k \in [p]$,  \\%[]
		\ \ $h_{k}(\mathbf{x})  \,= 0 \quad \, \forall k \in [q]$
		\end{tabular}
		\right\rbrace,
	\end{align}
	where $\mathbf{b}^l$ \& $\mathbf{b}^u$ are domain boundaries for each variable, $\mathbf{g}: \mathbb{R}^n \rightarrow \mathbb{R}^p$ are $p$ differentiable inequality constraints, and ${\mathbf{h}:\mathbb{R}^{n}\rightarrow \mathbb{R}^q}$ are $q$ differentiable equality constraints. 
    % The constraint functions are assumed to be differentiable.
    % and satisfy certain independence conditions (details in \S \ref{sec:constrained_moo}). 

    % \subsection{Incorporating Constraints.}
    % \label{sec:constraint}
	We check for the infeasibility of boundary, equality and inequality constraints in each iteration. If $x^{t}$ violates any of the boundary constraints, we project element-wise to $\pi(\mathbf{x}^t)$,
	to constrain them to remain within the bounds,
	where
	\begin{align}\label{eq:bd_project}
	\pi(x^{t}_{i}) = \begin{cases}
    	b^{l}_{i}, \quad \text{if } x^{t}_{i} < b^{l}_{i},\\
    	x^{t}_{i}, \quad \text{if } x^{t}_{i} \in [b^{l}_{i}, b^{u}_{i}],\\
    	b^{u}_{i}, \quad \text{if } x^{t}_{i} > b^{u}_{i}
    	\end{cases} \quad \text{for all } i \in [n].
	\end{align}
	
	Let the number of active inequality constraints be $p_a$, making $p_a+q$ total active constraints, since the equality constraints are always active. Without loss of generality, let the active inequality constraints be $g_{k}$ for $k= 1, \cdots, p_{a}$. Then the cone of first order feasible directions at $\mathbf{x}^{t}$ against which we can move to obtain $\mathbf{x}^{t+1}\in \mathbb{X}$ is given by
	\begin{align}
	\mathcal{F}_\mathbb{X}(\mathbf{x}^{t}) = \left\lbrace \mathbf{d} \in \mathbb{R}^{n} \ \middle| \ \mathbf{d}^{T}\,\nabla_{\!\mathbf{x}} g^{t}_{k} \geq 0 \ \ \forall \, k \in [p_{a}], \text{ and } \mathbf{d}^{T}\,\nabla_{\!\mathbf{x}} h^{t}_{k} = 0 \ \ \forall \, k \in [q] 
	\right\rbrace, \label{eq:feasibledir}
	\end{align}
	where $g^{t}_{k} = g_{k}(\mathbf{x}^{t})$ and $h^{t}_{k} = h_{k}(\mathbf{x}^{t})$. When there is no active constraint it is the same as the tangent plane: $\mathcal{F}_\mathbb{X}(\mathbf{x}^{t}) = \mathcal{T}_\mathbb{X}(\mathbf{x}^{t}) = \mathbb{R}^n$. Note, the tangent plane is the set of directions that keeps the next iterate in the feasible region:
    \begin{align}
        \mathcal{T}_\mathbb{X}(\mathbf{x}^t) := \left\{ \mathbf{d} \in \mathbb{R}^n \,\middle|\, \exists \eta > 0 \text{ s.t. } \mathbf{x}^t - \eta \mathbf{d} = \mathbf{x}^{t+1} \in \mathbb{X}\right\}.
    \end{align}
    When there are active constraints, the tangent plane becomes a tangent cone as certain directions would lead $\mathbf{x}^{t+1}$ out of the feasible region $\mathbb{X}$. However, the first order feasible directions $\mathcal{F}_\mathbb{X}(\mathbf{x}^t)$ may not be equal to $\mathcal{T}_\mathbb{X}(\mathbf{x}^t)$ always. 
	The tangent cone $\mathcal{T}_\mathbb{X}(\mathbf{x}^t)$ is unique and depends on the geometrical property of $\mathbb{X}$ at $\mathbf{x}^t$. 
    But, the $\mathcal{F}_\mathbb{X}(\mathbf{x}^t)$ cone is not unique and depends on the algebraic specification of the constraints (see \citep{NoceWrig06}[Ch 12.2]) through the functions $g_k$s and $h_k$s. E.g., although the constraints $g(\mathbf{x}) = x_i\leq 0$ and $g'(\mathbf{x})=x_i^3 \leq 0$ are algebraically different (at $\mathbf{x} = \mathbf{0}$, $\frac{\partial g}{\partial x_i} = 1$ whereas $\frac{\partial g'}{\partial x_i} = 0$), they are geometrically the same constraints.
    At a boundary point $\mathbf{x}^t\partial\mathbb{X}$, $\mathcal{F}_\mathbb{X}(\mathbf{x}^t)=\mathcal{T}_\mathbb{X}(\mathbf{x}^t)$ if the $p_a+q$ gradients of the active constraint function are linearly independent \citep{NoceWrig06}. This is called as \textit{Linear Independence Constraint Qualification} (LICQ). 
	Therefore, to make $\mathcal{F}_\mathbb{X}(\mathbf{x}^{t})$ same as the tangent cone $\mathcal{T}_\mathbb{X}(\mathbf{x}^{t})$, and render the conditions in \eqref{eq:feasibledir} useful, we assume the LICQ to be satisfied at every $\mathbf{x}\in\partial\mathbb{X}$.
 % \textit{Linear Independence Constraint Qualification} (LICQ) to be satisfied at $\mathbf{x}^t$. In other words, we assume that the gradients of active constraints are linearly independent.
	% Finally, with $\mathbf{d}=\mathrm{F}^T\bm{\beta}$ and the LICQ assumption, we constrain the coefficients $\bm{\beta}$ in the quadratic cost \eqref{eq:qp_cost} with $\bm{\beta}^T\mathrm{F}\, \nabla g_k  \geq 0$ for all $k\in [p_a]$  and $\bm{\beta}^T\mathrm{F}\, \nabla h_k  = 0$ for all $k\in [q]$ to obtain the search direction.

\subsection{Constrained EPO Search for Random Initialization}\label{sec:cepo_rand}
% When there are active constraints at $\mathbf{x}^t$, i.e. $\bm{\beta}$ is not freely decided to minimize the quadratic cost \eqref{eq:epo_qp}, we further constrain it to ensure convergence (see Lemma \ref{th:df_j*>0} and Theorem \ref{th:admissible_set}). Let $\mathrm{J}^*$ be the index set of maximum relative objective values:
% % 	$\mathrm{J}^* = \left\{j\ \middle|\ j = \arg\max_{j'\in[m]} \, f_{j'} r_{j'}\right\}$.
% 	\begin{align} \label{eq:J*}
% 	    \mathrm{J}^* = \left\{j\ \middle|\ j = \arg\max_{j'\in[m]} \, f_{j'} r_{j'}\right\}.
% 	\end{align}
% 	The final $m$ dimensional QP problem to be solved at $\mathbf{x}^t$ is given as 
We modify the QP \eqref{eq:qp_x0_random_unconst} into
	\begin{subequations}\label{eq:qp_x0_random}
	\begin{align}
	\bm{\beta}^{*} = \argmin_{\bm{\beta}\in \mathbb{R}^m\, |\, \|\bm{\beta}\|_1 \leq 1} \ \ & \|\mathrm{F}\mathrm{F}^{T}\bm{\beta} - \mathbf{a}\|^{2} \\
	\text{s.t.} \quad & \bm{\beta}^T\mathrm{F} \, \nabla g_k  \geq 0, \quad \text{for all}\ k \in [p_a], \label{eq:qp_ineq} \\
	& \bm{\beta}^T\mathrm{F} \, \nabla h_k = 0, \quad \text{for all}\ k \in [q],\label{eq:qp_eq} \\
	& \bm{\beta}^T\mathrm{F} \, \nabla f_{j} \geq 0, \quad \text{for all}\ j \in \mathrm{J} = \begin{cases}
            \ \mathrm{J}^*\ \quad \text{in balance mode} \\
            [m] \quad \text{in descent mode}
        \end{cases}, \label{eq:qp_J*}
	\end{align}
	\end{subequations}
where $\mathrm{J}^*$ is as defined in \eqref{eq:qp_J*_unconst}. This is similar to the gradient projection strategy for constrained SOO \citep{Luenberger2008}.
The search direction $\mathbf{d}_{nd}=\mathrm{F}^T\bm{\beta}$, can be considered as the projection of the  search direction obtained from QP \eqref{eq:qp_x0_random_unconst} onto the set $\mathcal{F}_\mathbb{X}(\mathbf{x}^t)$ \eqref{eq:feasibledir}. However in practice, at a boundary point $\mathbf{x}^t\in\partial \mathbb{X}$, the projected gradient may not guarantee to move the iterate to $\mathrm{Int}(\mathbb{X})$, the interior of $\mathbb{X}$, unless the step size is infinitesimal. Therefore, in our implementation we modify \eqref{eq:qp_ineq} to $\bm{\beta}^T\mathrm{F} \, \nabla g_k  \geq \gamma_k$ for a $\gamma_k > 0$ whenever the inequality constraint is violated $g_k(\mathbf{x}^t) > 0$. Similarly, we modify \eqref{eq:qp_eq} to $\bm{\beta}^T\mathrm{F} \, \nabla h_{k'}  \geq \gamma_{k'}$ if $h_{k'}(\mathbf{x}^t) > 0$ and $\bm{\beta}^T\mathrm{F} \, \nabla h_{k'}  \leq \gamma_{k'}$ if $h_{k'}(\mathbf{x}^t) < 0$ for a $\gamma_{k'}>0$. We increase the values of $\gamma$ if $\mathbf{x}^{t+1} \notin \mathbb{X}$, and increase further as $\gamma \leftarrow 2 * \gamma$,  until the iterate becomes feasible.

% We gradually increase the $\gamma$ value in the subsequent iterations until a feasible solution is found. 

Similar to the unconstrained MOO case, this QP also satisfies the following two Lemmas by construction:
    \begin{restatable}{lemma}{dndbalconst}\label{th:bal_const}
        If $\mathbf{x}^t$ is a non-PO regular point of the differentiable vector function $\mathbf{f}$ in a balance mode, i.e. $\omega(\mathbf{f}^t, \mathbf{r}^{-1}) > \epsilon_1$, then the non-dominating direction obtained from QP \eqref{eq:qp_x0_random} makes
        \begin{enumerate}
            \item \hspace{-1mm}non-negative \hspace{-0.3mm}angles \hspace{-0.3mm}with \hspace{-0.3mm}the \hspace{-0.3mm}gradients \hspace{-0.3mm}of \hspace{-0.3mm}maximum \hspace{-0.3mm}relative \hspace{-0.3mm}objectives: \hspace{-0.3mm}${\mathbf{d}_{nd}^T\nabla_{\!\mathbf{x}^t}\!f_j\!\geq\!0\ \forall\!j\!\in\!\mathrm{J}^*}$ \hspace{-1mm} \eqref{eq:qp_J*_unconst},
            \item \hspace{-1mm}a positive angle with the balancing anchor direction \eqref{eq:lgrn_anchor} in the objective space: $\mathbf{a}^T\mathrm{F}\mathbf{d}_{nd} > 0$.
        \end{enumerate}
	\end{restatable}
    \begin{restatable}{lemma}{dnddesconst}\label{th:des_const}
        If $\mathbf{x}^t$ is a non-PO regular point of the differentiable vector function $\mathbf{f}$ in a descent mode, i.e. $\omega(\mathbf{f}^t, \mathbf{r}^{-1}) \leq \epsilon_1$, 
	    then the non-dominating direction obtained from QP \eqref{eq:qp_x0_random} makes a non-negative angle with every gradient,  $\mathbf{d}_{nd}^T\, \nabla_{\!\mathbf{x}}f_j^t \geq 0\ \forall j\in [m]$, and a positive angle with at least one gradient.
	\end{restatable}

    % Note, In practice, when an inequality constraint is active, i.e., $g_k(\mathbf{x}^t) > 0$, 

 % \begin{subequations}\label{eq:qp_x0_random_unconst}
 %    \begin{align}
 %        \bm{\beta}^{*} = \argmin_{\|\bm{\beta}\|_1 \leq 1} \ \ &\|\mathrm{F}\mathrm{F}^{T}\bm{\beta} - \mathbf{a}\|^{2} \label{eq:qp_x0_random_unconst_cost}\\
 %        \text{s.t.} \quad & \bm{\beta}^T\mathrm{F} \, \nabla f_{j} \geq 0 \quad \forall\ j \in \mathrm{J} = \begin{cases}
 %            \ \mathrm{J}^*\ \quad \text{in balance mode} \\
 %            [m] \quad \text{in descent mode}
 %        \end{cases}, \label{eq:J_index} \\
 %        \text{where}\quad & \mathrm{J}^* = \left\{j \in [m]\ \middle|\ j = \arg\max_{j'\in[m]} \, f_{j'} r_{j'}\right\}  \label{eq:qp_J*_unconst}
 %    \end{align}
 %    \end{subequations}

    % Note that when $\mathbf{x}^t$ is in the interior of $\mathbb{X}$, there will be no constraints in the QP problem \eqref{eq:qp_x0_random} apart from $\|\bm{\beta}\| \leq 1$. The constraint in \eqref{eq:qp_J*} ensures that the search direction reduces the objective values of $\mathrm{J}^*$, irrespective of $\mathbf{x}^t\in \mathrm{Int}(\mathbb{X})$ (interior) or $\mathbf{x}^t\in\partial\mathbb{X}$ (boundary).

\subsection{Constrained EPO Search for Tracing the Pareto Front from $\mathbf{x}^0 \in \mathcal{P}$}\label{sec:cepo_trace}
We modify the QP \eqref{eq:qp_x0_po_unconst} into 
% For a constrained MOO problem, the QP problem solved in each iteration of the modified EPO search is given by 
    \begin{subequations}\label{eq:qp_x0_po}
	\begin{align}
	\bm{\beta}^{*} = \arg\min_{\bm{\beta} \in [-1,1]^{m}} \ \ & \|\mathrm{F}\mathrm{F}^{T}\bm{\beta} - \mathbf{a}\|^{2} \\
	\text{s.t.} \quad & \bm{\beta}^T\mathrm{F} \, \nabla g_k  \geq 0, \quad \text{for all}\ k \in [p_a],\\
	& \bm{\beta}^T\mathrm{F} \, \nabla h_k = 0, \quad \text{for all}\ k \in [q], \\
    &\mathbb{1}_{des}\bm{\beta}^T\mathrm{F}\mathrm{F}^T \mathbf{a}^{bal} \geq 0, \label{eq:des_pos_const} \\
        \text{and} \quad  &\mathbb{1}_{des}\bm{\beta}^T\mathrm{F} \, \nabla f_{j} \geq 0 \quad \forall\ j \in [m],
	\end{align}
	\end{subequations}
    where $\mathbf{a}^{bal}$ is the CSZ inequality based anchor direction \eqref{eq:cs_anchor}.
	Note that we have excluded the constraint \eqref{eq:qp_J*} associated to objectives of $\mathrm{J}^*$. Because, if the objective vector $\mathbf{f}^t$ is (approximately) on the Pareto front, then some objectives with highest relative value ($r_jf^t_j$) may require a further increase in their value to move towards an EPO solution.

 \subsubsection*{Pareto Criticality for constrained MOO:}\label{sec:cepo_convergence_x0_po}

    At a point $\mathbf{x}^t$, the set of descent directions is given by
    % \begin{align}
    $
        \mathcal{D}_\mathbb{X}^\mathbf{f}(\mathbf{x}^t) = \left\{\mathbf{d}\in\mathcal{T}_\mathbb{X}(\mathbf{x}^t) \,\middle|\, \mathbf{d^T}\nabla_{\!\mathbf{x}^t}f_j \geq 0, \, \forall \, j \in [m]\right\}.
    $
    % \end{align}
	At a local Pareto optimal point $\mathbf{x}^*$ there does not exist any non-zero feasible descent direction, i.e. $\mathcal{D}_\mathbb{X}^\mathbf{f}(\mathbf{x}^*) \,\bigcap\,\mathcal{F}_\mathbb{X}(\mathbf{x}^*) = \{\mathbf{0}\}$. Note that, if there are no active constraints at $\mathbf{x}^*$, $\mathcal{D}_\mathbb{X}^\mathbf{f}(\mathbf{x}^*) = \{\mathbf{0}\}$. A necessary condition to check if a point $\mathbf{x}^t$ is Pareto optimal, i.e.,  $\mathcal{D}_\mathbb{X}^\mathbf{f}(\mathbf{x}^t) \,\bigcap\,\mathcal{F}_\mathbb{X}(\mathbf{x}^t) = \{\mathbf{0}\}$, is given by Pareto Criticality:
% 	\begin{subequations}\label{eq:criticality}
	\begin{align}\label{eq:criticality}
	    \text{there exists a } \bm{\beta} \in \mathcal{S}^m,\ \bm{\rho} \in \mathbb{R}_+^{p_a}, \text{ and } \bm{\gamma} \in \mathbb{R}^{q},
	    \ \text{s.t. }\ \mathrm{F}^T\bm{\beta} + \mathrm{G}^T \bm{\rho} + \mathrm{H}^T\bm{\gamma} = \mathbf{0},
	\end{align}
% 	\end{subequations}
	where $\mathrm{G}$ and $\mathrm{H}$ are the Jacobians of inequality and equality constraints. 
	This is formulated by extending the KKT conditions to a multi-objective setup  (see \citep{NonlinearMOO}[Ch 4]).
	When there are no active constraints at an optimal point $\mathbf{x}^*\in \mathcal{P}$, the Pareto criticality condition in \eqref{eq:criticality} reduces to $\mathrm{F}^T\bm{\beta} = \mathbf{0}$ for some $\bm{\beta} \in \mathcal{S}^m$.

    \subsubsection*{Penetration Assumption:}\label{sec:assum_pen}
    We introduce the penetration assumption to guarantee non-convergence at a non-EPO point when there are active constraints, i.e, ${\mathbf{x}^*\in\partial\mathbb{X}}$ and $\mathcal{F}_\mathbb{X}(\mathbf{x}^*)\subsetneq \mathbb{R}^n$.
    We assume %we make an assumption: 
	there exists an $\eta_0 > 0$ such that ${\mathbf{f}^* + \eta \overrightarrow{\mathbf{f}^*} \in \mathrm{Int}(\mathcal{O})}$ for all $\eta \in [0, \eta_0]$, where $\mathbf{f}^* = \mathbf{f}(\mathbf{x}^*)$. In other words, an infinitesimal step along the direction of objective vector $\overrightarrow{\mathbf{f}^*}$ starting from $\mathbf{f}^*\in\partial\mathcal{O}$ will take it to the interior of $\mathcal{O}$. We call this as $\overrightarrow{\mathbf{f}^*}$ \textit{penetrates} $\mathcal{O}$. %With 
%\newpage
	\begin{figure}[h]
        \begin{minipage}{0.5\textwidth}
        \centering
		\def\svgwidth{\linewidth}	% if the document class is double column, use \columnwidth
		\begingroup\makeatletter\def\f@size{9}\check@mathfonts	% f@size decides the fontsize of "math" text in Figure
		\def\maketag@@@#1{\hbox{\m@th\large\normalfont#1}}%
		% \fontsize{9pt}{11pt}\selectfont % decides the overall fontsize
		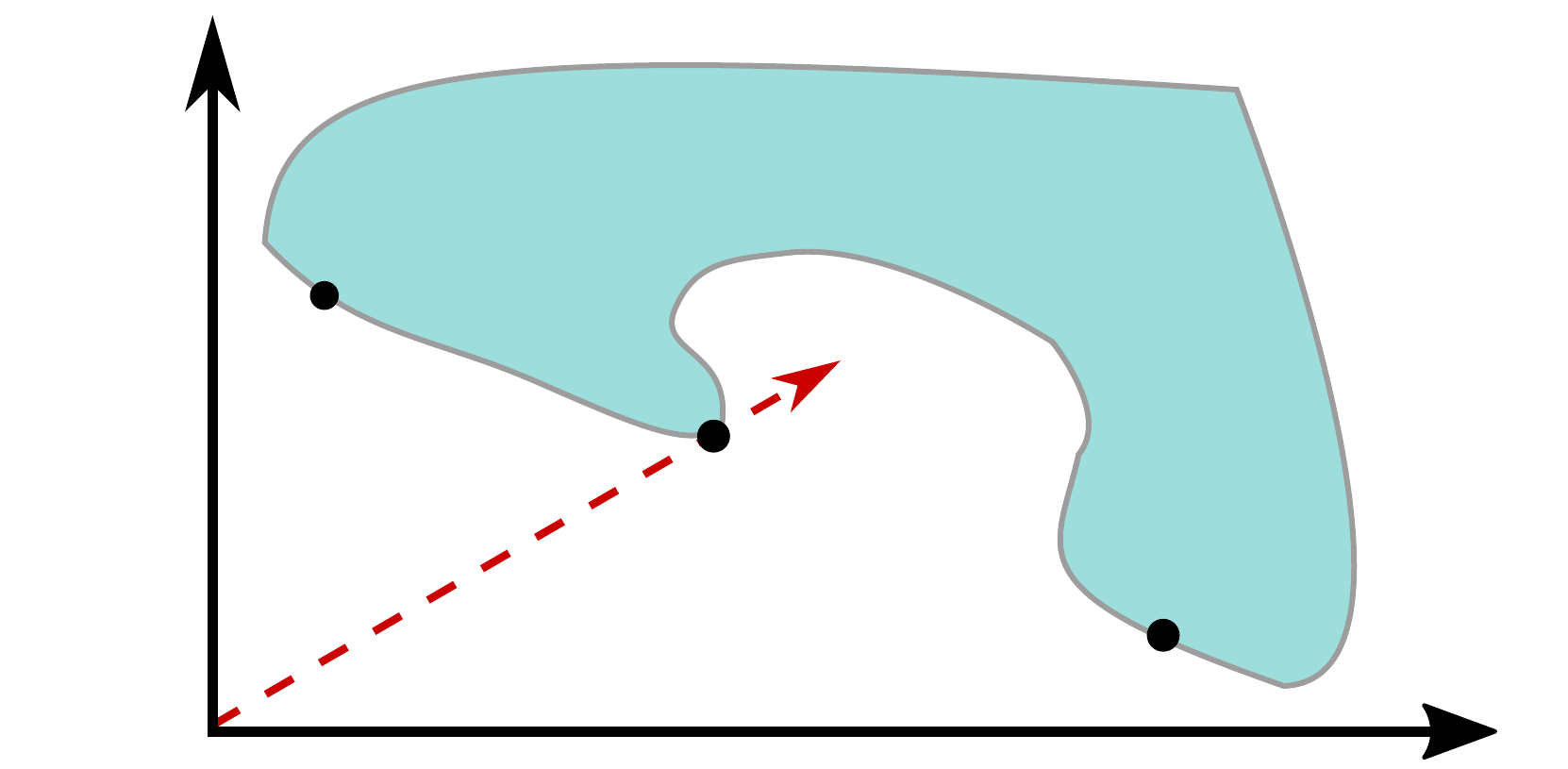
		\endgroup\vspace{-.2cm}
        \end{minipage}
		\begin{minipage}{0.49\textwidth}
		\caption{Penetration assumption is violated at $\mathbf{f}^2$. Therefore EPO search for tracing in Algorithm~\ref{alg:epo_search_x0_po} will not be able to trace from $\mathbf{f}^1$ to $\mathbf{f}^4$; the iterations will converge (stop prematurely) at $\mathbf{f}^2$. But it can trace from $\mathbf{f}^3$ to $\mathbf{f}^4$. Note that the set of points on boundary $\partial \mathcal{O}$ from $\mathbf{f}^2$ to $\mathbf{f}^3$ is not locally Pareto Optimal.
		\label{fig:penetration}
			% 		\vspace{-.2cm}
		}   
		\end{minipage}
	\end{figure}
     \begin{restatable}{theorem}{dndnc}		\label{th:d_nd}
		Let $\mathbf{x}^* \in \mathcal{P}$ such that, if 
		$\mathbf{x}^* \in \mathrm{Int}(\mathbb{X})$ then it is a regular Pareto optimal solution, and if $\mathbf{x}^* \in \partial\mathbb{X}$ then it is a regular point of $\mathbf{f}$ and $\overrightarrow{\mathbf{f}^*}$ penetrates $\mathcal{O}$. Then, at $\mathbf{x}^*$, the non\nobreakdash-dominating direction $\mathbf{d}_{nd} = \mathrm{F}^T\bm{\beta}^*$ found by the QP \eqref{eq:qp_x0_po} with Cauchy-Schwarz anchor \eqref{eq:cs_anchor} is $\mathbf{0} \in \mathbb{R}^n$ if and only if $\mathbf{x}^* \in \mathcal{P}_r$. 
    \end{restatable}
We consider the penetration assumption to be mild because, when the range set $\mathcal{O}$ is $m$ dimensional and its boundary $\partial\mathcal{O}$ is $m-1$ dimensional, almost all points in $\partial\mathcal{O}$ that violate the penetration assumption are not Pareto optimal, not even locally. 
Figure~\ref{fig:penetration} illustrates a scenario where it is violated.
If a point $\hat{\mathbf{f}}\in\partial\mathcal{O}$ violates penetration along with the points in any of its $m-1$ dimensional (relative) open neighbourhoods in $\partial\mathcal{O}$, then $-\overrightarrow{\hat{\mathbf{f}}}$ penetrates $\mathcal{O}$, and thus $\hat{\mathbf{f}}$ is dominated by $\hat{\mathbf{f}}-\eta\overrightarrow{\hat{\mathbf{f}}}$ for some small $\eta$.
	The assumption of $m-1$ dimensional boundary is fairly general since it is similar to the assumption of regular Pareto optimal in the unconstrained case.

\section{Proofs of Lemmas and Theorems}
\subsection{Proportionality Gauge}
\label{sec:proofs}
    \ancr*
    \begin{proof}
     
    We first prove \eqref{eq:mu+<mu} by considering $\omega\!\left(\mathbf{f}(\mathbf{x}), \, \mathbf{r}^{-1} \right)$ as a function of $\mathbf{x}$, $\omega_\mathbf{r}^\mathbf{f}(\mathbf{x})$. Taylor's expansion of this function can be written with the Peano's form of remainder as
    \begin{align}
        \omega_\mathbf{r}^\mathbf{f}(\mathbf{x} - \eta \mathbf{d}) = \omega_\mathbf{r}^\mathbf{f}(\mathbf{x}) - \eta\, \frac{\partial \omega_\mathbf{r}^\mathbf{f}}{\partial \mathbf{x}}\, \mathbf{d} + o(\eta),
    \end{align}
    where $\frac{\partial \omega_\mathbf{r}^\mathbf{f}}{\partial \mathbf{x}}$ is the transpose of gradient $\nabla_\mathbf{x}\omega_\mathbf{r}^\mathbf{f}$, and the asymptotic notation little-$o(\eta)$ represents a function that approaches $0$ faster than $\eta$. In particular, for every $\epsilon > 0$, there exists an $\eta_0>0$ such that
	\begin{align}\label{eq:o_prop}
	    \left|\frac{o(\eta)}{\eta}\right| < \epsilon, \text{ for } |\eta| < \eta_0.
	\end{align}
    Applying chain rule of differentiation on $\omega$, we get
    \begin{align*}
        \frac{\partial \omega_\mathbf{r}^\mathbf{f}}{\partial \mathbf{x}} = \frac{\partial \omega_\mathbf{r}}{\partial \mathbf{f}} \, \frac{\partial \mathbf{f}}{\partial \mathbf{x}} = \mathbf{a}^T \mathrm{F}.
    \end{align*}
    We know that $\mathbf{a}^T\mathrm{F}\mathbf{d}$ is non-negative from the statement of the Lemma~\ref{th:anchor}. Therefore, when positive, we treat $\mathbf{a}^T\mathrm{F}\mathbf{d}$ as $\epsilon$, and use the property of $o(\eta)$ as mentioned in \eqref{eq:o_prop} to conclude there exists a step size $\eta_0>0$ such that 
	\begin{align}
	    \left| \frac{o(\eta)}{\eta} \right| < \mathbf{a}^T\mathrm{F}\mathbf{d}, \forall \eta \in [0, \eta_0],
	\end{align}
	and hence $\omega_\mathbf{r}^\mathbf{f}(\mathbf{x} - \eta \mathbf{d}) \leq \omega_\mathbf{r}^\mathbf{f}(\mathbf{x})$; equality holds when $\mathbf{a}^T\mathrm{F}\mathbf{d}=0$. That proves \eqref{eq:mu+<mu}.
    
    The above strategy can be applied to prove there exists a step size $\eta_0>0$ such that 
    \begin{align}
        f_{j^*}(\mathbf{x} - \eta \mathbf{d}) &\leq f_{j^*}(\mathbf{x}),  \quad \forall \eta \in [0, \eta_0], \\
        \text{where} \quad j^* &= \argmax_{j \in [m]} \mathbf{d}^T\,\nabla_{\!\mathbf{x}}\!f_{j}\ .
    \end{align}
    This is true because of the assumption in Lemma~\ref{th:anchor} that $\mathbf{d}^T\nabla_{\!\mathbf{x}}\!f_{j^*} > 0$. And that proves \eqref{eq:f+n>f}. \hfill   
    \end{proof}
    
    % \fagtra*

    In the following Lemma \ref{th:fa>0>rinva}, we prove a property of scale invariant balancing anchor direction, that will be used in the proves of Theorem \ref{th:Fd=a}. % and Lemma \ref{th:df_j*>0}.
    \begin{restatable}{lemma}{fagtra}\label{th:fa>0>rinva}
        If $\omega$ is such that the anchoring direction is scale invariant to $\mathbf{r}$, i.e., $ \overrightarrow{\mathbf{a}}(\mathbf{f}, s\mathbf{r}) = \overrightarrow{\mathbf{a}}(\mathbf{f}, \mathbf{r})$ for all $s>0$, where $\overrightarrow{\mathbf{a}} = \frac{\mathbf{a}}{\|\mathbf{a}\|}$ then 
        \begin{align}
            \sum_{j=1}^m f_j a_j = \langle\mathbf{f}, \mathbf{a}\rangle \geq 0 \geq \langle\mathbf{r}^{-1}, \mathbf{a}\rangle = \sum_{j=1}^m \frac{a_j}{r_j}.
        \end{align}
    \end{restatable}
    \begin{proof}
     
    We use the second property of $\omega_r$ stated in \ref{num:mtct}, i.e. $\omega_r(\mathbf{r}^{-1} + \lambda (\mathbf{f} - \mathbf{r}^{-1}))$ increases monotonically with $\lambda \geq 0$. In other words, $\omega_\mathbf{r}'(\lambda) = \frac{d\omega_\mathbf{r}}{d\lambda} \geq 0$, for $\lambda\geq0$. At $\lambda=1$, the chain rule reveals
    \begin{align}
        \omega_\mathbf{r}'(1) = \langle \mathbf{f} - \mathbf{r}^{-1}, \mathbf{a}(\mathbf{f}, \mathbf{r}^{-1}) \rangle \ &\geq\ 0 \\
        \implies \langle \mathbf{f}, \overrightarrow{\mathbf{a}}(\mathbf{f}, \mathbf{r}^{-1}) \rangle \ &\geq\ \langle \mathbf{r}^{-1}, \overrightarrow{\mathbf{a}}(\mathbf{f}, \mathbf{r}^{-1}) \rangle. \label{eq:fa>rinva}
    \end{align}
    We apply the scale invariance property of the anchor direction to the preference vector in \eqref{eq:fa>rinva}:
    \begin{align}
        \langle \mathbf{f}, \overrightarrow{\mathbf{a}}(\mathbf{f}, s\mathbf{r}^{-1}) \rangle \ &\geq \ \langle s\mathbf{r}^{-1}, \overrightarrow{\mathbf{a}}(\mathbf{f}, s\mathbf{r}^{-1}) \rangle \\
        \implies \langle \mathbf{f}, \overrightarrow{\mathbf{a}}(\mathbf{f}, \mathbf{r}^{-1}) \rangle \ &\geq \ s\langle \mathbf{r}^{-1}, \overrightarrow{\mathbf{a}}(\mathbf{f}, \mathbf{r}^{-1}) \rangle \quad \forall s > 0. \label{eq:fa>srinva}
    \end{align}
    Applying $\lim_{s\rightarrow 0}$ to \eqref{eq:fa>srinva}, we get $\langle \mathbf{f}, \overrightarrow{\mathbf{a}} \rangle \geq 0$. Therefore, applying $\lim_{s\rightarrow \infty}$ to \eqref{eq:fa>srinva}, we get $\langle \mathbf{r}^{-1}, \overrightarrow{\mathbf{a}}\rangle \leq 0$, i.e. must not be positive. \hfill   
    \end{proof}

    \fdeqa*
    \begin{proof}
     
    From Lemma \ref{th:fa>0>rinva}, we know $\langle\mathbf{f}, \mathbf{a}\rangle \geq 0$. Therefore, $a_{j^+} > 0$ for at least one ${j^+} \in [m]$, because $f_j \geq 0$ for all $j\in[m]$. As $\mathrm{F}\mathbf{d}=s\mathbf{a}$ for some $s>0$, we can write
    \begin{enumerate}
        \item $0 < s a_{j^+} = \mathbf{d}^{T}\,\nabla_{\!\mathbf{x}}\!f_{j^+} < \max_{j} \{\mathbf{d}^{T}\,\nabla_{\!\mathbf{x}}\!f_{j}\}$, and
        \item $\mathbf{a}^T\mathrm{F}\mathbf{d} = s \|\mathbf{a}\|^2_2 > 0$
    \end{enumerate}
    So Lemma \ref{th:anchor} is applicable, and that concludes the proof. \hfill   
    \end{proof}

    \csaf*
    \begin{proof}
     
    The proof is apparent from the formula of the anchoring direction \eqref{eq:cs_anchor}.
    \hfill    
    \end{proof}
    
    \lgnaf*
    \begin{proof}
     
    The proof is apparent from the formula of the anchoring direction \eqref{eq:lgrn_anchor}.
    \hfill   
    \end{proof}
    
    \klaf*
    \begin{proof}
     
    We expand $c \mathbf{a}^{T} \, \mathbf{f}$, where $c=\|\mathbf{r} \odot \mathbf{f}\|_1$, as 
    \begin{align*}
        c\mathbf{a}^T\mathbf{f} &= \sum_{j=1}^m r_j \left(\log\left(m\hat{f}_j\right) - \omega_\mathbf{r}(\mathbf{f})\right) \times f_j \\
        &=\sum_{j=1}^{m} r_j f_j \left( \left(1-\hat{f}_j\right)\log\left(\hat{f}_j\right) - \sum_{j' \neq j} \hat{f}_{j'}\log\left(\hat{f}_{j'}\right)\right),
    \end{align*}
    where $\hat{f}_j = f_{j}r_{j} / \|\mathbf{f} \odot \mathbf{r}\|_{1}$.
    We use the fact that $\sum_{j=1}^m \hat{f}_j = 1$, and further expand as
    \begin{align*}
        c\mathbf{a}^T\mathbf{f} &=\sum_{j=1}^{m} r_j f_j \left( \sum_{j' \neq j} \hat{f}_{j'}\log\left(\hat{f}_j\right) - \sum_{j' \neq j} \hat{f}_{j'}\log\left(\hat{f}_{j'}\right)\right) \\
        &= \sum_{j=1}^{m} r_j f_j \left(\sum_{j' \neq j} \hat{f}_{j'}\log\left(\frac{\hat{f}_j}{\hat{f}_{j'}}\right) \right).
    \end{align*}
    In the inner summation we can now add the term for $j=j'$ as $\log\left(\frac{\hat{f}_j}{\hat{f}_{j'}}\right) = \log(1) = 0$ and write the above expression as 
    \begin{align*}
        c\mathbf{a}^T\mathbf{f} &= \frac{1}{\sum_{j=1}^m r_jf_j}  \sum_{j=1}^{m} \sum_{j'=1}^m r_j f_jr_{j'}f_{j'}\log\left(\frac{\hat{f}_j}{\hat{f}_{j'}}\right)
    \end{align*}
    The double summation in the numerator can be written as the inner product of a symmetric and a skew-symmetric matrix which is equal to $0$.
    \hfill   
    \end{proof}

    \subsection{EPO Search for Unconstrained MOO}\label{sec:proof_epo_unconst}
    % \dfjstr*
    \dndbal*
    \begin{proof}
      We denote the subset of all possible directions in the tangent space $\mathcal{T}_\mathcal{O}(\mathbf{f})$ of an objective vector $\mathbf{f} \in \mathcal{O} \subset \mathbb{R}^m$, which is constrained due to $\ell_1$ restrictions on the coefficients $\bm{\beta}$, as 
    \begin{align}\label{eq:chfpm}
         \mathcal{CH}_\mathbf{f}^{\pm} := \left\{\mathrm{F}\mathrm{F}^T\bm{\beta} \,\middle|\, \|\bm{\beta}\|_1 \leq 1 \right\} = \left\{\mathrm{F}\mathbf{d} \,\middle|\, \mathbf{d} \in \mathcal{CH}_\mathbf{x}^{\pm} \right\},
    \end{align}
    where $\mathcal{CH}_\mathbf{x}^{\pm}$ is defined in \eqref{eq:ch_pmgrads}. 
    We split the proof for two scenarios: 1) $\mathbf{a} \in \mathcal{CH}_\mathbf{f}^{\pm}$ and 2) $\mathbf{a} \notin \mathcal{CH}_\mathbf{f}^{\pm}$.

    In the first scenario, the the minimum value $0$ is achieved for the objective \eqref{eq:qp_x0_random_unconst_cost} with ${\bm{\beta}^* = (\mathrm{F}\mathrm{F}^T)^{-1}\mathbf{a}}$, $\mathbf{d}_{nd} = \mathrm{F}^T\bm{\beta}^*$, and hence 
    \begin{align}\label{eq:Fdnd=a}
        \mathrm{F}\mathbf{d}_{nd} = \mathbf{a}.
    \end{align}
    We can invert $\mathrm{F}\mathrm{F}^T$ due to the regularity assumption.
    Note that $a_j > 0$ for all $j\in \mathrm{J}^*$ when $\mathbf{f}^t$ and $\mathbf{r}^{-1}$ are not proportional. This can be deduced from its formula in \eqref{eq:lgrn_anchor} as follows
    \begin{enumerate}
        \item $\mathrm{sign}(a_j) = \mathrm{sign}(r_ja_j) \ \forall j \in \mathrm{J}^*$
        \hfill $\because r_j > 0$ for all $j\in [m]$.
        \item $\max_{j\in [m]} a_j > 0$.
        \hfill $\because$ If not true, then Claim \ref{th:lgrnar=0} is contradicted for any $\mathbf{r} \in \mathbb{R}_{++}^m$. 
        \item $\max_{j\in [m]} r_j a_j = \max_{j\in [m]} f_j^tr_j - \frac{\langle \mathbf{f}^t, \mathbf{r}^{-1}\rangle}{\|\mathbf{r}^{-1}\|^{2}} = r_{j^*} a_{j^*}$, for any $j^* \in \mathrm{J}^*$. 
        \hfill $\because$ definition of $\mathbf{a}$ in \eqref{eq:lgrn_anchor}.
        \item $\therefore$ 1, 2 and 3 $\implies \mathrm{sign}(a_{j}) > 0 $ for all $j\in \mathrm{J}^*$.
        % \item Since $r_j > 0\ \forall j\in[m]$, $a_j > 0$ for at least one $j\in [m]$. \hfill $\because$ Claim \ref{th:lgrnar=0}
    \end{enumerate}
    As a result, the constraint \eqref{eq:J_index} is inactive when $\mathbf{a} \in \mathcal{CH}_\mathbf{f}^{\pm}$, because $\bm{\beta}^T\mathrm{F} \, \nabla_{\!\mathbf{x}^t}\!f_{j} = a_j$.
    Now, with $\mathrm{F}\mathbf{d}_{nd} = \mathbf{a}$ we can invoke Theorem \ref{th:Fd=a}, and that proves Lemma \ref{th:bal} for the first scenario.

    In the second scenario, i.e., $\mathbf{a} \notin \mathcal{CH}_\mathbf{f}^{\pm}$, the $\mathbf{d}_{nd}$ satisfies the first property, i.e., $\mathbf{d}_{nd}^T \, \nabla_{\!\mathbf{x}^t}\! f_{j} \geq 0$ for all $j \in \mathbf{J}^*$, due to the constraint \eqref{eq:J_index}. We prove the second property, i.e., $\mathbf{a}^T\mathrm{F}\mathbf{d}_{nd} > 0$, by contradiction. Let $\bm{\beta}^*$ be the optimum of the QP \eqref{eq:qp_x0_random_unconst}, such that $\mathbf{a}^T\mathrm{F}\mathbf{d}_{nd} = \mathbf{a}^T\mathrm{F}\mathrm{F}^T\bm{\beta}^* \leq 0$. Then the optimal cost value of \eqref{eq:qp_x0_random_unconst_cost} can be written as 
    \begin{align}
        \|\mathrm{F}\mathrm{F}^T\bm{\beta}^*  - \mathbf{a}\|_2 = \|\mathrm{F}\mathrm{F}^T\bm{\beta}^* \|_2^2 + \|\mathbf{a}\|_2^2 - 2 \mathbf{a}^T\mathrm{F}\mathrm{F}^T\bm{\beta}^* > \|\mathbf{a}\|_2^2.
    \end{align}
    However, this is a contradiction due to the following counterexample that yields a lesser value of the cost than $\bm{\beta}^*$. The $\hat{\bm{\beta}} = \frac{(\mathrm{F}\mathrm{F}^T)^{-1}\mathbf{a}}{\|(\mathrm{F}\mathrm{F}^T)^{-1}\mathbf{a}\|_1}$ satisfies the constraints \eqref{eq:J_index} as proven for the previous scenario, and the cost value of this coefficient is $\left(1- \frac{1}{\|(\mathrm{F}\mathrm{F}^T)^{-1}\mathbf{a}\|_1}\right)^2 \|\mathbf{a}\|_2^2 < \|\mathbf{a}\|_2^2$.     
    $\therefore \mathbf{a}^T\mathrm{F}\mathbf{d}_{nd} > 0.$ 

    That concludes the proof.
    \hfill   
    \end{proof}
    
    \dnddes*
    \begin{proof}
      In the descent mode the anchor direction is $\mathbf{a}= \mathbf{f}^t$.
    Similar to the proof of Lemma \ref{th:bal}, we split this proof into two scenarios: 1) $\mathbf{a} \in \mathcal{CH}_\mathbf{f}^{\pm}$ and 2) $\mathbf{a} \notin \mathcal{CH}_\mathbf{f}^{\pm}$, where $\mathcal{CH}_\mathbf{f}^{\pm}$ is defined in \eqref{eq:chfpm}.

    For the first scenario, the proof is similar to that of Lemma \ref{th:bal}, where $\bm{\beta}^* = (\mathrm{F}\mathrm{F}^T)^{-1}\mathbf{a}$. And the constraint \eqref{eq:J_index} is redundant, since $\mathbf{d}_{nd}^T \, \nabla_{\!\mathbf{x}^t}\! f_{j} = f^t_j \geq 0$ for all $j\in [m]$. There exist at least one $j\in [m]$ such that $f_j^t > 0$, and therefore $\mathbf{d}_{nd}$ makes positive angle with the corresponding gradient. The point $\mathbf{0} \in \mathbb{R}^m$ is a utopia point in our formulation and cannot be attained.

    For the second scenario the first property is true due to the constraint \eqref{eq:J_index}. The second property of making a positive angle with at least one gradient can be proven by contradiction similar to the proof Lemma \ref{th:bal}. 
    \hfill    %   
    \end{proof}

	\textbf{Beyond Regularity Assumption:} Lemmas  \ref{th:bal} and \ref{th:des} can be true even without the full rank assumption of $\mathrm{F}$. In particular, when $\mathbf{x}^t$ is in the interior of the domain, \eqref{eq:Fdnd=a} can be satisfied if the anchor direction $\mathbf{a}\in \mathrm{Col}(\mathrm{F}\mathrm{F}^T)$, the column space of $\mathrm{F}\mathrm{F}^T$.

	% \vlta*
	% \proof{Proof.} 
 %    When $\omega(\mathbf{f}^t, \mathbf{r}^{-1}) > 0$,
	% \begin{align*} 
	%     \mathbf{f} \in \mathcal{V}_{\preccurlyeq \mathbf{f}^t}\
	%     \implies\ \mathbf{f} \preccurlyeq \mathbf{f}^t\ \implies\ \mathbf{r} \odot \mathbf{f} &\preccurlyeq \mathbf{r} \odot \mathbf{f}\ \implies\ \mathbf{r} \odot \mathbf{f} \preccurlyeq \lambda^t \mathbf{1}\ \implies\ \mathbf{f} \preccurlyeq \widecheck{\mathbf{f}}^t \\
	%     \therefore\ \mathbf{f} &\in  \mathcal{A}^\mathbf{r}_{\mathbf{f}^t}\ .
	% \end{align*}
	% When $\omega(\mathbf{f}^t, \mathbf{r}^{-1}) = 0$, $\widecheck{\mathbf{f}}^t = \mathbf{f}^t$. Therefore  $\mathcal{V}_{\preccurlyeq \mathbf{f}^t} = \mathcal{A}^\mathbf{r}_{\mathbf{f}^t}$.
	% \hfill   
	% \endproof
 
    \adms*
    \begin{proof}
     
    % \blue{
    % This proof holds true for both unconstrained and constrained MOO problem. 
    We prove for unconstrained MOO using Lemmas \ref{th:bal} and \ref{th:des} for the QP \eqref{eq:qp_x0_random_unconst}. However, the same proof holds true for constrained MOO by using Lemmas \ref{th:bal_const} and \ref{th:des_const} for the QP \eqref{eq:qp_x0_random}.
    % }
    
	We divide the proof into descent mode and balance mode. 

	From Lemma \ref{th:des}, in the descent mode, i.e. $\mathbf{a} = \mathbf{f}^t$, The QP in \eqref{eq:qp_x0_random_unconst}, produces a search direction that makes non-negative angle with each gradient, i.e. $\mathbf{d}_{nd}^T\nabla_\mathbf{x}f_j \geq 0$ for all $j\in[m]$. As a result, by applying the Taylor's expansion with Peano form of remainder to each $f_j$ along with the property of little-o notation, one can deduce that for every $j\in[m]$ there exists a step size $\eta_{0j} > 0$ such that $f_j(\mathbf{x}^t - \eta \mathbf{d}_{nd}) \leq f_j(\mathbf{x}^t)$ for all $\eta \in [0, \eta_{0j}]$. If we choose $\eta_0 = \min_j \{\eta_{0j}\}$, then for all $\eta \in [0, \eta_0]$, we have
	\begin{flalign}
	    & & \mathbf{f}(\mathbf{x}^t - \eta \mathbf{d}_{nd}) &\preccurlyeq \mathbf{f}(\mathbf{x}^t) &\nonumber\\
	    & & \implies \mathbf{f}(\mathbf{x}^{t+1}) &\in \mathcal{V}_{\preccurlyeq \mathbf{f}^t} &\text{($\because$ definition of $\mathcal{V}_{\preccurlyeq \mathbf{f}^t}$ in \eqref{eq:V<ft})} \\ %\label{eq:l+inV}  \\
        & & \implies \mathbf{f}(\mathbf{x}^{t+1}) &\in \mathcal{A}^\mathbf{r}_{\mathbf{f}^t} & \text{($\because$ definition of $\mathcal{A}^\mathbf{r}_{\mathbf{f}^t}$ in \eqref{eq:Arft})} 
	\end{flalign}
	% From Lemma~\ref{th:V<A} and \eqref{eq:l+inV} we conclude that $\mathbf{f}^{t+1} \in \mathcal{A}^\mathbf{r}_{\mathbf{f}^t}$ for all $\eta\in [0, \eta_0]$.

	Next, we consider the balance mode. Let $\mathrm{J}^+ = \{j\;|\; \mathbf{d}_{nd}^T\nabla_{\!\mathbf{x}}f_j \ge 0\}$ be the index set for descending objectives and $\mathrm{J}^- = [m]-\mathrm{J}$ for ascending ones. So, there exists an $\eta_{0j}>0$ for all $j\in \mathrm{J}^+$ such that
	\begin{align*}
	    f_j(\mathbf{x}^t - \eta \mathbf{d}_{nd})=  f_j^{t+1} \le f_j^t
	\end{align*}
	for all $\eta \in [0, \eta_{0j}]$. Let $\eta_0^{\mathrm{J}^+} = \min_{j\in\mathrm{J}^+} \{\eta_{0j}\}$, and $\tilde{\eta}_0 = \min \{\eta_0^{\omega_\mathbf{r}}, \eta_0^{J^+}\}$, where $\eta_0^{\omega_\mathbf{r}}$ is the maximum step size one can take so that $\omega_\mathbf{r}(\mathbf{f}^{t+1})\leq \omega_\mathbf{r}(\mathbf{f}^{t})$. 
	Then for all $\eta \in [0, \tilde{\eta}_0]$, and $\mathbf{f}^{t+1} = \mathbf{f}(\mathbf{x}^t - \eta \mathbf{d}_{nd})$
	\begin{align*}
	    \omega_\mathbf{r}(\mathbf{f}^{t+1}) \le \omega_\mathbf{r}(\mathbf{f}^t),
	    \;\text{ and }\ \mathbf{f}_j^{t+1} &\le   \mathbf{f}_j^t,  \quad \forall j \in J^+.\\
	    \implies r_jf_j^{t+1} &\le r_jf_j^t \le \lambda^t \\
	    \implies f_j^{t+1} &\le \widecheck{f}^t_j, \quad \forall j \in J^+
	\end{align*}
	Lemma \ref{th:bal} ensures that $\mathrm{J}^* \subset \mathrm{J}^+$. If all the other objectives in $\mathrm{J}^-$ also satisfy
	\begin{align*}
	    r_jf^{t+1}_j \le \lambda^t, \forall \eta \in [0, \tilde{\eta}_0]
	\end{align*}
	then $\tilde{\eta}_0$ can be used as the step size as it is. If this is not the case, i.e. there exists some $j'\in \mathrm{J}^-$ such that
	\begin{align*}
	    r_{j'}f_{j'}(\mathbf{x}^t - \tilde{\eta}_0 \mathbf{d}_{nd}) > \lambda^t,
	\end{align*}
	then continuity of the objective functions ensures that there must exists some $\eta_{0j'} < \tilde{\eta}_0$ such that 
	\begin{align*}
	    r_{j'}f^{t+1}_{j'} \le \lambda^t, \forall \eta \in [0, \eta_{0j'}].
	\end{align*}
	So choosing $\eta_0 = \min_{j'} \{\eta_{0j'}\}$ we finally get
	\begin{align*}
	    \mathbf{r} \odot \mathbf{f}(\mathbf{x}^t - \eta \mathbf{d}_{nd}) = \mathbf{r} \odot \mathbf{f}^{t+1} &\preccurlyeq  \lambda^t \mathbf{r} \\
	    \implies \mathbf{f}^{t+1} &\preccurlyeq \widecheck{\mathbf{f}}^t \\
	    \therefore\ \mathbf{f}^{t+1} &\in \mathcal{A}^\mathbf{r}_{\mathbf{f}^t}
	\end{align*}
	for all $\eta \in [0, \eta_0]$.
	\hfill   
	\end{proof}

    \mntnct*
    \begin{proof}
      We know from  Lemmas \ref{th:bal} and \ref{th:des} that $\mathbf{d}_{nd} \neq 0$ since it has positive angles angle with at least one gradient. Therefore, $\mathbf{f}^{t+1} \neq \mathbf{f}^t$. Now, using Lemma \ref{th:admissible_set} we can conclude that $\mathcal{A}^r_{\mathbf{f}^{t+1}} \subset \mathcal{A}^r_{\mathbf{f}^t}$, and everywhere regularity assumption ensures that $\{\mathcal{A}^r_{\mathbf{f}^t}\}$ converges.
    \hfill   
    \end{proof}
    
	% \mntnct*
    \unconstdndnc*
    \begin{proof}
     
	The necessity prove is trivial; because if $\mathbf{x}^*\in\mathcal{P}_\mathbf{r}$, then the CSZ inequality based anchor direction $\mathbf{a}$ in \eqref{eq:cs_anchor} is $\mathbf{0}_m\in \mathbb{R}^m$, resulting a $\mathbf{0}_m$ coefficient $\bm{\beta}$ from the QP \eqref{eq:qp_x0_po_unconst}, and $\mathbf{0}_n$ search direction.
	
	For the sufficiency, we prove the contra-positive: if $\mathbf{x}^*\notin \mathcal{P}_\mathbf{r}$, then $\mathbf{d}_{nd} \neq \mathbf{0}_n$.
	% First let us consider the case $\mathbf{x}^*\in \mathrm{Int}(\mathbb{X})$. 
    When $\mathbf{x}^*\notin \mathcal{P}_\mathbf{r}$, $\mathbf{a}$ is non-zero. For a non-zero $\mathbf{a}$, the $\bm{\beta}^*$ in the QP \eqref{eq:qp_x0_po_unconst} will be zero only when $\mathbf{a}$ lies in the null space of $\mathrm{F}\mathrm{F}^T$. But if $\mathbf{a}\in \mathrm{Null}(\mathrm{F}\mathrm{F}^T)$, then $\mathbf{f^* = f(x^*)} \in \mathrm{Col}(\mathrm{F}\mathrm{F}^T)$; because $\mathbf{f}^*$ is orthogonal to the CSZ inequality-based anchor from Claim \ref{th:fa_cs=0}, and $\mathrm{rank}(\mathrm{F}\mathrm{F}^T) = \mathrm{rank}(\mathrm{F}) = m - 1$ from the regularity condition. If $\mathbf{f}^*$ is in the column space of $\mathrm{F}\mathrm{F}^T$ then there exists a $\bm{\beta} \in \mathbb{R}^m$ such that $\mathrm{F}\mathrm{F}^T\bm{\beta} = \mathbf{f}^* \succ 0$, which makes $\mathrm{F}^T\bm{\beta}$ a descent direction. This is a contradiction, as no descent direction should exist at the Pareto optimal $\mathbf{x}^*$. 
    In fact, the eigenvector $\mathbf{v}_1$ of $\mathrm{F}\mathrm{F}^T$ corresponding to its Null space must be an all positive vector like the objective vector $\mathbf{f}^*$ in order to exclude all possible descent direction from the $\mathrm{Col}(\mathrm{F}\mathrm{F}^T)$. Note, the eigenvectors of $\mathrm{Col}(\mathrm{F}\mathrm{F}^T)$ are orthogonal to each other. Therefore $\mathbf{a} \notin \mathrm{Null}(\mathrm{F}\mathrm{F}^T)$ and $\mathrm{F}\mathrm{F}^T\bm{\beta}^*$ will not be zero from the QP \eqref{eq:qp_x0_po_unconst}, hence $\mathbf{d}_{nd}\neq \mathbf{0}_n$.
	\hfill   
    \end{proof}
    
    \textbf{Beyond Regularity Assumption:} In general, this Theorem is true even without the regularity assumptions when the orthogonal projection of the CSZ inequality-based anchor $\mathbf{a}$ \eqref{eq:cs_anchor} onto the hyperplane $\mathrm{Col(FF}^T)$ is non-zero.

\begin{figure}[h]
		\centering
        \begin{subfigure}[b]{0.49\textwidth}
        \centering
        \def\svgwidth{0.8\linewidth}	% if the document class is double column, use \columnwidth
		\begingroup\makeatletter\def\f@size{10}\check@mathfonts	% f@size decides the fontsize of "math" text in Figure
		\def\maketag@@@#1{\hbox{\m@th\large\normalfont#1}}%
		% \fontsize{9pt}{11pt}\selectfont % decides the overall fontsize
		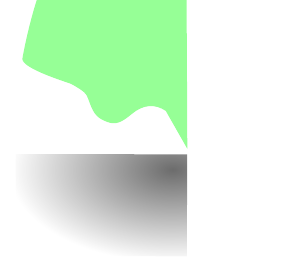
		\endgroup\vspace{-.2cm}
        \caption{Lagrange Identity anchor $\mathbf{a}$ \eqref{eq:lgrn_anchor} \label{fig:lgrn_admissible}}
        \end{subfigure}~
        \begin{subfigure}[b]{0.49\textwidth}
        \centering
        \def\svgwidth{0.8\linewidth}	% if the document class is double column, use \columnwidth
		\begingroup\makeatletter\def\f@size{10}\check@mathfonts	% f@size decides the fontsize of "math" text in Figure
		\def\maketag@@@#1{\hbox{\m@th\large\normalfont#1}}%
		% \fontsize{9pt}{11pt}\selectfont % decides the overall fontsize
		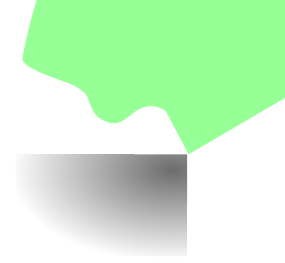
		\endgroup\vspace{-.2cm}
        \caption{CSZ inequality anchor $\mathbf{a}$ \eqref{eq:cs_anchor} \label{fig:cs_admisible}}
        \end{subfigure}
		\caption{(Color Online) Illustration of a corner case where the Lagrange identity based anchor may not escape a PO solution $\mathbf{f}^*$, whereas the CSZ inequality based anchor can escape it to reach the EPO solution $\mathbf{f}_\mathbf{r}^*$. The Null space of $\mathrm{F}\mathrm{F}^T$ is $\mathbf{v}_1$. The green filled region is the set $\mathcal{M}_{\mathbf{f}^*}^\mathbf{r} = \{\mathbf{f} \in \mathcal{O} \,|\, \omega_\mathbf{r}(\mathbf{f}) \leq \omega_\mathbf{r}(\mathbf{f}^*)\}$ 
        % contains objective vectors that are more proportional to the EPO point $\mathbf{f}_\mathbf{r}^*$ than $\mathbf{f}^*$, i.e., $\omega_\mathbf{r}(\mathbf{f}) \leq \omega_\mathbf{r}(\mathbf{f}^*)$, which is
         for CSZ inequality and Lagrange identity based proportionality gauges $\omega_\mathbf{r}$ defined in \eqref{eq:lgrn_prop} and \eqref{eq:cs_prop} respectively. 
			\label{fig:admissible_sets_comparison}}
	\end{figure}

    \textbf{Issue of Lagrange Identity based Balancing Anchor Direction:} The anchor direction in \eqref{eq:lgrn_anchor} is not orthogonal to the objective vector $\mathbf{f}^*$. Therefore, in a corner case, the anchor direction $\mathbf{a}$ may be very close to the $\mathrm{Null(FF}^T)$ as shown in Figure \ref{fig:lgrn_admissible}. Therefore its projection on to $\mathrm{Col(FF}^T)$ will vanish. However, CSZ inequality-based anchor \eqref{eq:cs_anchor} mitigates this issue as shown in Figure \ref{fig:cs_admisible}. Note, the set $\mathcal{M}_{\mathbf{f}^t}^\mathbf{r}$ defined in \eqref{eq:Mrft} is different for both the proportionality gauges. 
    In practice, although this extreme scenario may not arise, but the projection of $\mathbf{a}$ on to $\mathrm{Col(FF}^T)$ will have a lesser magnitude in Lagrange identity based anchor as compared to that of KL divergence based anchor.
 
    \plineq*
    \begin{proof}
    First we prove for the Cauchy-Schwarz inequality based $\omega_\mathbf{r}$ and $\mathbf{a}$.  
    \begin{flalign}
        & &\|\mathbf{a}\|^2 &= \|\overrightarrow{\mathbf{f}} \langle \overrightarrow{\mathbf{f}}, \overrightarrow{\mathbf{r}^{-1}}\rangle^2 - \overrightarrow{\mathbf{r}^{-1}} \langle \overrightarrow{\mathbf{f}}, \overrightarrow{\mathbf{r}^{-1}}\rangle\|^2 \\
        &\Rightarrow &  &= \langle \overrightarrow{\mathbf{f}}, \overrightarrow{\mathbf{r}^{-1}}\rangle^4 + \langle \overrightarrow{\mathbf{f}}, \overrightarrow{\mathbf{r}^{-1}}\rangle^2 - 2 \langle \overrightarrow{\mathbf{f}}, \overrightarrow{\mathbf{r}^{-1}}\rangle^4 \\
        &\Rightarrow & &= \langle \overrightarrow{\mathbf{f}}, \overrightarrow{\mathbf{r}^{-1}}\rangle^2 - \langle \overrightarrow{\mathbf{f}}, \overrightarrow{\mathbf{r}^{-1}}\rangle^4 \\
        &\Rightarrow & &= \langle \overrightarrow{\mathbf{f}}, \overrightarrow{\mathbf{r}^{-1}}\rangle^2(1-\langle \overrightarrow{\mathbf{f}}, \overrightarrow{\mathbf{r}^{-1}}\rangle^2) \\
        &\Rightarrow & &= \langle \overrightarrow{\mathbf{f}}, \overrightarrow{\mathbf{r}^{-1}}\rangle^2 2\omega_\mathbf{r}(\mathbf{f}) \\
        &\Rightarrow & \frac{1}{2} \|\mathbf{a}\|^2 &\geq  \langle \overrightarrow{\mathbf{f}^0}, \overrightarrow{\mathbf{r}^{-1}}\rangle^2 \omega_\mathbf{r}(\mathbf{f}) & \left(\because \langle \overrightarrow{\mathbf{f}}, \overrightarrow{\mathbf{r}^{-1}}\rangle \geq \langle \overrightarrow{\mathbf{f}^0}, \overrightarrow{\mathbf{r}^{-1}}\rangle \quad \forall \mathbf{f} \in \mathcal{M}_{\mathbf{f}^0}^\mathbf{r}\right) \\
        &\therefore & \frac{1}{2} \|\mathbf{a}\|^2 &\geq \tau \omega_\mathbf{r} & \text{where } \tau \leq \langle \overrightarrow{\mathbf{f}^0}, \overrightarrow{\mathbf{r}^{-1}}\rangle^2
    \end{flalign}

    Next we prove for the Lagrange anchor \eqref{eq:lgrn_anchor}.
    \begin{flalign}
        & & \|\mathbf{a}\|^2 &= \|\mathbf{f} -\langle \mathbf{f}, \overrightarrow{\mathbf{r}^{-1}} \rangle \overrightarrow{\mathbf{r}^{-1}}\|^2 & \text{where } \overrightarrow{\mathbf{r}^{-1}} = \frac{\mathbf{r}^{-1}}{\|\mathbf{r}^{-1}\|} \\
        &\Rightarrow & &= \|\mathbf{f}\|^2 + \langle \mathbf{f}, \overrightarrow{\mathbf{r}^{-1}} \rangle^2 - 2 \langle \mathbf{f}, \overrightarrow{\mathbf{r}^{-1}} \rangle^2 & \\
        & \Rightarrow & &= \|\mathbf{f}\|^2 - \langle \mathbf{f}, \overrightarrow{\mathbf{r}^{-1}} \rangle^2 & \\
        &\Rightarrow & &= 2 \omega_\mathbf{r}(\mathbf{f}) & (\because \text{definition of $\omega_\mathbf{r}$ in \eqref{eq:lgrn_prop}}) \\ 
        &\therefore & \frac{1}{2}\|\mathbf{a}\|^2 &\geq \tau \omega_\mathbf{r}(\mathbf{f}) & \text{where } \tau \leq 1.
    \end{flalign}
    \hfill   
	\end{proof}
 
    \cvnrt*
    \begin{proof}
     
    We use a property of Lipschitz smooth functions. If $g$ be a scalar valued function whose gradient is smooth with Lipschitz $L_g$, then (see Lemma 1.2.3 in \cite{Nesterov2004})
    \begin{align}\label{eq:lips}
        \left|g\left(\mathbf{y}^{t+1}\right) - \left(g\left(\mathbf{y}^t\right) + \langle \nabla_{\mathbf{y}^t}g,\, \mathbf{y}^{t+1} - \mathbf{y}^{t} \rangle\right)\right| \leq \frac{1}{2} L_g \| \mathbf{y}^{t+1} - \mathbf{y}^{t} \|^2. 
    \end{align}
    From Lipschitz smoothness Assumptions \ref{asm:smth}, we can write the following inequality for $\omega$:
    \begin{flalign}
        & &\omega_\mathbf{r}(\mathbf{f}^{t+1}) - \omega_\mathbf{r}(\mathbf{f}^{t})  &\leq \langle\nabla\omega_\mathbf{r}, \mathbf{f}^{t+1}-\mathbf{f}^{t}\rangle + \frac{L_\omega}{2} \|\mathbf{f}^{t+1}-\mathbf{f}^{t}\|^2, &(\because \eqref{eq:lips})\label{eq:omega_smth}
    \end{flalign}
    where $\mathbf{f}^t = \mathbf{f}(\mathbf{x}^t)$. Similarly, with $\mathbf{L} = [L_1, \cdots, L_m]^T$, Jacobian at $\mathbf{x}^t$ as $\mathrm{F}\in \mathbb{R}^{m\times n}$, stepsize $\eta > 0$, and
    \begin{align}
        \mathbf{x}^{t+1} &= \mathbf{x}^t - \eta \mathbf{d}, \label{eq:xtpt-xt} \\
        \text{where}\quad \mathbf{d} &=\mathrm{F}^T\bm{\beta}^* \label{eq:d=Fb}
    \end{align}
    % where the search direction $\mathbf{d}=\mathrm{F}^T\bm{\beta}^*$ 
    is obtained from solving the QP \eqref{eq:qp_x0_random_unconst} or \eqref{eq:qp_x0_po_unconst} in the balance mode, we can write the following inequalities for $\Delta\mathbf{f}^t = \mathbf{f}^{t+1} - \mathbf{f}^t$:
    % \begin{alignat}{2}
    \begin{flalign}
        % & & \mathrm{F}\,(\mathbf{x}^{t+1}-\mathbf{x}^{t}) - \frac{1}{2} \mathbf{L} \|\mathbf{x}^{t+1}-\mathbf{x}^{t}\|^2 &\preccurlyeq \mathbf{f}^{t+1} - \mathbf{f}^t \preccurlyeq \mathrm{F}\,(\mathbf{x}^{t+1}-\mathbf{x}^{t}) + \frac{1}{2} \mathbf{L} \|\mathbf{x}^{t+1}-\mathbf{x}^{t}\|^2 & (\because \eqref{eq:lips}) \nonumber \\
        & &\hspace{-2mm}  - \eta \mathrm{F}\mathbf{d} - \frac{\eta^2}{2} \mathbf{L} \|\mathbf{d}\|^2 &\preccurlyeq 
        % \mathbf{f}^{t+1} - \mathbf{f}^t 
        \Delta\mathbf{f}^t
        \preccurlyeq -\eta \mathrm{F}\mathbf{d} + \frac{\eta^2}{2}  \mathbf{L} \|\mathbf{d}\|^2 & 
        (\because \text{\eqref{eq:lips} and \eqref{eq:xtpt-xt}}) 
        % \& \|\mathbf{x}^{t+1}-\mathbf{x}^{t}\|_2^2 = \eta^2 \|\mathbf{d}\|_2^2) 
        \\
        &\Rightarrow &\hspace{-2mm} - \eta \mathrm{F}\mathbf{d} - \frac{\eta^2}{2} \mathbf{L} \langle\bm{\beta}^{*}, \mathrm{F}\mathbf{d}\rangle &\preccurlyeq 
        % \mathbf{f}^{t+1}  - \mathbf{f}^t 
        \Delta\mathbf{f}^t
        \preccurlyeq -\eta \mathrm{F}\mathbf{d} + \frac{\eta^2}{2} \mathbf{L} \langle\bm{\beta}^{*}, \mathrm{F}\mathbf{d}\rangle & (\because \text{\eqref{eq:d=Fb}})
        % \langle \mathbf{d}, \mathbf{d}\rangle = \langle \mathrm{F}^T\bm{\beta}^*, \mathbf{d}\rangle) 
        \label{eq:b_Fd}\\
        & \Rightarrow &\hspace{-2mm} - \eta \mathrm{F}\mathbf{d} - \frac{\eta^2}{2} \mathbf{L} \|\bm{\beta}^*\|\|\mathrm{F}\mathbf{d}\| &\preccurlyeq 
        % \mathbf{f}^{t+1}  - \mathbf{f}^t 
        \Delta\mathbf{f}^t
        \preccurlyeq  -\eta \mathrm{F}\mathbf{d} + \frac{\eta^2}{2}\mathbf{L} \|\bm{\beta}^*\|\|\mathrm{F}\mathbf{d}\| & (\because \langle \bm{\beta}^*, \mathrm{F}\mathbf{d} \rangle \leq \|\bm{\beta}^*\| \|\mathrm{F}\mathbf{d}\|) 
        \label{eq:cs_ineq} \\
        &\Rightarrow  &\hspace{-2mm} - \eta \mathrm{F}\mathbf{d} - \frac{\eta^2\sqrt{m}}{2} \mathbf{L}\|\mathrm{F}\mathbf{d}\| &\preccurlyeq 
        % \mathbf{f}^{t+1}  - \mathbf{f}^t 
        \Delta\mathbf{f}^t
        \preccurlyeq  -\eta \mathrm{F}\mathbf{d} + \frac{\eta^2\sqrt{m}}{2} \mathbf{L} \|\mathrm{F}\mathbf{d}\| & 
        (\because \|\beta\|_1 \leq 1 \Rightarrow \|\beta\|_2 \leq \sqrt{m})
        \label{eq:m_Fd} \\
        & \Rightarrow &\hspace{-2mm}  - \eta \mathrm{F}\mathbf{d} - \frac{\eta^2\sqrt{m}L_\mathbf{f}\|\mathrm{F}\mathbf{d}\|}{2} \mathbf{1} &\preccurlyeq 
        % \mathbf{f}^{t+1}  - \mathbf{f}^t 
        \Delta\mathbf{f}^t
        \preccurlyeq -\eta \mathrm{F}\mathbf{d} + \frac{\eta^2\sqrt{m}L_\mathbf{f}\|\mathrm{F}\mathbf{d}\|}{2}  \mathbf{1}, & (\because L_\mathbf{f} = \max_{j\in [m]} L_j)\label{eq:Lf_m_Fd}
    \end{flalign}
    % \end{alignat}
    where $\mathbf{1}\in\mathbb{R}^m$ has all ones. If no subscript is mentioned explicitly then $\|\cdot\|$ means $\ell_2$ norm. In \eqref{eq:cs_ineq}, we have applied Cauchy-Schwarz inequality. 
    % % We get \eqref{eq:m_Fd} from \eqref{eq:b_Fd} by using
    % The step \eqref{eq:b_Fd}$\implies$\eqref{eq:m_Fd} because
    % $\langle \bm{\beta}^*, \mathrm{F}\mathbf{d} \rangle \leq \|\bm{\beta}^*\| \|\mathrm{F}\mathbf{d}\| \leq \sqrt{m} \|\mathrm{F}\mathbf{d}\|,$
    % where the first one is Cauchy-Schwarz inequality and the second inequality is due to $\|\beta\|_1 \leq 1$. The step \eqref{eq:m_Fd}$\implies$\eqref{eq:Lf_m_Fd}, 
    
    Next, we upper bound the term $\langle\nabla\omega_\mathbf{r}, \mathbf{f}^{t+1}-\mathbf{f}^{t}\rangle $ in \eqref{eq:omega_smth} as 
    \begin{flalign}
        & & \langle\nabla\omega_\mathbf{r}, \Delta\mathbf{f}^{t}\rangle &\leq -\eta \langle\nabla\omega_\mathbf{r}, \mathrm{F}\mathbf{d}\rangle + \frac{K}{2} \|\mathrm{F}\mathbf{d}\|
        \|\nabla\omega_\mathbf{r}\|_1,
        % \langle \nabla\omega_\mathbf{r}, \mathrm{sign}( \nabla\omega_\mathbf{r})\rangle 
        & \text{where } K=L_\mathbf{f}\eta^2\sqrt{m} \quad\qquad (\because \text{\eqref{eq:Lf_m_Fd}})  \label{eq:w_f}\\ 
        & &\Rightarrow \qquad\qquad &\leq -\eta \langle\nabla\omega_\mathbf{r}, \mathrm{F}\mathbf{d}\rangle + \frac{K'}{2} \|\mathrm{F}\mathbf{d}\| \|\nabla\omega_\mathbf{r}\|, & K'=K\sqrt{m}\quad (\because 
        \|\nabla\omega_\mathbf{r}\|_1 \leq \sqrt{m}\|\nabla\omega_\mathbf{r}\|_2 )
        % \langle \nabla\omega_\mathbf{r}, \mathrm{sign}( \nabla\omega_\mathbf{r})\rangle \leq \|\nabla\omega_\mathbf{r}\|\|\mathbf{1}\|)
        \label{eq:w_f_m}\\
        & & \Rightarrow \qquad\qquad &= -\frac{\eta}{s} \langle\mathbf{a}, \mathrm{F}\mathbf{d}\rangle + \frac{K'}{2s} \|\mathrm{F}\mathbf{d}\| \|\mathbf{a}\|, & \left(\because\mathbf{a} = s\nabla\omega_\mathbf{r}, \ s= \begin{cases}
            \|\mathbf{f}^t\| \  \text{in $a=$\eqref{eq:cs_anchor}} \\
            \ 1 \quad \text{in $a=$\eqref{eq:lgrn_anchor}}
        \end{cases} \right) \label{eq:a/s}\\
        & & \Rightarrow \qquad\qquad &\leq -\frac{\eta}{s_0} \langle\mathbf{a}, \mathrm{F}\mathbf{d}\rangle + \frac{K'}{2s_1} \|\mathrm{F}\mathbf{d}\| \|\mathbf{a}\|, & (\because s_1 \leq s \leq s_0 \text{ from } \text{Assumption \ref{asm:comp}}) \label{eq:s0s1} \\
        & &\Rightarrow \qquad\qquad &\leq -\frac{\eta}{s_0} \langle\mathbf{a}, \mathrm{F}\mathbf{d}\rangle + \frac{K'}{2s_1}\|\mathbf{a}\|^2 & (\because \|\mathrm{F}\mathbf{d}\| \leq \|\mathbf{a}\| \text{ by QP \eqref{eq:qp_x0_random_unconst} or \eqref{eq:qp_x0_po_unconst}}) \label{eq:wf_w}\\
        & &\Rightarrow \qquad\qquad &\leq -\frac{\eta\delta^2}{s_0\sqrt{m}W^2} \|\mathbf{a}\|^2+ \frac{L_\mathbf{f}\eta^2m}{2s_1}\|\mathbf{a}\|^2 & (\because \text{Lemma \ref{th:afd_angle}}) \label{eq:wdfub}
    \end{flalign}
    % The $\mathrm{sign}$ function in \eqref{eq:w_f} is applied element-wise. 
    In \eqref{eq:w_f}, we use the right (left) inequality in \eqref{eq:Lf_m_Fd} if $\frac{\partial\omega_\mathbf{r}}{\partial f_j}$ is positive (negative) for $j\in [m]$. 
    In \eqref{eq:s0s1}, the factors $s_0$ and $s_1$ are as given in the statement of the theorem.
    The step \eqref{eq:w_f}$\Rightarrow$\eqref{eq:w_f_m} is similar to \eqref{eq:b_Fd}$\Rightarrow$\eqref{eq:m_Fd}. 
    The step \eqref{eq:s0s1}$\Rightarrow$\eqref{eq:wf_w} because, by construction, the solution of QP  in \eqref{eq:qp_x0_random_unconst} or \eqref{eq:qp_x0_po_unconst} gives $\|\mathrm{F}\mathbf{d}\| \leq = \|\mathbf{a}\|$.
    
    Similarly, we upper bound the term $\|\mathbf{f}^{t+1}-\mathbf{f}^{t}\|^2$ in \eqref{eq:omega_smth} as
    \begin{flalign}
        & &\|\mathbf{f}^{t+1}-\mathbf{f}^{t}\|^2 &\leq \eta^2\|\mathrm{F}\mathbf{d}\|^2 + \frac{1}{4}\eta^4 L_\mathbf{f}^2m^2 \|\mathrm{F}\mathbf{d}\|^2 + \eta^3L_\mathbf{f}\sqrt{m}\, 
        \|\mathrm{F}\mathbf{d}\|_1
        % \langle\mathrm{F}\mathbf{d}, \mathrm{sign}(\mathrm{F}\mathbf{d})\rangle 
        &(\because \text{ \eqref{eq:Lf_m_Fd}})\\
        &\Rightarrow &  &\leq \eta^2\|\mathrm{F}\mathbf{d}\|^2 + \frac{1}{4}\eta^4 L_\mathbf{f}^2m^2 \|\mathrm{F}\mathbf{d}\|^2 + \eta^3L_\mathbf{f}m\|\mathrm{F}\mathbf{d}\|^2 & (\because \|\mathrm{F}\mathbf{d}\|_1 \leq \sqrt{m}\|\mathrm{F}\mathbf{d}\|_2)\\
        & \Rightarrow & &\leq \eta^2\|\mathbf{a}\|^2 + \frac{1}{4}\eta^4 L_\mathbf{f}^2m^2 \|\mathbf{a}\|^2 + \eta^3L_\mathbf{f}m\|\mathbf{a}\|^2. & (\because \|\mathrm{F}\mathbf{d}\| \leq \|\mathbf{a}\|)\label{eq:f_f}
    \end{flalign}
    Finally, we can rewrite \eqref{eq:omega_smth} as 
    \begin{flalign}
        & &\omega_\mathbf{r}(\mathbf{f}^{t+1}) - \omega_\mathbf{r}(\mathbf{f}^{t})  &\leq - \eta p(\eta) \|\mathbf{a}\|^2 &  (\because \text{\eqref{eq:omega_smth}, \eqref{eq:wdfub} and \eqref{eq:f_f}}) \\
        & \Rightarrow& \omega_\mathbf{r}(\mathbf{f}^{t+1}) - \omega_\mathbf{r}(\mathbf{f}^{t}) &\leq - 2\tau \eta p(\eta) \omega_\mathbf{r}(\mathbf{f}^{t}) & (\because \text{Lemma \eqref{th:pl_ineq}}) \\
        & \therefore & \omega_\mathbf{r}(\mathbf{f}^{t+1}) &\leq \left(1 - 2\tau \eta p(\eta) \right) \omega_\mathbf{r}(\mathbf{f}^{t}) & 
        \\
        & \Rightarrow & \omega_\mathbf{r}(\mathbf{f}^{t+1}) &\leq \left(1 - 2\tau \eta p(\eta) \right)^{t+1} \omega_\mathbf{r}(\mathbf{f}^{0}) \label{eq:wtptinw0}
    \end{flalign}
    where the polynomial $p(\eta) = c_0 - c_1\eta - c_2 \eta^2 - c_3 \eta^3$ has the coefficients as given in the statement of the theorem. The ideal stepsize $\eta^*$ that minimizes $\left(1 - 2\tau \eta p(\eta) \right)$ is given by the positive root of 
    % $\frac{d}{d\eta}\eta p(\eta) = c_0 - 2c_1 \eta - 3c_2\eta^2 - 4c_3 \eta^3$.
    \begin{align}
         \frac{d}{d\eta}\eta p(\eta) = c_0 - 2c_1 \eta - 3c_2\eta^2 - 4c_3 \eta^3 \label{eq:poly}
    \end{align}
    % Note, since $c_0, c_1, c_2,$ and $c_3$ are positive, $\frac{d}{d\eta}\eta p(\eta) = c_0 - 2c_1 \eta - 3c_2\eta^2 - 4c_3 \eta^3 $ has only one positive root that can be used as the maximum step size. 
    % Let that ideal step size be $\eta^*>0$, i.e.,  $p(\eta^*0)$. 
    The $\eta_0 > 0$ in the statement of the theorem is such that $\eta_0 \leq \eta^*$. We obtain this lower bound using the following properties of a polynomial.
    \begin{enumerate}[label=(\roman*)]
        \item\label{it:cauchy_bound} Cauchy's upper bound \citep{10.2307/43679074} $U$ for all the absolute roots $|z|$ of a polynomial $b_3 + b_2 x + b_1x^2 + b_0 x^3$, i.e., $|z|_i \leq U \ \forall i\in[3]$, is given by $U = 1 + \max_{i=1}^3 \left| \frac{b_i}{b_0} \right|$
        \item\label{it:root_lb}  $1/U$ is the lower bound for all the absolute roots of polynomial $b_0 + b_1 x + b_2x^2 + b_3 x^3$.
    \end{enumerate}
    Finally, applying Properties \ref{it:cauchy_bound} and \ref{it:root_lb} on the polynomial \eqref{eq:poly}, we get the $\eta_0$ as given in the statement of the theorem.

    The iteration complexity, i.e., the maximum number of iterations until which $\omega_\mathbf{r}(\mathbf{f}^{t})\geq \epsilon$ is given by
    \begin{flalign}
        & & \epsilon &\leq \omega_\mathbf{r}(\mathbf{f}^{t})    & \\
        &\Rightarrow & \epsilon &\leq  \left(1 - 2\tau \eta_0 p(\eta_0) \right)^{t} \omega_\mathbf{r}(\mathbf{f}^{0}) & (\because \text{\eqref{eq:wtptinw0}}) \\
        & \Rightarrow & \log(\epsilon) &\leq t \log(\rho) + \log(\omega_\mathbf{r}(\mathbf{f}^{0})), & \text{where } \rho =\left(1 - 2\tau \eta_0 p(\eta_0) \right) \\
        & \Rightarrow & t\log\left(\frac{1}{\rho}\right) &\leq \log\left(\frac{\omega_\mathbf{r}(\mathbf{f}^{0})}{\epsilon}\right) \\
        & \therefore & t &\leq \log\left(\frac{\omega_\mathbf{r}(\mathbf{f}^{0})}{\epsilon}\right)/ \log\left(\frac{1}{\rho}\right) \leq O(\log(1/\epsilon)) & (\because \rho < 1 \Rightarrow \log(1/\rho) > 0) 
    \end{flalign}
    % \footnote{\url{https://en.wikipedia.org/wiki/Geometrical_properties_of_polynomial_roots}} 
    % \hfill   
    \end{proof}

    \begin{lemma}\label{th:afd_angle}
    If Assumption \ref{asm:comp} is satisfied for an $W$ and Assumption \ref{asm:reg} is satisfied for a $\delta$, then search direction $\mathbf{d}$ obtained in the balance mode of QP \eqref{eq:qp_x0_random_unconst} or \eqref{eq:qp_x0_po_unconst} at a point $\mathbf{x}$ with Jacobian $\mathrm{F}$ satisfies
    \begin{align}
        \langle\nabla\omega_\mathbf{r}, \mathrm{F}\mathbf{d}\rangle \geq \frac{\delta^2}{\sqrt{m}W^2} \|\nabla\omega_\mathbf{r}\|^2
    \end{align}
    \end{lemma}
    \begin{proof}
          
        We can write
        \begin{align}
            \langle\nabla\omega_\mathbf{r}, \mathrm{F}\mathbf{d}\rangle = \|\nabla\omega_\mathbf{r}\| \|\mathrm{F}\mathbf{d}\| \cos(\theta),
        \end{align}
        where $\theta$ is the angle between the anchor direction $\mathbf{a} = \nabla\omega_\mathbf{r}$ and the direction $\mathrm{F}\mathbf{d} = \mathrm{F}\mathrm{F}^T\bm{\beta}^*$. In this prove we find lower bound for $\|\mathrm{F}\mathbf{d}\| \cos(\theta)$. 
        Note, if $\mathbf{a} \in \mathcal{CH}_\mathbf{f}^\pm$ (defined in \eqref{eq:chfpm}), then $\|\mathrm{F}\mathbf{d}\| \cos(\theta) = \|\nabla\omega_\mathbf{r}\|$  (e.g., see proof of Lemma \ref{th:bal}), which is an upper bound for $\|\mathrm{F}\mathbf{d}\| \cos(\theta)$. Therefore, we only consider for $\mathbf{a} \notin \mathcal{CH}_\mathbf{f}^\pm$.

        First, we lower bound $\cos(\theta)$. The maximum possible value of $\cos(\theta)=1$; it occurs when $\bm{\beta}^* = \frac{(\mathrm{F}\mathrm{F}^T)^{-1}\mathbf{a}}{\|(\mathrm{F}\mathrm{F}^T)^{-1}\mathbf{a}\|}$, which results in $\mathrm{F}\mathbf{d} = s \mathbf{a}$, where $s= \frac{1}{\|(\mathrm{F}\mathrm{F}^T)^{-1}\mathbf{a}\|_1}$. 
    This is a feasible solution even for the constrained balance mode in the QP \eqref{eq:qp_x0_random_unconst} as proven in Lemma \ref{th:bal}. Here, the residue $\mathbf{a} - \mathrm{F}\mathrm{F}^T\bm{\beta}^*$ is aligned with $\mathrm{F}\mathbf{d}$.
    The minimum possible value of $\cos(\theta)$ (or the maximum angle between $\mathrm{F}\mathbf{d}$ and $\mathbf{a}$)  will occur when the residue $\mathbf{a} - \mathrm{F}\mathrm{F}^T\bm{\beta}^*$ is orthogonal to $\mathrm{F}\mathrm{F}^T\bm{\beta}^*$. In other words, $\mathrm{F}\mathbf{d}$ and  $\mathbf{a}$ are the base and hypotenuse of a right triangle. Therefore,
    \begin{flalign}
        & & \cos{\theta} &\geq \frac{\|\mathrm{F}\mathbf{d}\|}{\|\mathbf{a}\|} & \\
        % & & \Rightarrow \qquad &\geq \frac{\delta}{\sqrt{m} \|\mathbf{a}\|} & (\because \text{\eqref{eq:Fdlb}}) \\
        & & \Rightarrow \qquad &\geq \frac{\|\mathrm{F}\mathbf{d}\|}{W} & (\because \|\mathbf{a}\| \leq W \text{ from Assumption \eqref{asm:comp}}). \label{eq:coslb}
    \end{flalign}
    % $\therefore$ Using \eqref{eq:Fdlb} and \eqref{eq:coslb} we get the required lower bound stated in the Lemma. 

        % We split
        % \begin{enumerate}
        %     \item QP \eqref{eq:qp_x0_po_unconst} when $\mathbf{x} \in \mathbb{X} \backslash \mathcal{P}^\delta$
        %     \item QP \eqref{eq:qp_x0_po_unconst} when $\mathbf{x} \in \mathcal{P}^\delta$
        %     \item QP \eqref{eq:qp_x0_random_unconst} when $\mathbf{x} \in \mathbb{X} \backslash \mathcal{P}^\delta$
        %     \item QP \eqref{eq:qp_x0_random_unconst} when $\mathbf{x} \in \mathcal{P}^\delta$. 
        % \end{enumerate}

        Next, we lower bound $\|\mathrm{F}\mathbf{d}\|$.
        % In both cases 1 and 3, 
        % first case, the only constraint is $\|\beta\|_1$. Therefore, 
        To minimize the quadratic cost $\|\mathrm{F}\mathrm{F}^T\bm{\beta} - \mathbf{a}\|^2$, the vector $\mathrm{F}\mathbf{d}$ must lie on the boundary of $\mathcal{CH}_\mathbf{f}^\pm$,   
        %$\mathrm{F}\mathbf{d}=\mathrm{F}\mathrm{F}^T\bm{\beta}^* \in \partial \mathcal{CH}_\mathbf{f}^\pm$, the boundary of $\mathcal{CH}_\mathbf{f}^\pm$, 
        resulting in
        \begin{align}
            \|\bm{\beta}^*\|_1 = 1 \label{eq:b*=1}
        \end{align}
        % , which is the transformation of $\ell_1$ ball after applying $\mathrm{F}\mathrm{F}^T$. 
        We first lower bound $\|\mathrm{F}\mathbf{d}\|$ when $\mathbf{x} \in \mathbb{X} \backslash \mathcal{P}^\delta$:
        \begin{flalign}
            & & \|\mathrm{F}\mathbf{d}\| &= \|\mathrm{F}\mathrm{F}^T\bm{\beta}^*\|  & \\
            & \qquad \Rightarrow & & \geq \|\mathrm{F}\mathrm{F}^T\|_\circ \, \|\bm{\beta}^*\|_2 & \text{where } \| A\|_\circ = \min_{\bm{\beta}} \frac{\|A\bm{\beta}\|_2}{\|\bm{\beta}\|_2}. \label{eq:mineigen1}\\
            & \qquad \Rightarrow & & = \sigma_1^2 \|\bm{\beta}^*\|_2  &  \text{where }  \sigma_1 \text{ is the smallest singular value of } \mathrm{F} \label{eq:mineigen2}\\
            & \qquad \Rightarrow & & \geq \delta^2 \|\bm{\beta}^*\|_2  & (\because \text{Assumption \ref{asm:reg}}) \\
            & \qquad \Rightarrow & & \geq \delta^2 \frac{\|\bm{\beta}^*\|_1}{\sqrt{m}} & (\because \|\bm{\beta}^*\|_2 \geq \frac{\|\bm{\beta}^*\|_1}{\sqrt{m}}) \\
            & \qquad \Rightarrow & & = \frac{\delta^2}{\sqrt{m}} & (\because \text{\eqref{eq:b*=1}}) \label{eq:Fdlb1}
            %\\
            % & \qquad \Rightarrow & & = \frac{\delta}{\sqrt{m}} \frac{\|\mathbf{a}\|}{\|\mathbf{a}\|} \geq \frac{\delta}{\sqrt{m}} \frac{\|\mathbf{a}\|}{W} & (\because \text{Assumption \ref{asm:comp}}) \label{eq:Fdlb4cos}
            % \geq \min_{\|\bm{\beta}\|_2 = 1} \|\mathrm{F}\mathrm{F}^T\bm{\beta}\| = \sigma_1^2 
        \end{flalign}
    Now, we consider the case $\mathbb{x}\in \mathcal{P}^\delta$. From the definition of $\mathcal{P}^\delta$ \eqref{eq:P_delta}, we know that the lowest singular value $\sigma_1$ vanishes as $\mathbf{x}$ approaches the PF. As a result, the corresponding singular vector (eigenvector of $\mathrm{F}\mathrm{F}^T$) ${\mathbf{v}_1:\mathbb{X}\rightarrow \mathbb{R}^m}$ turns into the null space of $\mathrm{F}\mathrm{F}^T$ as $\mathbf{x}$ approaches\footnote{Note, similar to the singular values, the singular vectors are also smooth w.r.t. variations in $\mathbf{x} \in \mathbb{X}$ due to the smoothness of Jacobian $\mathrm{F}:\mathbb{X} \rightarrow \mathbb{R}^{m\times n}$.} the PF,  
    because according to Assumption \ref{asm:reg}, the second singular value $\sigma_2$ does not vanish in $\mathcal{P}^\delta$. 
    In other words, the column space of $\mathrm{F}\mathrm{F}^T$ becomes equal to the columns space of $\{\mathbf{v}_2, \cdots, \mathbf{v}_m\}$.
    The QP \eqref{eq:qp_x0_po_unconst} is used in such cases. Its balancing anchor direction $\mathbf{a}\notin \mathrm{Null}(\mathrm{F}\mathrm{F}^T)$ as proven in Theorem \ref{th:d_nd_unconst}, where we deduced that $\mathbf{v}_1$ must be an all positive vector like the descending anchor direction $\mathbf{f}$. 
    Moreover, since $\mathbf{a}$ and $\mathbf{f}$ are orthogonal (see Claim \ref{th:fa_cs=0}), to minimize $\|\mathrm{F}\mathrm{F}^T\mathbf{d} - \mathbf{a}\|$, $\mathbf{a}$ gets projected onto the $\mathrm{span}(\{\mathbf{v}_2, \cdots, \mathbf{v}_m\})$ in $\mathcal{CH}_\mathbf{f}^\pm$. Therefore, instead of $\sigma_1$ in \eqref{eq:mineigen1} and \eqref{eq:mineigen2}, the second singular value $\sigma_2$ is applicable, which results in
    \begin{flalign}
            & & \|\mathrm{F}\mathbf{d}\| &\geq \sigma_2^2 \|\bm{\beta}^*\|_2  & \\
            & \qquad \Rightarrow & & \geq \delta^2 \|\bm{\beta}^*\|_2  & (\because \text{Assumption \ref{asm:reg}}) \\
            & \qquad \Rightarrow & & \geq \delta^2 \frac{\|\bm{\beta}^*\|_1}{\sqrt{m}} & (\because \|\bm{\beta}^*\|_2 \geq \frac{\|\bm{\beta}^*\|_1}{\sqrt{m}}) \\
            & \qquad \Rightarrow & & = \frac{\delta^2}{\sqrt{m}} & (\because \text{\eqref{eq:b*=1}}) \label{eq:Fdlb2}
            %\\
            % & \qquad \Rightarrow & & = \frac{\delta}{\sqrt{m}} \frac{\|\mathbf{a}\|}{\|\mathbf{a}\|} \geq \frac{\delta}{\sqrt{m}} \frac{\|\mathbf{a}\|}{W} & (\because \text{Assumption \ref{asm:comp}}) \label{eq:Fdlb4cos}
            % \geq \min_{\|\bm{\beta}\|_2 = 1} \|\mathrm{F}\mathrm{F}^T\bm{\beta}\| = \sigma_1^2 
        \end{flalign}
    Therefore, for any $\mathbf{x}\in\mathbb{X}$ we can write 
    \begin{align}
         \|\mathrm{F}\mathbf{d}\| \geq \frac{\delta^2}{\sqrt{m}} \label{eq:Fdlb}
    \end{align}
    % Let $\mathbf{a}^{\perp \mathbf{v}_1}$ be the orthogonal projection of $\mathbf{a}$ onto the $\mathrm{Col}(\{\mathbf{v}_2, \cdots, \mathbf{v}_m\})$.
    % % , and $\mathbf{a}^{\mathbf{v}_1}$ be the projection onto $\mathbf{v}_1$. 
    % If $\mathbf{a}^{\perp \mathbf{v}_1} \in \mathcal{CH}_\mathbf{f}^\pm$ then $\mathrm{F}\mathbf{d} = \mathbf{a}^{\perp \mathbf{v}_1}$ and $\langle \mathbf{a}, \mathrm{F}\mathbf{d} \rangle = \|\mathbf{a}\| $
    
    Finally, we can lower bound the inner product as 
    \begin{flalign}
        & & \langle\nabla\omega_\mathbf{r}, \mathrm{F}\mathbf{d}\rangle & = \|\nabla\omega_\mathbf{r}\| \|\mathrm{F}\mathbf{d}\| \cos(\theta) & \\
        & \Rightarrow & &\geq \|\nabla\omega_\mathbf{r}\| \frac{\delta^2}{\sqrt{m}W} & (\because \text{\eqref{eq:Fdlb} and \eqref{eq:coslb}}) \\
        & \Rightarrow & &= \|\nabla\omega_\mathbf{r}\|^2 \frac{\delta^2}{\|\nabla\omega_\mathbf{r}\|\sqrt{m}W} & \\
        & \therefore & &\geq \|\nabla\omega_\mathbf{r}\|^2 \frac{\delta^2}{\sqrt{m}W^2} & (\because \text{Assumption \ref{asm:comp}})
    \end{flalign}
    
    \hfill   
    \end{proof}
    
    \subsection{EPO Search for Constrained MOO}
    \dndbalconst*
    \begin{proof} 
      The statement is true by construction of the QP \eqref{eq:qp_x0_random}.
    \hfill   
    \end{proof}
    
    \dnddesconst*
    \begin{proof}
      The statement is true by construction of the QP \eqref{eq:qp_x0_random}.
    \hfill   
    \end{proof}
    
    \dndnc*
    \begin{proof}
     
    The proof for $\mathbf{x}^* \in \mathrm{Int}(\mathbb{X})$ is same as the unconstrained case in Theorem \eqref{th:d_nd_unconst}. 

    When $\mathbf{x}^* \in \partial \mathbb{X}$, the penetration assumption ensures that the orthogonal projection of the CSZ inequality-based balancing anchor direction $\mathbf{a}$\eqref{eq:cs_anchor} into the cone $\{\mathrm{F}\mathbf{d} \,|\, \mathbf{d}\in \mathcal{F}_\mathbb{X}(\mathbf{x}^*)\} \subset \mathbb{R}^m$ is non-zero, because $\mathbf{a}$ and $\mathbf{f}^*$ are orthogonal according to Claim \eqref{th:fa_cs=0}.

    \end{proof}
    
\section{Useful variations in EPO Search}
\label{sec:epo_var}

    In practice, for high-dimensional solution spaces $\mathbb{X}\subset\mathbb{R}^n$, e.g. DNN parameters, we use a fixed step size instead of adaptively deciding by line search. So, for proper movement in the objective space, we introduce few variations in the EPO search algorithm.
    
    \subsection{Momentum in Anchor While Tracing}
    While tracing the Pareto front, i.e. starting from a Pareto optimal $\mathbf{x}^0 \in \mathcal{P}$, we use the Cauchy-Schwarz anchor direction \eqref{eq:cs_anchor}, which is always perpendicular to the objective vector (see Claim \ref{th:fa_cs=0}). The first order change in the objective space created by the search direction found from QP \eqref{eq:qp_x0_po} is  $\delta\mathbf{f} = \mathrm{F}\mathbf{d}_{nd} = \mathrm{FF}^T\bm{\beta}^*$. This change $\delta \mathbf{f}$ is same as the orthogonal projection of $\mathbf{a}$ onto the cone $\mathrm{Col(FF}^T) \bigcap \mathrm{F}\mathcal{F}$. When the Pareto front is connected, at any point on the Pareto front, ${\mathbf{a}\in \mathrm{Col(FF}^T) \bigcap \mathrm{F}\mathcal{F}}$, so the magnitude of $\delta \mathbf{f}$ is significant enough to move the iterate with a small step size. But when, the Pareto front is disconnected, e.g. ZDT3 in Figure \ref{fig:zdt3}, and the iterate is at a boundary point outside the Pareto front,  $\mathbf{f}^t \in \partial \mathcal{O} - \mathbf{f}(\mathcal{P})$, then ${\mathbf{a}\notin \mathrm{Col(FF}^T) \bigcap \mathrm{F}\mathcal{F}}$, and the magnitude of its projection $\delta \mathbf{f}$ may not be enough to propel the iterate ahead with a small step size. The movements in objective slows down. To mitigate this we use a momentum term in the anchor,
    \begin{align}
    \mathbf{a}_m = \mathbf{a} + (\mathbf{f}^t - \mathbf{f}^{t-1}),
    \end{align}
    and use $\mathbf{a}_m$ in the QP. Using this anchor if the next iterate $\mathbf{f}^{t+1} \sim \mathbf{f}^t + \eta \Delta \mathbf{f}$ is dominated by the current one, i.e. $\mathbf{f}^{t+1} \succ \mathbf{f}^t$, then we conclude that $\mathbf{f}^t \in \partial\mathcal{O} - \mathbf{f}(\mathcal{P})$, and don't enter the descent mode in the subsequent iterations and only operate in the balance mode. As soon as a non-dominated iterate is found, i.e. $\mathbf{f}^{t+1} \nsucc \mathbf{f}^t$, we resume alternating the modes of operation.
    
    \subsubsection{Importance of Alternating Mode of Operation While Tracing.}\label{sec:trace_wo_des}
\begin{figure}[h]
        \centering
        \includegraphics{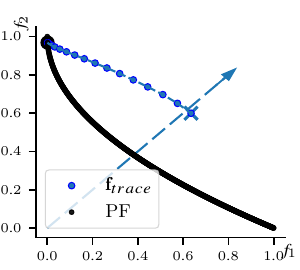}
        \caption{Tracing the Pareto front of ZDT1 without the descent mode: the trajectory drifts away from PF.}
        \label{fig:trace_wo_des}
    \end{figure}

    It is important to use the descent mode of operation in every other iteration to keep the iterate close to the Pareto front, especially in case of convex objectives. Otherwise the trajectory will drift away from the Pareto front. This is shown in Figure \ref{fig:trace_wo_des} for ZDT1 problem.
    \subsection{Restricting Trajectory in Descent Mode When not Tracing} \label{sec:restricted_descent}
\begin{figure}[h]
        \centering
	    \begin{subfigure}[b]{0.3\linewidth}
           \centering
            \includegraphics[width=\linewidth]{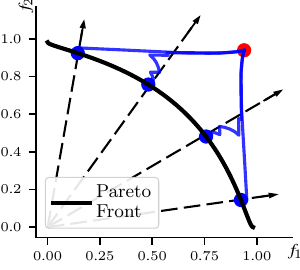}
            \caption{\label{fig:relaxed} Relaxed Descent}
        \end{subfigure}~
        \begin{subfigure}[b]{0.3\linewidth}
            \centering
            \includegraphics[width=\linewidth]{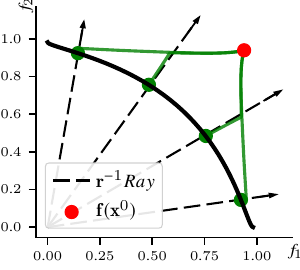}
            \caption{\label{fig:restricted}Restricted Descent}
        \end{subfigure}
%    	\vspace{-.3cm}
	    \caption{
	    Restricting the QP with the constraint \eqref{eq:res_des} eliminates fluctuations in descent mode.}
	    \label{fig:descent}
	\end{figure}

    When reaching the EPO solution starting from an arbitrary initialization, ideally the algorithm should enter the descent mode only when the iterate $\mathbf{f}^t$ reaches exactly onto the $\mathbf{r}^{-1}$ ray. Because, when $\omega_\mathbf{r}(\mathbf{f}^t) = 0$, the anchor direction of descent becomes $\overrightarrow{\mathbf{a}}=\overrightarrow{\mathbf{f}^t}=\overrightarrow{\mathbf{r}^{-1}}$, and the iterates descend along the $\mathbf{r}^{-1}$ to reach the EPO solution. 
    But achieving $\omega_\mathbf{r}(\mathbf{f}^t) = 0$ while using a fixed step size is less likely.
    Therefore we perform a descent mode operation whenever the objective vector $\mathbf{f}^t$ lies in the cone
    \begin{align} \label{eq:Meps}
	    \mathcal{M}^\mathbf{r}_{\epsilon} = \left\{\mathbf{f}\in \mathbb{R}^m_+ \;\middle|\; \omega_\mathbf{r}(\mathbf{f}) \le \epsilon\right\},
	\end{align}
	for a small $\epsilon > 0$. As a result, the descending anchor direction, and hence the first order change in objective space $\delta \mathbf{f} = \mathrm{FF}^T\bm{\beta}^*$, will no longer be aligned with the $\mathbf{r}^{-1}$ ray. This causes oscillations around the $\mathbf{r}^{-1}$ ray while descending, as shown in Figure \ref{fig:relaxed}. To mitigate this, we add the following equality constraint to the QP \eqref{eq:epo_qp} for unconstrained MOO and \eqref{eq:qp_x0_random} for constrained MOO:
    \begin{align}
        \delta \mathbf{f} &= \overrightarrow{\mathbf{r}^{-1}}\, \langle \overrightarrow{\mathbf{r}^{-1}}, \, \delta \mathbf{f} \rangle \nonumber\\
        \implies \mathrm{FF}^T \bm{\beta} &=  \overrightarrow{\mathbf{r}^{-1}} \, \overrightarrow{\mathbf{r}^{-1}}^T \mathrm{FF}^T \bm{\beta} \nonumber\\
        \implies (\mathbf{I}_m - \overrightarrow{\mathbf{r}^{-1}} \, \overrightarrow{\mathbf{r}^{-1}}^T) \mathrm{FF}^T \bm{\beta} &= 0, \label{eq:res_des}
    \end{align}
    where $\overrightarrow{\mathbf{r}^{-1}}$ is the $\ell_2$ normalized vector, and $\mathbf{I}_m$ is the $m\times m$ identity matrix. This constraint ensures that the movement in the objective space will be aligned with $\mathbf{r}^{-1}$ ray. We apply this equality constraint only when there are no active constraints, i.e. $\mathbf{x}^t \in \mathrm{Int}(\mathbb{X})$. The restriction in \eqref{eq:res_des} makes the trajectory of descent mode non-oscillatory as shown in Figure \ref{fig:restricted}. The objective functions used in Figure \ref{fig:descent} is described in section~\ref{sec:exp_toy_moo}.

	\subsection{Further Comparison with Pareto MTL}\label{sec:pmtl_info}
% 	The %above 
% 	restricted descent 
    Our EPO Search algorithm switches from balance mode to descent mode after entering a narrow conical region around the $\mathbf{r}^{-1}$ ray. This may appear similar to Pareto MTL \citep{NIPS2019pmtl} (described in section~\ref{sec:mtl_back}), where in the first phase one finds a solution $\mathbf{x}^0_r \in \Omega_k$, where
    \begin{align}\label{eq:decomp}
	   \Omega_k := \left\{\mathbf{x} \in \mathbb{R}^n \;\middle|\; \langle \mathbf{u}^k,\, \mathbf{f(x)} \rangle \, \ge\, \langle \mathbf{u}^{k'},\, \mathbf{f(x)} \rangle, \ \forall k'\neq k \right\},
	\end{align}
    such that the EPO solution is in $\Omega_k$, and in the second phase one does pure descent. The construction of $\Omega_k$ is such that $\mathbf{f}(\Omega_k)$ is also a cone.
    % as defined in \eqref{eq:decomp}. 

	However, their method does not guarantee that the outcome of second phase $\mathbf{x}^*$ also lies in $\Omega_k$. Because while descending, the objective vector may go outside the cone $\mathbf{f}(\Omega_k)$. On the other hand, our method guarantees that the objective vector of the final solution will be inside the cone $\mathcal{M}^\mathbf{r}_\epsilon$ in \eqref{eq:Meps}. Because, if in some iteration the $\mathbf{f}^t \notin \mathcal{M}^\mathbf{r}_\epsilon$, then a balancing anchor direction is used in the QP to bring it back inside the cone $\mathcal{M}^\mathbf{r}_\epsilon$ in the subsequent iterations. 
	
	Moreover, the angular fineness of their cone $\mathbf{f}(\Omega_k)$, which dictates the accuracy of the final solution, is dependent on how many reference vectors $\mathbf{u}^k, \; k=1,\cdots, K$ are used, which increases exponentially with the number of objectives $m$. On other hand, the angular fineness of our cone $\mathcal{M}^\mathbf{r}_\epsilon$ can be set by merely choosing a small value of $\epsilon$.

\section{Case Study: Personalized Medicine}
\label{sec:drug_casestudy}

Drug development is an elaborate, long and expensive process -- estimates show the average duration between discovery and market launch to be around 15 years \citep{ng2015drugs} and
the cost per approved new drug to be roughly \$2.6 billion in 2013 dollars
\citep{dimasi2016innovation}
and is continuing to increase \citep{kiriiri2020exploring}.
    The entire process consists of multiple stages that can be broadly categorized into: (i) Drug Discovery (ii) Pre-clinical Development (iii) Clinical Trials and (iv) Post-marketing surveillance \citep{huang2012principles,beninger2018pharmacovigilance}.
    The first stage involves finding and validating a biological entity (called ``target'', e.g., a gene) and a chemical (which may or may not be a previously used drug) such that their interaction has a therapeutic effect on the considered disease.
    After such a pair is found, pre-clinical studies are performed to understand the mechanism of drug action, e.g., through experimental studies on response of administering the drug on animals or cells in laboratory conditions.
    If a drug is found to be safe and efficacious, multiple clinical trials are conducted on increasing number of human subjects.
    Post-marketing surveillance or pharmacovigilance continues even after the drug is approved and in clinical use, to identify adverse side effects not found in previous stages.
    Each stage in turn has multiple steps, with several regulatory constraints and complex technical challenges; predictive models are utilized for many tasks at each stage \citep{vamathevan2019applications}.
    %We consider three prediction tasks, Drug Target, Drug Response, and Drug Side-effect prediction, described in more detail below, 
    %that are undertaken at 
    %from different stages of this process.
%    \subsubsection*{Drug Targets, Response, and Side Effects Prediction}

\subsection{Task Details}
    \begin{itemize}
    \item {\it Drug Target Prediction.}
    Drugs that act on specific disease-causing genes are instrumental for personalized medicine, and are actively studied \citep{schenone2013target}, particularly in early stages of drug discovery.
    Lab-based genomic techniques to find such drug targets are expensive, time-consuming and have high failure rates; hence, computational methods to predict new targets of drugs are used to prioritize drug-target pairs before subsequent testing \citep{chen2020idrug}.
    Large collections of experimental data of known drug targets have been made publicly available that can be used for supervised learning.
    Using such data, the binary classification task is to predict, for a given drug-gene pair, whether or not the gene can be targeted by the drug. %which is a binary classification problem.
    \item
    {\it Drug Response Prediction.}
    Here we consider a single disease, cancer, 
    that is caused by genetic aberrations leading to uncontrolled cell reproduction.
    Cells with differing genomic profiles may differ in their response to treatment \citep{senft2017precision}.
    To enable personalized cancer treatment based on individual genomic profiles,
    %requires deep knowledge of the genetic factors that influence the efficacy of drugs \citep{senft2017precision}.
    the effect of anti-cancer drugs on many cancer cells under laboratory conditions are being actively studied and documented \citep{Yang2013-vm,Rees2016-su}. 
    Broadly, in these experiments, cancer cells are subjected to varying concentrations of drugs, whose capacity to inhibit the reproduction of cancer cells is measured.
    A widely adopted measure of drug efficacy is half maximal inhibitory concentration (IC50) which is the concentration required to inhibit 50\% of the cells \citep{huang2012principles}.
    %; thus, lower the IC50 value, better the efficacy of the drug.
    Genomic profiles of cells can be constructed through various measurements and used as feature vectors, containing information relevant to cancer, such as mutations and indicators of gene activity.
    Experimental data on drug efficacy can be used to build a regression model that predicts, for a given drug and genomic profile, the efficacy of the drug on cells with such a profile.
    Such models can provide deeper insights into the mechanism of action of the drugs by uncovering the genomic factors that enable therapeutic action of drugs.
    %ad vector of expression values (which indicate biological activity) of the genes considered.
    They can also potentially be used to personalize drug recommendation for cancer patients, whose genomic profiles 
    %are routinely being for cancer care 
    can be measured in hospitals \citep{ma2021few}.
    %In such settings, it is also important to obtain the most effective drugs for a given genomic profile.
    For the genomic profile of a patient, and a given list of drugs (approved for use in the hospital), clinicians typically seek the top ranking drugs that are predicted to be most effective, which in turn can support subsequent decision-making for treatment planning.
    Thus, in addition to predicting the drug efficacy, it is also important to obtain accurate ranking of drugs based on efficacy for a given genomic profile.
    \item
    {\it Drug Side Effect Prediction.}
    Adverse drug events (ADEs) are unintended side effects of drugs that often lead to emergency visits, prolonged hospital stays, and worse patient outcomes \citep{ventola2018big}. 
    Worldwide, they remain a leading cause of morbidity and mortality, posing substantial clinical and economic burden \citep{watanabe2018cost}. 
    Clinical trials are limited by the number and characteristics of patients tested as well as the duration of the observation period, and they may not detect all ADEs, especially those with long latency or those that affect only certain patient groups \citep{coloma2013reference}.
    So, pharmacovigilance is routinely conducted to document reports of ADEs of approved drugs.
    Large databases have been created
    %, such as SIDER and OFFSIDES, 
    \citep{Kuhn2016-vy,Tatonetti2012-ml} that correct for biases and omissions in the reports due to concomitant medication, patient demographics and medical histories.
    This enables development of supervised learning models that can be used to predict potential ADEs of a drug, which can be further investigated by pharmacovigilance teams.
    The problem is that of binary classification, where given a disease-drug pair, the model predicts whether the disease may be a side-effect of a drug.
    \end{itemize}

\subsection{Data}
    Data from multiple publicly available drug-related databases have been collected and preprocessed by \citet{Jiang2020DrugOrchestraJP} for integrative analysis.
    For Drug Target (DT) prediction we use data from 
    STITCH \citep{Szklarczyk2016-ui}, DrugBank \citep{Wishart2018-fo} and Repurposing Hub \citep{repur};
    for Drug Response (DR) prediction, we use GDSC \citep{Yang2013-vm} and CCLE \citep{Rees2016-su} databases;
    for Drug Side-effect (DS) prediction, we use SIDER \citep{Kuhn2016-vy} and OFFSIDES \citep{Tatonetti2012-ml}. 
    %The three datasets used for Drug Target (DT) prediction, where an input is a drug gene-target pair and target (expected output) is a binary label {\color{blue} representing their association}, are STITCH \citep{Szklarczyk2016-ui}, DrugBank \citep{Wishart2018-fo} and Repurposing Hub \citep{repur}. 
    %The two datasets used for Drug Response (DR) prediction, where an input is a drug cell-line pair and target is the corresponding IC50 value (z-score), are GDSC \citep{Yang2013-vm} and CCLE \citep{Rees2016-su}. The two datasets used for Drug Side-effect (DS) prediction, where an input is a drug disease pair and target is a binary label {\color{blue}representing their association}, are SIDER \citep{Kuhn2016-vy}, OFFSIDE \citep{Tatonetti2012-ml}. 
    A summary of the datasets is given in Table \ref{tab:dsdetail}; more details can be found in 
    %For more details on the datasets, such as feature representation and drug overlap among the datasets, see 
    \cite{Jiang2020DrugOrchestraJP}.
  
\subsection{Additional Details of Model and Training}

We used ReLU activation function for all the layers, except the final layers of the predictors; sigmoid activation is used for DT and DS classifiers, and an identity map for DR regressor.
    % \red{layers, losses}.
The training hyper parameters used are identical for all methods: Adam optimizer with learning rate $0.001$, mini-batch size of $256$ for each training dataset, and $10$K iterations. 
We did not compare with PMTL, because, as noted in \cite{NIPS2019pmtl} and observed in our experiment in \S \ref{sec:exp_toy_moo}, it fails to scale for more than two objectives when the solution space is high-dimensional (in this case,
    %the dimensionality of our model's DNN parameter space is 
    more than 0.73 million).

\section{Additional Experimental Results}
    \label{sec:more_results}

    \subsection{Multi-Task Learning in Hydrometeorology}
    \label{sec:expt_regr}
    
    Flooding and other hydrological threats pose critical risks to lives and property, and can impact multiple industries such as agriculture, fishing, forestry, transportation and construction \citep{adams2016flood}.
    Billions of dollars are lost in major disasters, including floods, in the US alone \citep{disaster}.
    Timely river flow forecasts play a crucial role in mitigating the adverse effects of such events  \citep{chang2019flood}.
    River flow forecasts are also used for water supply management, reservoir operations, and navigation planning.
    Forecasts are required at strategic sites along rivers where flow levels along with other meteorological variables (e.g., rainfall and temperature) are regularly recorded.
    A predictive model for all the considered sites can be developed jointly using MTL, where a task is a site-specific prediction.
    
    We use the River Flow dataset  \citep{Spyromitros-Xioufis2016} that has $m=8$ tasks: predicting the flow at $8$ sites in the Mississippi River network.
    %  $48$ hours in the future. 
    Each sample %in the dataset 
    contains, for each site, the most recent and time-lagged flow measurements from $6, 12, 18, 24, 36, 48, 60$ hours in the past. 
    Thus, there are $64$ features and $8$ target variables.
    \begin{figure}[h]
        \centering
        \includegraphics[width=0.5\linewidth]{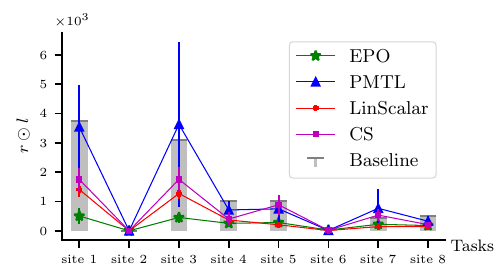}
        %\vspace{-.2cm}
        \caption{(Color Online) Comparison of mean RLP (with standard deviation; lower is better) of MTL methods after training the same neural network model to predict  flow at $8$ sites in the Mississippi River.}
        \vspace{-0.3cm}
        \label{fig:mtr}
    \end{figure}
    We remove samples with missing values and use $6,300$ samples for training and $2,700$ for testing. 
    We use a fully connected feed-forward neural network (FNN) with $4$ layers (layerwise sizes: $64\!\rightarrow\!32\!\rightarrow\!16\!\rightarrow\!8\!\rightarrow\!8$)  with $n=6,896$ parameters to fit the data. 
    We randomly choose $20$ input priority vectors $\mathbf{r}\in \mathbb{R}_+^8$ (with $\sum_j r_j = 1$) and train the FNN  using EPO search, PMTL, Linear Scalarization (LinScalar) and Chebychev Scalarization (CS).
    We use each of the $8$ objectives trained separately as baselines.
    % The same hyper-parameters are used for each method as done in the previous experiment. 
    We used Mean Squared Error (MSE) as the loss for each task.
    For all the methods
    %, including the baseline,
    stochastic gradient descent is used for training with the same hyperparameters: number of epochs, number of mini-batches and learning rate. 
    Since visualization is difficult for 8 dimensions, we compare the methods using the 
%    For comparison, we use the 
    relative loss profile (RLP) $\mathbf{r}\odot \mathbf{f}$ on the test data as
    %The mean and standard deviations of the RLP %$r_jl_j$s 
   %(lower is better) are
    shown in Figure~\ref{fig:mtr}. 
    
    We observe that EPO Search outperforms the other methods, indicating that it complies better with the input user priorities; the RLP of EPO search is more uniform (in the sense of definition \eqref{eq:non-uniformity}).    
    % Compared to the previous experiment with 2 tasks, 
    The improvement over PMTL is higher in this experiment compared to the e-commerce experiment in \ref{sec:exp_real_moo} with two tasks. 
    This is expected since the number of reference vectors required by PMTL, to reach a desired $\mathbf{r}^{-1}$ ray, grows exponentially with $m$. The problem with the min-max strategy of CS gets highlighted in this experiment showing how combining information from all the task gradients in each iteration is important in an MTL setup. LS does that, therefore performs better than CS in this experiment. 
    % Note that, here the preferences are randomly selected, not set disproportionately as in \ref{sec:exp_pharma}. 
    However, it does not comply with the priority specification, resulting in a non-uniform RLP in Figure \ref{fig:mtr}.
    Interestingly, except PMTL, all other MOO based MTL methods improve over the baseline which shows the advantage of MTL for correlated tasks
    % This experiment also 
    % shows that when the tasks are related, MTL outperforms 
    over learning each task independently: 
    predicting river flow at one site helps improving the prediction at other sites as all the sites are from the same river.

    \subsection{Multi-Task Learning in E-Commerce}
    \label{sec:exp_real_moo}
    %\subsubsection{Classification.}
    
    Fashion and lifestyle items, such as bags, footwear and apparel, constitute a large portion of e-commerce sales \citep{elahi2020fashion,deldjoo2022review}. Visual appearance of these items plays an influential role in purchase decisions. Search and recommendation engines in e-commerce websites, that are traditionally based on text-based keywords and descriptions, are increasingly using systems that can utilize images directly \citep{cheng2021fashion}. An important element in such systems is a classifier that can classify the input image into various categories, in order to restrict subsequent search and recommendation within the category. 
    In many cases the image may have multiple items, for instance in bundled sales, from the same or different categories.
    When multiple items are present in the input image 
    MTL can be effectively used, by considering each task as a classification problem for an item in the image \citep{Lin2020}.
    
     We use three benchmark classification datasets: (1) MultiMNIST, (2) MultiFashion, and (3) Multi-Fashion+MNIST.    
    In the MultiMNIST dataset \citep{NIPS2017_6975}, two images of different digits are randomly picked from the original MNIST dataset \citep{726791}, and combined to form a new image, where one is in the top-left and the other is in the bottom-right. There is zero padding in the top-right and bottom-left. 
    The MultiFashion dataset is generated in a similar manner from the FashionMNIST dataset \citep{DBLP:journals/corr/abs-1708-07747}. In Multi-Fashion+MNIST dataset, one image is from MNIST (top-left) and the other image is from FashionMNIST (bottom-right). 
    In each dataset, there are $120,000$ samples in the training set and $20,000$ samples in the test set. 
    %These are the same datasets used by \citet{NIPS2019pmtl}
    % \footnote{Downloaded from: \ \url{https://github.com/Xi-L/ParetoMTL}}.
    
    \begin{figure}[h]
        \centering
        \begin{subfigure}[b]{0.32\textwidth}
            \centering
            \includegraphics[width=\linewidth]{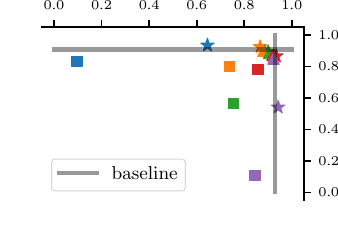}
            % \caption{\label{fig:mnist_acc}Multi-MNIST accuracies}
        \end{subfigure}
        \begin{subfigure}[b]{0.32\textwidth}
            \centering
            \includegraphics[width=\linewidth]{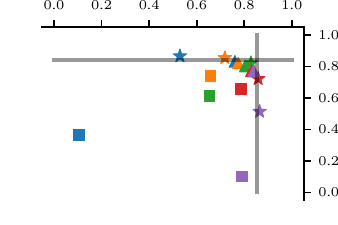}
            % \caption{\label{fig:fashion_acc}Multi-Fashion accuracies}
        \end{subfigure}
        \begin{subfigure}[b]{0.32\textwidth}
            \centering
            \includegraphics[width=\linewidth]{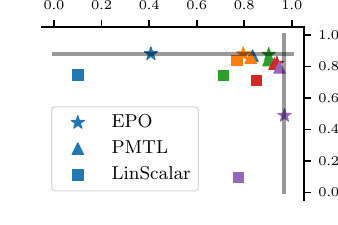}
            % \caption{\label{fig:fandm_acc} Multi Fashion$+$MNIST accuracies}
        \end{subfigure} \\ [-2mm]
        \begin{subfigure}[b]{0.32\textwidth}
            \centering
            \includegraphics[width=\linewidth]{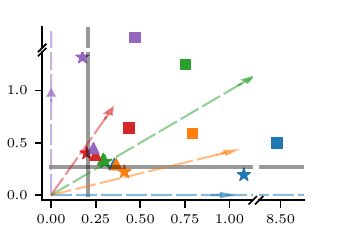}
            \caption{\label{fig:mnist_loss}Multi-MNIST}
        \end{subfigure}
        \begin{subfigure}[b]{0.32\textwidth}
            \centering
            \includegraphics[width=\linewidth]{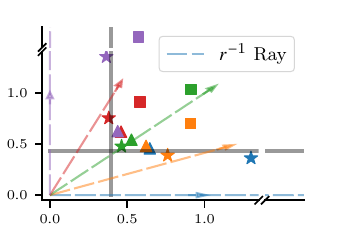}
            \caption{\label{fig:fashion_loss}Multi-Fashion}
        \end{subfigure}
        \begin{subfigure}[b]{0.32\textwidth}
            \centering
            \includegraphics[width=\linewidth]{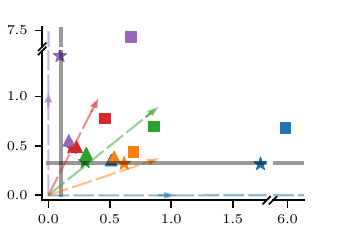}
            \caption{\label{fig:fandm_loss} Multi Fashion$+$MNIST}
        \end{subfigure}
      \caption{
      \label{fig:Multi-Fashion-MNIST} (Color Online) The top row show the accuracies, and the bottom row losses for 3 datasets. In each Figure, $x$ axis corresponds to task-1
      %, classifying the top-left image, 
      while $y$ axis corresponds to task-2.
      %, classifying the bottom-right image. 
      Different colors indicate different $\mathbf{r}^{-1}$ vectors, which are shown with corresponding $r^{-1}$ rays.
      EPO solutions have the highest per-task accuracy and are closest to the $\mathbf{r}^{-1}$ vectors.
      }
    \end{figure}
    
    For each dataset, there are two tasks: 1) classifying the top-left image, and 2) classifying the bottom-right image. Cross entropy losses are used for training. 
    %For a fair comparison with PMTL, 
    We use the same network (LeNet \citep{726791} with 31,910 trainable free parameters) used in \cite{NIPS2019pmtl}
    as the MTL neural network.
    The baseline for comparison is training the network for individual tasks. In addition we compare with the results from linear scalarization (LinScalar) and Chebyshev Scalarization. 
    % , which were worse than ours but better than PMTL and LinScalar. 
    
    For all the methods
    %, including the baseline,
    stochastic gradient descent is used for training with the same hyperparameters: number of epochs, number of mini-batches and learning rate. 
    We test the performance of all methods for the {\it same} 5 $\mathbf{r}^{-1}$ vectors, shown as rays in the bottom row of Figure~\ref{fig:Multi-Fashion-MNIST}. Ideal solutions should lie on these rays. Thus, each method has exactly 5 points corresponding to the test set losses in the bottom row and the top row shows the test set accuracies of the corresponding 5 DNN solutions.

    The results in Figure~\ref{fig:Multi-Fashion-MNIST}. show that the per-task accuracy of EPO search is higher than that of PMTL in every single run (top).
    The test set losses (bottom) show that the solutions from EPO search are closer to the corresponding $\mathbf{r}^{-1}$ vectors, compared to the solutions from PMTL. 
    %The preference specification of our method is same as that of CS and 
    For the reasons discussed in previous experiments, our method outperforms CS. 
    To avoid clutter in the Figures \ref{fig:Multi-Fashion-MNIST}, the results of CS are not shown.
    We observe that the performance of LinScalar is worse than all the other methods. 
    %Recall that search direction is a convex combination of the gradients. In PMTL and EPO search, this combination is optimally chosen in each iteration. In LinScalar it is fixed, in every iteration, to the input user preference (L1 normalized). This causes opposing gradients to cancel each other and decrease the magnitude of the resulting update. Thus, update magnitude of LinScalar is lesser than that of other methods in almost every iteration. With learning rate and number of iterations fixed across methods, update magnitude finally determines proximity to the Pareto front.

Apart from the domains, the 3 applications we consider also differ in terms of predictive models developed for the tasks, data usage, and subsequent use of the model. 
In application 1 (personalized medicine), the tasks considered are different and the corresponding predictive models are typically developed independently with  data sources that are partly shared and partly task-specific.
In contrast, in application 3 (e-commerce), the tasks are identical, use the same data sources for model building and, most often, are used together in subsequent applications.
Application 2 (hydrometeorology) is similar to the third one in terms of the prediction task and data used for model building. However, models for each task may or may not be used together subsequently.    

\section{Preference Elicitation}
\label{sec:pe_det}

PE methods may be classified based on
% %differ in models for
% %major aspects of PE are 
how the utility function is modeled. 
% and
% (b) selection of 
%b) how it is approximated after receiving each response to find  the next alternative in each iteration for the DM's consideration. 

%\subsubsection
{\bf Deterministic PE.} 
% In this category, both the aspects of PE 
Here the utility function is modelled deterministically, i.e., %The utility is modelled 
as a parametric function of the objective values. 
E.g., 
\cite{doi:10.1287/mnsc.22.6.652} 
used linear scalarization of the objectives as the utility function, which cannot model all Pareto optimal solutions as DM's preferred solution for non-convex MOO problems. 
Since LS cannot model all PO solutions for non-convex MOO problems, 
%the Chebyshev utility, based on CS, was developed and used in 
many interactive methods, such as 
\cite{10.1007/978-3-642-74919-3_8,doi:10.1287/mnsc.39.10.1255,dellkarwan1990,ozbeykarwan2014,REEVES19991311}, were developed that use Chebyshev scalarization as utility function due to its ability to model all the Pareto optimal solutions, even for non-convex objectives. 
Chebyshev scalarization can be considered as a proxy utility function to reach the best preferred alternative for the DM. 
These \textit{Weight Space Reduction Methods} reduce the parameter space of the utility, after every interaction, to be consistent with the DM's responses. 
However, interactive MCDM with Chebyshev utility is inefficient, because
%With Chebyshev utility
the reduced weight space consists of several disconnected components, which grows exponentially with the number of interactions with the DM. Therefore, selecting a weight from the fragmented weight space to probe for the next alternative 
%(that should be different from the existing set of solutions) 
becomes computationally expensive %, and unrealistic 
for large and complex problems \citep{Miettinen1998}.

Another deterministic PE approach is 
Preference Robust Optimization (PRO), where 
the set of all possible utility functions are restricted to only those that are consistent with the previous pair-wise comparisons, and the worst utility in this restricted set is optimized to find the next alternative. 
E.g., \cite{https://doi.org/10.48550/arxiv.2003.01899} model the utility as a linear function and \cite{haskell2018preference} model it as a quasi-concave function, using several support (``hockey stick'') functions, 
where the number of parameters of the utility function grows linearly with the number of interactions. PRO methods assume the MOO to be convex and the utility to be quasi-convex in the most general case. They cannot model non-convex PE problems, where either the MOO problem or the utility function could be non-convex. 
% Different deterministic models of the utility function 
% \begin{itemize}
%     \item multilinear form: Le and Haddawy, 1998, AAAI
% \end{itemize}

%\subsubsection
{\bf Probabilistic PE.} 
Here, the utility function is modelled using probabilistic methods. 
There are parametric models of utility for probabilistic PE, e.g., \cite{doi:10.1287/mnsc.2014.2059} which does not model non-convex utility functions.
In contrast, non-parametric methods, such as Gaussian processes can model any class of functions -- see \S \ref{sec:gp} and \ref{sec:interactive_back}. 

% and to the best of our knowledge has been utilized in single criteria decision making (SOO) problems.

% considers parametric modelling of utility function for probabilistic PE in the context of single criteria decision making (SOO) problems. However, it cannot model non-convex utility functions.

%\blue{The Gaussian Process based probabilistic PE is described in \S \ref{sec:interactive_back}} 

\section{Bayesian optimization with Gaussian Process}
\label{sec:gp}

Bayesian optimization (BO) is a sequential model-based approach for solving black-box function optimization problems (see, e.g. \cite{shahriari2016taking}). 
The key idea is to learn a {\it surrogate probabilistic model} $P(u)$ 
that captures our beliefs about the unknown %utility 
function $u(\textbf{x})$.
The functional form of $u$ is unknown but it is assumed that the function can be evaluated at any given point $\textbf{x}$.
The model is learnt from {\it data}, $\mathcal{D}_t$ = $\{(\textbf{x}^1,u(\mathbf{x}^1),\ldots,(\textbf{x}^t,u(\textbf{x}^t)\}$, that consists of sequential evaluations of $u(\textbf{x})$ for different values of $\textbf{x}$.\footnote{Note, in our setup of PE, we do not have direct evaluation of the utility at a point $\mathbf{x}^t$. Instead the data is in the form of pairwise comparisons $\mathcal{D}_t = \{(\mathbf{x}^i, \mathbf{x}^j)\}$ such that $u(\mathbf{x}^i) > u(\mathbf{x}^j)$.}

Generating this data sequence requires making the decision of which $\textbf{x}$ to evaluate next, at each step.
This decision is made through an {\it acquisition function} $\alpha$.
These functions are designed to have optima at points with high uncertainty in the surrogate model (thus facilitating exploration) and/or at points with high predictive values in the surrogate model (thus facilitating exploitation).
Acquisition functions have known functional forms and are usually easier to optimize than the original objective function.
%Algorithm \ref{algo:bo} (adapted from \cite{shahriari2016taking}) shows the general framework of Bayesian Optimization for generating sequential data ${(\textbf{p}_1,\textbf{L}_1),\ldots,(\textbf{p}_n,\textbf{L}_n)}$ while sequentially learning the surrogate model $\mathcal{M}$.
The surrogate model is updated sequentially with each observed data point.
Over multiple steps, the landscape of the black-box function $u(\textbf{x})$ is learnt by the surrogate model and can be exploited by the acquisition function to yield values of $\textbf{x}$ that are, on average, closer to the optimal $\textbf{x}^*$.

Many different choices of surrogate models and acquisition functions have been explored. 
A Gaussian process (GP) may be used to model priors over functions. 
GPs can be viewed as an infinite-dimensional extension to a multivariate Gaussian distribution \citep{rasmussen2004gaussian}, and can approximate general non-linear functions.
A $\mathrm{GP}(\mu, \kappa)$ is specified by a mean function $\mu:\mathbb{R}^m \rightarrow \mathbb{R}$ and a covariance function $\kappa:\mathbb{R}^m \times \mathbb{R}^m \rightarrow \mathbb{R}_+$, e.g., Gaussian or Mat\'ern kernel.
This models the uncertainty in the unknown objective function value at a particular solution $\mathbf{x}^i$ as a Gaussian distribution, $u(\mathbf{f}^i) \sim \mathcal{N}\left(\mu(\mathbf{f}^i), \kappa(\mathbf{f}^i, \mathbf{f}^i)\right)$. 
The joint probability of the function values at more than one solution, e.g., $[\mathbf{x}^i, \mathbf{x}^j]^T$, is modelled as multivariate Gaussian, $[u(\mathbf{f}^i), u(\mathbf{f}^j)]^T \sim \mathcal{N}([\mu(\mathbf{f}^i), \mu(\mathbf{f}^j)]^T, K(\mathbf{f}^i, \mathbf{f}^j))$, where $K$ is the gram matrix of kernel $\kappa$ evaluated at $\mathbf{f}^i, \mathbf{f}^j$.

A common choice for the acquisition function is {\it Expected Improvement (EI)} \citep{jones2001taxonomy}, that has a closed form for GP, does not require its own tuning parameter and has been shown to perform well 
% in minimization settings 
\citep{snoek2012practical}.
%EI is the expectation that $\alpha$ will exceed (negatively) some threshold, 
EI is the expectation that $\textbf{x}^{t+1}$ will improve $u$ 
% (negatively, as we would like to minimize a loss) 
over ${\textbf{x}^{t}}^{*}$ which is 
%chosen to be 
the best observation from $t$ steps of BO so far, 
%i.e. $\alpha^t^{best} = \min_{i \leq n} \alpha_i(\mathcal{D}_n,\mathcal{M})$, and
i.e. ${\textbf{x}^{t}}^{*} = \argmax_{\textbf{x}^{i \leq t}} u(\textbf{x}^i)$, and
%\begin{eqnarray*}
%$EI_n = E_n[\max{\{(\alpha_{n}^{best} - \alpha_n), 0\}}]$,
%\end{eqnarray*}
$EI^t(\textbf{x}^{t+1}) = \mathbb{E}_t\left[\max\{u({\textbf{x}^{t}}^{*}) - u(\textbf{x}^{t+1}), 0\}\right]$,
where the expectation $\mathbb{E}_t$ is under the posterior distribution given evaluations of 
%$\alpha$ 
$u$ at $\textbf{x}^1,\dots,\textbf{x}^t$. The next value is chosen by $\textbf{x}^{t+1} = \argmax EI^t(\textbf{p}_{n+1})$.
For a GP with predictive variance $\kappa^t(\mathbf{x}) = \kappa(\textbf{x};\mathcal{D}_t)$ and predictive mean $\mu^t(\mathbf{x}) = \mu(\textbf{x};\mathcal{D}_t)$: 
%where $p_1,\dots,p_n$ are scalars, 
%this is given by:
\begin{equation}
\label{eq:EI}
%EI_n = \sigma [\gamma \Phi(\gamma) + \phi(\gamma)] \text{ where } \gamma = (\alpha_{n}^{best} -\mu)/\sigma,  
%\alpha_n(\textbf{p}_{n+1}) = 
EI^t(\textbf{x}^{t+1}) = \kappa [\gamma(\textbf{x}^{t+1}) \Phi(\gamma(\textbf{x}^{t+1})) + \phi(\gamma(\textbf{x}^{t+1}))]
\end{equation}

where, $\gamma(\textbf{x}^{t+1}) = (u({\textbf{x}^{t}}^{*}) -\mu^t(\textbf{x}^{t+1}))/\sigma^t(\textbf{x}^{t+1})$, and $\Phi$ and $\phi$ denote the CDF and PDF of the standard normal distribution respectively.

\section{Optimization in Neural Networks}
\label{sec:nn}
% \red{EXPAND}: 
A neural network is a parametric function $L: \mathbb{R}^{d_I} \rightarrow \mathbb{R}^{d_T}$, where $d_I$ is the input dimension and $d_T$ is the target dimension, created by composition of multiple constituent functions $L_k$ represented by layers in the network. 
For example a three layered network can be written as
\begin{align}
    L(\mathbf{x}; \theta_1, \theta_2, \theta_3) = L_3(L_2(L_1(\mathbf{x}; \theta_1); \theta_2); \theta_3),
\end{align}
where $\theta_k$ is the set of parameters for the $k^\text{th}$ layer $L_k:\mathbb{R}^{d_{k-1}}\rightarrow\mathbb{R}^{d_{k}}$, $d_{k-1}$ and $d_{k}$ are the input and output dimensions of the $k^\text{th}$ layer (here $d_0=d_I$, and $d_3=d_T$). A layer $L_k$ performs an affine transformation followed by an elementwise nonlinear transformation, e.g., sigmoid or ReLU. 
The function parameters $\Theta = \{\theta_k\}$ (aka network weights) are learnt by optimizing the training objective which is determined by the learning task and targets given by a training dataset 
${\mathcal{D} =\{(\mathbf{x}^i,\mathbf{y}^i)\}_{i=1}^N}$:
\begin{align}
    \min_\Theta f(\Theta; \mathcal{D}) = \sum_{i=1}^N l(L(\mathbf{x}^i; \Theta), \mathbf{y}^i),
\end{align}
where $l$ is a differentiable loss function measuring the deviation of the output $L(\mathbf{x}; \Theta)$ from the expected value $\mathbf{y}$. Common loss functions include the mean-squared error loss for regression tasks and cross-entropy loss for classification tasks.
Note, here $\mathbf{x}_i$'s are not the variables of optimization, they are input data features and $\mathbf{y}_i$'s are the targets. It is customary in the neural network literature to denote the model parameters, which are the optimization variables, as $\theta$. 
% Therefore the solution space (or decision space) is $\Theta\subset\mathbb{R}^n$ where $n$ is the total number of parameters. 
% Common loss functions include the mean-squared error loss for regression tasks and cross-entropy loss for classification tasks.
%Note that in optimization for machine learning models, interest lies in local optima that can lead to model generalization over unseen test data rather than global optima of the functions.
A neural network with three or more layers is generally considered deep. 
Training such a model requires learning large number of parameters, which
makes it prohibitively expensive to use second-order methods. 
First order optimization techniques, based on gradient descent, are most widely used.
%and can effectively train deep networks.
More details can be found in books on deep learning, e.g., by \cite{Goodfellow-et-al-2016}.

\subsection{Deep Neural Networks for Multi-task Learning} 
In MTL, there are more than one training objectives each stemming from a task. The MTL datasets may have multiple targets, for different tasks, e.g., $\mathcal{D}= \{(\mathbf{x}^i, \mathbf{y}^i_1, \mathbf{y}^i_2, \cdots, \mathbf{y}^i_m)\}_{i=1}^N$ has targets for $m$ different tasks and a common input for all of them, like in our experiments in \S \ref{sec:more_results}. Another example of a MTL dataset is $\mathcal{D}= \{(\mathbf{x}^i_s, \mathbf{x}^i_1, \cdots, \mathbf{x}^i_m, \mathbf{y}^i_1, \mathbf{y}^i_2, \cdots, \mathbf{y}^i_m)\}_{i=1}^N$, where there is a common input $\mathbf{x}_s$ for all the tasks, task specific inputs $\mathbf{x}_j$,  $j\in[m]$ for each task, and the corresponding targets  $\mathbf{y}_j$,  $j\in[m]$, like in our experiment in \S \ref{sec:expt_real}.

MTL models are designed according to the format of data. For example, the model in Figure~\ref{fig:mtl_moo} can be considered as two parametric functions designed for the first type of MTL dataset:
\begin{align}
    \mathbf{L}(\mathbf{x}; \Theta) = \begin{bmatrix}
                                L_1(L_s(\mathbf{x}; \theta_s); \theta_1) \\
                                L_2(L_s(\mathbf{x}; \theta_s); \theta_2)
                            \end{bmatrix},
\end{align}
where  $\theta_s$ is the network parameter for embedding the common input, and $\theta_j$'s are for the task specific layers. The $m$ simultaneous optimization problems for training this model is given by
\begin{align}
    \min_\Theta f_j(\Theta; \mathcal{D}) = \sum_{i=1}^N l_j(L(\mathbf{x}^i; \Theta), \mathbf{y}^i_j),  \quad \forall j\in[m],
\end{align}
where $l_j$'s are task-specific loss functions.

% \putbib    
% \end{bibunit}

\end{document}

%% file: diagrams/admissible_sets_lgrn.pdf_tex
%% Creator: Inkscape 1.2.1 (9c6d41e4, 2022-07-14), www.inkscape.org
%% PDF/EPS/PS + LaTeX output extension by Johan Engelen, 2010
%% Accompanies image file 'admissible_sets_lgrn.pdf' (pdf, eps, ps)
%%
%% To include the image in your LaTeX document, write
%%   \input{<filename>.pdf_tex}
%%  instead of
%%   \includegraphics{<filename>.pdf}
%% To scale the image, write
%%   \def\svgwidth{<desired width>}
%%   \input{<filename>.pdf_tex}
%%  instead of
%%   \includegraphics[width=<desired width>]{<filename>.pdf}
%%
%% Images with a different path to the parent latex file can
%% be accessed with the `import' package (which may need to be
%% installed) using
%%   \usepackage{import}
%% in the preamble, and then including the image with
%%   \import{<path to file>}{<filename>.pdf_tex}
%% Alternatively, one can specify
%%   \graphicspath{{<path to file>/}}
%% 
%% For more information, please see info/svg-inkscape on CTAN:
%%   http://tug.ctan.org/tex-archive/info/svg-inkscape
%%
\begingroup%
  \makeatletter%
  \providecommand\color[2][]{%
    \errmessage{(Inkscape) Color is used for the text in Inkscape, but the package 'color.sty' is not loaded}%
    \renewcommand\color[2][]{}%
  }%
  \providecommand\transparent[1]{%
    \errmessage{(Inkscape) Transparency is used (non-zero) for the text in Inkscape, but the package 'transparent.sty' is not loaded}%
    \renewcommand\transparent[1]{}%
  }%
  \providecommand\rotatebox[2]{#2}%
  \newcommand*\fsize{\dimexpr\f@size pt\relax}%
  \newcommand*\lineheight[1]{\fontsize{\fsize}{#1\fsize}\selectfont}%
  \ifx\svgwidth\undefined%
    \setlength{\unitlength}{230.26268906bp}%
    \ifx\svgscale\undefined%
      \relax%
    \else%
      \setlength{\unitlength}{\unitlength * \real{\svgscale}}%
    \fi%
  \else%
    \setlength{\unitlength}{\svgwidth}%
  \fi%
  \global\let\svgwidth\undefined%
  \global\let\svgscale\undefined%
  \makeatother%
  \begin{picture}(1,0.71976808)%
    \lineheight{1}%
    \setlength\tabcolsep{0pt}%
    \put(0,0){\includegraphics[width=\unitlength,page=1]{admissible_sets_lgrn.pdf}}%
    \put(1.00869251,0.6402467){\color[rgb]{0,0,0}\makebox(0,0)[lt]{\begin{minipage}{0.19303764\unitlength}\raggedright $\mathcal{V}_{\preccurlyeq \mathbf{f}^t} $\end{minipage}}}%
    \put(0.2066815,0.14617474){\color[rgb]{0,0,0}\makebox(0,0)[lt]{\begin{minipage}{0.28156607\unitlength}\end{minipage}}}%
    \put(0,0){\includegraphics[width=\unitlength,page=2]{admissible_sets_lgrn.pdf}}%
    \put(1.00699366,0.47017594){\color[rgb]{0,0,0}\makebox(0,0)[lt]{\begin{minipage}{0.19496271\unitlength}\raggedright $\mathcal{M}^\mathbf{r}_{\mathbf{f}^t} $\end{minipage}}}%
    \put(0,0){\includegraphics[width=\unitlength,page=3]{admissible_sets_lgrn.pdf}}%
    \put(0.48235182,0.38978643){\color[rgb]{0,0,0}\makebox(0,0)[lt]{\begin{minipage}{0.20827366\unitlength}\raggedright $\mathbf{f}^t$\end{minipage}}}%
    \put(0.20984059,0.27526242){\color[rgb]{0,0,0}\makebox(0,0)[lt]{\begin{minipage}{0.25001576\unitlength}\raggedright $\mathbf{f}^*$\end{minipage}}}%
    \put(0.51510553,0.73204216){\color[rgb]{0,0,0}\makebox(0,0)[lt]{\begin{minipage}{0.42310028\unitlength}\raggedright $\mathbf{r}^{-1}$ Ray \end{minipage}}}%
    \put(0,0){\includegraphics[width=\unitlength,page=4]{admissible_sets_lgrn.pdf}}%
    \put(1.01700367,0.34012184){\color[rgb]{0,0,0}\makebox(0,0)[lt]{\begin{minipage}{0.36750485\unitlength}\raggedright $\widecheck{\mathbf{f}}^t$\end{minipage}}}%
    \put(0.49080314,0.65899776){\color[rgb]{0,0,0}\makebox(0,0)[lt]{\begin{minipage}{0.36750485\unitlength}\raggedright $\widecheck{\mathbf{f}}^t$\end{minipage}}}%
    \put(1.01343091,0.11767781){\color[rgb]{0,0,0}\makebox(0,0)[lt]{\begin{minipage}{0.19872363\unitlength}\raggedright $\mathcal{A}^\mathbf{r}_{\mathbf{f}^t}$\end{minipage}}}%
    \put(0,0){\includegraphics[width=\unitlength,page=5]{admissible_sets_lgrn.pdf}}%
    \put(0.2813938,0.20693121){\color[rgb]{0,0,0}\makebox(0,0)[lt]{\begin{minipage}{0.11201656\unitlength}\raggedright Pareto\\ Front\end{minipage}}}%
    \put(0.74954254,0.2190473){\color[rgb]{0,0,0}\makebox(0,0)[lt]{\begin{minipage}{0.57102796\unitlength}\end{minipage}}}%
    \put(0,0){\includegraphics[width=\unitlength,page=6]{admissible_sets_lgrn.pdf}}%
    \put(0.02988829,0.73725721){\color[rgb]{0,0,0}\makebox(0,0)[lt]{\begin{minipage}{0.1109754\unitlength}\raggedright $f_2$\end{minipage}}}%
    \put(0.7763829,0.06919206){\color[rgb]{0,0,0}\makebox(0,0)[lt]{\begin{minipage}{0.11363105\unitlength}\raggedright $f_1$\end{minipage}}}%
  \end{picture}%
\endgroup%

%% file: diagrams/penetration.pdf_tex
%% Creator: Inkscape 1.1 (c4e8f9e, 2021-05-24), www.inkscape.org
%% PDF/EPS/PS + LaTeX output extension by Johan Engelen, 2010
%% Accompanies image file 'penetration.pdf' (pdf, eps, ps)
%%
%% To include the image in your LaTeX document, write
%%   \input{<filename>.pdf_tex}
%%  instead of
%%   \includegraphics{<filename>.pdf}
%% To scale the image, write
%%   \def\svgwidth{<desired width>}
%%   \input{<filename>.pdf_tex}
%%  instead of
%%   \includegraphics[width=<desired width>]{<filename>.pdf}
%%
%% Images with a different path to the parent latex file can
%% be accessed with the `import' package (which may need to be
%% installed) using
%%   \usepackage{import}
%% in the preamble, and then including the image with
%%   \import{<path to file>}{<filename>.pdf_tex}
%% Alternatively, one can specify
%%   \graphicspath{{<path to file>/}}
%% 
%% For more information, please see info/svg-inkscape on CTAN:
%%   http://tug.ctan.org/tex-archive/info/svg-inkscape
%%
\begingroup%
  \makeatletter%
  \providecommand\color[2][]{%
    \errmessage{(Inkscape) Color is used for the text in Inkscape, but the package 'color.sty' is not loaded}%
    \renewcommand\color[2][]{}%
  }%
  \providecommand\transparent[1]{%
    \errmessage{(Inkscape) Transparency is used (non-zero) for the text in Inkscape, but the package 'transparent.sty' is not loaded}%
    \renewcommand\transparent[1]{}%
  }%
  \providecommand\rotatebox[2]{#2}%
  \newcommand*\fsize{\dimexpr\f@size pt\relax}%
  \newcommand*\lineheight[1]{\fontsize{\fsize}{#1\fsize}\selectfont}%
  \ifx\svgwidth\undefined%
    \setlength{\unitlength}{797.53916823bp}%
    \ifx\svgscale\undefined%
      \relax%
    \else%
      \setlength{\unitlength}{\unitlength * \real{\svgscale}}%
    \fi%
  \else%
    \setlength{\unitlength}{\svgwidth}%
  \fi%
  \global\let\svgwidth\undefined%
  \global\let\svgscale\undefined%
  \makeatother%
  \begin{picture}(1,0.49011325)%
    \lineheight{1}%
    \setlength\tabcolsep{0pt}%
    \put(0,0){\includegraphics[width=\unitlength,page=1]{penetration.pdf}}%
    \put(0.07218293,0.44831654){\makebox(0,0)[lt]{\lineheight{1.25}\smash{\begin{tabular}[t]{l}$f_2$\end{tabular}}}}%
    \put(0.9378052,0.05441653){\makebox(0,0)[lt]{\lineheight{1.25}\smash{\begin{tabular}[t]{l}$f_1$\end{tabular}}}}%
    \put(0.20879162,0.32401718){\makebox(0,0)[lt]{\lineheight{1.25}\smash{\begin{tabular}[t]{l}$\mathbf{f}^1$\end{tabular}}}}%
    \put(0.46615488,0.17753386){\makebox(0,0)[lt]{\lineheight{1.25}\smash{\begin{tabular}[t]{l}$\mathbf{f}^2$\end{tabular}}}}%
    \put(0.42454138,0.33077441){\makebox(0,0)[lt]{\lineheight{1.25}\smash{\begin{tabular}[t]{l}$\mathbf{f}^3$\end{tabular}}}}%
    \put(0,0){\includegraphics[width=\unitlength,page=2]{penetration.pdf}}%
    \put(0.74611959,0.11346882){\makebox(0,0)[lt]{\lineheight{1.25}\smash{\begin{tabular}[t]{l}$\mathbf{f}^4$\end{tabular}}}}%
    \put(0.69893371,0.38543474){\makebox(0,0)[lt]{\lineheight{1.25}\smash{\begin{tabular}[t]{l}$\mathcal{O}$\end{tabular}}}}%
  \end{picture}%
\endgroup%

%% file: diagrams/admissible_sets_lgrn_corner.pdf_tex
%% Creator: Inkscape 1.2.1 (9c6d41e4, 2022-07-14), www.inkscape.org
%% PDF/EPS/PS + LaTeX output extension by Johan Engelen, 2010
%% Accompanies image file 'admissible_sets_lgrn_corner.pdf' (pdf, eps, ps)
%%
%% To include the image in your LaTeX document, write
%%   \input{<filename>.pdf_tex}
%%  instead of
%%   \includegraphics{<filename>.pdf}
%% To scale the image, write
%%   \def\svgwidth{<desired width>}
%%   \input{<filename>.pdf_tex}
%%  instead of
%%   \includegraphics[width=<desired width>]{<filename>.pdf}
%%
%% Images with a different path to the parent latex file can
%% be accessed with the `import' package (which may need to be
%% installed) using
%%   \usepackage{import}
%% in the preamble, and then including the image with
%%   \import{<path to file>}{<filename>.pdf_tex}
%% Alternatively, one can specify
%%   \graphicspath{{<path to file>/}}
%% 
%% For more information, please see info/svg-inkscape on CTAN:
%%   http://tug.ctan.org/tex-archive/info/svg-inkscape
%%
\begingroup%
  \makeatletter%
  \providecommand\color[2][]{%
    \errmessage{(Inkscape) Color is used for the text in Inkscape, but the package 'color.sty' is not loaded}%
    \renewcommand\color[2][]{}%
  }%
  \providecommand\transparent[1]{%
    \errmessage{(Inkscape) Transparency is used (non-zero) for the text in Inkscape, but the package 'transparent.sty' is not loaded}%
    \renewcommand\transparent[1]{}%
  }%
  \providecommand\rotatebox[2]{#2}%
  \newcommand*\fsize{\dimexpr\f@size pt\relax}%
  \newcommand*\lineheight[1]{\fontsize{\fsize}{#1\fsize}\selectfont}%
  \ifx\svgwidth\undefined%
    \setlength{\unitlength}{136.72062551bp}%
    \ifx\svgscale\undefined%
      \relax%
    \else%
      \setlength{\unitlength}{\unitlength * \real{\svgscale}}%
    \fi%
  \else%
    \setlength{\unitlength}{\svgwidth}%
  \fi%
  \global\let\svgwidth\undefined%
  \global\let\svgscale\undefined%
  \makeatother%
  \begin{picture}(1,0.92196052)%
    \lineheight{1}%
    \setlength\tabcolsep{0pt}%
    \put(0,0){\includegraphics[width=\unitlength,page=1]{admissible_sets_lgrn_corner.pdf}}%
    \put(0.31734945,0.84331414){\color[rgb]{0,0,0}\makebox(0,0)[lt]{\begin{minipage}{0.32835307\unitlength}\raggedright $\mathcal{M}^\mathbf{r}_{\mathbf{f}^*} $\end{minipage}}}%
    \put(0.67329141,0.16468104){\color[rgb]{0,0,0}\makebox(0,0)[lt]{\begin{minipage}{0.79567061\unitlength}\raggedright {\color{blue}$\mathrm{Col}(\mathrm{F}\mathrm{F}^T)$}\end{minipage}}}%
    \put(0.34884627,0.34597622){\color[rgb]{0,0,0}\makebox(0,0)[lt]{\begin{minipage}{0.35942408\unitlength}\raggedright {\color{yellow}$-\mathbf{v}_1$}\end{minipage}}}%
    \put(0,0){\includegraphics[width=\unitlength,page=2]{admissible_sets_lgrn_corner.pdf}}%
    \put(0.6953118,0.43112003){\color[rgb]{0,0,0}\makebox(0,0)[lt]{\begin{minipage}{0.42107253\unitlength}\raggedright $\mathbf{f}^*$\end{minipage}}}%
    \put(0.07988819,0.98249584){\color[rgb]{0,0,0}\makebox(0,0)[lt]{\begin{minipage}{0.46406699\unitlength}\raggedright $\mathbf{r}^{-1}$ Ray \end{minipage}}}%
    \put(0.6953118,0.43112003){\color[rgb]{0,0,0}\makebox(0,0)[lt]{\begin{minipage}{0.42107253\unitlength}\raggedright $\mathbf{f}^*$\end{minipage}}}%
    \put(0.11842633,0.75332824){\color[rgb]{0,0,0}\makebox(0,0)[lt]{\begin{minipage}{0.30530812\unitlength}\raggedright $\mathbf{f}_\mathbf{r}^*$\end{minipage}}}%
    \put(0,0){\includegraphics[width=\unitlength,page=3]{admissible_sets_lgrn_corner.pdf}}%
    \put(0.16269694,0.45311084){\color[rgb]{0,0,0}\makebox(0,0)[lt]{\begin{minipage}{0.30530812\unitlength}\raggedright {\color{red}$-\mathbf{a}$}\end{minipage}}}%
    \put(0,0){\includegraphics[width=\unitlength,page=4]{admissible_sets_lgrn_corner.pdf}}%
    \put(-0.01315493,0.79108677){\color[rgb]{0,0,0}\makebox(0,0)[lt]{\begin{minipage}{0.186903\unitlength}\raggedright $f_2$\end{minipage}}}%
    \put(0.84081705,0.08845298){\color[rgb]{0,0,0}\makebox(0,0)[lt]{\begin{minipage}{0.1913756\unitlength}\raggedright $f_1$\end{minipage}}}%
    \put(0,0){\includegraphics[width=\unitlength,page=5]{admissible_sets_lgrn_corner.pdf}}%
    \put(0.7259699,0.32350128){\color[rgb]{0,0,0}\rotatebox{-15.47678444}{\makebox(0,0)[lt]{\lineheight{1.25}\smash{\begin{tabular}[t]{l}Pareto Front\end{tabular}}}}}%
    \put(0,0){\includegraphics[width=\unitlength,page=6]{admissible_sets_lgrn_corner.pdf}}%
  \end{picture}%
\endgroup%

%% file: diagrams/admissible_sets_kl_corner.pdf_tex
%% Creator: Inkscape 1.2.1 (9c6d41e4, 2022-07-14), www.inkscape.org
%% PDF/EPS/PS + LaTeX output extension by Johan Engelen, 2010
%% Accompanies image file 'admissible_sets_kl_corner.pdf' (pdf, eps, ps)
%%
%% To include the image in your LaTeX document, write
%%   \input{<filename>.pdf_tex}
%%  instead of
%%   \includegraphics{<filename>.pdf}
%% To scale the image, write
%%   \def\svgwidth{<desired width>}
%%   \input{<filename>.pdf_tex}
%%  instead of
%%   \includegraphics[width=<desired width>]{<filename>.pdf}
%%
%% Images with a different path to the parent latex file can
%% be accessed with the `import' package (which may need to be
%% installed) using
%%   \usepackage{import}
%% in the preamble, and then including the image with
%%   \import{<path to file>}{<filename>.pdf_tex}
%% Alternatively, one can specify
%%   \graphicspath{{<path to file>/}}
%% 
%% For more information, please see info/svg-inkscape on CTAN:
%%   http://tug.ctan.org/tex-archive/info/svg-inkscape
%%
\begingroup%
  \makeatletter%
  \providecommand\color[2][]{%
    \errmessage{(Inkscape) Color is used for the text in Inkscape, but the package 'color.sty' is not loaded}%
    \renewcommand\color[2][]{}%
  }%
  \providecommand\transparent[1]{%
    \errmessage{(Inkscape) Transparency is used (non-zero) for the text in Inkscape, but the package 'transparent.sty' is not loaded}%
    \renewcommand\transparent[1]{}%
  }%
  \providecommand\rotatebox[2]{#2}%
  \newcommand*\fsize{\dimexpr\f@size pt\relax}%
  \newcommand*\lineheight[1]{\fontsize{\fsize}{#1\fsize}\selectfont}%
  \ifx\svgwidth\undefined%
    \setlength{\unitlength}{136.72062551bp}%
    \ifx\svgscale\undefined%
      \relax%
    \else%
      \setlength{\unitlength}{\unitlength * \real{\svgscale}}%
    \fi%
  \else%
    \setlength{\unitlength}{\svgwidth}%
  \fi%
  \global\let\svgwidth\undefined%
  \global\let\svgscale\undefined%
  \makeatother%
  \begin{picture}(1,0.92196052)%
    \lineheight{1}%
    \setlength\tabcolsep{0pt}%
    \put(0,0){\includegraphics[width=\unitlength,page=1]{admissible_sets_kl_corner.pdf}}%
    \put(0.31734945,0.84331414){\color[rgb]{0,0,0}\makebox(0,0)[lt]{\begin{minipage}{0.32835307\unitlength}\raggedright $\mathcal{M}^\mathbf{r}_{\mathbf{f}^*} $\end{minipage}}}%
    \put(0,0){\includegraphics[width=\unitlength,page=2]{admissible_sets_kl_corner.pdf}}%
    \put(0.70130933,0.40527584){\color[rgb]{0,0,0}\makebox(0,0)[lt]{\begin{minipage}{0.42107253\unitlength}\raggedright $\mathbf{f}^*$\end{minipage}}}%
    \put(0.07988819,0.98249584){\color[rgb]{0,0,0}\makebox(0,0)[lt]{\begin{minipage}{0.46406699\unitlength}\raggedright $\mathbf{r}^{-1}$ Ray \end{minipage}}}%
    \put(0.11842633,0.75332824){\color[rgb]{0,0,0}\makebox(0,0)[lt]{\begin{minipage}{0.30530812\unitlength}\raggedright $\mathbf{f}_\mathbf{r}^*$\end{minipage}}}%
    \put(0,0){\includegraphics[width=\unitlength,page=3]{admissible_sets_kl_corner.pdf}}%
    \put(0.50048085,0.74904046){\color[rgb]{0,0,0}\makebox(0,0)[lt]{\begin{minipage}{0.30530812\unitlength}\raggedright {\color{red}$-\mathbf{a}$}\end{minipage}}}%
    \put(0.6760111,0.16658044){\color[rgb]{0,0,0}\makebox(0,0)[lt]{\begin{minipage}{0.79567061\unitlength}\raggedright {\color{blue}$\mathrm{Col}(\mathrm{F}\mathrm{F}^T)$}\end{minipage}}}%
    \put(0.35156596,0.34787562){\color[rgb]{0,0,0}\makebox(0,0)[lt]{\begin{minipage}{0.35942408\unitlength}\raggedright {\color{yellow}$-\mathbf{v}_1$}\end{minipage}}}%
    \put(0,0){\includegraphics[width=\unitlength,page=4]{admissible_sets_kl_corner.pdf}}%
    \put(-0.01315493,0.79108677){\color[rgb]{0,0,0}\makebox(0,0)[lt]{\begin{minipage}{0.186903\unitlength}\raggedright $f_2$\end{minipage}}}%
    \put(0.84081705,0.08845298){\color[rgb]{0,0,0}\makebox(0,0)[lt]{\begin{minipage}{0.1913756\unitlength}\raggedright $f_1$\end{minipage}}}%
    \put(0,0){\includegraphics[width=\unitlength,page=5]{admissible_sets_kl_corner.pdf}}%
    \put(0.73660958,0.31838372){\color[rgb]{0,0,0}\rotatebox{-15.47678444}{\makebox(0,0)[lt]{\lineheight{1.25}\smash{\begin{tabular}[t]{l}Pareto Front\end{tabular}}}}}%
    \put(0,0){\includegraphics[width=\unitlength,page=6]{admissible_sets_kl_corner.pdf}}%
  \end{picture}%
\endgroup%